\documentclass[10pt,twocolumn]{article} 
\usepackage{simpleConference}
\usepackage{times}
\usepackage{graphicx}
\usepackage{amssymb}
\usepackage[hyphens]{url}
\usepackage{hyperref}
\usepackage[numbers]{natbib}
\usepackage{caption}
\usepackage{subcaption}
\usepackage{amsthm}
\usepackage{multirow}
\usepackage{booktabs}
\usepackage{amsmath}
\usepackage[]{algorithm2e}
\usepackage{color}
\usepackage{multicol}
\usepackage{bm}
\usepackage{authblk}
\usepackage{commath}    
\usepackage[capitalize]{cleveref}
\usepackage[outline]{contour}
\usepackage{todonotes}
\usepackage[export]{adjustbox}

\DeclareMathOperator*{\argmax}{argmax}

\crefname{section}{Sec.}{Secs.}
\Crefname{section}{Section}{Sections}
\Crefname{table}{Table}{Tables}
\crefname{table}{Tab.}{Tabs.}
\crefname{appendix}{Appx.}{Appxs.}
\crefname{figure}{Fig.}{Figs.}
\crefname{theorem}{Theorem}{Theorem}

\title{Class Attribute Inference Attacks: \\ Inferring Sensitive Class Information by Diffusion-Based Attribute Manipulations}

\makeatletter
\newcommand{\printfnsymbol}[1]{%
  \textsuperscript{\@fnsymbol{#1}}%
}
\makeatother

\author[1]{Lukas Struppek\thanks{Contact: lukas.struppek@cs.tu-darmstadt.de}}
\author[1]{Dominik Hintersdorf}
\author[1]{Felix Friedrich}
\author[1]{Manuel Brack}
\author[1,3,4,5]{\newline Patrick Schramowski}
\author[1,2,3,4]{Kristian Kersting}
\affil[1]{Department of Computer Science, Technical University of Darmstadt, Germany}
\affil[2]{Centre for Cognitive Science, Technical University of Darmstadt}
\affil[3]{German Research Center for Artificial Intelligence (DFKI)}
\affil[4]{Hessian Center for AI (hessian.AI), Germany}
\affil[5]{LAION}

\begin{document}
\maketitle
\thispagestyle{empty}

\begin{abstract}
    Neural network-based image classifiers are powerful tools for computer vision tasks, but they inadvertently reveal sensitive attribute information about their classes, raising concerns about their privacy. To investigate this privacy leakage, we introduce the first \textbf{C}lass \textbf{A}ttribute \textbf{I}nference \textbf{A}ttack (\textsc{Caia}), which leverages recent advances in text-to-image synthesis to infer sensitive attributes of individual classes in a black-box setting, while remaining competitive with related white-box attacks. Our extensive experiments in the face recognition domain show that \textsc{Caia} can accurately infer undisclosed sensitive attributes, such as an individual's hair color, gender, and racial appearance, which are not part of the training labels. Interestingly, we demonstrate that adversarial robust models are even more vulnerable to such privacy leakage than standard models, indicating that a trade-off between robustness and privacy exists.
\end{abstract}

\section{Introduction}
Classifying images with neural networks is widely adopted in various domains~\citep{esteva21medical, ibrahim22cancer, baumann18roads, barua21covid19}. Face recognition systems~\citep{guo19face_rec}, for example, take facial images as input and attempt to predict the depicted person's identity. In the pursuit of enhancing a model's predictive performance, privacy concerns of the acquired knowledge are often disregarded and moved into the background. However, correctly assessing and mitigating the risk of compromising private information is crucial in privacy-sensitive domains, as neglect can lead to the disclosure of private information~\citep{shokri2017, struppek22_mia, hintersdorf_clip}. Face recognition techniques are a fundamental component of many real-life applications, e.g., criminal identification, video surveillance, access control, and autonomous vehicles.

Smart home devices~\citep{abuassoa21accesscontrol}, for example, contain face recognition models for access control and user authorization. Users expect these models to recognize them reliably, but at the same time to not reveal information about their appearance to third parties. However, this assumption does not necessarily hold true, and malicious parties could extract sensitive features about users without any further information about their appearance by simply interacting with the trained classification model in a black-box fashion.

We investigate the privacy leakage of image classifiers and demonstrate that models indeed leak sensitive class information even without any specific information about the classes, training samples, or attribute distributions required by the adversary. We focus our investigation on face recognition models that are trained on labeled datasets where each class corresponds to a different person's identity. During the training stage, the victim has access to a private dataset of facial images and trains an image classifier on it. After training, the face recognition model makes a prediction on the identity of a given input facial image. 

\begin{figure*}[ht]
     \centering
     \begin{subfigure}[b]{0.42\textwidth}
         \centering
         \includegraphics[height=2.9cm]{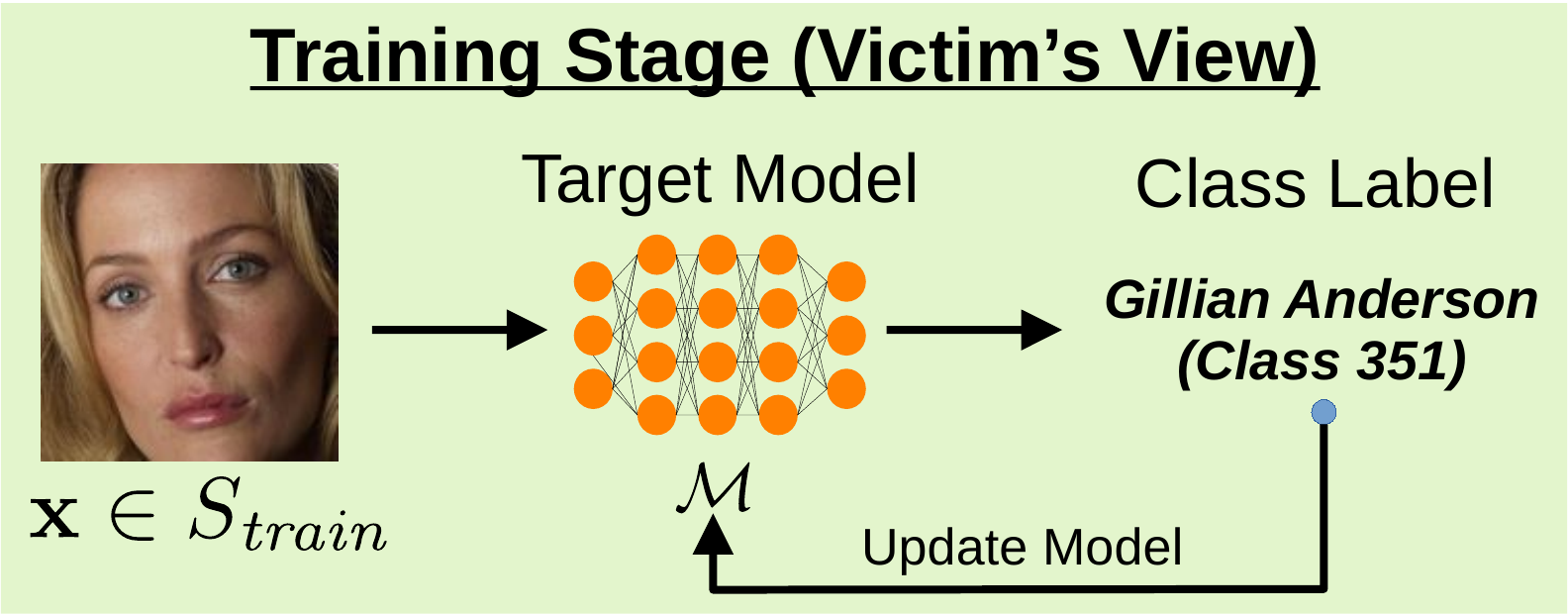}
         \caption{The target model is trained on private identities.}
         \label{fig:caia_overview_victim}
     \end{subfigure}
     \hfill
     \begin{subfigure}[b]{0.575\textwidth}
         \centering
         \includegraphics[height=2.9cm]{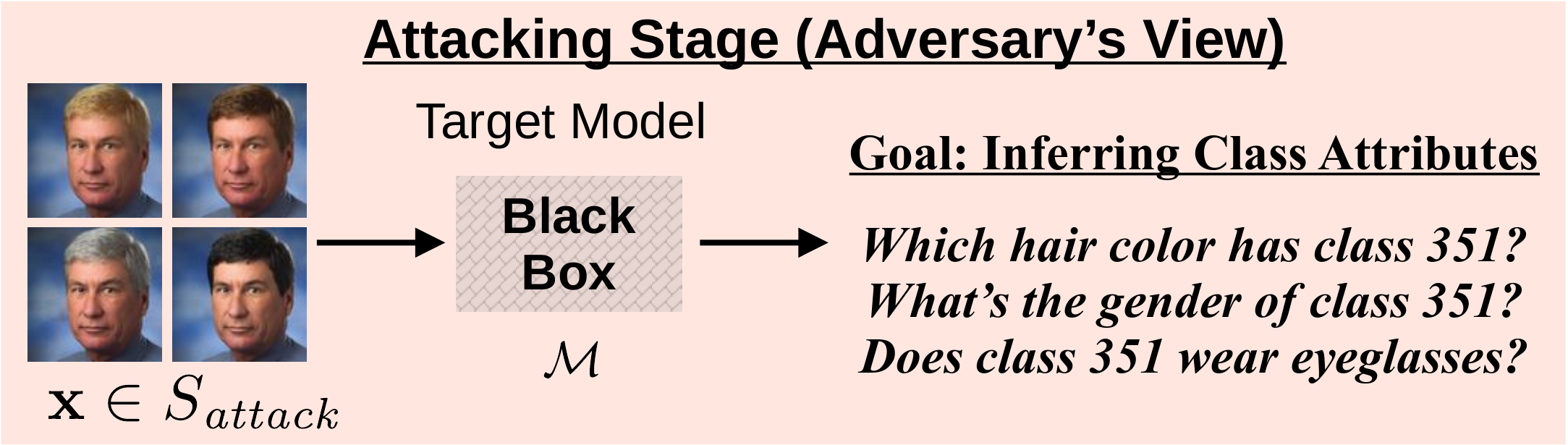}
         \caption{The adversary tries to infer sensitive information about the private identities.}
         \label{fig:caia_overview_adv}
     \end{subfigure}
        \caption{Overview of the setting of our proposed Class Attribute Inference Attack (\textsc{Caia}). We investigate the setting of face recognition systems, which is a multi-class classification task for which each class corresponds to the identity of an individual. The victim (\ref{fig:caia_overview_victim}) first trains a model on a private training set $S_{train}$ to predict the identity of a given input image. Whereas the victim has access to a labeled training set and knows all the identities of the training samples, the adversary (\ref{fig:caia_overview_adv}) only has black-box access to the trained model without any information about the individual classes or training samples. The adversary then tries to infer sensitive attributes about the individual class identities, e.g., their hair color, gender, or racial appearance. For this, the adversary has access to a separate attack dataset $S_{attack}$ of publicly available facial images but has no information about $S_{train}$.}
        \label{fig:caia_overview}
\end{figure*}

Our research shows that these models reveal sensitive details about the different identities within their outputs, such as gender, hair color, and racial appearance. \cref{fig:caia_overview} illustrates the basic setting of our investigation, in which the adversary has only black-box access to the target model and no specific information about the appearance of individual identities. The goal of the attacking stage is then to infer sensitive information about the identities from the target model's training data.

Within our analysis, we introduce a \textit{\textbf{C}lass \textbf{A}ttribute \textbf{I}nference \textbf{A}ttack} (\textsc{Caia}), which enables an adversary to infer sensitive attributes of a specific class from a trained image classifier with high accuracy. Phrased differently, \textsc{Caia} allows the creation of a profile of the individual classes by only interacting with the trained classifier through the extraction of class attributes that have not been explicitly part of the training objective. For the attack, we utilize recent advances in text-to-image synthesis to craft images that only differ in one attribute by editing real images with textual guidance. We then exploit that image classifiers, as we show, assign higher logits to inputs that share the same sensitive attribute with the training samples of a class, which allows us to infer class information by only observing the input-output relation of the trained target model. 

Compared to related inference attacks~\citep{fredrikson14pharmacogenetics,fredriskon15mia,secret_revealer, struppek22_mia}, \textsc{Caia} is model-agnostic and requires only black-box access and basic domain knowledge. Once the attack images are crafted, attribute inference requires only a single model forward pass of the generated samples.

In our extensive evaluation, we demonstrate the success of \textsc{Caia} and show that robust models trained with adversarial training~\citep{goodfellow15adv, madry18pgd} are even more susceptible to these attacks. They leak more information about their individual classes than non-robust models, even if their prediction accuracy is substantially lower. We provide a formal explanation based on the insights that robust models tend to lay their focus on robust image features, which are connected to sensitive attributes. This indicates a trade-off for model designers: making a model robust to adversarial examples comes at the expense of higher privacy leakage.

In summary, we make the following contributions:
\begin{itemize}
    \item We introduce a novel \textit{\textbf{C}lass \textbf{A}ttribute \textbf{I}nference \textbf{A}ttack} (\textsc{Caia}) to infer sensitive attributes of specific classes from image classifiers.
    \item By utilizing recent image manipulation capabilities of diffusion models, \textsc{Caia} can infer sensitive attributes with high precision.
    \item We show that robustly-trained models even leak more information, indicating a trade-off between model robustness and privacy leakage.
\end{itemize}

The paper is structured as follows: We first provide in \cref{sec:background} an overview of the background and related work, including attribute inference and model inversion attacks. We then introduce the setting and procedure of our Class Attribute Inference Attack (\textsc{Caia}) in \cref{sec:caia}. An extensive experimental evaluation in \cref{sec:experiments} demonstrates the strong performance of \textsc{Caia} and investigates the trade-off between robust models and their privacy leakage. An ablation and sensitivity analysis concludes our experiments. We discuss our findings together with open challenges in \cref{sec:discussion}, as well as ethical considerations of our research in \cref{sec:ethical_considerations}. A summary and overview of future avenues conclude the paper in \cref{sec:conclusion}.
\\
\\
\noindent\textbf{Disclaimer:} \textit{This paper investigates the extraction of sensitive identity information, including gender and racial appearance of people. The groups used (Asian, Black, Indian, White) are in line with current research and follow the taxonomy of \citet{karkkainen21fairface}, which itself is adopted from the US Census Bureau~\citep{uscensus22}. Importantly, we emphasize that this work's goal is to investigate the leakage of sensitive attributes in image classifiers. We do not intend to discriminate against identity groups or cultures in any way.}

\section{Background and Related Work}\label{sec:background}
We start by introducing the background and related work for our analysis. We focus here on attribute inference attacks, which exploit models trained on tabular data to infer hidden attributes for a given input sample, and model inversion attacks, which aim to reconstruct the appearance of training images with the help of generative models. We then highlight the novelty of \textsc{Caia} compared to existing inference attacks. Furthermore, we present adversarial training to make a model robust against adversarial examples.

\subsection{Attribute Inference Attacks}
In recent years, various types of inference attacks have been proposed. Those include membership inference attacks~\citep{shokri2017, yeom2018privacy, choquette2021, hintersdorf22trust, hintersdorf_clip}, which attempt to identify training samples from a larger set of candidates, and property inference attacks~\citep{ganju18property, parisot21propertycnn, zhou22property, wang22group}, which try to infer general properties and statistical information about a model's training data. Most related to our work are attribute inference attacks (AIAs)~\citep{fredrikson14pharmacogenetics}, which aim to infer sensitive attribute values of an incomplete data record in the context of classification and regression models. More specifically, the adversary has access to a target model $\mathcal{M}$, which has been trained on a tabular dataset $S_\mathit{train}$, sampled from the distribution $\mathcal{D}$. Each training sample is a triplet $(x_s, x_n, y)$ and consists of some sensitive attributes $x_s$ and non-sensitive attributes $x_n$ together with a ground-truth label $y$. The adversary has access to a set of candidates $(x_n, y)\subseteq S_\mathit{train}$ with the sensitive attribute values missing. AIAs try to infer the sensitive values $x_s$ by exploiting $\mathcal{M}$ and its learned information about distribution $\mathcal{D}$. 

Fredrikson et al.~\citep{fredrikson14pharmacogenetics, fredriskon15mia} proposed maximum-a-posterior AIAs that, assuming all attributes are independent, predict the sensitive attribute value that minimizes the adversary's expected misclassification rate. \citet{yeom2018privacy} extended the approach and combined attribute inference with membership inference attacks. \citet{mehnaz22attributes} introduced an attack based on the softmax scores of the target model, assuming that the model's prediction is more likely to be correct and confident if the input sample contains the true sensitive attribute value. Common AIAs make strong assumptions regarding the adversary's knowledge that is generally hard to gain under realistic assumptions, e.g., the adversary knows the marginal prior of sample attributes~\citep{fredrikson14pharmacogenetics, fredriskon15mia, yeom2018privacy} or the target model's confusion matrix on its training data~\citep{fredriskon15mia,mehnaz22attributes}. \citet{jayaraman22imputation} questioned previous black-box AIAs and empirically showed that those attacks could not reveal more private information than a comparable adversary without access to the target model. They conclude that black-box AIAs perform similarly well as data imputation techniques to fill the missing attribute values. To improve AIAs over data imputation, they proposed a white-box attack that outperforms imputation in settings with limited data and skewed distributions.

\subsection{Model Inversion Attacks}
All presented AIAs are limited to tabular data and are not applicable to image classification since the variation of single image attributes, e.g., changing the hair color in a facial image, is not trivially possible. Moreover, the AIA setting itself is not transferable to the vision domain, since it is unclear how an adversary can have access to incomplete images with only one attribute missing. This fact also makes it impossible to directly compare our \textit{\textbf{C}lass \textbf{A}ttribute \textbf{I}nference \textbf{A}ttack} (\textsc{Caia}) with common AIAs and corresponding defenses.

Another class of attacks, so-called model inversion attacks (MIAs), try to fill this gap for image classification. We note that the notion of MIAs is not consistent in the literature, and the term is sometimes also used for AIAs. Generally, given a classification model, an adversary attempts to create synthetic input samples that either reconstruct samples from the model's training data~\citep{fredriskon15mia, secret_revealer,knowledge_mia,kahla22labelmia} or craft synthetic samples that reflect the characteristics of a specific class~\citep{variational_mia,struppek22_mia}. Whereas most MIAs require access to samples from the target training distribution for training a custom generative adversarial network (GAN), \citet{struppek22_mia} recently proposed Plug \& Play (PPA) MIAs, which make the attacks agnostic to the target model, increase their flexibility, and enhance their inference speed by utilizing pre-trained GANs. We will, therefore, use PPA as a baseline for comparing with our \textsc{Caia}.

Formally, generative model inversion attacks rely on a generative model $G:W\to X$, which is trained to map latent vectors $w\in W$ sampled from a probability distribution to the image space $X$ to synthesize facial images. The adversary then optimizes the sampled latent vectors using the target model $\mathcal{M}$ to find meaningful representations of the target identity on the generative model's learned image manifold. By analyzing the corresponding generated images, the adversary is able to infer sensitive visual features of the individual identities for which $\mathcal{M}$ has been trained to recognize. The optimization goal can be formulated as
\begin{equation}
\min_{\hat{w}} \mathcal{L}(\mathcal{M}(G(\hat{w}), y),
\end{equation}
which optimizes latent vector $\hat{w}$ for the target class $y$ using a suitable loss function $\mathcal{L}$. For the optimization, PPA minimizes a Poincaré loss~\citep{poincare} between the generated images and the target identity to overcome the vanishing gradient problem and also performs random transformations on the generated images to avoid overfitting and misleading results. The basic underlying generative model used in PPA is a pre-trained StyleGAN2~\citep{Karras2019stylegan2}.

\subsection{Novelty of \textsc{Caia}} 
In contrast to previous work on AIAs, we move the scope of inference attacks from the sample level to a class level in the vision domain, with the goal of inferring sensitive information about the distinct classes learned by a model. To achieve this, we use the latest advancements in text-to-image synthesis to manipulate single attributes of input images, resulting in consistent images that differ only in the targeted attribute. Our approach is more efficient than MIAs as it requires only black-box access to the target model and no extensive knowledge of the training data distribution. Furthermore, the inference step is done in seconds, since \textsc{Caia} does not require any further sample optimization after constructing the initial attack dataset. This makes \textsc{Caia} more flexible and target-independent than previous AIAs and MIAs. Throughout this work, we focus on the privacy-sensitive domain of face recognition systems. The goal is to infer sensitive information about identities, e.g., their gender or racial appearance, without any specific information about the individual identity or underlying training set distributions available to the adversary. 

\subsection{Adversarial Robust Training}
Besides privacy attacks, neural networks are known to be susceptible to adversarial examples~\citep{szegedy14intriguing, goodfellow15adv}. Formally, adversarial examples are crafted by adding an optimized perturbation $\delta$ with $\|\delta\|\leq \epsilon$ to a model input $x$ to maximize the model's loss $\mathcal{L}(\mathcal{M}(x+\delta),y)$ for the true label $y$. One of the most reliable and commonly used defenses is adversarial training~\citep{goodfellow15adv, madry18pgd}, which updates a model's weights $\theta$ on adversarial examples. During each training step, a sample-wise worst-case perturbation is computed in an $\epsilon$-environment around the clean samples to maximize the model's confusion. These perturbations are then added to the clean samples to train the model and make it robust against such manipulations. Formally, this comes down to a min-max optimization: 

\begin{equation}
    \min_\theta \sum_{(x,y)\in S_\mathit{train}} \max_{\|\delta\| \leq \epsilon} \mathcal{L}(\mathcal{M}(x+\delta), y) \,.
\end{equation}

Since the inner maximization problem cannot be solved numerically in tractable time, local search algorithms are applied to craft adversarial examples, e.g., FGSM~\citep{goodfellow15adv, wong20ffgsm} or PGD~\citep{madry18pgd}. By training on adversarial examples, the model becomes more robust against adversarial perturbations. In our experiments, we also investigate the influence of model robustness on its privacy leakage. 
\section{Class Attributes Inference Attacks}\label{sec:caia}
We now introduce the novel \textit{\textbf{C}lass \textbf{A}ttribute \textbf{I}nference \textbf{A}ttack} (\textsc{Caia}), which consists of two steps. First, the adversary generates a set of attack samples by creating different versions of images through a generative approach that alters the sensitive attribute values. In the second step, these attack samples are used to infer the sensitive attribute values for the identities in a face recognition model. The underlying assumption of the attack is that image classifiers assign higher prediction scores to samples that share the sensitive attribute value with the training samples of a class. Before delving into the details, we outline our general threat model. 

\paragraph{Adversary's Goal.}
Let $\mathcal{M}\colon X \to \mathbb{R}^{|Y|}$ denote the trained target image classifier, which takes input images $x\in X$ and computes prediction scores for each class label $y \in Y$. The model's training data $S_\mathit{train}$ consisted of labeled data samples $(x, y) \sim \mathcal{D}$. The underlying attack assumption is that samples of a certain class share a constant sensitive attribute value $z \in Z$, which is not part of the class label but is implicitly encoded in the image features. For example, a face recognition model is a classifier trained to predict the identity $y$ of each facial image $x$. A sensitive attribute $z$ in this context might be the gender appearance or hair color of a specific identity. The attack goal is to infer the value of this sensitive attribute for each individual class. \cref{fig:caia_overview} demonstrates the general setting, in which the victim trains the target model on labeled images of various identities. In the depicted case, class $y$ corresponds to the identity of the actress \textit{Gillian Anderson}. The adversary then tries to predict sensitive attributes about the class $y$ without any further information such as the name or training samples of the person's available. We note that multiple classes can share the same attribute, e.g., having the same hair color.

\paragraph{Adversary's Capabilities.} The adversary has only black-box access to the trained target model $\mathcal{M}$, i.e., the adversary can query the target model and observe its output logits. Furthermore, the adversary knows the domain of the target model's training data, e.g., facial images, but not the exact training data distribution or labels. Instead, the adversary can sample images from a data distribution $\hat{\mathcal{D}}$ from the same domain. Note that the images available to the adversary contain no identity labels and have no overlapping with the target models training data. For the sensitive attributes, the adversary defines an individual set of possible values to infer. We emphasize that \textsc{Caia} is model- and task-agnostic and does not require white-box access to the target model. Information about the prior distribution of the sensitive attributes is neither available nor required.

\begin{figure*}[t]
    \centering
    \includegraphics[width=\linewidth]{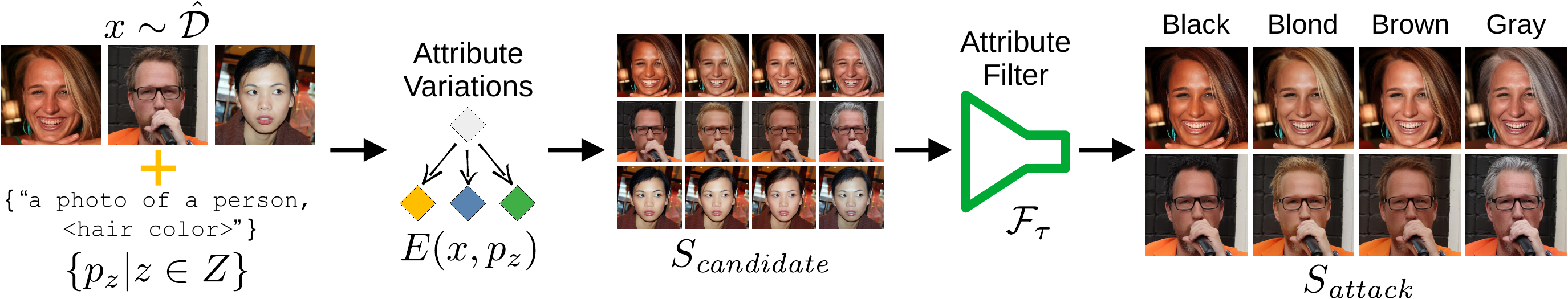}
    \caption{Overview of our attack dataset crafting process for the sensitive attribute \textit{hair color}, which has four possible values. Real images are used to generate image variations by modifying characteristics associated with the sensitive attribute. The resulting candidate images are then filtered to ensure that each sample accurately reflects the intended attribute values. The final output of this process is the set of attack samples.}
    \label{fig:synthesis_concept}
\end{figure*}

\subsection{Crafting Attack Samples}\label{sec:craft_attack_samples}
While it is easy for tabular data to vary the values of attributes in an inference sample, this is a non-trivial task on images since attributes are encoded in pixels, and multiple pixels define not only a single attribute but a set of possibly entangled attributes. For example, human faces contain information such as age and skin color, among many more. Furthermore, changing the value for any of these attributes requires semantically consistent changes in multiple pixels. 

To enable meaningful image manipulations, we utilize recent advances in text-to-image synthesis. Recent generative adversarial models (GANs)~\citep{goodfellow_gan, Karras_stylegan1, Karras2019stylegan2} also offer attribute manipulations by first projecting images into the latent space and then changing the individual latent vector by moving into directions correlated with the target attribute. However, this procedure requires a faithful inversion process to compute the corresponding latent vectors~\citep{subramanyam22styleganinversion,alaluf22hyperstyle,dinh22hyperinverter} and a preceding analysis of the latent space to discover directions~\citep{parihar22exploration,pajouheshgar22optimizing,abdal22clip2stylegan} for the desired content changes. Both components are challenging on their own. We, therefore, choose text-guided image synthesis, which allows us to describe desired content changes with natural language and does not require explicit exploration of the latent space.

Text-to-image synthesis systems like Stable Diffusion~\citep{Rombach2022} are able to generate high-quality images following a user-provided text description $p$. \citet{mokady22nullinversion} recently proposed \textit{Null-text Inversion} to encode real images into the domain of diffusion models and enable text-based editing while keeping the overall image composition and content fixed. In combination with \textit{Prompt-to-Prompt}~\citep{hertz2022prompt}, it allows a user to instruct image edits $x_\mathit{edit}=E(x, p)$ on images $x$ conditioned on a description $p$. We apply \textit{Null-text Inversion} to generate variations of existing images by changing only the sensitive attribute values, such as the hair color or gender appearance while aiming to leave other image aspects unchanged. 

\cref{fig:synthesis_concept} illustrates the crafting process for the attack samples. Formally, the adversary has access to a data distribution $\mathcal{\hat{D}}$ from which to sample images $x$. Note that the attack does not require the attack distribution $\mathcal{\hat{D}}$ to be the same as the training data distribution $\mathcal{D}$ but only that both data distributions are from the same domain, e.g., facial images. As we will show in our experimental evaluation, even if the style, size, and quality of images between both distributions vary significantly, the attack is still highly successful. 

The adversary defines the target attribute with a set of $k$ possible attribute values $Z=\{z_1,\ldots, z_k \}$ and corresponding edit prompts $p_z$ that describe the general domain and explicitly state an attribute value $z\in Z$. For example, $p_z=\text{"A photo of a person, } \langle \text{gender}\rangle\text{"}$, where $\langle\text{gender}\rangle$ is replaced by $z\in \{\text{female appearance}, \text{male appearance}\}$. The candidate dataset $S_\mathit{candidate}$ is then constructed as
\begin{equation}
    S_\mathit{candidate}=\{E(x, p_z) | z\in Z, x \sim \mathcal{\hat{D}}\},
\end{equation} 
consisting of image tuples $\mathbf{x}=(x_1,\ldots,x_k)$, each containing $k$ images with different sensitive attribute values. 

However, the attribute manipulation might not always succeed in changing an image attribute to the desired value. This can be due to interfering concepts already present in the image that are strongly entangled with the target attribute. For example, changing the hair color of a person with a dark skin tone to blonde often fails because such attribute combinations are rather rare in the actual training data, which is also reflected in the underlying diffusion model. We, therefore, employ a filtering approach, i.e., filtering out all sample tuples $\mathbf{x}$ that are not correctly depicting the various attribute values. For this, we use a trained attribute classifier $\mathcal{F}_\tau \colon X \to Z$ to create a subset 
\begin{equation}
    S_\mathit{attack}=\{\mathbf{x} \in S_\mathit{candidate} | \mathcal{F}_\tau(x_z)=z, \forall z\in Z \}
\end{equation}
of sample tuples $\mathbf{x}=(x_1,\ldots,x_k)$. The attribute classifier computes for each input image a probability score that the attribute value $z\in \mathcal{Z}$ is present in the image. We also add a threshold $\tau$ on the softmax scores of the attribute classifier and classify only predictions with a softmax score $\geq \tau$ as correct. This removes images for which the attribute classifier has only low confidence in its prediction. For the example of hair color, each resulting tuple $\mathbf{x}\in S_{attack}$ consists of four facial images $(x_1, \ldots, x_4)$ that only differ in the depicted hair color of the person. This use case is also depicted in the attack dataset crafting process in \cref{fig:synthesis_concept}.

\begin{figure*}[t]
    \centering
    \includegraphics[width=\linewidth]{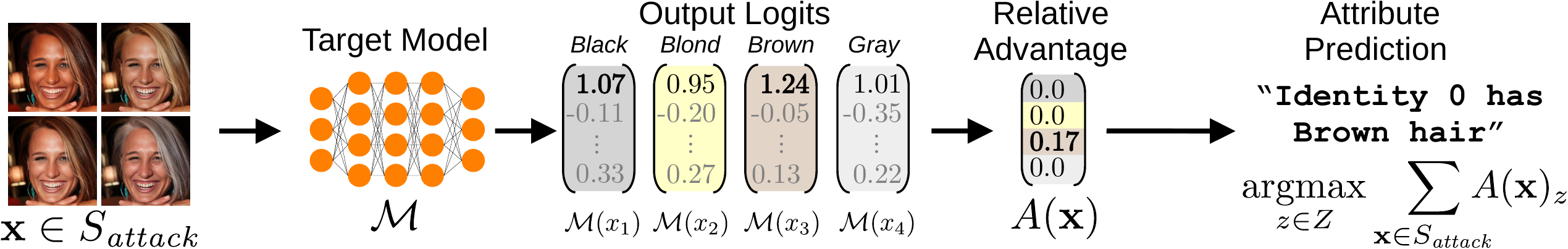}
    \caption{Overview of the class attribute inference step. Each image tuple from the attack set is fed sequentially into the target model to compute the logits for the target class. The relative advantage is then computed by subtracting the second-highest logit value from the maximum value, and this difference is added to a running sum for each sensitive attribute value. The final prediction for the sensitive attribute is the value with the highest relative advantage computed across all attack samples.}
    \label{fig:inferring_concept}
\end{figure*}

\subsection{Revealing Sensitive Attributes}
After crafting the attack samples, we can now begin inferring the sensitive class attributes learned by the target classifier. We recall that the adversary has no specific information about the distinct classes and particularly no access to the training data samples. \cref{fig:inferring_concept} illustrates the basic concept of the inference process. In the depicted case, the adversary tries to infer the sensitive attribute \textit{hair color} of the first identity, which corresponds to the first class of the face recognition model.

While our filtering approach ensures that the variations of an image depict the different values of the sensitive attribute, other confounding behavior might still remain unintentionally. For example, changing a person's hair color to gray might also influence the depicted age. To mitigate such influences, we predict the sensitive attribute with a variety of different attack samples to, in turn, reduce the influence of these confounding factors over a larger number of samples.

Be $\mathcal{M}(x)_y\colon X \to \mathbb{R}$ the pre-softmax logits computed by the target model $\mathcal{M}$ on input image $x$ for class $y$. To infer the sensitive attribute value $z$ of class $y$, we query the target model consecutively with multiple sample tuples $\mathbf{x}\in S_\mathit{attack}$. We then compute for each tuple $\mathbf{x}$ the relative advantage $A(\mathbf{x})\in \mathbb{R}^{|Z|}$. For each $x_z \in \mathbf{x}$, the relative advantage component $A(\mathbf{x})_z$ is defined by
\begin{equation}
    A(\mathbf{x})_z = \max\left(0, \, \mathcal{M}(x_z)_y - \max_{\tilde{x} \in \mathbf{x}, \tilde{x}\neq x_z}{\mathcal{M}(\tilde{x})_y} \right).
\end{equation}

The relative advantage computes the difference between the highest and the second-highest logit values and assigns this difference to the attribute sample $x_z \in \mathbf{x}$ with the highest logit. For all other samples $\tilde{x}\in\mathbf{x}$ with $\tilde{x} \neq x_z$, the relative advantage is set to zero. \cref{fig:inferring_concept} illustrates the relative advantage computation for a single $\mathbf{x}$. In the depicted case, the sample depicting \textit{brown hair color} achieves the highest logit value and its relative advantage of $A(\mathbf{x})_{brown}=0.17$ describes the difference to the second highest logit value assigned to the sample with the attribute \textit{black hair color}. The relative advantage of the other three attribute values is consequently set to zero.

The final attribute prediction $\hat{z}$ is then done by taking the attribute with the highest relative advantage summed up over all attack samples: 
\begin{equation}
    \hat{z}=\argmax_{z\in Z} \sum_{\mathbf{x} \in S_\mathit{attack}}A(\mathbf{x})_z \, .
\end{equation} 

We emphasize that only a single forward pass on all attack samples is sufficient to compute the relative advantage for all target classes, which makes the inferring step computationally very cheap, e.g., the gender inference for a ResNet-101 model and 500 identities took roughly 2 seconds in our experiments.
\begin{table}[t]
\centering
\resizebox{\linewidth}{!}{  
\begin{tabular}{lllllc}
\toprule
    \textbf{Attribute} & \textbf{Value} & \textbf{Prompt} \\
\midrule
    \multirow{2}{*}{Gender}     & Female & \textit{female appearance} \\
                                & Male & \textit{male appearance} \\
\midrule
    \multirow{2}{*}{Glasses}    & No Glasses & \textit{no eyeglasses} \\
                                & Glasses & \textit{wearing eyeglasses} \\
\midrule
                                & Asian & \textit{with asian appearance} \\
    Racial                      & Black & \textit{with black skin} \\
    Appearance                  & Indian & \textit{with indian appearance} \\
                                & White &  \textit{with white skin} \\
\midrule
    \multirow{4}{*}{Hair Color} & Black & \textit{with black hair} \\
                                & Blond & \textit{with blond hair} \\
                                & Brown &  \textit{with brown hair} \\
                                & Gray & \textit{with gray hair} \\
\bottomrule
\end{tabular}
}
\caption{Prompts for attack dataset generation. Each prompt is appended to the string \textit{"A photo of a person, }$\langle\text{ }\rangle$\textit{"} by replacing $\langle\text{ }\rangle$ with the attribute-specific prompt.}
\label{tab:attack_prompts}
\end{table}

\section{Experimental Evaluation}\label{sec:experiments}
Next, we experimentally investigate the information leakage of face recognition models with \textsc{Caia}. We provide our source code with all hyperparameters to reproduce the experiments and facilitate future research. More experimental details are provided in \cref{appx:experimental_details}.

\subsection{Experimental Setup}
\paragraph{Training Datasets.} We used the CelebA facial attributes dataset~\citep{celeba} to train our target face recognition systems. CelebA contains labeled images of 10,177 individuals and additional 40 binary attribute annotations per image. We selected the following sensitive attributes:
\begin{itemize}
    \item \textit{gender} $=\{\text{female}, \text{male}\}$
    \item \textit{eyeglasses} $=\{\text{no eyeglasses}, \text{eyeglasses}\}$
    \item \textit{hair color} $= \{ \text{black}, \text{blond}, \text{brown}, \text{gray}\}$
    \item \textit{racial appearance} $= \{\text{Asian}, \text{Black}, \text{Indian}, \text{White}\}$
\end{itemize}

For the first three attributes, the CelebA dataset already provides corresponding labels. To infer the attribute \textit{racial appearance}, we applied a pretrained \textit{FairFace}~\citep{karkkainen21fairface} classifier to label each image. Since the provided attributes and labels are often inconsistent for samples of one identity, e.g., people do not always wear eyeglasses or might dye their hair, we created custom subsets for each sensitive attribute group by selecting an equal number of identities for each attribute value and removed samples with inconsistent labels. This is important for evaluation because otherwise the training samples of an identity depict various values for the sensitive attribute, and no clear ground-truth value can be defined. 

We trained various target models on these datasets with the standard cross-entropy loss to predict a person's identity. Note that the attribute labels were not part of the training process. Since not every attribute is present with every identity, we selected the 100 identities for each attribute value of \textit{hair color}, \textit{eyeglasses}, and \textit{racial appearance}, respectively, with the most training samples available. For \textit{gender}, we selected 250 identities per attribute value. We also created larger training datasets with a total of 1,000 identities to see if more identities influence the attack's success. 

Additional target models were trained on the FaceScrub~\citep{facescrub} facial image dataset, which contains images of 530 identities with equal gender split. Images are available in a cropped version that only contains a person's face, and an uncropped version that contains the original image before cropping. Those uncropped images often not only depict a single person but also shows other people in the background which makes training a face recognition system more challenging. We split all datasets into 90\% for training and 10\% for testing the models' prediction accuracy. Further details on the different dataset statistics are provided in \cref{appx:dataset_details}.

\paragraph{Attack Datasets.}
To craft the attack datasets, we used the Flickr-Faces-HQ (FFHQ)~\cite{Karras_stylegan1} and CelebAHQ~\citep{karras18progressive} datasets. Both dataset contain facial images, but compared to CelebA and FaceScrub, these datasets have a much higher resolution, depict a person's whole head and additional background information. Since the FFHQ dataset consists of images collected from the image hosting platform Flickr, we simulate an adversary who collects samples from public sources, e.g., social media platforms. We then generated and filtered images with attribute manipulations to collect 300 attack image tuples for each attribute group. \cref{tab:attack_prompts} states the editing prompts used to manipulate the features of the attack dataset samples. We further set the filter threshold $\tau=0.6$ in all experiments. We visualize randomly selected training and attack samples in \cref{appx:sample_visualizations}.

\paragraph{Training Hyperparameters.}
We trained ResNet-18, ResNet-101, ResNet-152~\citep{resnet_he}, ResNeSt-101~\citep{zhang2020resnest}, and DenseNet169~\citep{densenet} target models. For each dataset-architecture combination, we trained three models with different seeds.  We emphasize that we did not aim for achieving state-of-the-art performances but rather tried to train models with generally good prediction performance.

To investigate the effects of model robustness on information leakage, we trained adversarially robust models on FaceScrub with standard adversarial training~\citep{goodfellow15adv}. We perturbed training samples with Projected Gradient Descent~\citep{madry18pgd} with 7 steps, $\epsilon=4/255$, and step size $\alpha=1/255$\footnote{Corresponds to images with pixel values in range $[0,1]$.}.

We further trained ResNet-50 models on the CelebA attributes for gender, eyeglasses, and hair color for filtering the candidate images. To mitigate the overconfidence of neural networks, we trained all filter models with label smoothing~\citep{szegedy16labelsmoothing, mueller19labelsmoothing} with a smoothing factor of $\alpha=0.1$ to calibrate the models and make them more suitable for the confidence threshold filtering. Since some attributes are more frequent in the CelebA dataset, we draw the same number of samples for each attribute value by oversampling/undersampling from the attribute images. To filter images depicting different \textit{racial appearances}, we relied on the pre-trained \textit{FairFace} classifier. Details on the training of FairFace for filtering the racial appearance images can be found at \url{https://github.com/dchen236/FairFace}.

We state additional training details, including image preprocessing, augmentation, and optimization parameters in \cref{appx:training_hyperparameters}.

\paragraph{Metrics.} We computed the standard classification metrics of precision, recall, and F1 score for each attribute value, together with the overall prediction accuracy for all attributes. All experimental results are averaged over three independently trained models and three disjoint subsets resulting in nine runs per configuration.

\begin{figure*}[t]
    \centering
    \includegraphics[width=\textwidth]{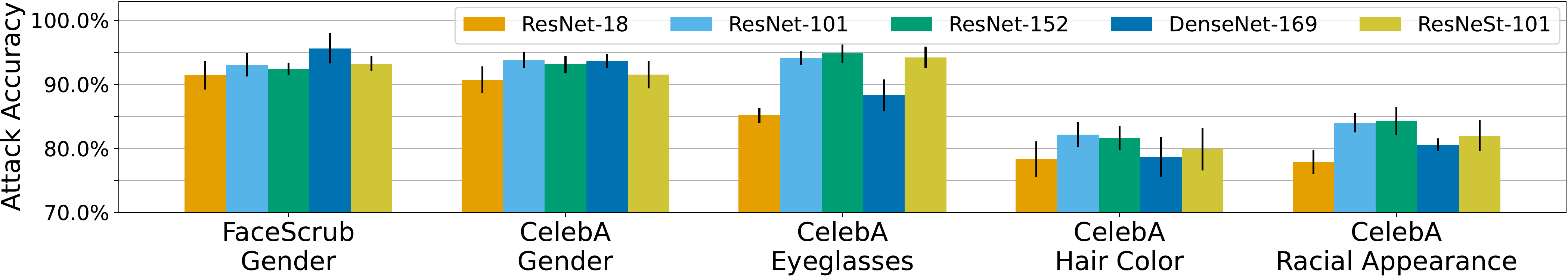}
    \caption{Attack accuracy for different target model architectures and CelebA attribute datasets. Results are averaged over three models and three attack datasets. Except for ResNet-18, the attacks are comparably successful on the different models. }
    \label{fig:accuracy_results}
    \vspace{0.2cm}
\end{figure*}

\begin{figure*}[t]
     \begin{subfigure}[c]{\textwidth}
         \centering
         \includegraphics[height=0.025\textwidth]{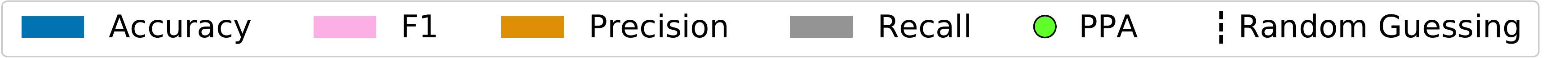}
     \end{subfigure}
     \begin{subfigure}[b]{0.48\textwidth}
        \centering
         \includegraphics[width=\textwidth]{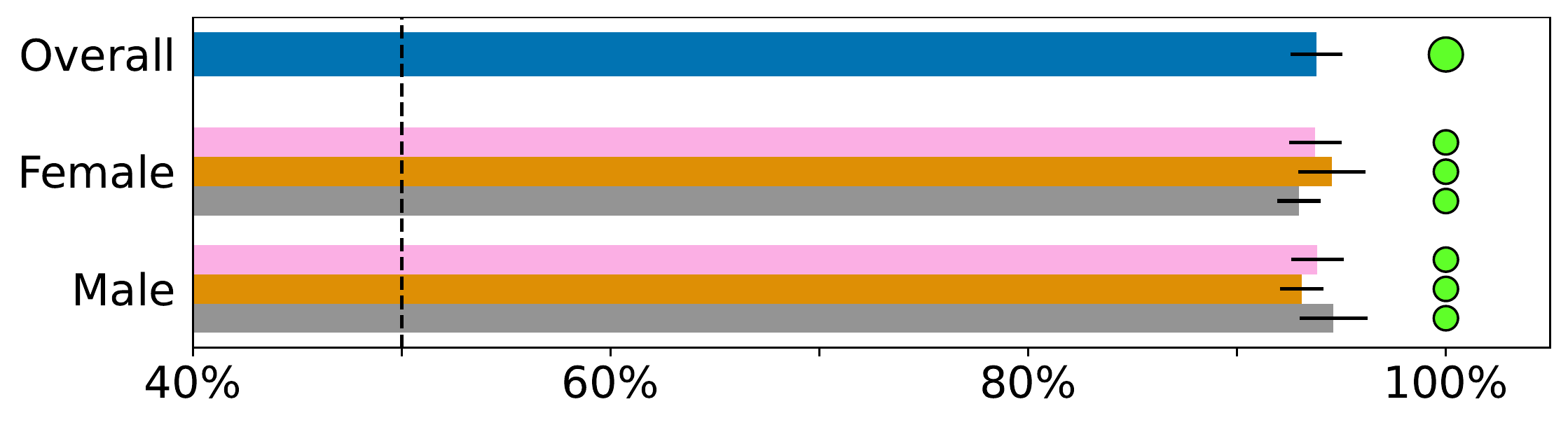}
         \caption{Gender (CelebA)}
         \vspace{0.2cm}
     \end{subfigure}
     \hfill
     \begin{subfigure}[b]{0.48\textwidth}
         \centering
         \includegraphics[width=\textwidth]{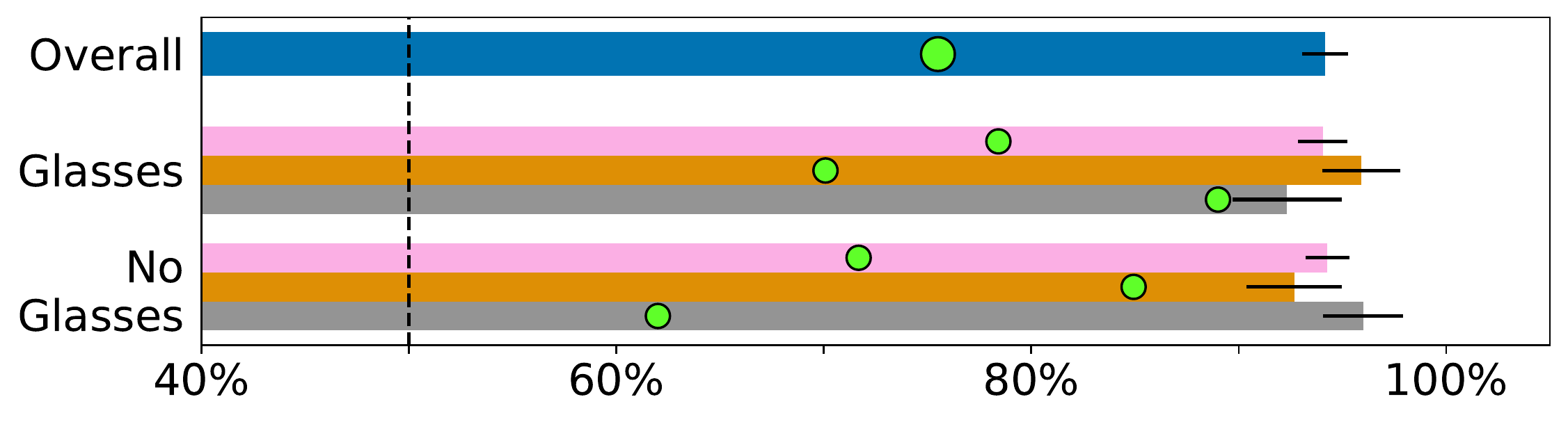}
         \caption{Eyeglasses (CelebA)}
         \vspace{0.2cm}
     \end{subfigure}
     \begin{subfigure}[b]{0.48\textwidth}
         \centering
         \includegraphics[width=\textwidth]{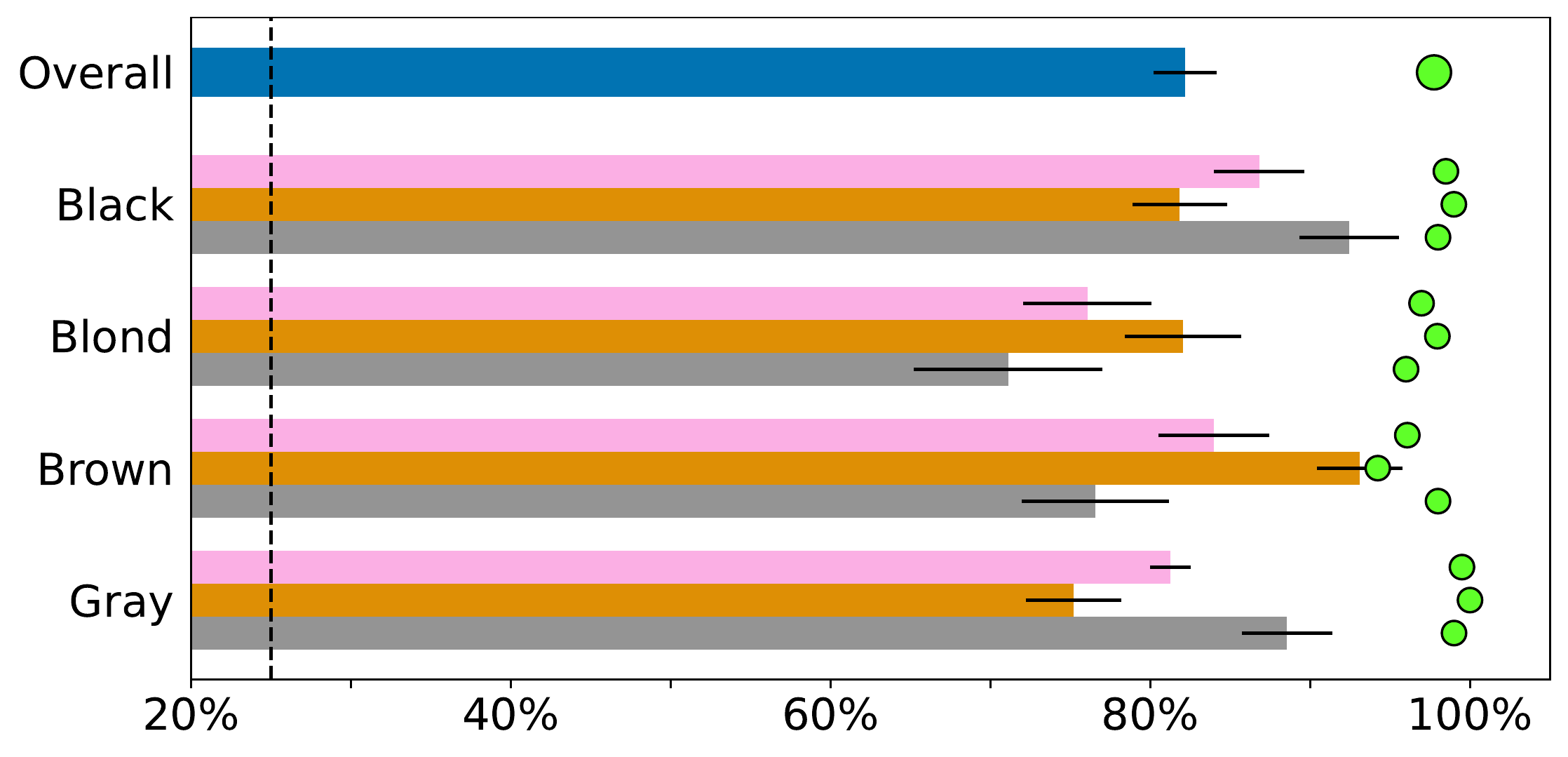}
         \caption{Hair Color (CelebA)}
     \end{subfigure}
     \hfill
     \begin{subfigure}[b]{0.48\textwidth}
         \centering
         \includegraphics[width=\textwidth]{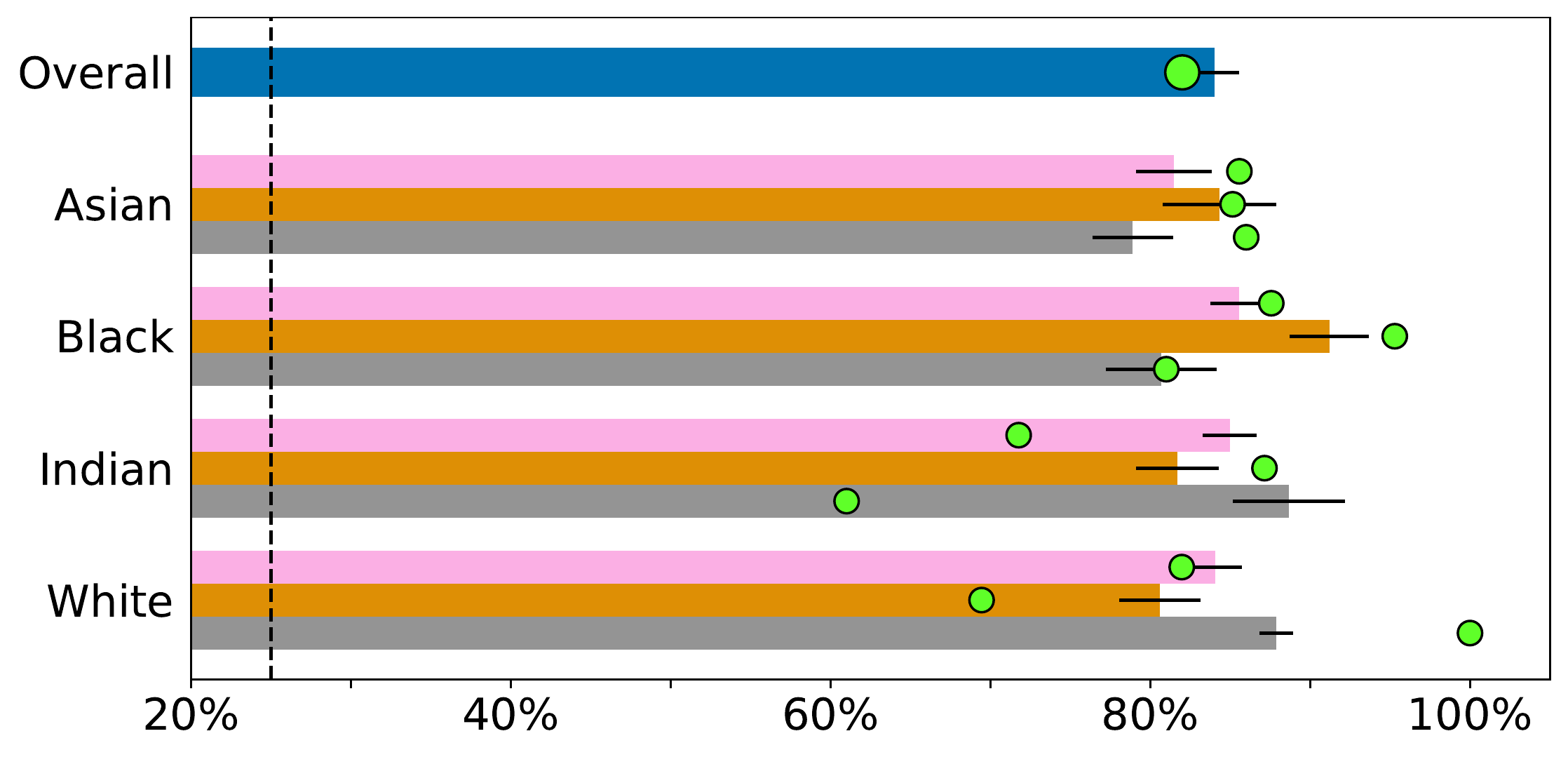}
         \caption{Racial Appearance (CelebA)}
     \end{subfigure}
    \caption{Evaluation results for \textsc{Caia} performed on ResNet-101 CelebA models to infer four different target attributes. The black horizontal lines denote the standard deviation over nine runs. We further state random guessing (dashed line) and Plug and Play Attacks (PPA, green dots) for comparison. While \textsc{Caia} outperforms random guessing by a large margin, it extracts information on racial appearance and if someone is wearing eyeglasses even more reliably than the white-box PPA attack.}
    \label{resnet101_results}
\end{figure*}

\paragraph{Attack Hyperparameters.}
To create the attack dataset, we first applied Null-Text Inversion~\citep{mokady22nullinversion} with 50 DDIM steps and a guidance scale of 7.5 on Stable Diffusion v1.5\footnote{Available at \url{https://huggingface.co/runwayml/stable-diffusion-v1-5}.}. We further used the generic prompt \textit{“A photo of a person”} for all samples. After the inversion, we generated image variations by adding the sensitive attribute values to the prompt using prompt-to-prompt, e.g., \textit{“A photo of a person, female appearance”} and \textit{“A photo of a person, male appearance”} to generate gender variations. See \cref{tab:attack_prompts} for all prompts used for generating the different feature representations. For prompt-to-prompt, the cross-replace steps were set to 1.0, and the self-replace steps to 0.4 for gender and eyeglasses and 0.6 for hair color and racial appearance. The confidence threshold for the filter models was set to $0.6$ for all models. We then generated and filtered attribute variations one after another until we collected a total of $300$ candidates for each attribute category. If not stated otherwise, all attacks were then performed on subsets of $100$ candidates.

\paragraph{Baselines.} We took random guessing as a naive baseline. Because the sensitive attributes are equally distributed among the different identities, random guessing corresponds to the underlying prior attribute distribution. Since existing AIAs are not applicable to our use case, we instead performed the state-of-the-art Plug \& Play model inversion attack (PPA)~\citep{struppek22_mia} to synthesize characteristic samples for each class. We then used our filter models to predict the sensitive attribute value for each sample and take the majority vote for each targeted identity as the attribute prediction. We emphasize that, unlike \textsc{Caia}, PPA requires white-box access to the target model and a pre-trained GAN, for which we used the StyleGAN2~\citep{Karras2019stylegan2} FFHQ model. Due to the high computational effort, we limit the PPA comparison to individual ResNet-101 models for each setting. Based on the results stated in the paper~\citep{struppek22_mia}, we expect PPA to perform comparably on the other architectures.

\begin{figure*}[t]
     \begin{subfigure}[b]{0.24\textwidth}
        \captionsetup{justification=centering}
        \centering
         \includegraphics[width=\textwidth]{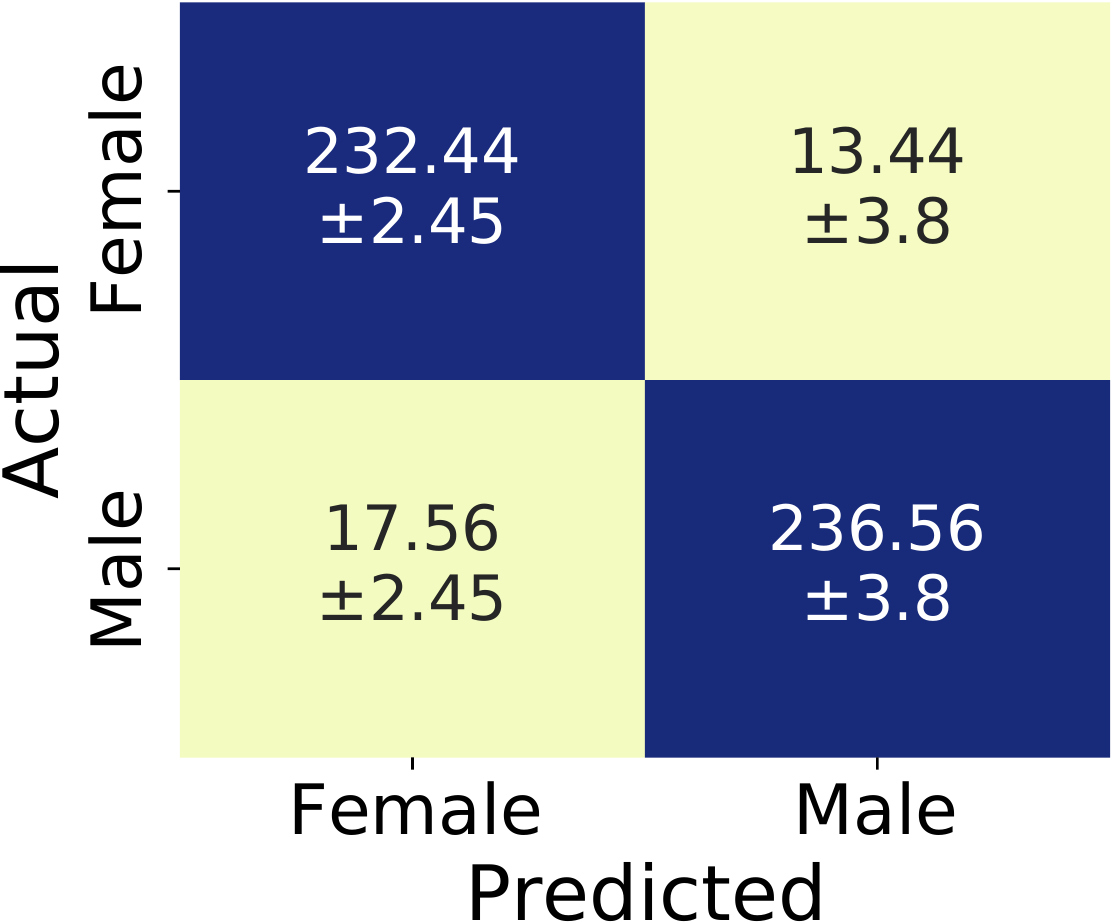}
         \caption{Gender (FFHQ) \\ (250 true positives per class)}
     \end{subfigure}
     \hfill
     \begin{subfigure}[b]{0.24\textwidth}
        \captionsetup{justification=centering}
         \centering
         \includegraphics[width=\textwidth]{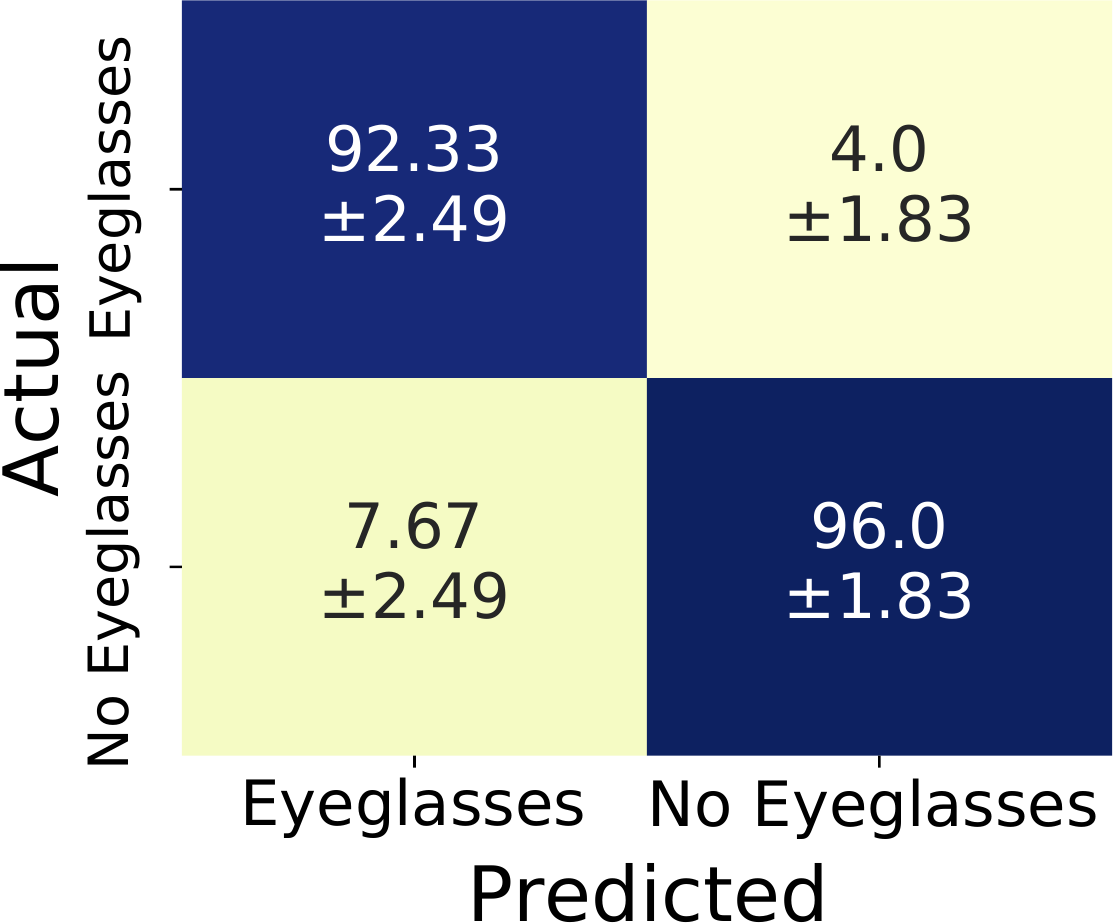}
         \caption{Eyeglasses (FFHQ) \\ (100 true positives per class)}
     \end{subfigure}
     \hfill
     \begin{subfigure}[b]{0.24\textwidth}
        \captionsetup{justification=centering}
         \centering
         \includegraphics[width=\textwidth]{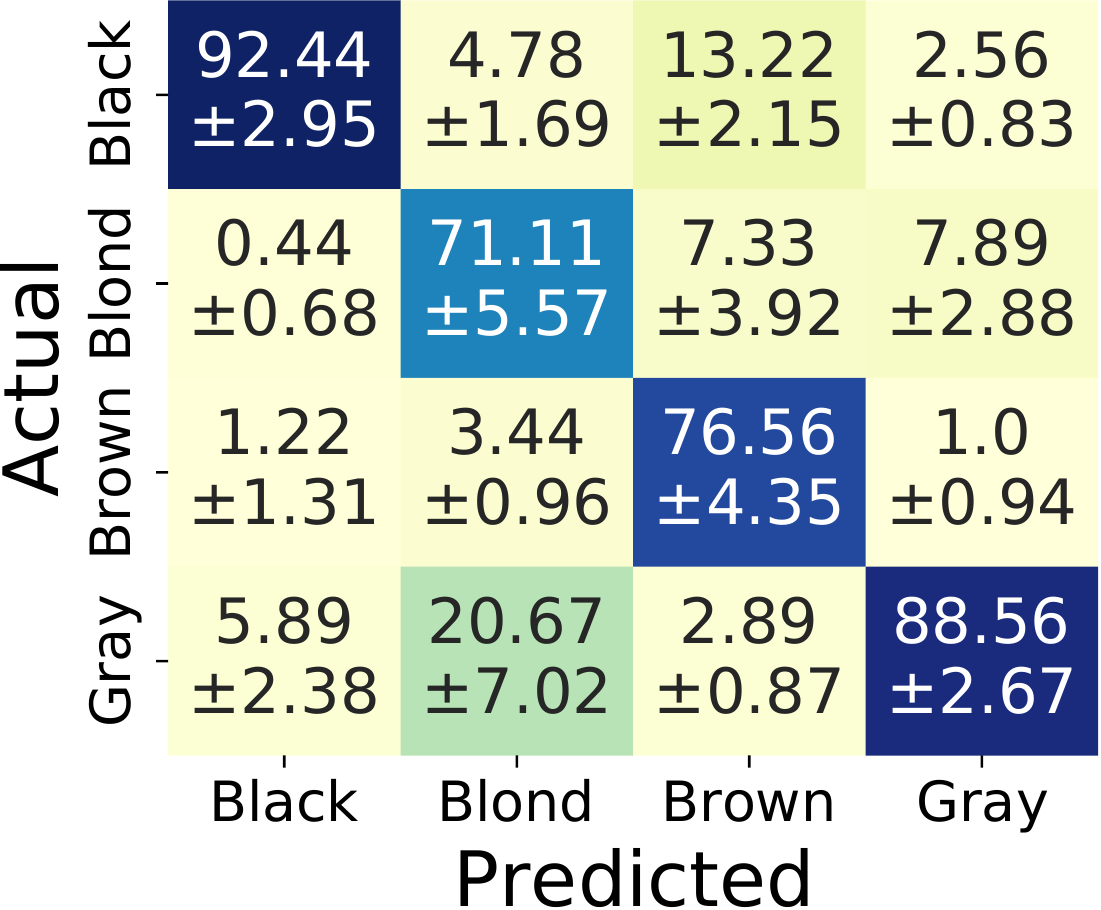}
         \caption{Hair Color (FFHQ) \\ (100 true positives per class)}
     \end{subfigure}
    \hfill
     \begin{subfigure}[b]{0.24\textwidth}
        \captionsetup{justification=centering}
         \centering
         \includegraphics[width=\textwidth]{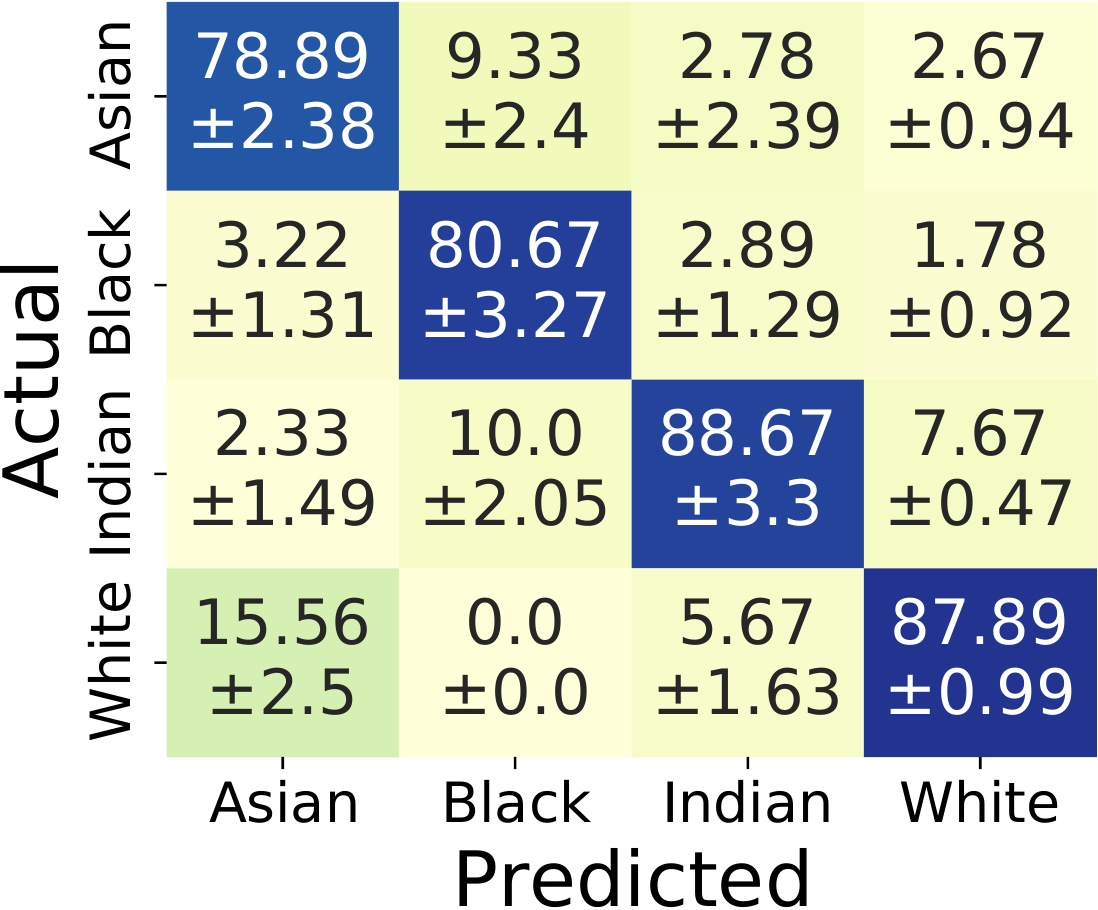}
         \caption{Racial Appearance (FFHQ) \\ (100 true positives per class)}
     \end{subfigure}
    \caption{Confusion Matrices for ResNet-101 models trained on the four different CelebA subsets. The results demonstrate that most attribute values can be inferred with similar precision. However, the attacks sometimes tend to falsely predict \textit{blond} hair color as \textit{gray} and \textit{Asian} appearance as \textit{white}.}
    \label{fig:confusion_matrizes}
\end{figure*}

\subsection{Extracting Sensitive Class Information}
In the main part of the paper, we focus on results on ResNet-101 models using FFHQ attack samples. We provide more detailed results for the other architectures, CelebAHQ as attack dataset, and experiments with 1,000 target identities in \cref{appx:add_results_celeba}. We note that the attack success for an increased number of identities did not change substantially, even if fewer samples of each identity were available during training. Also, the attack success using CelebAHQ instead of FFHQ as dataset to craft the attack samples has no significant impact on the results and demonstrates that \textsc{Caia} is not dependent on a specific attack dataset.

The attack accuracy for different models and target attributes in \cref{fig:accuracy_results} demonstrates that \textsc{Caia} performed comparably well on different architectures and predicted the sensitive attribute values correctly in over 90\% of the cases for the attributes \textit{gender} and \textit{eyeglasses} and about 80\% for the \textit{hair color} and \textit{racial appearance}. Only the attack results of ResNet-18 stand out and are a few percentage points lower than those of the other architectures, which we attribute to the small number of model parameters (only about a quarter of ResNet-101). Still, all attacks reliably inferred the sensitive attributes in most cases.

Next, we investigate the attribute leakage more closely. Therefore, we performed a more detailed analysis of the attribute leakage of ResNet-101 models, for which the results are depicted in \cref{resnet101_results}. For all four attributes, \textsc{Caia} significantly beats the random guessing baseline by a large margin. Whereas \textit{gender} and \textit{eyeglasses} were predicted correctly in about 94\% of the cases, \textit{racial appearance} could be inferred correctly in 84\%. The attack accuracy for \textit{hair color} was also about 82\% on average, but the attack success varied substantially between the different attribute values. \cref{fig:confusion_matrizes} shows the confusion matrices for each attribute. For the hair color, blond hair seems to be the hardest value to predict and is frequently confused with gray hair, which is not unexpected since hair colors have different shades, of which blond might be the broadest one. Another reason for the confusion in these attributes is that the CelebA dataset also contains numerous training images of poor quality and sometimes disturbing lighting, which could interfere with the ground-truth identity attributes.

\subsection{Comparison with White-Box Attacks}
We also investigated gradient-based model inversion attacks, here PPA, to compare to state-of-the-art white-box model inversion methods that reconstruct characteristic class inputs. On the ResNet-101 models, PPA achieved perfect attack results for inferring an identity's \textit{gender}. It also precisely revealed \textit{hair color} and outperformed the black-box \textsc{Caia} for both settings. However, for inferring whether an identity is wearing \textit{eyeglasses}, PPA fell significantly behind \textsc{Caia}. Regarding \textit{racial appearance}, PPA's attack accuracy was comparable to \textsc{Caia} but less consistent between different attribute values. We suspect the reason to be the uneven distribution of the \textit{racial appearance} and \textit{eyeglasses} attribute values in the underlying StyleGAN2's training data~\citep{karakas22fairstyle}. Since PPA starts from randomly sampled latent vectors, the biased attribute distributions also influence the attack's success in terms of generating and inferring sensitive attributes. Nevertheless, \textsc{Caia} shows competitive and impressive performance given that it accesses the target model only in a black-box fashion and has no access to internal gradient information.

\begin{figure*}[t]
     \begin{subfigure}[c]{\textwidth}
         \centering
        \includegraphics[width=\textwidth]{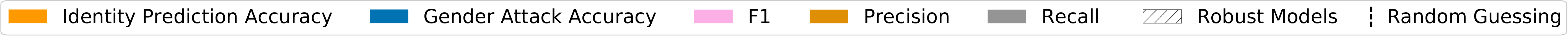}
     \end{subfigure}
     \begin{subfigure}[c]{0.48\textwidth}
         \centering
         \includegraphics[width=\textwidth]{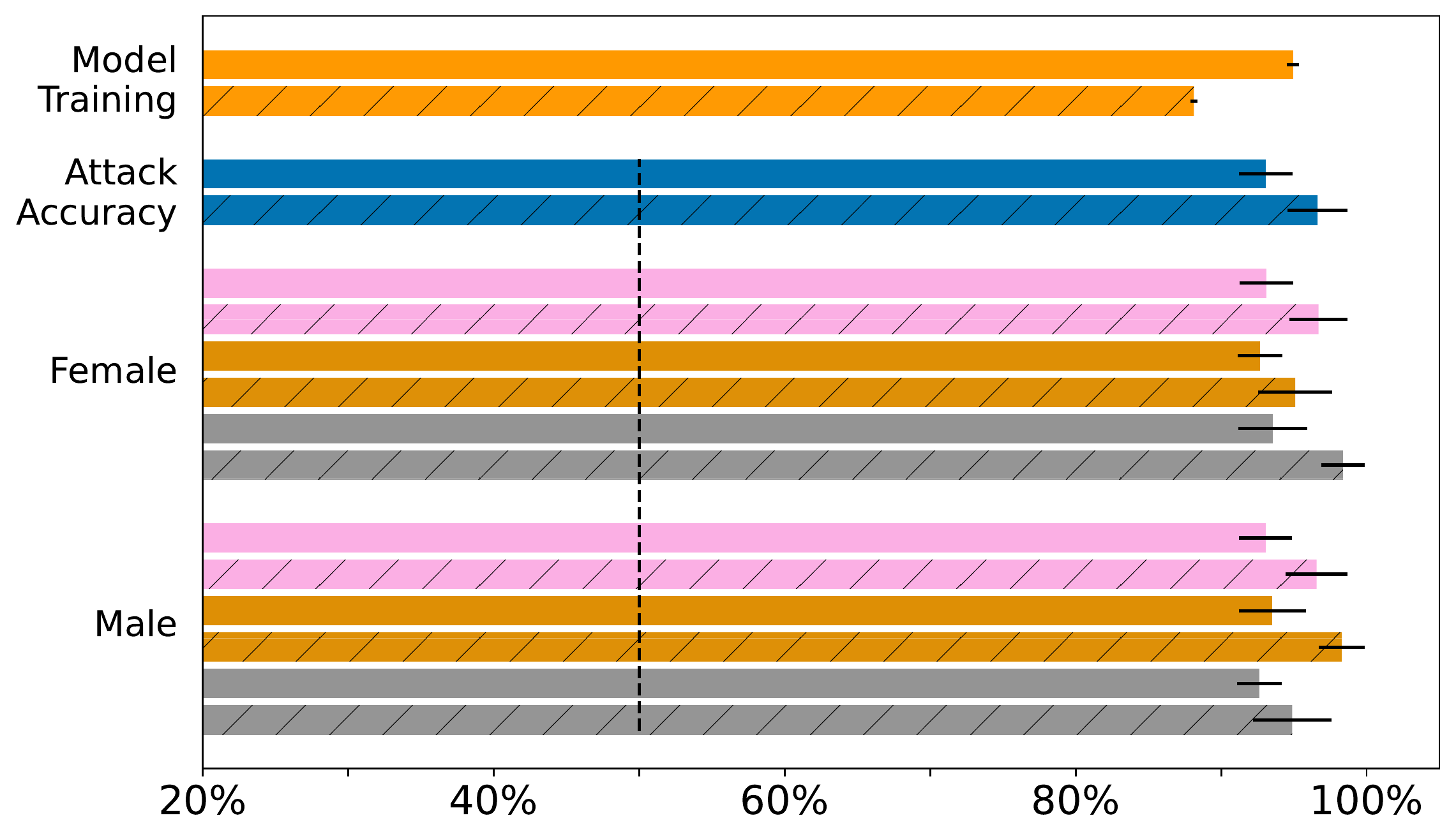}
         \caption{Gender (FaceScrub Cropped)}
         \label{fig:facescrub_cropped}
     \end{subfigure}
     \hfill
    \begin{subfigure}[c]{0.48\textwidth}
         \centering
         \includegraphics[width=\textwidth]{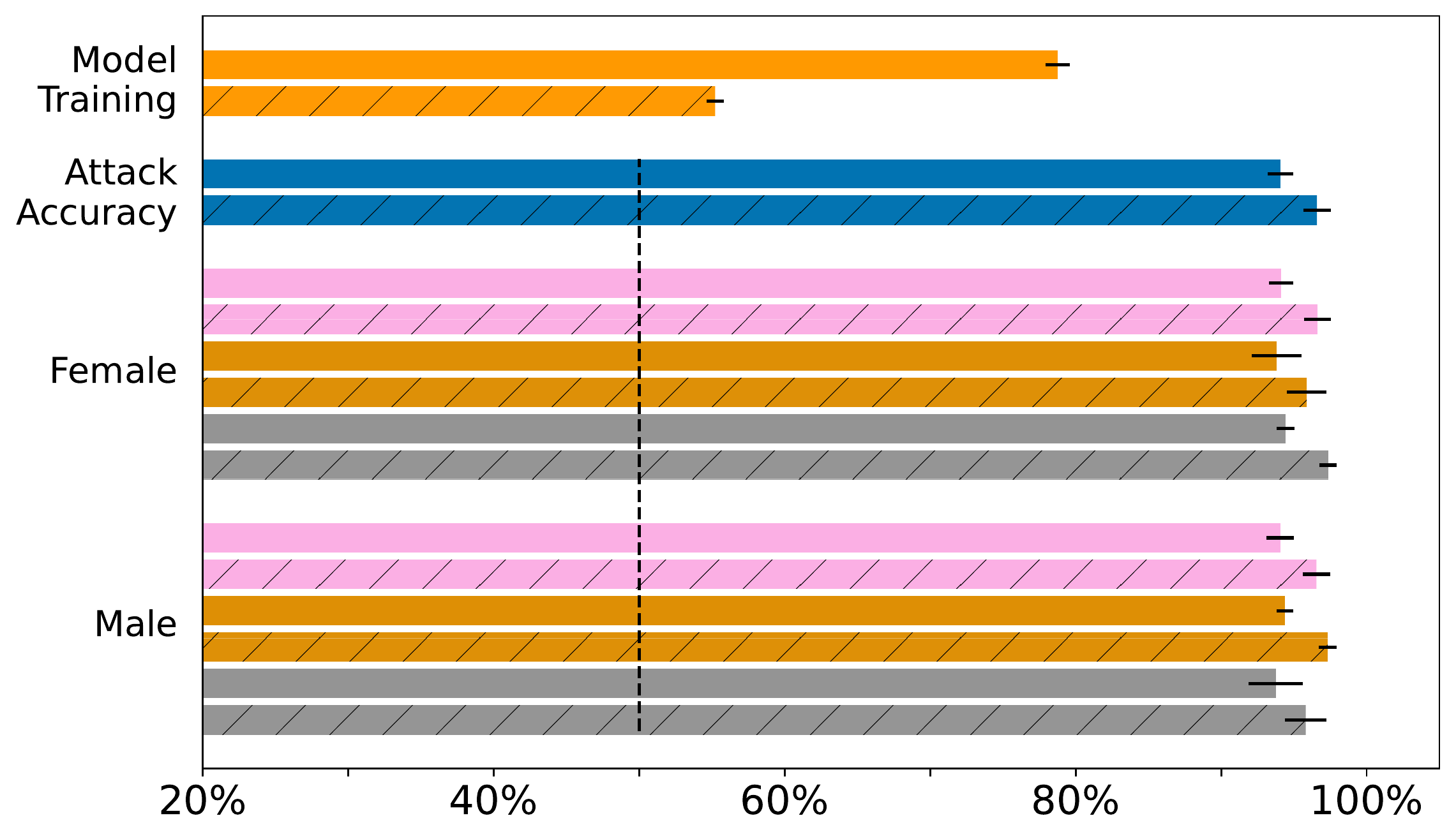}
         \caption{Gender (FaceScrub Uncropped)}
         \label{fig:facescrub_uncropped}
     \end{subfigure}

    \caption{Gender appearance inference results for \textsc{Caia} performed on ResNet-101 models trained on cropped \textbf{(a)} and uncropped \textbf{(b)} FaceScrub samples. We compared standard models to robust models trained with adversarial training (hatched bars). The results demonstrate that robust models indeed increase the information leakage, even if the underlying models' prediction accuracy (orange bars) is significantly below that of non-robust models.}
    \label{fig:facescrub_results}
\end{figure*}

\subsection{Robustness Increases Privacy Leakage}
Since adversarial examples are a common weakness of deep learning models, adversarial training is widely adopted to make models robust against this security threat. To establish a connection between privacy and security, we now extend our investigation to robustly trained models and show that robustness even increases privacy leakage. However, training robust models requires larger datasets due to increased sample complexity~\citep{schmidt18advrobust}. The limited size of the CelebA dataset, which provides approx. 30 samples per identity, makes it difficult to train stable and robust models. Therefore, this section focuses on models trained on FaceScrub, which provides an average of 70 samples per identity and, by this, facilitates stable adversarial training. We trained ResNet-101 models on both cropped images containing only faces and uncropped images, showing a large amount of content unrelated to an identity. For a visual comparison of the datasets, we refer the reader to \cref{appx:dataset_samples}.

The attack results for ResNet-101 models are shown in \cref{fig:facescrub_results}, while \cref{appx:add_results_Facescrub} presents results for other model architectures. Comparing the attack accuracy of non-robust models ($93.08\%$) against that of robust models ($96.62\%$) trained on cropped images suggests that robust models tend to leak more \textit{gender} information about their learned identities than non-robust models, even if the models' clean prediction accuracy is roughly five percentage points lower. A similar pattern holds for models trained on uncropped images, with robust models still exhibiting higher information leakage. However, it is important to note that the standard model's prediction accuracy ($78.75\%$) on uncropped images is significantly higher than that of robust models ($55.23\%$). This suggests that a model's prediction accuracy is not necessarily indicative of its privacy leakage. We hypothesize that robust models tend to exhibit higher privacy leakage due to their concentration on more robust features for their predictions. We formalize and investigate this hypothesis in the remaining section.

\begin{figure*}[ht]
    \centering
    \begin{subfigure}[t]{0.49\textwidth}
         \centering
         \includegraphics[height=0.6\textwidth, keepaspectratio]{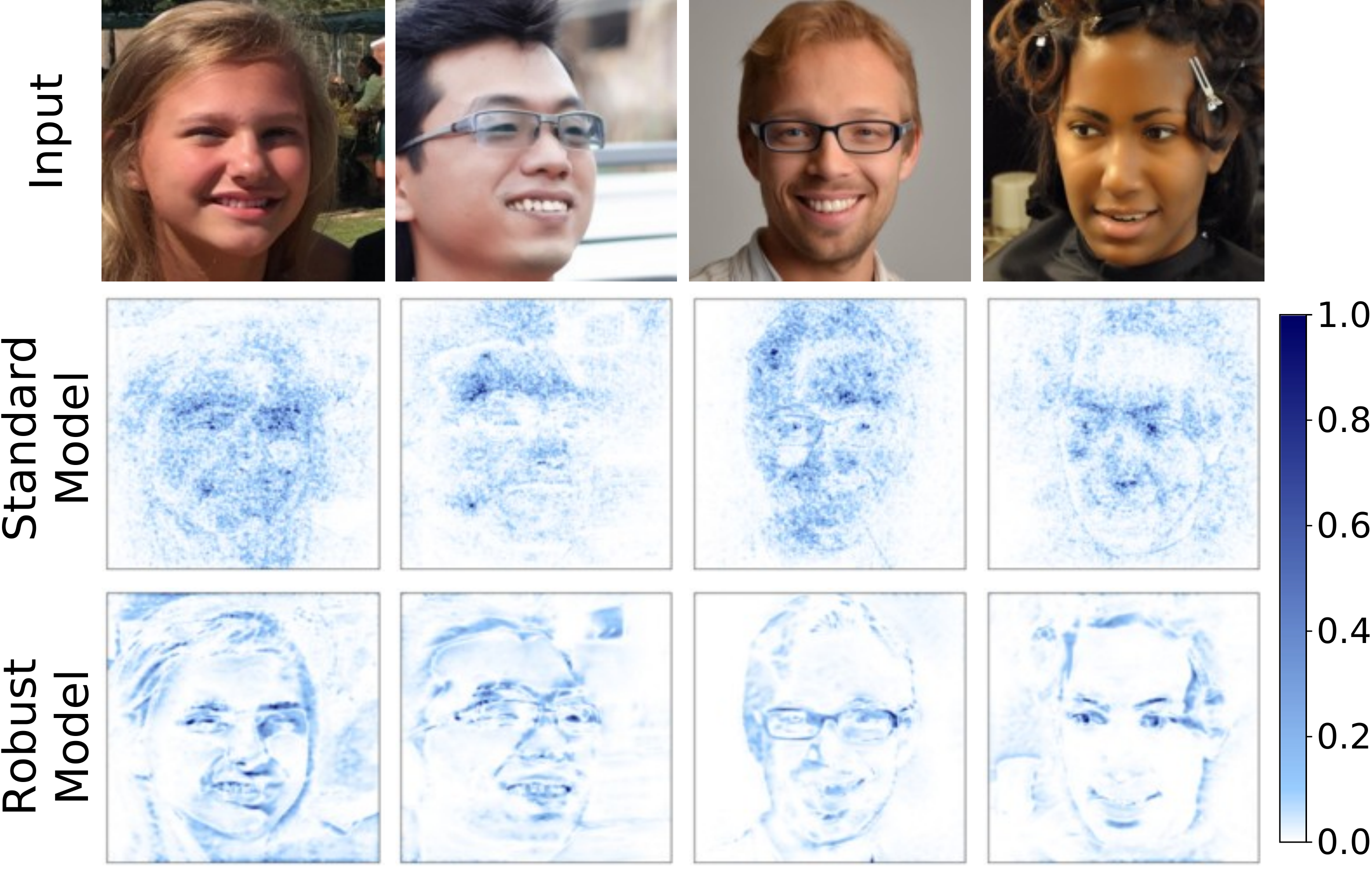}
          \caption{Visualized Attribution with Integrated Gradients.}
          \label{fig:qualitative_ig}
     \end{subfigure}
     \hfill
    \begin{subfigure}[t]{0.49\textwidth}
         \centering
         \includegraphics[height=0.6\textwidth, keepaspectratio]{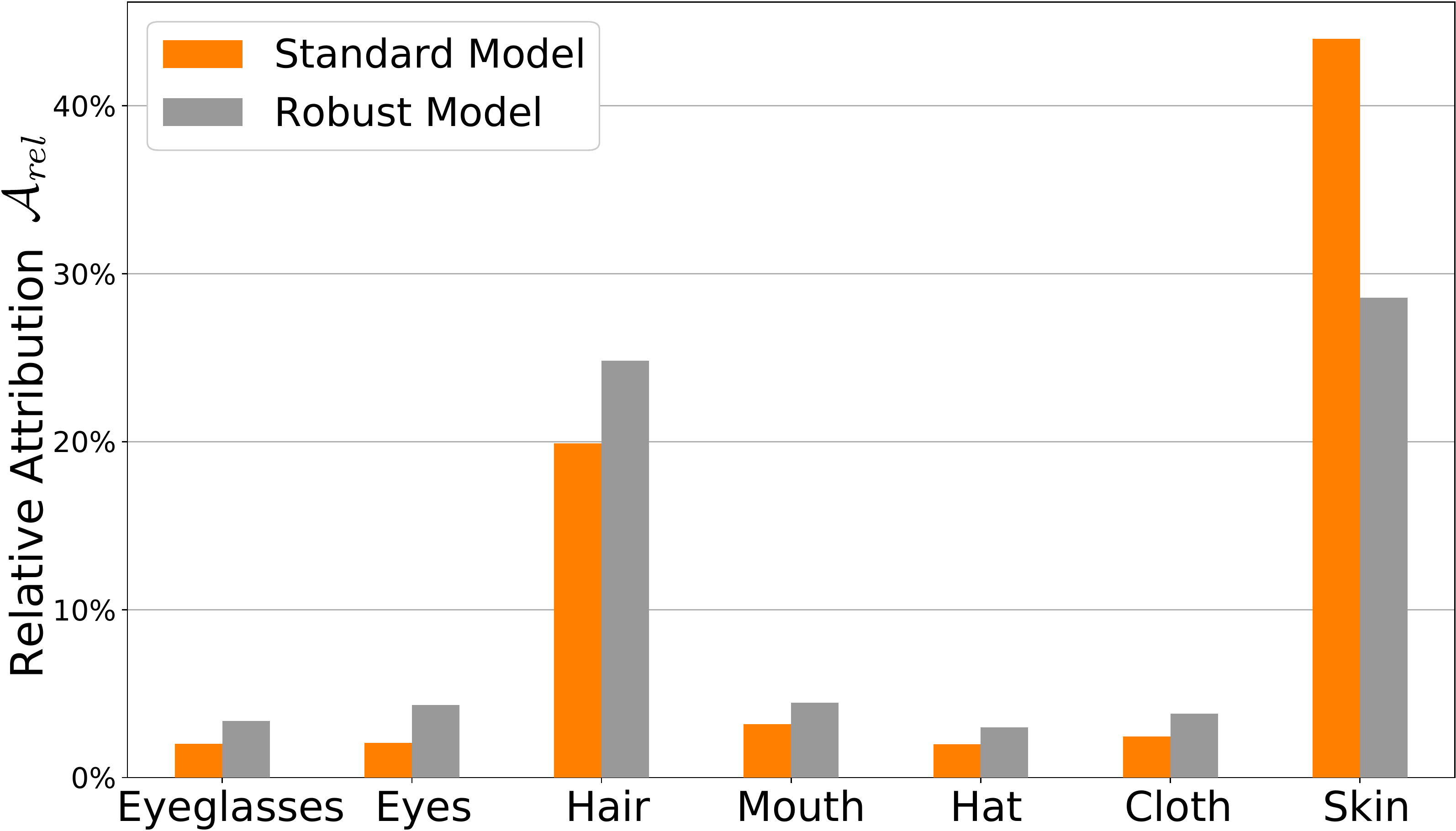}
          \caption{Relative Attribution for facial features.}
          \label{fig:quantative_ig}
     \end{subfigure}
    \caption{The comparison of absolute attribution based on integrated gradients between a robust and a standard ResNet-101 model in \textbf{(a)} shows that robust models assign perceptually more attuned attribution to facial features. A quantitative comparison of relative attribution in \textbf{(b)} further highlights that the robust model assigns more attribution to unique facial features, such as eyes and hair, while the standard model assigns most attribution to non-specific parts of the skin.}
    \label{fig:integrated_gradients}
\end{figure*}

Our formal analysis builds upon the robust feature model for binary classifiers of \citet{ilyas19bugs}, which we extend to the privacy leakage setting. Let $\mathcal{M}:X \to Y$ be a model trained to predict a label $y\in Y$ for each input $x\in X$. We divide a model's inputs $x=(\tilde{x},\bar{x})$ into predictive features $\tilde{x}$ and non-predictive features $\bar{x}$. The $i$-th predictive feature $\tilde{x}^i$ is positively correlated with a sample's true label: $E_{(x,y)\sim \mathcal{D}} \left[ \, \hat{y}\cdot\mathcal{M}(\tilde{x}^i)_y \, \right] \geq \rho$ for a sufficiently large $\rho$. Here, we deviate slightly from our previous notation of $y$ and define $\hat{y}\in \{-1,1\}$ for each class, with $\hat{y}=1$ indicating the ground-truth label. We further assume that all model outputs $\mathcal{M}_y$ are centered around zero. Predictive features are (ideally) learned by a model to predict the label. Conversely, a non-predictive feature $\bar{x}^j$ does not correlate with the true label and should be ignored during inference if $E_{(x,y)\sim \mathcal{D}} \left[ \, \hat{y}\cdot\mathcal{M}(\bar{x}^j)_y \, \right] < \rho$. 

We further explore the nature of the predictive features by categorizing them into two groups: robust features $\tilde{x}_\textit{robust}$ and non-robust features $\tilde{x}_\textit{non-robust}$. Robust predictive features remain predictive even under adversarial perturbations $\delta$ with $\| \delta \| \leq \epsilon$, satisfying $E_{(x,y)\sim \mathcal{D}} \left[ \, \hat{y}\cdot\mathcal{M}(\tilde{x}_{robust}^i + \delta)_y \, \right] \geq \rho$, while non-robust predictive features lose their predictive power under adversarial perturbations. Neural networks rely not only on salient features, such as hair color and other facial characteristics, for image processing, but also on subtle, non-robust image features that can still be highly predictive.

Adversarial perturbations can significantly disrupt non-robust models' predictions by manipulating $\tilde{x}_\textit{non-robust}$ in an $\epsilon$-environment that is imperceptible to humans. Adversarial training can help reduce the impact of adversarial examples by focusing the model's prediction on $\tilde{x}_\textit{robust}$. However, we stress that adversarial training's potential to enhance model robustness comes at the cost of increased privacy leakage. To explain why adversarial training increases a model's privacy leakage, we hypothesize that human-recognizable, sensitive attributes are part of $\tilde{x}_\textit{robust}$. These attributes are highly predictive because they enable discrimination between samples of different classes based on the assumption that they remain constant within samples of a specific class. We also argue that these attributes are robust features since they are easily identifiable by humans, and small adversarial perturbations are unlikely to affect their value.

In the context of face recognition, \textit{gender} is a sensitive attribute that is highly predictive, enabling us to distinguish individuals from people with other \textit{gender} appearances. Furthermore, it is a robust feature, as altering a person's \textit{gender} appearance in an image would require significant changes to the pixels. To showcase the importance of robust features in robust models, we utilized the axiomatic attribution method Integrated Gradients~\citep{sundararajan17ig} on two ResNet-101 models, one trained using standard training and the other with adversarial training. \cref{fig:qualitative_ig} visualizes the computed gradients for four input images from our attack datasets. While the standard model relies on noisy, non-robust features, such as image background, the robust model concentrates more on human-recognizable facial attributes, including hair, eyes, eyeglasses, and facial structure.

To also provide quantitative support for our analysis, we used Integrated Gradients to measure a model's relative attribution to specific image parts. For this, we applied a pre-trained face segmentation model~\citep{lee20maskgan} to locate various attributes in facial images. Be $H_{Z}(x) \in \{0,1\}^{H\times W}$ the binary segmentation mask for image $x$ and attribute class $Z$, such as a person's hair. Be further $\mathcal{A}(\mathcal{M}, x)\in \mathbb{R}^{H\times W}$ the absolute pixel-wise attribution by Integrated Gradients for model $\mathcal{M}$ and image $x\in \mathbb{R}^{H\times W}$. Let also $\| \cdot \|_1$ denote the sum norm and $\odot$ the Hadamard product. The relative attribution for a dataset $X$ is then computed by: 
\begin{equation}
    \mathcal{A}_\mathit{rel}(\mathcal{M}, X) = \frac{1}{|X|} \sum_{x\in X} \frac{\| \mathcal{A}(\mathcal{M}, x) \odot H_Z(x) \|_1}{\| \mathcal{A}(\mathcal{M}, x) \|_1} \, .
\end{equation}
The relative attribution computes the average share of total importance a model assigns to a particular facial attribute.

By analyzing the relative attribution of seven attributes in 100 FFHQ samples and over 530 target identities of ResNet-101 models, \cref{fig:quantative_ig} demonstrates that robust models assign more importance to distinct features like eyes or hair, while standard models prioritize general skin areas. This highlights the greater importance placed by robustly trained models on sensitive attributes encoded in images and reflected in their outputs. As \textsc{Caia} exploits the differences in logits, robust models appear to be more susceptible to such privacy attacks and leak more sensitive attribute information in their outputs. Thus, there exists a trade-off between model robustness and sensitive class information leakage.

\begin{figure*}[t]
    \centering
    \includegraphics[width=\textwidth]{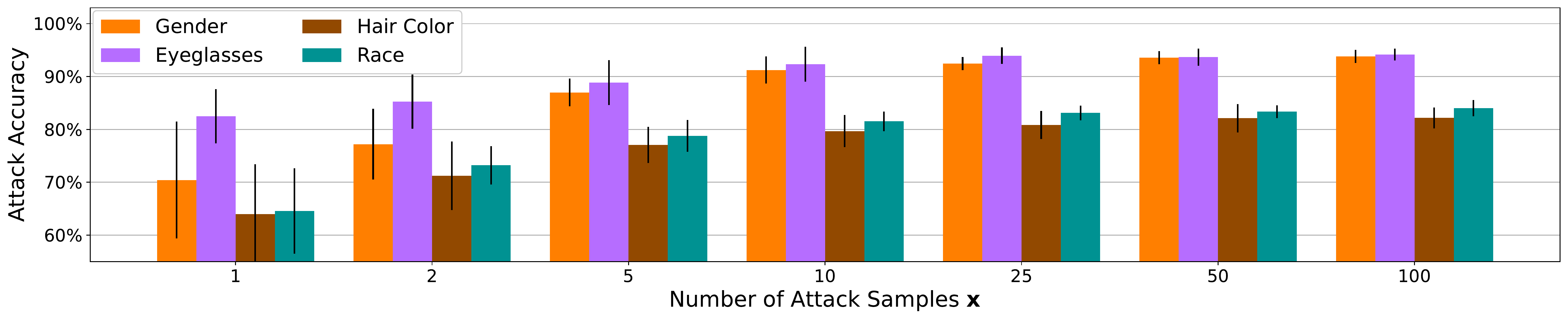}
    \caption{Attack accuracy achieved with a varying number of attack samples $\mathbf{x}$ available. Each result has been computed across three ResNet-101 CelebA models and with different subsets of the attack dataset. The results demonstrate that already 10 to 25 attack examples are sufficient to infer the sensitive attributes reliably.}
    \label{fig:num_attack_samples}
\end{figure*}

\begin{figure*}[t]
    \centering
    \includegraphics[width=\textwidth]{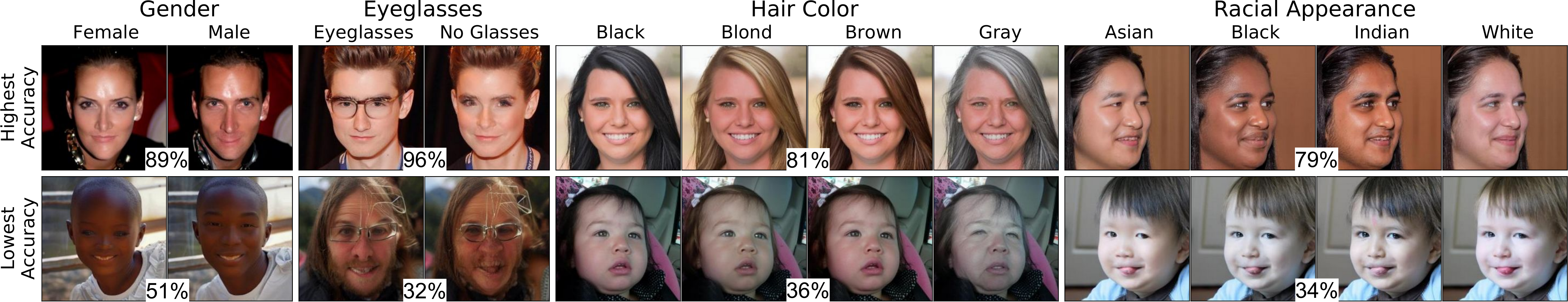}
    \caption{Attack accuracy and illustration of the attack samples $\mathbf{x}$ that achieve the highest and lowest accuracy, respectively, on ResNet-101 CelebA models. It becomes clear that samples that represent the different attribute values in a clear way perform best. On the contrary, samples with inconsistent attribute representation are far less informative and fail to extract sensitive values.}
    \label{fig:performing_samples}
\end{figure*}

\begin{figure*}[t]
    \centering
    \includegraphics[width=\textwidth]{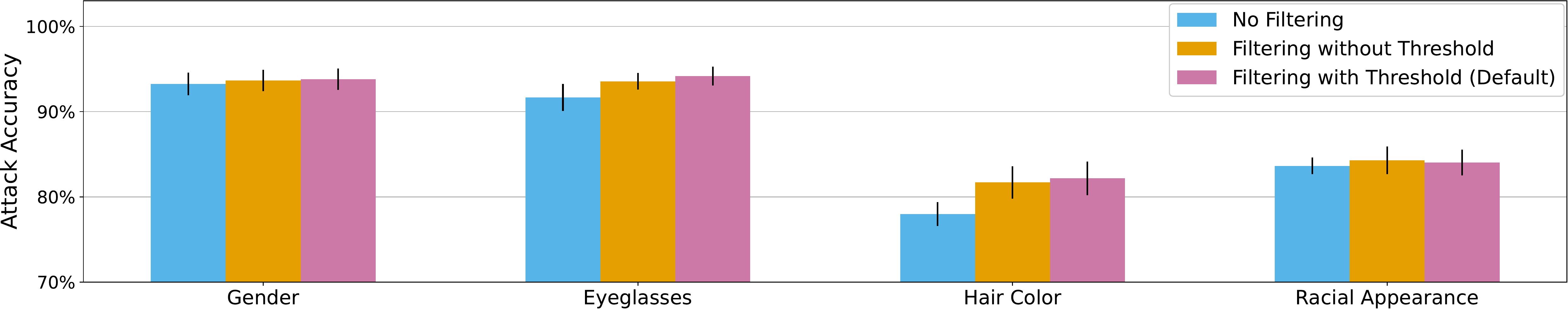}
    \caption{Attack accuracy achieved with unfiltered samples and samples filtered without any confidence threshold compared to our default approach with a confidence threshold of $\tau=0.6$. The filtering approach mainly improves the results for the attributes \textit{Eyeglasses} and \textit{Hair Color}, and only has a small impact on the attack success for the attributes \textit{Gender} and \textit{Racial Appearance}.}
    \label{fig:filtering}
\end{figure*}

\subsection{Ablation and Sensitivity Analysis}
To provide a complete picture of \textsc{Caia}, we also performed an ablation and sensitivity analysis. We first investigated the influence of the number of attack samples available to the adversary. For this, we repeated the attacks against the ResNet-101 CelebA models and varied the number of attack samples used to compute the relative advantage. As before, results are averaged across three models. We then built nine disjoint attack datasets for each setting and used three of them for each target model. Our experimental results, which are shown in \cref{fig:num_attack_samples}, demonstrate that 10 to 25 attack samples $\mathbf{x}=(x_1, \ldots, x_k)$ are already sufficient to infer the sensitive attributes reliably. Increasing the number any further only improves the results slightly. Even using a single attack example already beats the random guessing baselines significantly. But we also observe a high variance in the results for attacks with only a few attack samples available.

To gain further insights, we selected the attack samples $\mathbf{x}$ that achieved the highest and lowest attack accuracy when performing the attacks only with a single sample. \cref{fig:performing_samples} shows the corresponding images and their achieved attack accuracy for all four attribute settings. The best-performing samples (top row) all clearly portray the different characteristics of the respective target attributes. For example, the \textit{hair color} samples mainly differ in the depicted color of the hair, whereas the face of the woman has barely changed. The samples with the lowest attack accuracy (bottom row) generally have low-quality attribute representations and, in some cases, fail to depict any attribute differences. The results also demonstrate that even if carefully editing the images with \textit{Prompt-to-Prompt} might lead to spurious attribute changes, e.g., removing the eyeglasses could make the person's appearance more female. Similarly, changing the hair color to gray is also entangled with the perceived age of the depicted person. Future, more reliable synthesis models and image editing techniques will probably overcome such feature entanglements and allow attribute manipulations without introducing unwanted image changes. We expect this to improve the performance of \textsc{Caia} even further.

We also measured the impact of the filtering approach during the attack dataset synthesis and repeated the process using the FFHQ dataset with two modifications. First, we set the filter threshold to $\tau=0.0$ and accepted all generated images for which the attribute classifiers predict the target attributes independent of the assigned softmax scores. And second, we completely removed the filtering step and accepted all generated image variations of the first 300 FFHQ samples. We then repeated the attacks on the ResNet-101 CelebA models and compared the results to the ones with the standard filtering approach using a threshold of $\tau=0.6$ applied. \cref{fig:filtering} illustrates the mean attack accuracy for each setting. The results show that the filtering approach significantly improves the attack accuracy for the attributes \textit{Eyeglasses} and \textit{Hair Color}, whereas the accuracy for \textit{Gender} and \textit{Racial Appearance} stays comparable even without any filtering applied. We assume the reason to be that, e.g., adding or removing eyeglasses to a human face is much harder to realize for the generative model than changing a person's gender appearance. Therefore, the importance of the filter model ensuring that only consistent image variations make it into the attack dataset is higher and has a larger influence on the attack results.

\section{Discussion and Challenges}\label{sec:discussion}
In this paper, we introduced \textsc{Caia}, a novel privacy attack that enables the inference of sensitive attribute information about classes learned by standard image classifiers. Our extensive experiments demonstrate that classifiers indeed leak sensitive class information, with significant implications for the secure application of machine learning models. While black-box models have traditionally been practically secure against MIAs and AIAs, our research shows that models in the vision domain are not immune to attribute inference. By leveraging recent advances in text-to-image synthesis, \textsc{Caia} infers sensitive information from image classifiers with high accuracy, comparable to white-box MIAs. Furthermore, our findings suggest a trade-off between adversarial training and a model's information leakage, and future adversarial defenses should account for the privacy leakage of models to avoid creating new vulnerabilities.

Although \textsc{Caia} offers reliable attribute inference capabilities, it still faces some challenges and limitations. Accurately evaluating class information leakage requires high-quality data and consistent labeling, which heavily relies on the underlying dataset. We note that the sample quality of CelebA and FaceScrub images varies widely in terms of resolution, sharpness, and coloring, and the labeling is not always consistent within an identity or attribute class, with also falsely labeled samples contained in the datasets. For instance, the boundaries between gray and blond hair are not well-defined, which can lead to reduced inference accuracy for these attributes. However, \textsc{Caia} is generally successful in inferring sensitive attribute values in most cases. The attack metrics might even underestimate its effectiveness since false attribute predictions might still provide some identity information, e.g., black skin tone predictions make white as ground truth rather unlikely. Although, the attack may be more powerful for more consistently labeled datasets with higher image quality.
\section{Ethical Considerations}\label{sec:ethical_considerations}
Our work demonstrates that common face recognition models leak sensitive information about individuals from the training data. The results of our research might be misused to infer sensitive information through unethical or illegal means. Since face recognition models are already widespread in real-life and privacy-critical applications, e.g., mobile devices, smart home applications, and law enforcement, our proposed Class Attribute Inference Attack (\textsc{Caia}) could be applied to infer sensitive information from such systems and, therefore, lead to a privacy breach and potential harm to individuals and service providers.

However, we believe it is critical to enhance user awareness and educate practitioners about the existence and feasibility of such attacks. Moreover, \textsc{Caia} paves the way for future research into the factors facilitating such attacks and the development of potential countermeasures. Understanding such threats enables researchers and service providers to respond early, develop potential security measures, and create more reliable models. These advantages, in our opinion, outweigh any possible concerns. With our research, we also hope to highlight the necessity for users to keep a model's potential privacy leakage in mind, even if its training data is successfully protected from unauthorized access.

\section{Conclusion}\label{sec:conclusion}
To summarize, our research provides novel insights into the privacy of image classifiers and shows that models leak more sensitive information than previously assumed. Our experimental results and formal analysis demonstrate that robust models are particularly susceptible to such attacks, which introduce a novel trade-off between model robustness and privacy. It combines both areas of research, which have largely been studied separately in previous research. We hope our work motivates future security research and defense endeavors in building secure and private models.

Expanding on the mentioned challenges, we expect that with the upcoming developments in text-guided image manipulation, we propose to enhance \textsc{Caia} and extend its use to continuous features like age or more detailed skin tone grading. For example, the attack samples could be generated to reflect various shades of dark skin color or fine-grained age representations. Moreover, in its current implementation, \textsc{Caia} infers each attribute independently. However, in the real world, different attributes often correlate with each other. For example, information inferred on racial appearance influences the probability distribution of hair colors. To this end, \textsc{Caia} could leverage already inferred attributes as a prior for inferring additional attributes. We also envision the use of \textsc{Caia} beyond privacy analyses. For example, it is exciting to explore it in the context of explainable AI to determine the importance of features by analyzing prediction scores for different feature characteristics. Similarly, it could be employed to assess the fairness of models and identify potential biases associated with certain classes.

\vskip 0.1in
\paragraph{Reproducibility Statement.} Our source code is publicly at \url{https://github.com/LukasStruppek/Class_Attribute_Inference_Attacks} to reproduce the experiments and facilitate further analysis.

\vskip 0.1in
\paragraph{Acknowledgments} This research has benefited from the Federal Ministry of Education and Research (BMBF) project KISTRA (reference no. 13N15343), the Hessian Ministry of Higher Education, Research, Science and the Arts (HMWK) cluster projects “The Third Wave of AI” and hessian.AI, from the German Center for Artificial Intelligence (DFKI) project “SAINT”, as well as from the joint ATHENE project of the HMWK and the BMBF “AVSV”.

\newpage

\bibliographystyle{plainnat}
\bibliography{references}

\begin{thebibliography}{58}
\providecommand{\natexlab}[1]{#1}
\providecommand{\url}[1]{\texttt{#1}}
\expandafter\ifx\csname urlstyle\endcsname\relax
  \providecommand{\doi}[1]{doi: #1}\else
  \providecommand{\doi}{doi: \begingroup \urlstyle{rm}\Url}\fi

\bibitem[Abdal et~al.(2022)Abdal, Zhu, Femiani, Mitra, and
  Wonka]{abdal22clip2stylegan}
Rameen Abdal, Peihao Zhu, John Femiani, Niloy~J. Mitra, and Peter Wonka.
\newblock Clip2stylegan: Unsupervised extraction of stylegan edit directions.
\newblock In \emph{Special Interest Group on Computer Graphics and Interactive
  Techniques Conference (SIGGRAPH)}, pages 48:1--48:9, 2022.

\bibitem[Abuassba and Farha(2021)]{abuassoa21accesscontrol}
Adnan Abuassba and Fadi Farha.
\newblock Access control and authorization in smart homes: A survey.
\newblock \emph{Tsinghua Science \& Technology}, 26:\penalty0 906–917, 2021.

\bibitem[Alaluf et~al.(2022)Alaluf, Tov, Mokady, Gal, and
  Bermano]{alaluf22hyperstyle}
Yuval Alaluf, Omer Tov, Ron Mokady, Rinon Gal, and Amit Bermano.
\newblock Hyperstyle: Stylegan inversion with hypernetworks for real image
  editing.
\newblock In \emph{Conference on Computer Vision and Pattern Recognition
  (CVPR)}, pages 18490--18500, 2022.

\bibitem[Barua et~al.(2021)Barua, Muhammad~Gowdh, Rahmat, Ramli, Ng, Chan,
  Kuluozturk, Dogan, Baygin, Yaman, Tuncer, Wen, Cheong, and
  Acharya]{barua21covid19}
Prabal~Datta Barua, Nadia~Fareeda Muhammad~Gowdh, Kartini Rahmat, Norlisah
  Ramli, Wei~Lin Ng, Wai~Yee Chan, Mutlu Kuluozturk, Sengul Dogan, Mehmet
  Baygin, Orhan Yaman, Turker Tuncer, Tao Wen, Kang~Hao Cheong, and U.~Rajendra
  Acharya.
\newblock Automatic covid-19 detection using exemplar hybrid deep features with
  x-ray images.
\newblock \emph{International Journal of Environmental Research and Public
  Health}, 18\penalty0 (15), 2021.

\bibitem[Baumann et~al.(2018)Baumann, Huang, Gl{\"{a}}ser, Herman, Banzhaf, and
  Z{\"{o}}llner]{baumann18roads}
Ulrich Baumann, Yuan{-}Yao Huang, Claudius Gl{\"{a}}ser, Michael Herman, Holger
  Banzhaf, and J.~Marius Z{\"{o}}llner.
\newblock Classifying road intersections using transfer-learning on a deep
  neural network.
\newblock In \emph{International Conference on Intelligent Transportation
  Systems ({ITSC})}, pages 683--690, 2018.

\bibitem[Bureau(2022)]{uscensus22}
U.S.~Census Bureau.
\newblock About the topic of race, 2022.
\newblock URL \url{https://www.census.gov/topics/population/race/about.html}.
\newblock Accessed: 22-February-2023.

\bibitem[Chen et~al.(2021)Chen, Kahla, Jia, and Qi]{knowledge_mia}
Si~Chen, Mostafa Kahla, Ruoxi Jia, and Guo-Jun Qi.
\newblock {Knowledge-Enriched Distributional Model Inversion Attacks}.
\newblock In \emph{International Conference on Computer Vision (ICCV)}, pages
  16178--16187, 2021.

\bibitem[Choquette{-}Choo et~al.(2021)Choquette{-}Choo, Tram{\`{e}}r, Carlini,
  and Papernot]{choquette2021}
Christopher~A. Choquette{-}Choo, Florian Tram{\`{e}}r, Nicholas Carlini, and
  Nicolas Papernot.
\newblock Label-only membership inference attacks.
\newblock In \emph{International Conference on Machine Learning (ICML)}, pages
  1964--1974, 2021.

\bibitem[Dinh et~al.(2022)Dinh, Tran, Nguyen, and Hua]{dinh22hyperinverter}
Tan~M. Dinh, Anh~Tuan Tran, Rang Nguyen, and Binh{-}Son Hua.
\newblock Hyperinverter: Improving stylegan inversion via hypernetwork.
\newblock In \emph{Conference on Computer Vision and Pattern Recognition
  (CVPR)}, pages 11379--11388, 2022.

\bibitem[Esteva et~al.(2021)Esteva, Chou, Yeung, Naik, Madani, Mottaghi, Liu,
  Topol, Dean, and Socher]{esteva21medical}
Andre Esteva, Kat Chou, Serena Yeung, Nikhil Naik, Ali Madani, Ali Mottaghi,
  Yun Liu, Eric Topol, Jeff Dean, and Richard Socher.
\newblock Deep learning-enabled medical computer vision.
\newblock \emph{npj Digital Medicine}, 2021.

\bibitem[Fredrikson et~al.(2015)Fredrikson, Jha, and
  Ristenpart]{fredriskon15mia}
Matt Fredrikson, Somesh Jha, and Thomas Ristenpart.
\newblock {Model Inversion Attacks that Exploit Confidence Information and
  Basic Countermeasures}.
\newblock In \emph{Conference on Computer and Communications Security (CCS)},
  pages 1322--1333, 2015.

\bibitem[Fredrikson et~al.(2014)Fredrikson, Lantz, Jha, Lin, Page, and
  Ristenpart]{fredrikson14pharmacogenetics}
Matthew Fredrikson, Eric Lantz, Somesh Jha, Simon~M. Lin, David Page, and
  Thomas Ristenpart.
\newblock Privacy in pharmacogenetics: An end-to-end case study of personalized
  warfarin dosing.
\newblock In \emph{USENIX Security Symposium}, pages 17--32, 2014.

\bibitem[Ganju et~al.(2018)Ganju, Wang, Yang, Gunter, and
  Borisov]{ganju18property}
Karan Ganju, Qi~Wang, Wei Yang, Carl~A. Gunter, and Nikita Borisov.
\newblock Property inference attacks on fully connected neural networks using
  permutation invariant representations.
\newblock In \emph{Conference on Computer and Communications Security ({CCS})},
  pages 619--633, 2018.

\bibitem[Goodfellow et~al.(2014)Goodfellow, Pouget-Abadie, Mirza, Xu,
  Warde-Farley, Ozair, Courville, and Bengio]{goodfellow_gan}
Ian Goodfellow, Jean Pouget-Abadie, Mehdi Mirza, Bing Xu, David Warde-Farley,
  Sherjil Ozair, Aaron Courville, and Yoshua Bengio.
\newblock {Generative Adversarial Nets}.
\newblock In \emph{Conference on Neural Information Processing Systems
  (NeurIPS)}, 2014.

\bibitem[Goodfellow et~al.(2015)Goodfellow, Shlens, and
  Szegedy]{goodfellow15adv}
Ian~J. Goodfellow, Jonathon Shlens, and Christian Szegedy.
\newblock Explaining and harnessing adversarial examples.
\newblock In \emph{International Conference on Learning Representations
  ({ICLR})}, 2015.

\bibitem[Guo and Zhang(2019)]{guo19face_rec}
Guodong Guo and Na~Zhang.
\newblock A survey on deep learning based face recognition.
\newblock \emph{Computer Vision and Image Understanding}, 189, 2019.

\bibitem[He et~al.(2016)He, Zhang, Ren, and Sun]{resnet_he}
Kaiming He, Xiangyu Zhang, Shaoqing Ren, and Jian Sun.
\newblock Deep residual learning for image recognition.
\newblock In \emph{Conference on Computer Vision and Pattern Recognition
  (CVPR)}, pages 770--778, 2016.

\bibitem[Hertz et~al.(2022)Hertz, Mokady, Tenenbaum, Aberman, Pritch, and
  Cohen-Or]{hertz2022prompt}
Amir Hertz, Ron Mokady, Jay Tenenbaum, Kfir Aberman, Yael Pritch, and Daniel
  Cohen-Or.
\newblock Prompt-to-prompt image editing with cross attention control.
\newblock \emph{arXiv preprint}, arXiv:2208.01626, 2022.

\bibitem[Hintersdorf et~al.(2022{\natexlab{a}})Hintersdorf, Struppek, and
  Kersting]{hintersdorf22trust}
Dominik Hintersdorf, Lukas Struppek, and Kristian Kersting.
\newblock To trust or not to trust prediction scores for membership inference
  attacks.
\newblock In \emph{International Joint Conference on Artificial Intelligence
  ({IJCAI})}, pages 3043--3049, 2022{\natexlab{a}}.

\bibitem[Hintersdorf et~al.(2022{\natexlab{b}})Hintersdorf, Struppek, and
  Kersting]{hintersdorf_clip}
Dominik Hintersdorf, Lukas Struppek, and Kristian Kersting.
\newblock Clipping privacy: Identity inference attacks on multi-modal machine
  learning models.
\newblock \emph{arXiv preprint}, arXiv:2209.07341, 2022{\natexlab{b}}.

\bibitem[Huang et~al.(2017)Huang, Liu, van~der Maaten, and
  Weinberger]{densenet}
Gao Huang, Zhuang Liu, Laurens van~der Maaten, and Kilian~Q. Weinberger.
\newblock {Densely Connected Convolutional Networks}.
\newblock In \emph{Conference on Computer Vision and Pattern Recognition
  (CVPR)}, pages 2261--2269, 2017.

\bibitem[Ibrahim et~al.(2022)Ibrahim, Mohamed, Maher, and
  Zhang]{ibrahim22cancer}
Ahmad Ibrahim, Hoda~K. Mohamed, Ali Maher, and Baochang Zhang.
\newblock A survey on human cancer categorization based on deep learning.
\newblock \emph{Frontiers in Artificial Intelligence}, 5, 2022.

\bibitem[Ilyas et~al.(2019)Ilyas, Santurkar, Tsipras, Engstrom, Tran, and
  Madry]{ilyas19bugs}
Andrew Ilyas, Shibani Santurkar, Dimitris Tsipras, Logan Engstrom, Brandon
  Tran, and Aleksander Madry.
\newblock Adversarial examples are not bugs, they are features.
\newblock In \emph{Conference on Neural Information Processing Systems
  (NeurIPS)}, volume~32, 2019.

\bibitem[Jayaraman and Evans(2022)]{jayaraman22imputation}
Bargav Jayaraman and David Evans.
\newblock Are attribute inference attacks just imputation?
\newblock In \emph{Conference on Computer and Communications Security ({CCS})},
  pages 1569--1582, 2022.

\bibitem[Kahla et~al.(2022)Kahla, Chen, Just, and Jia]{kahla22labelmia}
Mostafa Kahla, Si~Chen, Hoang~Anh Just, and Ruoxi Jia.
\newblock Label-only model inversion attacks via boundary repulsion.
\newblock In \emph{Conference on Computer Vision and Pattern Recognition
  (CVPR)}, pages 15025--15033, 2022.

\bibitem[Karakas et~al.(2022)Karakas, Dirik, Yalcinkaya, and
  Yanardag]{karakas22fairstyle}
Cemre Karakas, Alara Dirik, Eylul Yalcinkaya, and Pinar Yanardag.
\newblock Fairstyle: Debiasing stylegan2 with style channel manipulations.
\newblock In \emph{European Conference on Computer Vision ({ECCV})}, volume
  13673, pages 570--586, 2022.

\bibitem[Karkkainen and Joo(2021)]{karkkainen21fairface}
Kimmo Karkkainen and Jungseock Joo.
\newblock Fairface: Face attribute dataset for balanced race, gender, and age
  for bias measurement and mitigation.
\newblock In \emph{Winter Conference on Applications of Computer Vision
  (WACV)}, pages 1548--1558, 2021.
\newblock Pretrained classifier available at
  \url{https://github.com/dchen236/FairFace}.

\bibitem[Karras et~al.(2018)Karras, Aila, Laine, and
  Lehtinen]{karras18progressive}
Tero Karras, Timo Aila, Samuli Laine, and Jaakko Lehtinen.
\newblock Progressive growing of gans for improved quality, stability, and
  variation.
\newblock In \emph{International Conference on Learning Representations
  ({ICLR})}, 2018.

\bibitem[Karras et~al.(2019)Karras, Laine, and Aila]{Karras_stylegan1}
Tero Karras, Samuli Laine, and Timo Aila.
\newblock {A Style-Based Generator Architecture for Generative Adversarial
  Networks}.
\newblock In \emph{Conference on Computer Vision and Pattern Recognition
  (CVPR)}, pages 4401--4410, 2019.

\bibitem[Karras et~al.(2020)Karras, Laine, Aittala, Hellsten, Lehtinen, and
  Aila]{Karras2019stylegan2}
Tero Karras, Samuli Laine, Miika Aittala, Janne Hellsten, Jaakko Lehtinen, and
  Timo Aila.
\newblock {Analyzing and Improving the Image Quality of {StyleGAN}}.
\newblock In \emph{Conference on Computer Vision and Pattern Recognition
  (CVPR)}, 2020.
\newblock Pretrained FFHQ model available at
  \url{https://github.com/NVlabs/stylegan2-ada-pytorch}.

\bibitem[Kingma and Ba(2015)]{adam_optimizer}
Diederik~P. Kingma and Jimmy Ba.
\newblock {Adam: Method for Stochastic Optimization}.
\newblock In \emph{International Conference on Learning Representations
  (ICLR)}, 2015.

\bibitem[Lee et~al.(2020)Lee, Liu, Wu, and Luo]{lee20maskgan}
Cheng-Han Lee, Ziwei Liu, Lingyun Wu, and Ping Luo.
\newblock Maskgan: Towards diverse and interactive facial image manipulation.
\newblock In \emph{IEEE Conference on Computer Vision and Pattern Recognition
  (CVPR)}, 2020.
\newblock Pre-trained model available at
  \url{https://github.com/switchablenorms/CelebAMask-HQ}.

\bibitem[Liu et~al.(2015)Liu, Luo, Wang, and Tang]{celeba}
Ziwei Liu, Ping Luo, Xiaogang Wang, and Xiaoou Tang.
\newblock {Deep Learning Face Attributes in the Wild}.
\newblock In \emph{International Conference on Computer Vision (ICCV)}, 2015.

\bibitem[Madry et~al.(2018)Madry, Makelov, Schmidt, Tsipras, and
  Vladu]{madry18pgd}
Aleksander Madry, Aleksandar Makelov, Ludwig Schmidt, Dimitris Tsipras, and
  Adrian Vladu.
\newblock Towards deep learning models resistant to adversarial attacks.
\newblock In \emph{International Conference on Learning Representations
  ({ICLR})}, 2018.

\bibitem[Mehnaz et~al.(2022)Mehnaz, Dibbo, Kabir, Li, and
  Bertino]{mehnaz22attributes}
Shagufta Mehnaz, Sayanton~V. Dibbo, Ehsanul Kabir, Ninghui Li, and Elisa
  Bertino.
\newblock Are your sensitive attributes private? novel model inversion
  attribute inference attacks on classification models.
\newblock In \emph{USENIX Security Symposium}, pages 4579--4596, 2022.

\bibitem[Mokady et~al.(2022)Mokady, Hertz, Aberman, Pritch, and
  Cohen{-}Or]{mokady22nullinversion}
Ron Mokady, Amir Hertz, Kfir Aberman, Yael Pritch, and Daniel Cohen{-}Or.
\newblock Null-text inversion for editing real images using guided diffusion
  models.
\newblock \emph{arXiv preprint}, arXiv:2211.09794, 2022.

\bibitem[M{\"{u}}ller et~al.(2019)M{\"{u}}ller, Kornblith, and
  Hinton]{mueller19labelsmoothing}
Rafael M{\"{u}}ller, Simon Kornblith, and Geoffrey~E. Hinton.
\newblock When does label smoothing help?
\newblock In \emph{Conference on Neural Information Processing Systems
  (NeurIPS)}, pages 4696--4705, 2019.

\bibitem[Ng and Winkler(2014)]{facescrub}
Hongwei Ng and Stefan Winkler.
\newblock {A data-driven approach to cleaning large face datasets}.
\newblock In \emph{IEEE International Conference on Image Processing (ICIP)},
  pages 343--347, 2014.

\bibitem[Nickel and Kiela(2017)]{poincare}
Maximilian Nickel and Douwe Kiela.
\newblock {Poincar\'{e} Embeddings for Learning Hierarchical Representations}.
\newblock In \emph{Conference on Neural Information Processing Systems
  (NeurIPS)}, pages 6341--6350, 2017.

\bibitem[Pajouheshgar et~al.(2022)Pajouheshgar, Zhang, and
  S{\"{u}}sstrunk]{pajouheshgar22optimizing}
Ehsan Pajouheshgar, Tong Zhang, and Sabine S{\"{u}}sstrunk.
\newblock Optimizing latent space directions for gan-based local image editing.
\newblock In \emph{International Conference on Acoustics, Speech and Signal
  Processing (ICASSP)}, pages 1740--1744, 2022.

\bibitem[Parihar et~al.(2022)Parihar, Dhiman, Karmali, and
  R]{parihar22exploration}
Rishubh Parihar, Ankit Dhiman, Tejan Karmali, and Venkatesh R.
\newblock Everything is there in latent space: Attribute editing and attribute
  style manipulation by stylegan latent space exploration.
\newblock In \emph{ACM International Conference on Multimedia}, pages
  1828--1836, 2022.

\bibitem[Parisot et~al.(2021)Parisot, Pej{\'{o}}, and
  Spagnuelo]{parisot21propertycnn}
Mathias P.~M. Parisot, Bal{\'{a}}zs Pej{\'{o}}, and Dayana Spagnuelo.
\newblock Property inference attacks on convolutional neural networks:
  Influence and implications of target model's complexity.
\newblock In \emph{International Conference on Security and Cryptography
  ({SECRYPT})}, pages 715--721, 2021.

\bibitem[Paszke et~al.(2019)Paszke, Gross, Massa, Lerer, Bradbury, Chanan,
  Killeen, Lin, Gimelshein, Antiga, Desmaison, K{\"{o}}pf, Yang, DeVito,
  Raison, Tejani, Chilamkurthy, Steiner, Fang, Bai, and Chintala]{pytorch}
Adam Paszke, Sam Gross, Francisco Massa, Adam Lerer, James Bradbury, Gregory
  Chanan, Trevor Killeen, Zeming Lin, Natalia Gimelshein, Luca Antiga, Alban
  Desmaison, Andreas K{\"{o}}pf, Edward Yang, Zachary DeVito, Martin Raison,
  Alykhan Tejani, Sasank Chilamkurthy, Benoit Steiner, Lu~Fang, Junjie Bai, and
  Soumith Chintala.
\newblock {PyTorch: An Imperative Style, High-Performance Deep Learning
  Library}.
\newblock In \emph{Conference on Neural Information Processing Systems
  (NeurIPS)}, pages 8024--8035, 2019.

\bibitem[Rombach et~al.(2022)Rombach, Blattmann, Lorenz, Esser, and
  Ommer]{Rombach2022}
Robin Rombach, Andreas Blattmann, Dominik Lorenz, Patrick Esser, and Bj\"orn
  Ommer.
\newblock High-resolution image synthesis with latent diffusion models.
\newblock In \emph{Conference on Computer Vision and Pattern Recognition
  (CVPR)}, pages 10684--10695, 2022.

\bibitem[Schmidt et~al.(2018)Schmidt, Santurkar, Tsipras, Talwar, and
  Madry]{schmidt18advrobust}
Ludwig Schmidt, Shibani Santurkar, Dimitris Tsipras, Kunal Talwar, and
  Aleksander Madry.
\newblock Adversarially robust generalization requires more data.
\newblock In \emph{Conference on Neural Information Processing Systems
  (NeurIPS)}, pages 5019--5031, 2018.

\bibitem[Shokri et~al.(2017)Shokri, Stronati, Song, and Shmatikov]{shokri2017}
Reza Shokri, Marco Stronati, Congzheng Song, and Vitaly Shmatikov.
\newblock Membership inference attacks against machine learning models.
\newblock In \emph{Symposium on Security and Privacy (S\&P)}, pages 3--18,
  2017.

\bibitem[Struppek et~al.(2022)Struppek, Hintersdorf, Correia, Adler, and
  Kersting]{struppek22_mia}
Lukas Struppek, Dominik Hintersdorf, Antonio De~Almeida Correia, Antonia Adler,
  and Kristian Kersting.
\newblock Plug {\&} play attacks: Towards robust and flexible model inversion
  attacks.
\newblock In \emph{International Conference on Machine Learning ({ICML})},
  volume 162, pages 20522--20545, 2022.

\bibitem[Subramanyam et~al.(2022)Subramanyam, Narayanaswamy, Naufel, Spanias,
  and Thiagarajan]{subramanyam22styleganinversion}
Rakshith Subramanyam, Vivek~Sivaraman Narayanaswamy, Mark Naufel, Andreas
  Spanias, and Jayaraman~J. Thiagarajan.
\newblock Improved stylegan-v2 based inversion for out-of-distribution images.
\newblock In \emph{International Conference on Machine Learning (ICML)}, volume
  162, pages 20625--20639, 2022.

\bibitem[Sundararajan et~al.(2017)Sundararajan, Taly, and
  Yan]{sundararajan17ig}
Mukund Sundararajan, Ankur Taly, and Qiqi Yan.
\newblock Axiomatic attribution for deep networks.
\newblock In \emph{International Conference on Machine Learning (ICML)},
  volume~70, pages 3319--3328, 2017.

\bibitem[Szegedy et~al.(2014)Szegedy, Zaremba, Sutskever, Bruna, Erhan,
  Goodfellow, and Fergus]{szegedy14intriguing}
Christian Szegedy, Wojciech Zaremba, Ilya Sutskever, Joan Bruna, Dumitru Erhan,
  Ian~J. Goodfellow, and Rob Fergus.
\newblock Intriguing properties of neural networks.
\newblock In \emph{International Conference on Learning Representations
  ({ICLR})}, 2014.

\bibitem[Szegedy et~al.(2016)Szegedy, Vanhoucke, Ioffe, Shlens, and
  Wojna]{szegedy16labelsmoothing}
Christian Szegedy, Vincent Vanhoucke, Sergey Ioffe, Jonathon Shlens, and
  Zbigniew Wojna.
\newblock Rethinking the inception architecture for computer vision.
\newblock In \emph{Conference on Computer Vision and Pattern Recognition
  (CVPR)}, pages 2818--2826, 2016.

\bibitem[Wang et~al.(2021)Wang, Fu, Khisti, Zemel, and
  Makhzani]{variational_mia}
Kuan-Chieh Wang, Yan Fu, Ke~Liand~Ashish Khisti, Richard Zemel, and Alireza
  Makhzani.
\newblock {Variational Model Inversion Attacks}.
\newblock In \emph{Conference on Neural Information Processing Systems
  (NeurIPS)}, 2021.

\bibitem[Wang and Wang(2022)]{wang22group}
Xiuling Wang and Wendy~Hui Wang.
\newblock Group property inference attacks against graph neural networks.
\newblock In \emph{Conference on Computer and Communications Security ({CCS})},
  pages 2871--2884, 2022.

\bibitem[Wong et~al.(2020)Wong, Rice, and Kolter]{wong20ffgsm}
Eric Wong, Leslie Rice, and J.~Zico Kolter.
\newblock Fast is better than free: Revisiting adversarial training.
\newblock In \emph{International Conference on Learning Representations
  ({ICLR})}, 2020.

\bibitem[Yeom et~al.(2018)Yeom, Giacomelli, Fredrikson, and
  Jha]{yeom2018privacy}
Samuel Yeom, Irene Giacomelli, Matt Fredrikson, and Somesh Jha.
\newblock Privacy risk in machine learning: Analyzing the connection to
  overfitting.
\newblock In \emph{Computer Security Foundations Symposium (CSF)}, pages
  268--282, 2018.

\bibitem[Zhang et~al.(2020{\natexlab{a}})Zhang, Wu, Zhang, Zhu, Lin, Zhang,
  Sun, He, Mueller, Manmatha, Li, and Smola]{zhang2020resnest}
Hang Zhang, Chongruo Wu, Zhongyue Zhang, Yi~Zhu, Haibin Lin, Zhi Zhang, Yue
  Sun, Tong He, Jonas Mueller, R.~Manmatha, Mu~Li, and Alexander Smola.
\newblock Resnest: Split-attention networks.
\newblock \emph{CoRR}, abs/2004.08955, 2020{\natexlab{a}}.

\bibitem[Zhang et~al.(2020{\natexlab{b}})Zhang, Jia, Pei, Wang, Li, and
  Song]{secret_revealer}
Yuheng Zhang, Ruoxi Jia, Hengzhi Pei, Wenxiao Wang, Bo~Li, and Dawn Song.
\newblock {The Secret Revealer: Generative Model-Inversion Attacks Against Deep
  Neural Networks}.
\newblock In \emph{Conference on Computer Vision and Pattern Recognition
  (CVPR)}, pages 250--258, 2020{\natexlab{b}}.

\bibitem[Zhou et~al.(2022)Zhou, Chen, Shen, and Zhang]{zhou22property}
Junhao Zhou, Yufei Chen, Chao Shen, and Yang Zhang.
\newblock Property inference attacks against gans.
\newblock In \emph{Annual Network and Distributed System Security Symposium
  ({NDSS})}, 2022.

\end{thebibliography}

\cleardoublepage
\newpage

\onecolumn
\appendix
\section{Experimental Details}\label{appx:experimental_details}

\subsection{Hard- and Software Details}\label{app:hardware_details}
We performed our experiments on NVIDIA DGX machines running NVIDIA DGX Server Version 5.1.0 and Ubuntu 20.04.4 LTS. The machines have 2TB of RAM and contain NVIDIA A100-SXM4-40GB GPUs and AMD EPYC 7742 CPUs. We further relied on CUDA 11.4, Python 3.10.0, and PyTorch 1.12.1 with Torchvision 0.13.1~\cite{pytorch} for our experiments. If not stated otherwise, we used the model architecture implementations and pre-trained ImageNet weights provided by Torchvision. We further provide a Dockerfile together with our code to make the reproduction of our results easier. In addition, all training and attack configuration files are available to reproduce the results stated in this paper.

To perform Plug \& Play Attacks (PPA), we used a NVIDIA DGX machine running NVIDIA DGX Server Version 5.2.0 and Ubuntu 20.04.4 LTS. The machines has 2TB of RAM and contain NVIDIA A100-SXM4-80GB GPUs and AMD EPYC 7742 CPUs. The experiments were performed with CUDA 11.4, Python 3.8.10, and PyTorch 1.10.0 with Torchvision 0.11.0. We further used PPA in combination with the pre-trained FFHQ StyleGAN-2 and relied on the standard CelebA attack parameters, as stated in \citet{struppek22_mia} and \url{https://github.com/LukasStruppek/Plug-and-Play-Attacks}. We only changed the number of candidates to 50 and the final samples per target to 25, to speed up the attack process.

\subsection{Dataset Details}\label{appx:dataset_details}
We state the number of identities and samples for each custom CelebA subset in \cref{tab:celeba_subsets}. We further state the lowest, median and maximum number of samples of a single identity per attribute and for the total dataset. For the filter models, the datasets contained 155,304 samples (gender and eyeglasses) and 93,942 (hair color), respectively.

\begin{table}[ht]
\centering
\resizebox{0.8\linewidth}{!}{  
\begin{tabular}{llcccccc}
\toprule
\multirow{2}{*}{\textbf{Attribute}} & \multirow{2}{*}{\textbf{Value}}     & \multirow{2}{*}{\textbf{Identities}} & \textbf{Total} & \textbf{Min} & \textbf{Median} & \textbf{Max} & \textbf{Avg}\\
          &           &            & \textbf{Samples} & \textbf{Samples} & \textbf{Samples} & \textbf{Samples} & \textbf{Samples}\\
\midrule
    \parbox[t]{1mm}{\multirow{3}{*}{\rotatebox[origin=c]{90}{\parbox{1.2cm}{\centering \textbf{CelebA} \\ \textbf{Gender}}}}}  & Female & 250 & 7,639 & 30 & 30 & 35 & 30.56\\
    & Male        & 250 & 7,652 & 30 & 30 & 36 & 30.61 \\
    \cmidrule{2-8}
    & Total       & 500 & 15,291 & 30 & 30 & 36 & 30.58 \\
\midrule
    \parbox[t]{1mm}{\multirow{3}{*}{\rotatebox[origin=c]{90}{\parbox{1.2cm}{\centering \textbf{CelebA} \\ \textbf{Glasses}}}}} & No Glasses & 100 & 3186 & 31 & 32 & 36 & 31.86\\
    & Glasses        & 100 & 2,346 & 20 & 22 & 31 & 23.46 \\
     \cmidrule{2-8}
    & Total          & 200 & 5532 & 20 & 31 & 36 & 27.66 \\
\midrule
    \parbox[t]{1mm}{\multirow{5}{*}{\rotatebox[origin=c]{90}{\parbox{1.2cm}{\centering \textbf{CelebA} \\ \textbf{Race}}}}} & Asian & 100 & 2944 & 28 & 30 & 31 & 29.44 \\
    & Black   & 100 & 2876    & 26   & 29    & 34 & 28.76 \\
    & Indian  & 100 & 2281    & 19   & 22    & 30 & 22.81 \\
    & White   & 100 & 3164    & 31   & 31    & 35 & 31.64\\
     \cmidrule{2-8}
    & Total   &  400 & 11,265  & 19   & 29    & 35 & 28.16\\
\midrule
    \parbox[t]{1mm}{\multirow{5}{*}{\rotatebox[origin=c]{90}{\parbox{1.2cm}{\centering \textbf{CelebA} \\\textbf{Hair Color}}}}} & Black Hair & 100 & 2,840 & 23 & 28 & 32 & 28.40\\
    & Blond Hair & 100 & 2,867 & 25 & 29 & 31 & 28.67 \\
    & Brown Hair & 100 & 2,564 & 17 & 26 & 30 & 25.64 \\
    & Gray Hair  & 100 & 2,117 & 14 & 20 & 31 & 21.17 \\
     \cmidrule{2-8}
   &  Total      & 400 & 10,388 & 14 & 27 & 32 & 25.97\\
\midrule
    \parbox[t]{1mm}{\multirow{3}{*}{\rotatebox[origin=c]{90}{\parbox{1.5cm}{\centering \textbf{FaceScrub} \\ \textbf{Gender}}}}} & Female & 265 & 17,743 & 9 & 65 & 140 & 66.95 \\
    & Male        &  265 & 20,136 & 25 & 77 & 128 & 75.98\\
     \cmidrule{2-8}
    & Total          & 530 & 37,879 & 9 & 72 & 140 & 71.47 \\
\bottomrule
\end{tabular}
}
\label{tab:celeba_subsets}
\caption{Dataset identity statistics}
\end{table}

\newpage
\subsection{Training Hyperparameters}\label{appx:training_hyperparameters}
We emphasize that we did not aim for achieving state-of-the-art performances but rather tried to train models with generally good prediction performance. All non-robust models were initialized with pre-trained ImageNet weights and trained for 80 epochs with an initial learning rate of $0.1$ and a batch size of 128. We multiplied the learning rate by factor $0.1$ after 60 and 70 epochs. For the models trained with adversarial training, set the number of epochs to 100 and reduced the learning rate after 80 and 90 epochs. Optimization was done with SGD and a momentum of $0.9$. Images were resized to $224\times224$ and normalized with $\mu=\sigma=0.5$ to set the pixel value range to $[-1, 1]$. As augmentation, we applied random horizontal flipping with $p=0.5$ flipping probability and random resized cropping with $\text{scale}=[0.8, 1.0]$ and $\text{ratio}=[0.8, 1.25]$. Cropped images were resized to size $224\times 224$. Due to the limited number of training samples available for each identity, we did not use a validation set and early stopping. For each architecture and dataset, we trained three models with different seeds. Performance evaluation was done on a holdout test set.

We further trained ResNet-50 models on the CelebA attributes for gender, eyeglasses, and hair color for filtering the candidate images. All models were initialized with pre-trained ImageNet weights and trained for 10 epochs with an initial learning rate of $1e-3$ and a batch size of 128. We multiplied the learning rate by factor 0.1 after 9 epochs. We used the Adam optimizer~\citep{adam_optimizer} with $\beta=(0.9, 0.999)$ and no weight decay. Normalization and augmentation were identical to training the target models, except the scale and ratio parameters of the random resized cropping, which we set to $[0.9, 1.0]$ and $[1.0, 1.0]$ to prevent cutting out the attributes from the training samples. We again used no validation set for early stopping. To calibrate the models, we applied label smoothing with smoothing factor $\alpha=0.1$ during training. Since some attributes are more frequent in the CelebA dataset, we draw the same number of samples for each attribute by oversampling / undersampling from the attribute images. Details on the training of FairFace for filtering the racial appearance images can be found at \url{https://github.com/dchen236/FairFace}.

\begin{table}[ht]
\begin{subtable}[h]{0.45\textwidth}
\centering
\resizebox{!}{4.4cm}{  
\begin{tabular}{lllc}
\toprule
    \textbf{Dataset} & \textbf{Num Classes} &  \textbf{Architecture} & \textbf{Accuracy} \\
\midrule
    \parbox[t]{1mm}{\multirow{5}{*}{\rotatebox[origin=c]{90}{\footnotesize\textbf{Gender}}}} & \multirow{5}{*}{{500}} 
    &  ResNet-18 & $85.99\%\pm0.21$ \\
    & & ResNet-101 & $86.78\%\pm0.66$ \\
    & & ResNet-152 &  $86.62\%\pm0.62$ \\
    & & DenseNet-169 &  $75.93\%\pm5.54$ \\
    & & ResNeSt-101 &  	$81.53\%\pm4.40$ \\
\midrule
    \parbox[t]{1mm}{\multirow{5}{*}{\rotatebox[origin=c]{90}{\footnotesize\textbf{Eyeglasses}}}} & \multirow{5}{*}{{200}} 
    &  ResNet-18 & $93.08\%\pm0.58$ \\
    & & ResNet-101 & $90.97\%\pm1.41$ \\
    & & ResNet-152 & $91.10\%\pm0.85$ \\
    & & DenseNet-169 & $93.56\%\pm0.38$ \\
    & & ResNeSt-101 & $90.73\%\pm0.21$ \\
\midrule
    \parbox[t]{1mm}{\multirow{5}{*}{\rotatebox[origin=c]{90}{\footnotesize\textbf{Race}}}} &
    \multirow{5}{*}{{400}} 
    &  ResNet-18 & $84.65\%\pm0.44$ \\
    & & ResNet-101 & $82.99\%\pm0.40$ \\
    & & ResNet-152 & $84.06\%\pm1.19$ \\
    & & DenseNet-169 & $85.00\%\pm0.96$ \\
    & & ResNeSt-101 & $81.13\%\pm1.04$ \\
\midrule
    \parbox[t]{1mm}{\multirow{5}{*}{\rotatebox[origin=c]{90}{\footnotesize\textbf{Hair Color}}}} &
    \multirow{5}{*}{{400}} 
    &  ResNet-18 & $87.94\%\pm0.56$ \\
    & & ResNet-101 & $87.65\%\pm0.70$ \\
    & & ResNet-152 & $87.20\%\pm0.67$ \\
    & & DenseNet-169 & $88.45\%\pm1.68$ \\
    & & ResNeSt-101 &  $85.53\%\pm0.53$ \\
\bottomrule
\end{tabular}
}
\caption{CelebA}
\end{subtable}
\hfill
\begin{subtable}[h]{0.45\textwidth}
\centering
\resizebox{!}{4.4cm}{  
\begin{tabular}{lllc}
\toprule
    \textbf{Dataset} & \textbf{Num Classes} &  \textbf{Architecture} & \textbf{Accuracy} \\
\midrule
    \parbox[t]{1mm}{\multirow{5}{*}{\rotatebox[origin=c]{90}{\footnotesize \parbox{1.5cm}{\centering \textbf{FaceScrub} \\ \textbf{Cropped}}}}} &
    \multirow{5}{*}{{530}} 
    &  ResNet-18 & $93.92\%\pm0.17$ \\
    & & ResNet-101 & $94.93\%\pm0.43$ \\
    & & ResNet-152 & $95.32\%\pm0.22$ \\
    & & DenseNet-169 & $91.93\%\pm0.65$ \\
    & & ResNeSt-101 & $93.50\%\pm0.70$ \\
\midrule
    \parbox[t]{1mm}{\multirow{5}{*}{\rotatebox[origin=c]{90}{\footnotesize \parbox{1.5cm}{\centering \textbf{FaceScrub} \\ \textbf{Cropped} \\ \textbf{Robust}}}}} &
    \multirow{5}{*}{{530}} 
    &  ResNet-18 & $83.79\%\pm0.21$ \\
    & & ResNet-101 & $88.12\%\pm0.24$ \\
    & & ResNet-152 & $87.11\%\pm0.28$ \\
    & & DenseNet-169 & $88.53\%\pm0.76$ \\
    & & ResNeSt-101 & $90.04\%\pm0.48$ \\
\midrule
    \parbox[t]{1mm}{\multirow{5}{*}{\rotatebox[origin=c]{90}{\footnotesize \parbox{1.5cm}{\centering \textbf{FaceScrub} \\ \textbf{Uncropped}}}}} &
    \multirow{5}{*}{{530}} 
    &  ResNet-18 & $70.29\%\pm2.71$ \\
    & & ResNet-101 & $78.75\%\pm0.84$ \\
    & & ResNet-152 & $78.52\%\pm0.58$ \\
    & & DenseNet-169 & $\,\,\,69.11\%\pm12.73$ \\
    & & ResNeSt-101 & $77.32\%\pm0.54$ \\
\midrule
    \parbox[t]{1mm}{\multirow{5}{*}{\rotatebox[origin=c]{90}{\footnotesize \parbox{1.5cm}{\centering \textbf{FaceScrub} \\ \textbf{Uncropped} \\ \textbf{Robust}}}}} &
    \multirow{5}{*}{{530}} 
    &  ResNet-18 & $41.58\%\pm0.63$ \\
    & & ResNet-101 & $55.23\%\pm0.60$ \\
    & & ResNet-152 & $55.69\%\pm1.24$ \\
    & & DenseNet-169 & $54.65\%\pm2.98$ \\
    & & ResNeSt-101 & $58.74\%\pm0.84$ \\
\bottomrule
\end{tabular}
}
\caption{FaceScrub}
\end{subtable}

\caption{Prediction accuracy of the target models on the individual test sets.}
\end{table}

\newpage
\subsection{Attack Dataset Generation}
To create the attack dataset, we first applied Null-Text Inversion~\citep{mokady22nullinversion} with 50 DDIM steps  and a guidance scale of 7.5 on Stable Diffusion v1.5\footnote{Available at \url{https://huggingface.co/runwayml/stable-diffusion-v1-5}.}. We further used the generic prompt \textit{“A photo of a person”} for all samples. After the inversion, we generated image variations by adding the sensitive attribute values to the prompt using prompt-to-prompt, e.g., \textit{“A photo of a person, female appearance”} and \textit{“A photo of a person, male appearance”} to generate gender variations. We set the cross replace steps to 1.0 and the self replace steps to 0.4 for gender and eyeglasses, and to 0.6 for hair color and racial appearance. The confidence threshold for the filter models was set to $0.6$ for all models. We then generated and filtered attribute variations one after another, until we collected 300 candidates for each attribute category. \cref{tab:attack_prompts} states all prompts used for generating.

\begin{table}[ht]
\centering
\resizebox{0.4\linewidth}{!}{  
\begin{tabular}{lllllc}
\toprule
    \textbf{Attribute} & \textbf{Value} & \textbf{Prompt} \\
\midrule
    \multirow{2}{*}{Gender}     & Female & \textit{female appearance} \\
                                & Male & \textit{male appearance} \\
\midrule
    \multirow{2}{*}{Glasses}    & No Glasses & \textit{no eyeglasses} \\
                                & Glasses & \textit{wearing eyeglasses} \\
\midrule
                                & Asian & \textit{with asian appearance} \\
    Racial                      & Black & \textit{with black skin} \\
    Appearance                  & Indian & \textit{with indian appearance} \\
                                & White &  \textit{with white skin} \\
\midrule
    \multirow{4}{*}{Hair Color} & Black & \textit{with black hair} \\
                                & Blond & \textit{with blond hair} \\
                                & Brown &  \textit{with brown hair} \\
                                & Gray & \textit{with gray hair} \\
\bottomrule
\end{tabular}
}
\caption{Prompts for attack dataset generation. Each prompt is appended to the string \textit{"A photo of a person, }$\langle\text{ }\rangle$\textit{"} by replacing $\langle\text{ }\rangle$ with the attribute-specific prompt.}
\end{table}

\begin{table}[ht]
\centering
\resizebox{0.6\linewidth}{!}{  
\begin{tabular}{llllc}
\toprule
    \textbf{Attribute}  & \textbf{Num Classes} & \textbf{Training Samples} & \textbf{Architecture} & \textbf{Accuracy} \\
\midrule
    Gender & 2 & 155,304 & ResNet-50 &  $98.98\%$ \\
    Glasses & 2 & 155,304 & ResNet-50 & $98.81\%$ \\
    Hair Color & 4 & 93,940 & ResNet-50 & $93.51\%$ \\
\bottomrule
\end{tabular}
}
\caption{Prediction accuracy of the filter models on the individual test sets.}
\end{table}

\clearpage
\section{Additional CelebA Experimental Results}\label{appx:add_results_celeba}

\subsection{Confusion Matrices for CelebA ResNet-101 Models}\label{appx:confusion_matrices}
\begin{figure*}[h!]
     \begin{subfigure}[b]{0.45\textwidth}
        \centering
         \includegraphics[width=\textwidth]{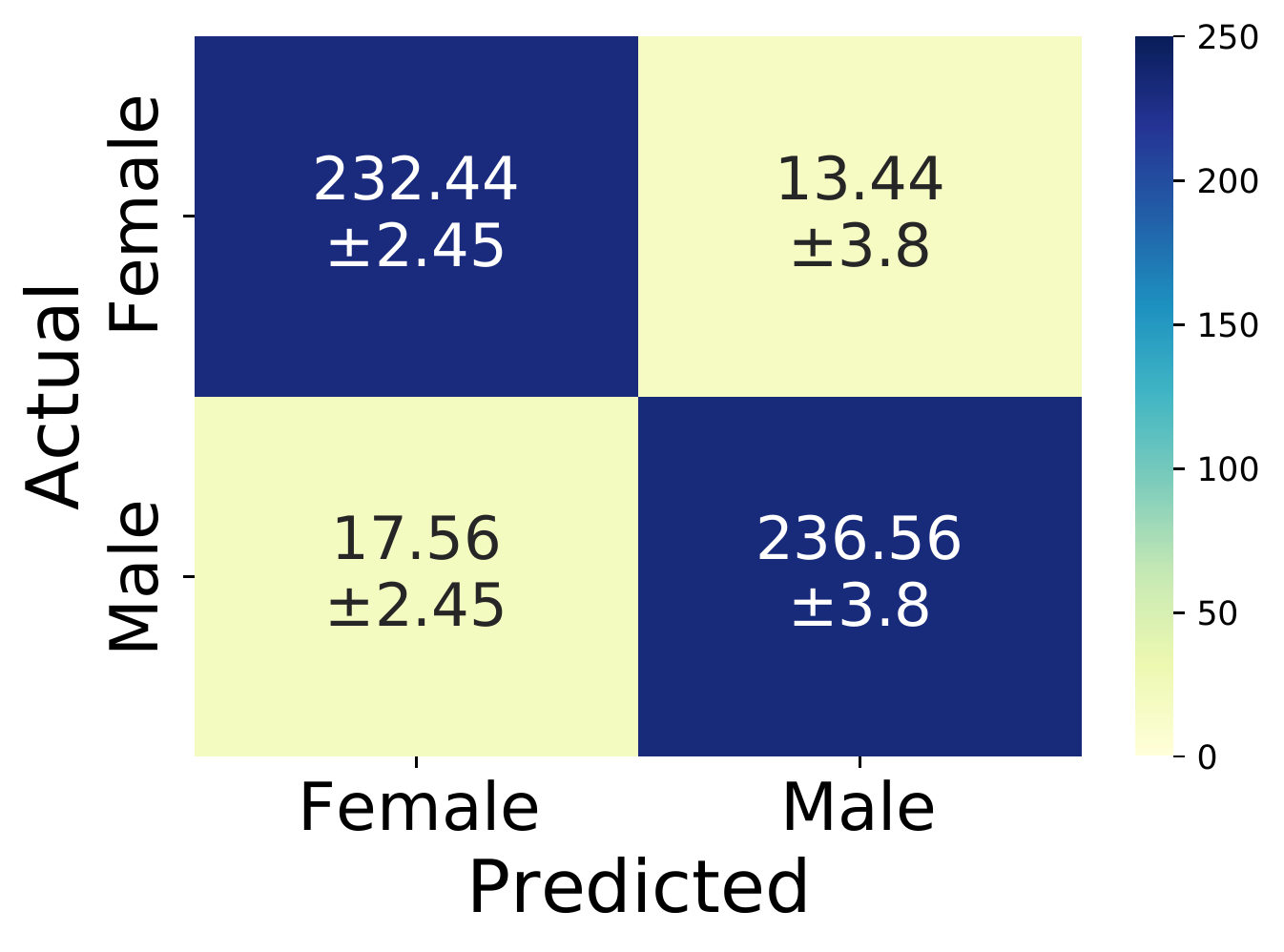}
         \caption{Gender (FFHQ)}
     \end{subfigure}
     \hfill
     \begin{subfigure}[b]{0.45\textwidth}
         \centering
         \includegraphics[width=\textwidth]{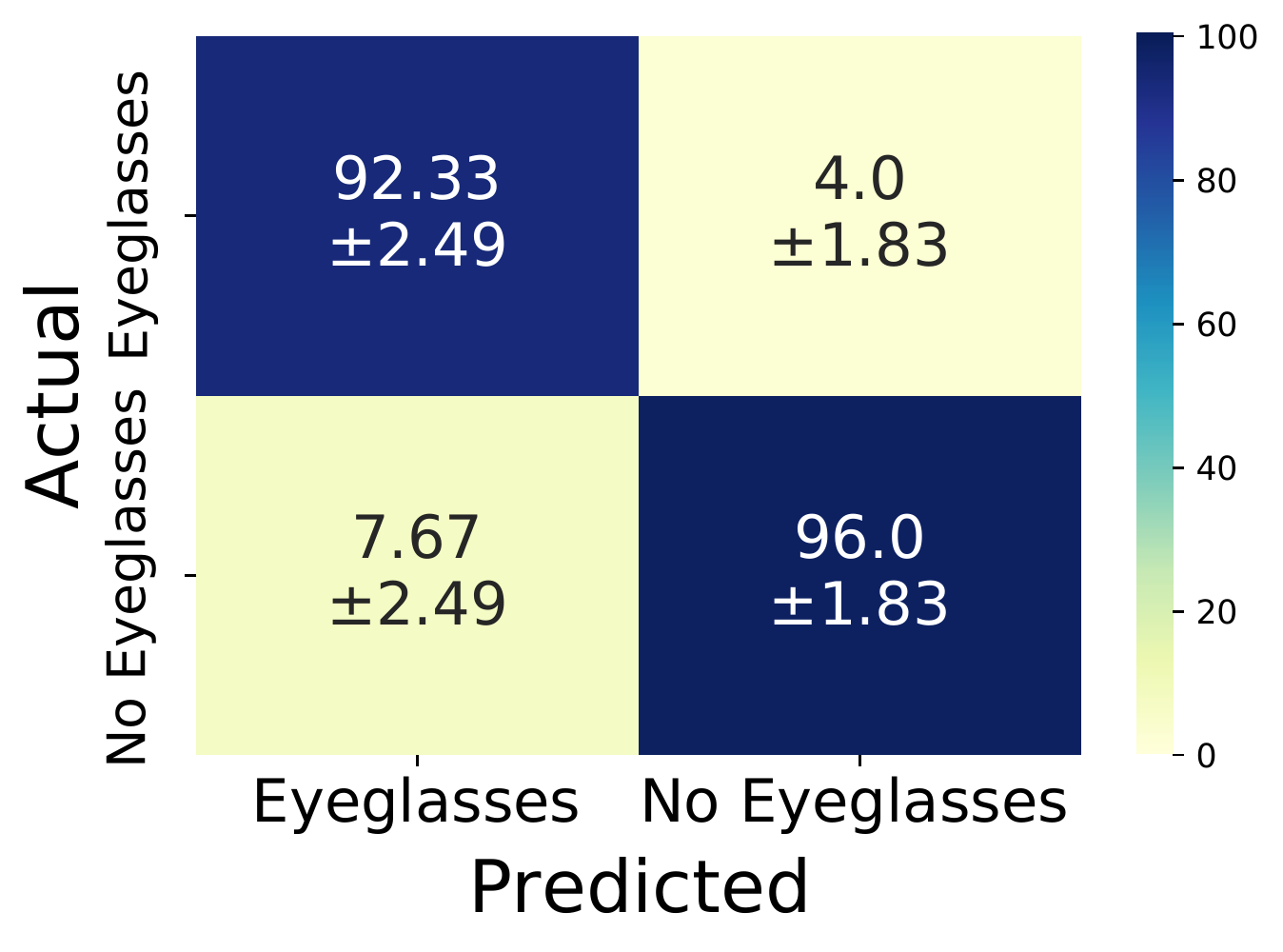}
         \caption{Eyeglasses (FFHQ)}
     \end{subfigure}
     \par\medskip
     \begin{subfigure}[b]{0.45\textwidth}
         \centering
         \includegraphics[width=\textwidth]{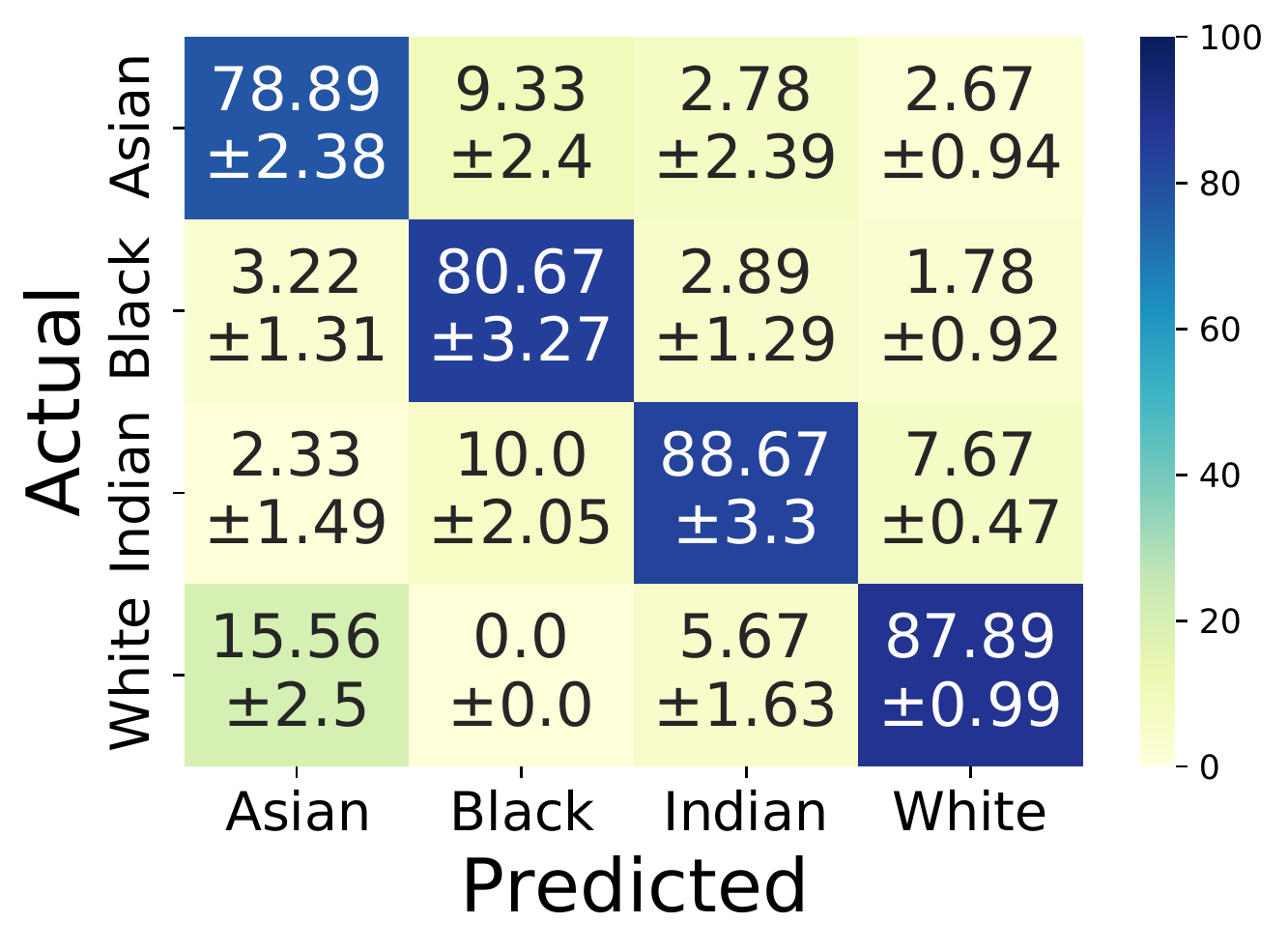}
         \caption{Racial Appearance (FFHQ)}
     \end{subfigure}
     \hfill
     \begin{subfigure}[b]{0.45\textwidth}
         \centering
         \includegraphics[width=\textwidth]{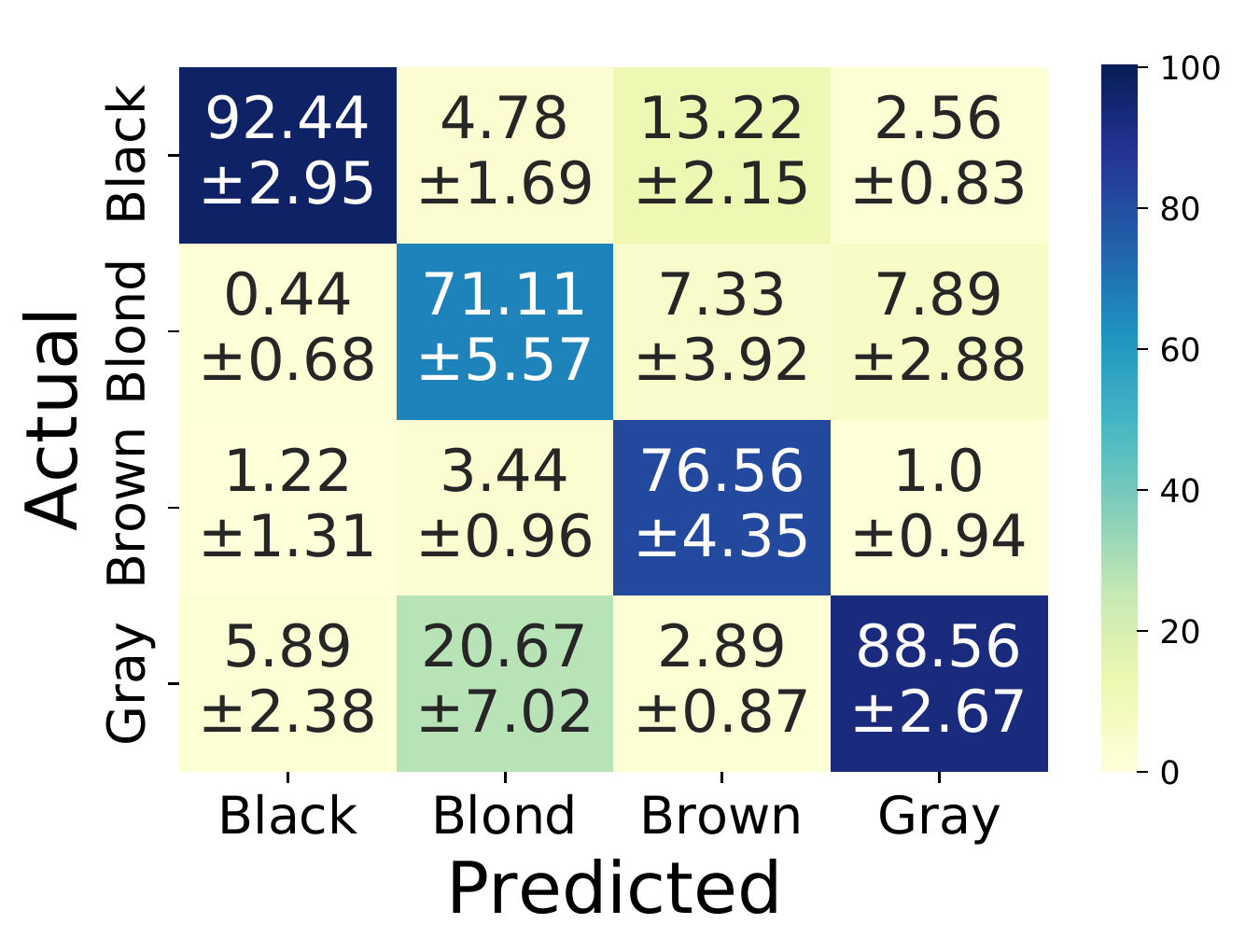}
         \caption{Hair Color (FFHQ)}
     \end{subfigure}
    \caption{Confusion Matrices for ResNet-101 models trained on the four different CelebA subsets. All attacks were performed against three target models and with three different attack datasets.}
\end{figure*}
\clearpage

\subsection{ResNet-18 - CelebA}
\begin{figure*}[h!]
\centering
     \begin{subfigure}[c]{\textwidth}
         \centering
         \includegraphics[width=0.75\textwidth]{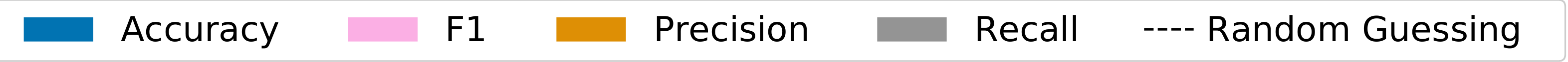}
     \end{subfigure}
     
     \begin{subfigure}[b]{0.48\textwidth}
        \centering
         \includegraphics[width=\textwidth]{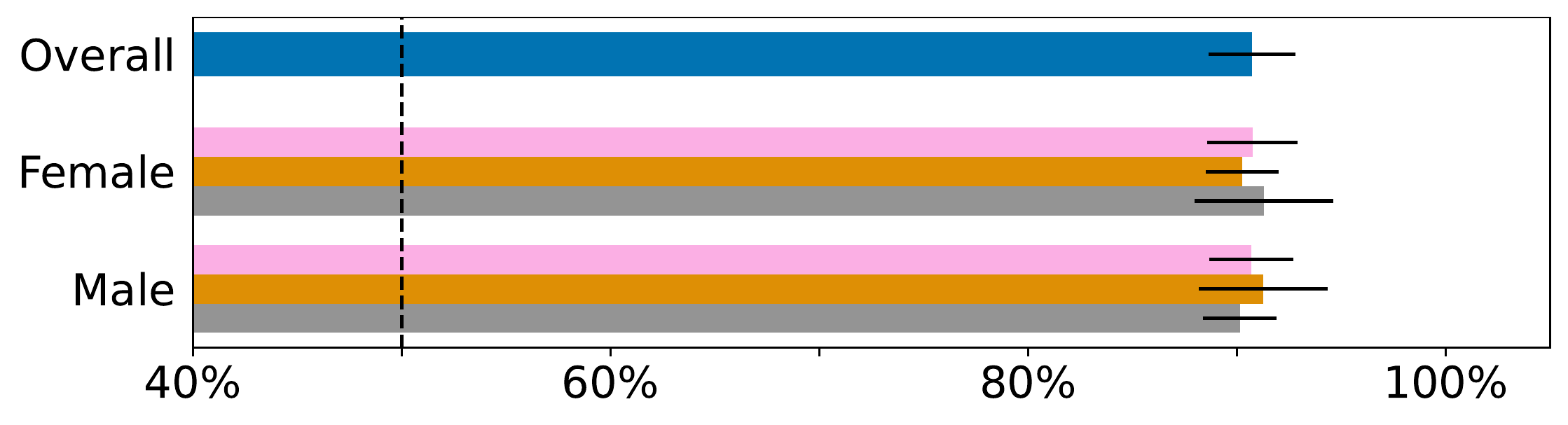}
         \caption{Gender (FFHQ)}
     \end{subfigure}
     \hfill
     \begin{subfigure}[b]{0.48\textwidth}
        \centering
         \includegraphics[width=\textwidth]{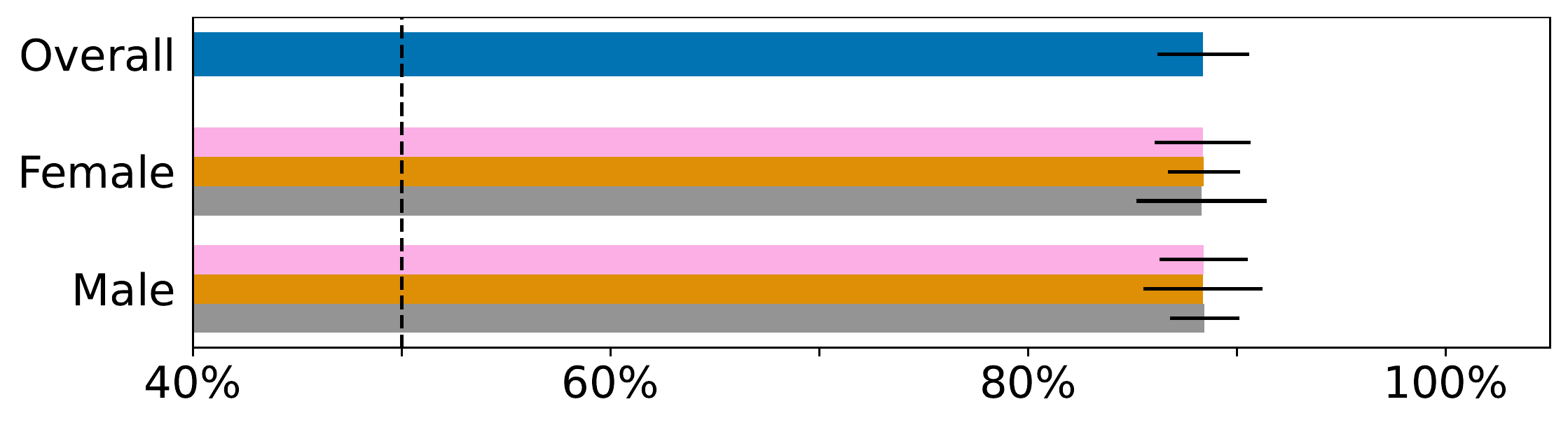}
         \caption{Gender (CelebAHQ)}
     \end{subfigure}

     \par\medskip
     \begin{subfigure}[b]{0.48\textwidth}
         \centering
         \includegraphics[width=\textwidth]{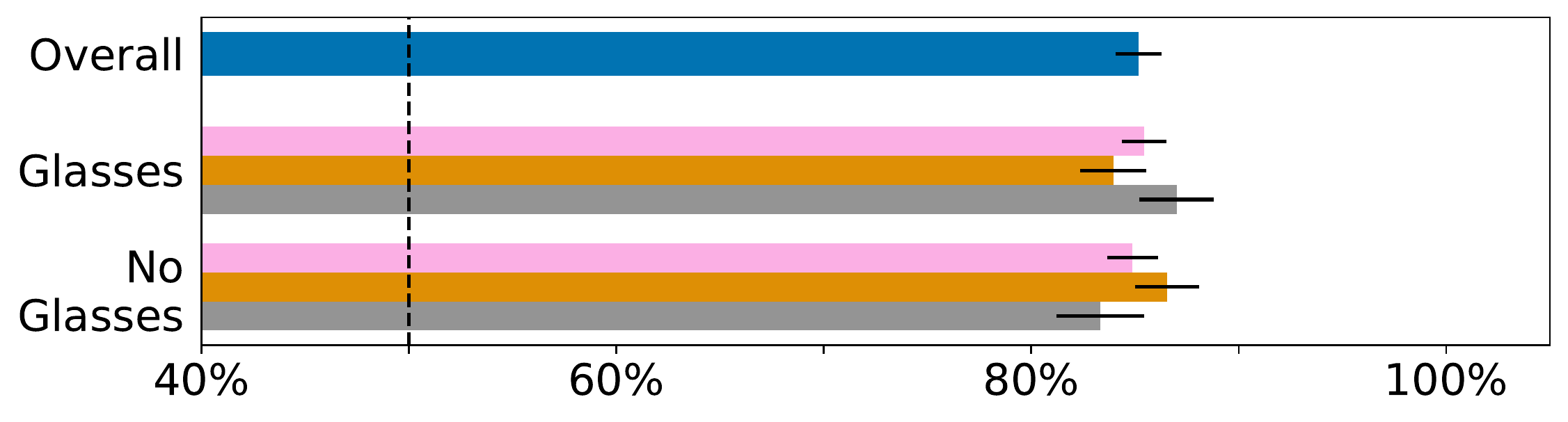}
         \caption{Eyeglasses (FFHQ)}
     \end{subfigure}
     \hfill
     \begin{subfigure}[b]{0.48\textwidth}
         \centering
         \includegraphics[width=\textwidth]{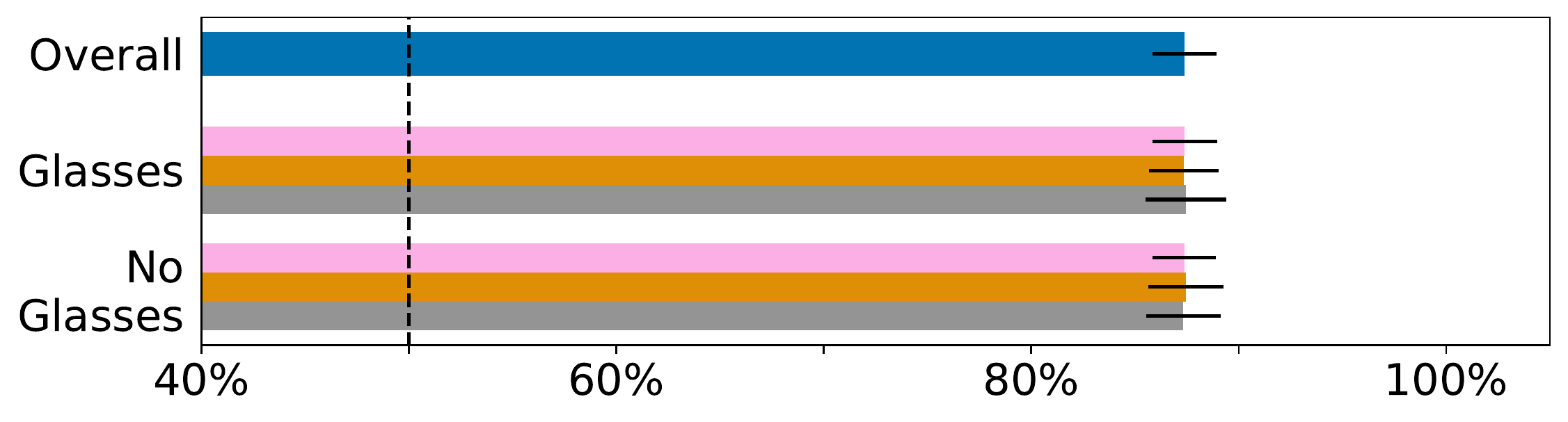}
         \caption{Eyeglasses (CelebAHQ)}
     \end{subfigure}

     \par\medskip
     \begin{subfigure}[b]{0.48\textwidth}
         \centering
         \includegraphics[width=\textwidth]{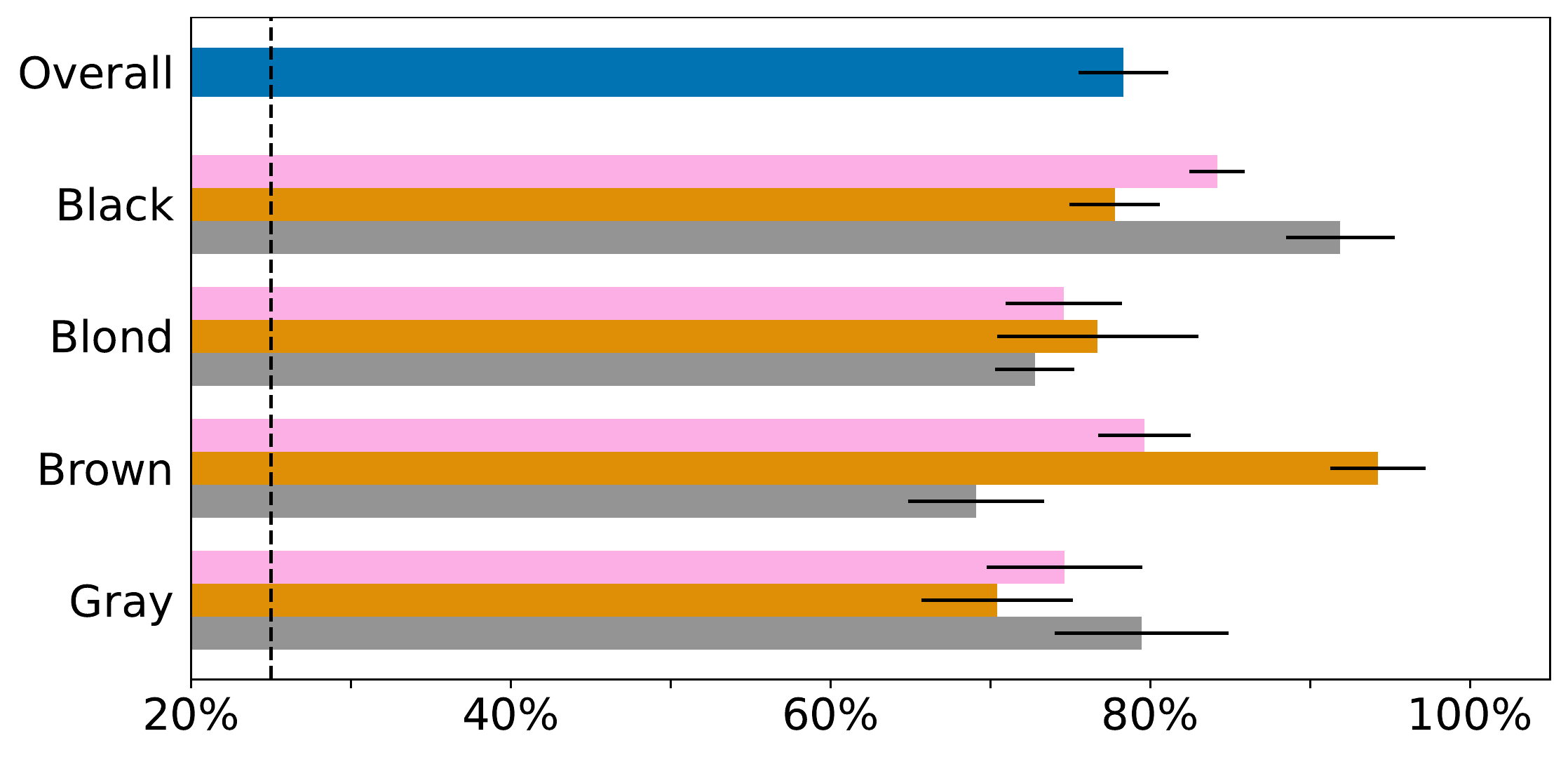}
         \caption{Hair Color (FFHQ)}
     \end{subfigure}
     \hfill
     \begin{subfigure}[b]{0.48\textwidth}
         \centering
         \includegraphics[width=\textwidth]{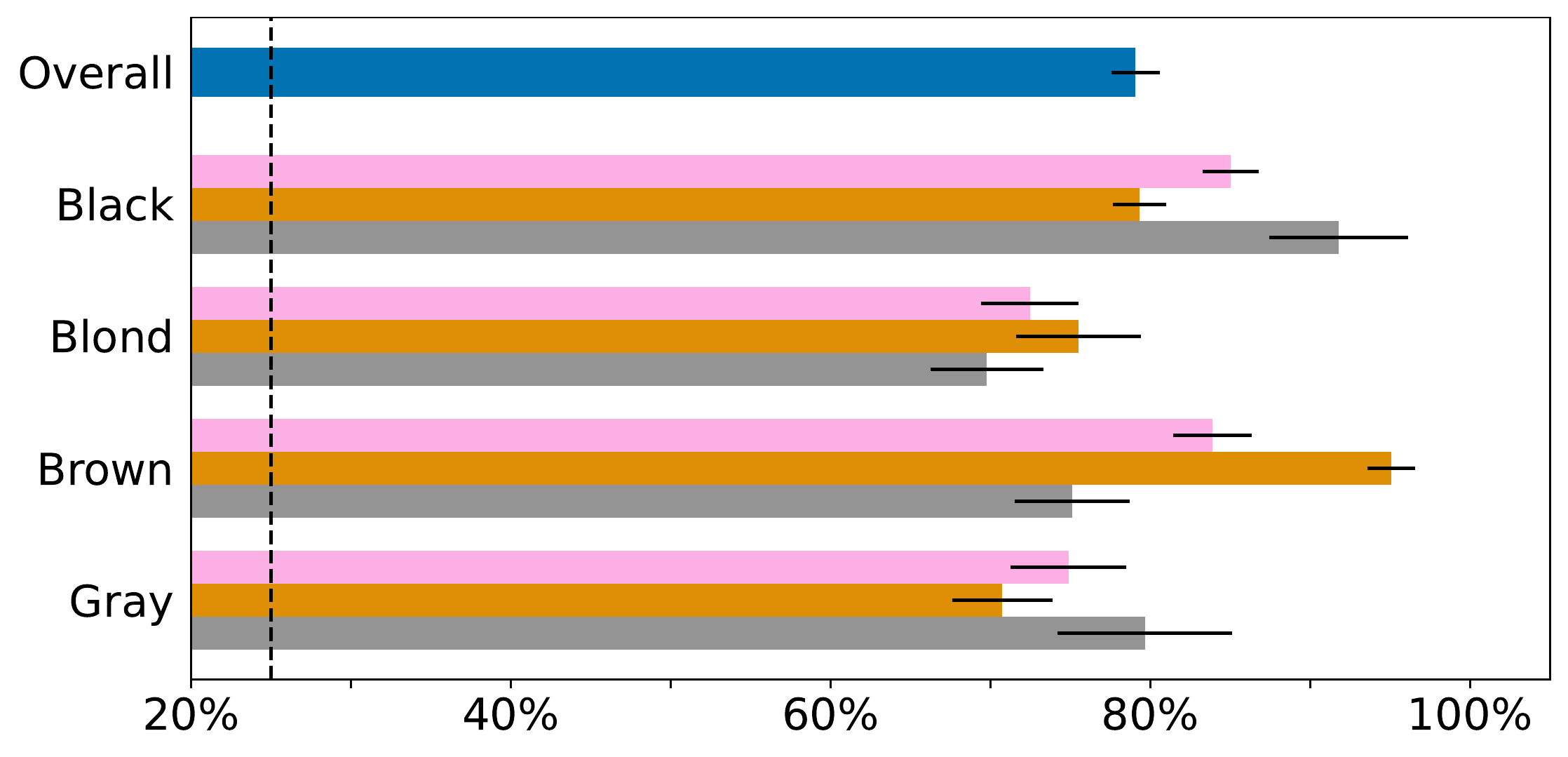}
         \caption{Hair Color (CelebAHQ)}
     \end{subfigure}

     \par\medskip
     \begin{subfigure}[b]{0.48\textwidth}
         \centering
         \includegraphics[width=\textwidth]{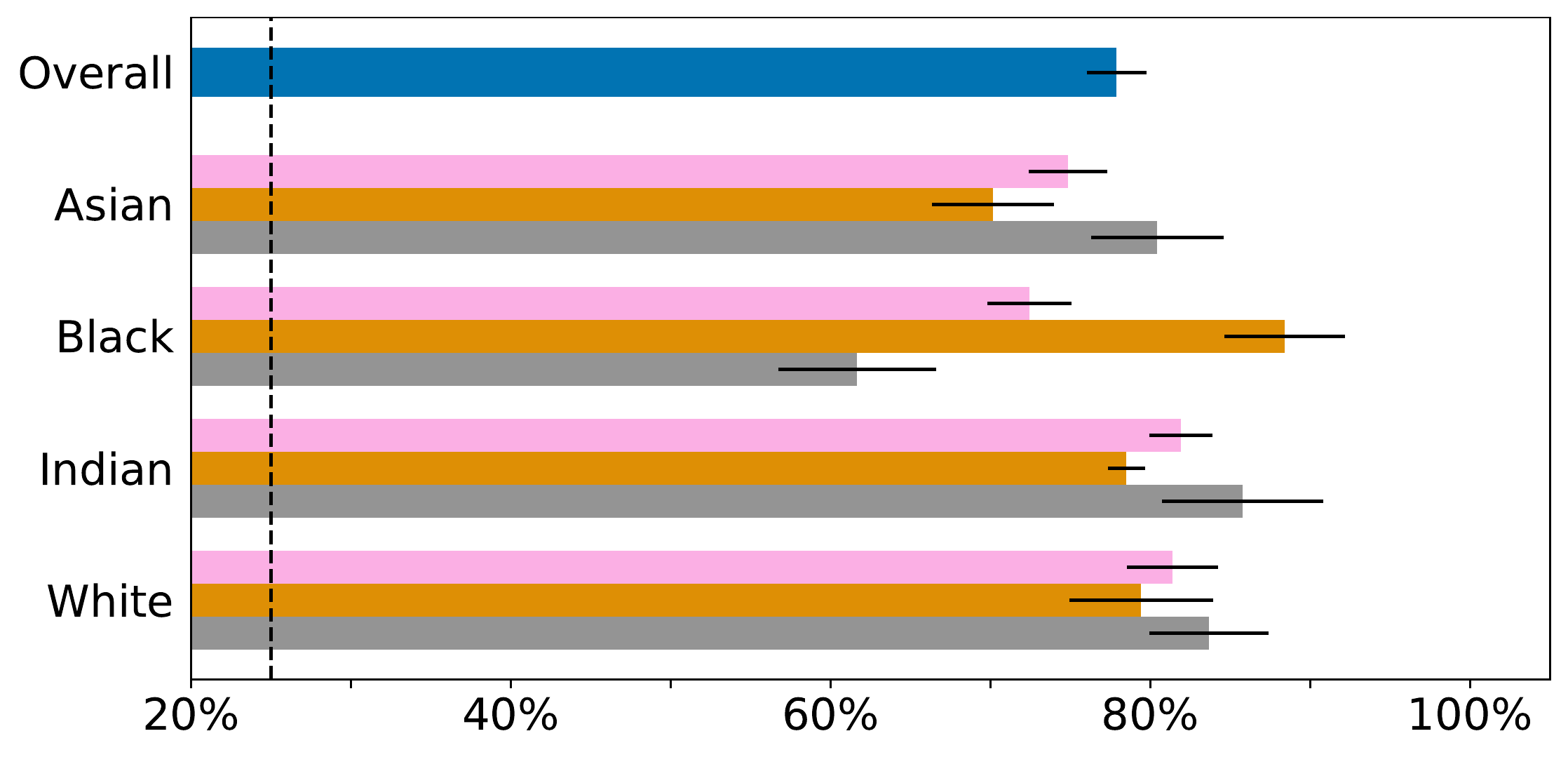}
         \caption{Racial Appearance (FFHQ)}
     \end{subfigure}
     \hfill
     \begin{subfigure}[b]{0.48\textwidth}
         \centering
         \includegraphics[width=\textwidth]{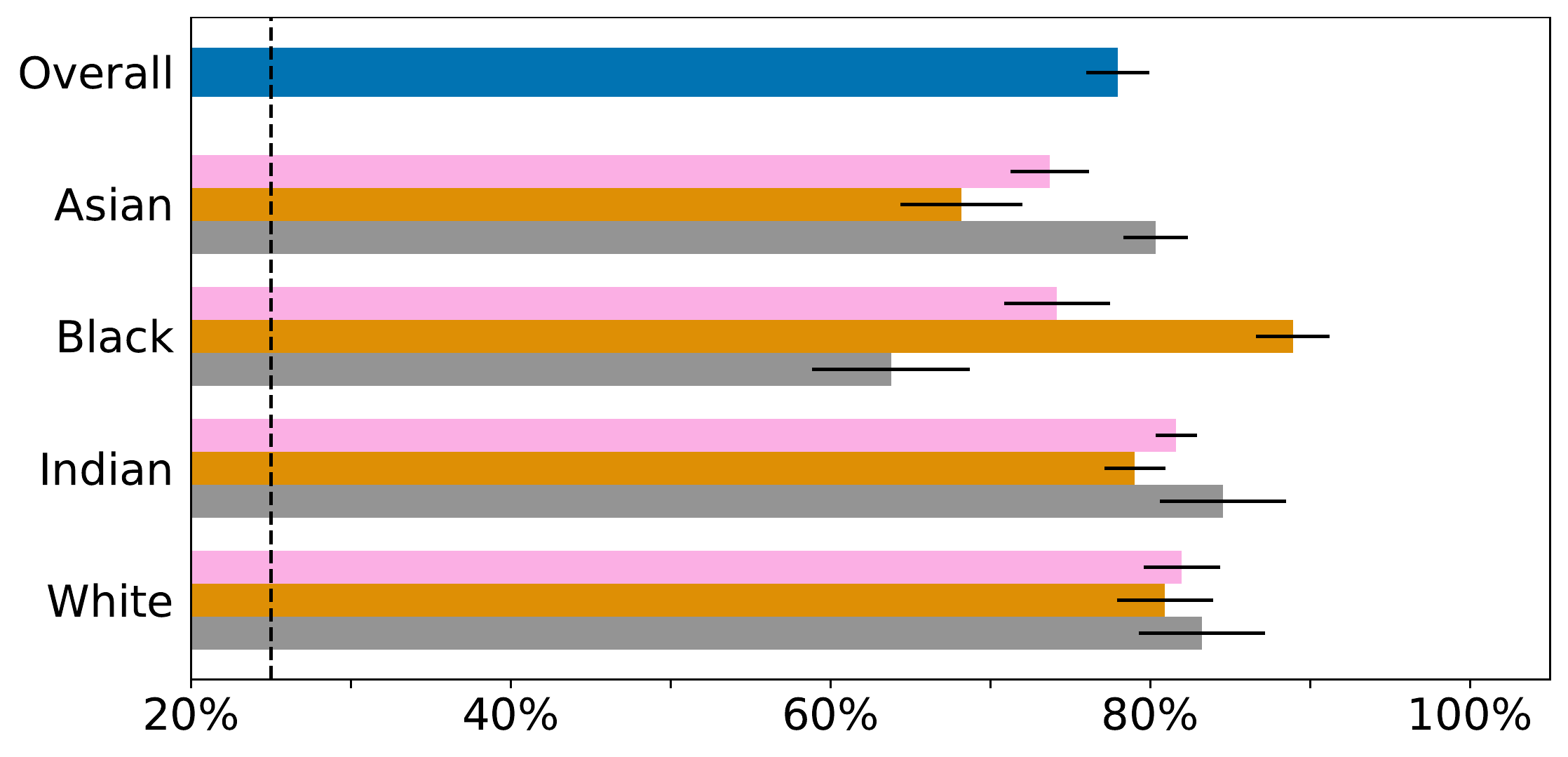}
         \caption{Racial Appearance (CelebAHQ)}
     \end{subfigure}

    \caption{Evaluation results for CAIA performed on ResNet-18 models to infer four different target attributes. The black horizontal lines denote the standard deviation over nine runs. We further state random guessing (dashed line) for comparison.}
\end{figure*}

\clearpage

\subsection{ResNet-18 - CelebA (1000 Targets)}
\begin{figure*}[h!]
\centering
     \begin{subfigure}[c]{\textwidth}
         \centering
         \includegraphics[width=0.75\textwidth]{images/barplots/legend_small.pdf}
     \end{subfigure}
     
     \begin{subfigure}[b]{0.48\textwidth}
        \centering
         \includegraphics[width=\textwidth]{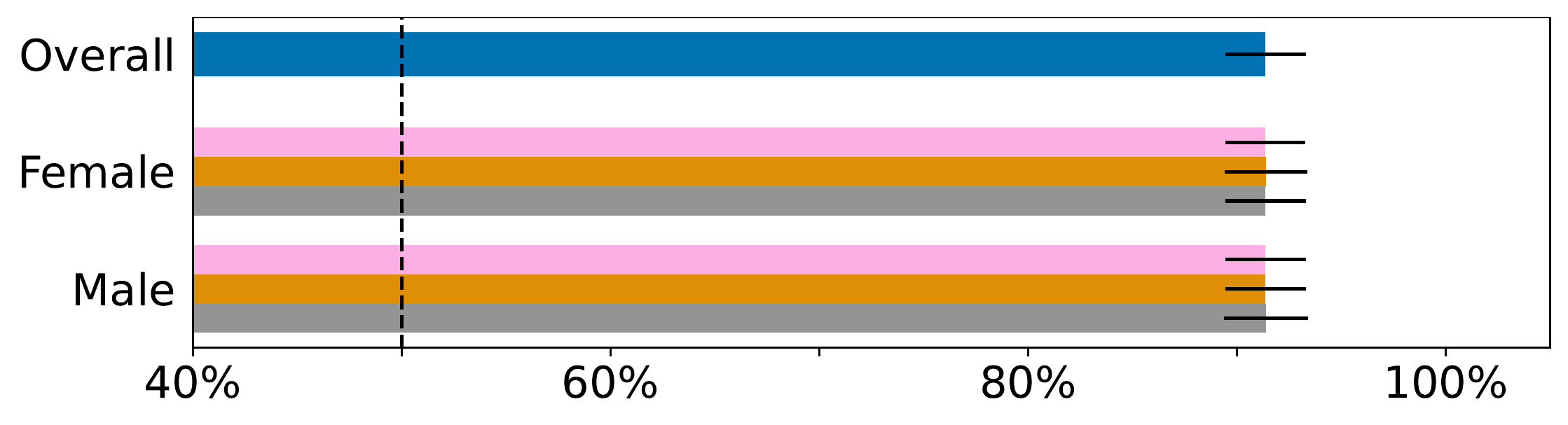}
         \caption{Gender (FFHQ)}
     \end{subfigure}
     \hfill
     \begin{subfigure}[b]{0.48\textwidth}
        \centering
         \includegraphics[width=\textwidth]{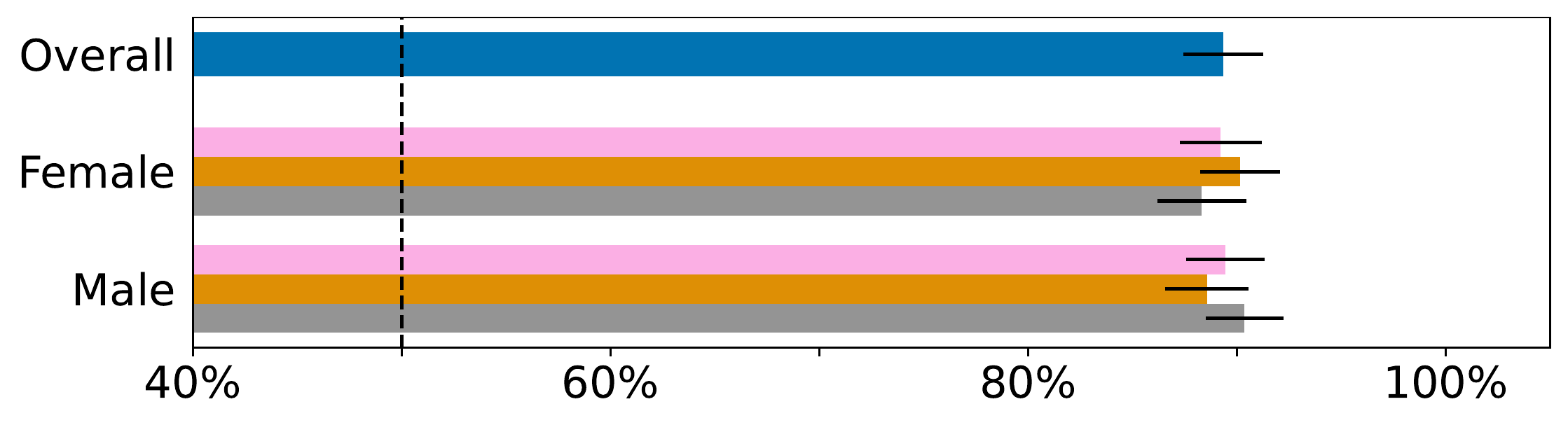}
         \caption{Gender (CelebAHQ)}
     \end{subfigure}

     \par\medskip
     \begin{subfigure}[b]{0.48\textwidth}
         \centering
         \includegraphics[width=\textwidth]{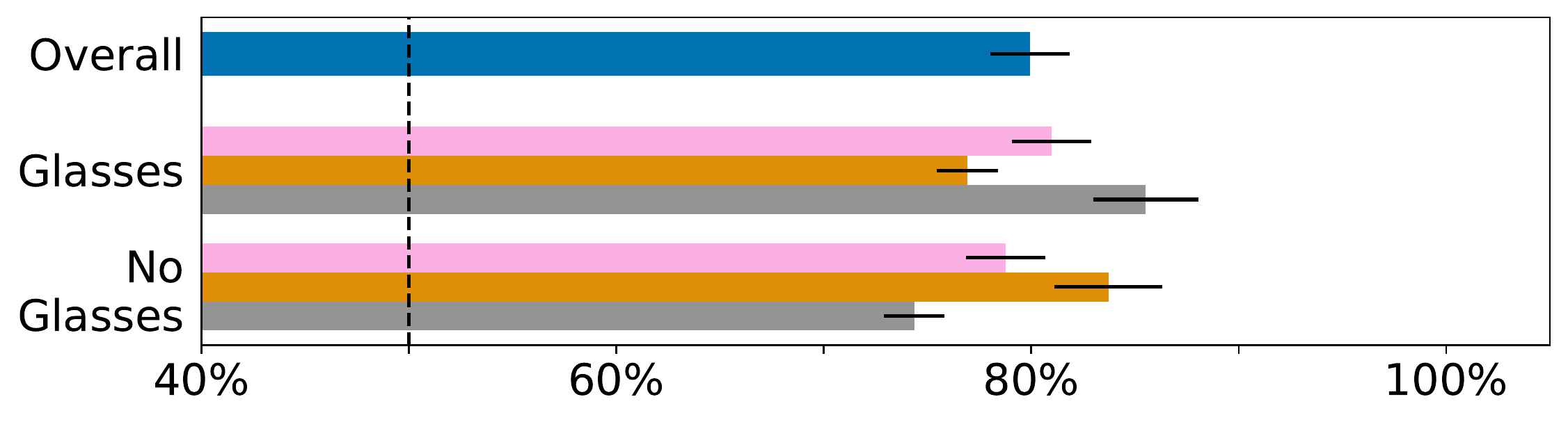}
         \caption{Eyeglasses (FFHQ)}
     \end{subfigure}
     \hfill
     \begin{subfigure}[b]{0.48\textwidth}
         \centering
         \includegraphics[width=\textwidth]{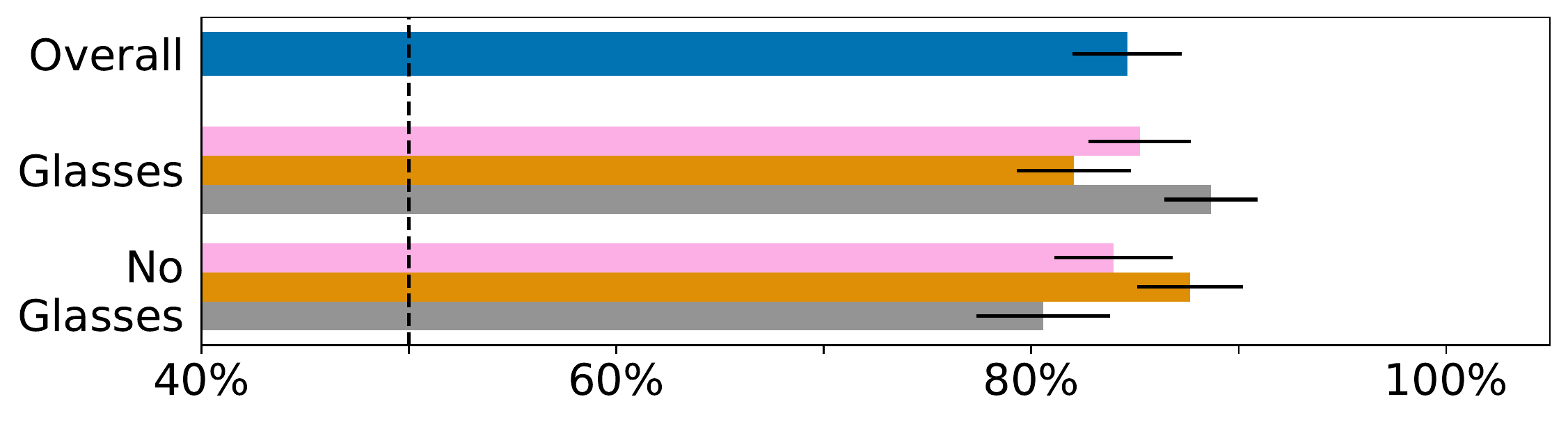}
         \caption{Eyeglasses (CelebAHQ)}
     \end{subfigure}

     \par\medskip
     \begin{subfigure}[b]{0.48\textwidth}
         \centering
         \includegraphics[width=\textwidth]{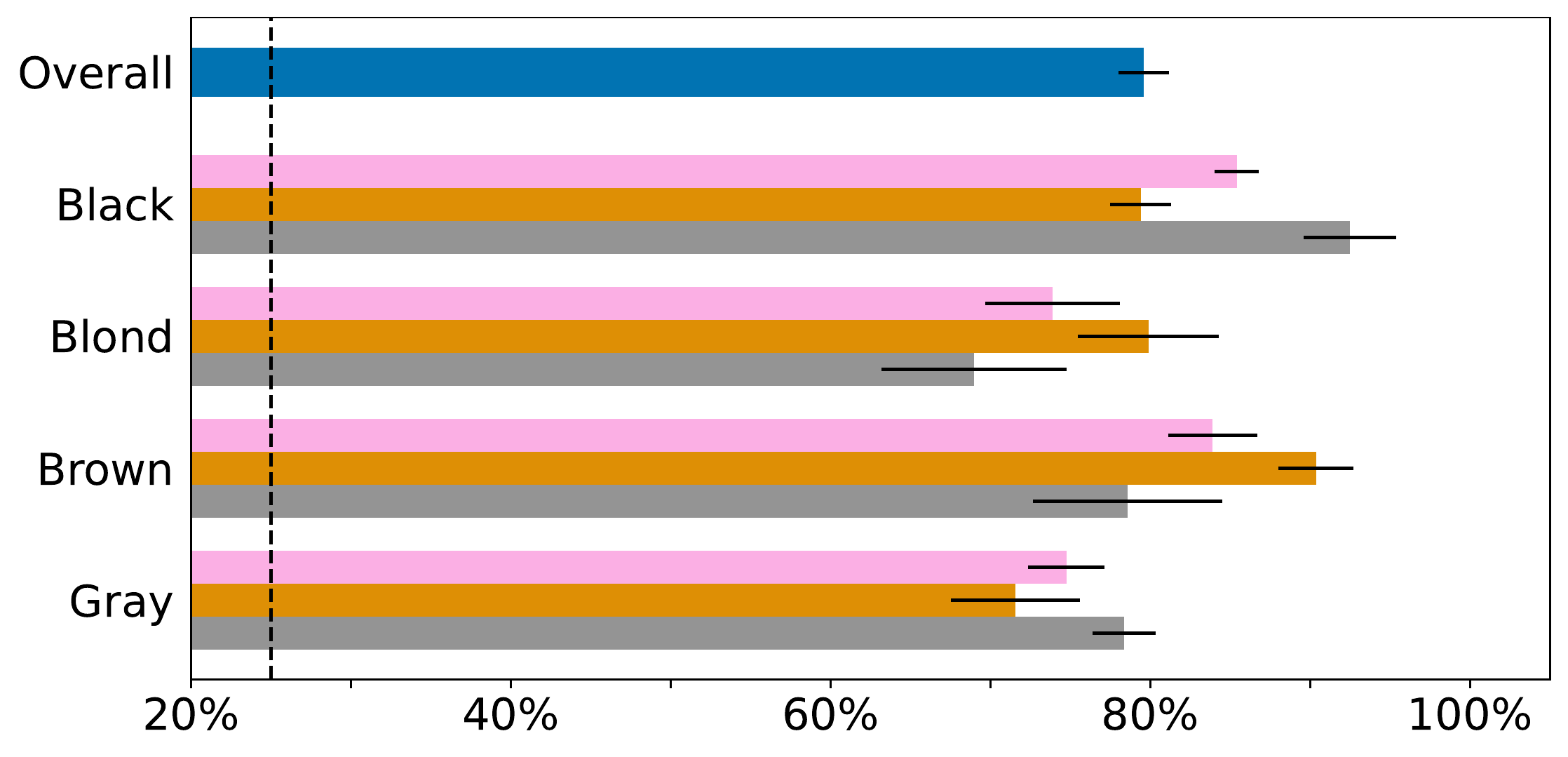}
         \caption{Hair Color (FFHQ)}
     \end{subfigure}
     \hfill
     \begin{subfigure}[b]{0.48\textwidth}
         \centering
         \includegraphics[width=\textwidth]{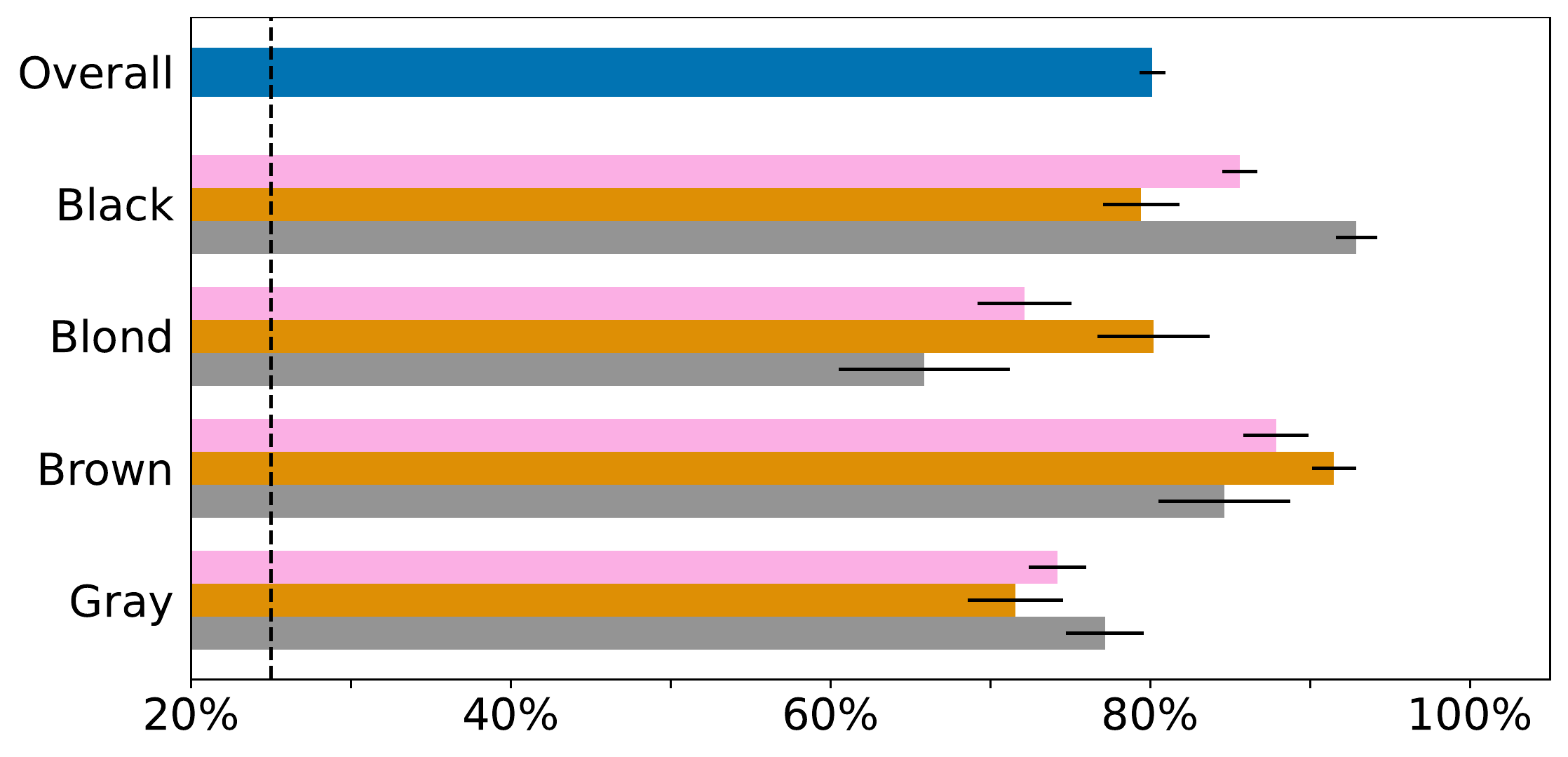}
         \caption{Hair Color (CelebAHQ)}
     \end{subfigure}

     \par\medskip
     \begin{subfigure}[b]{0.48\textwidth}
         \centering
         \includegraphics[width=\textwidth]{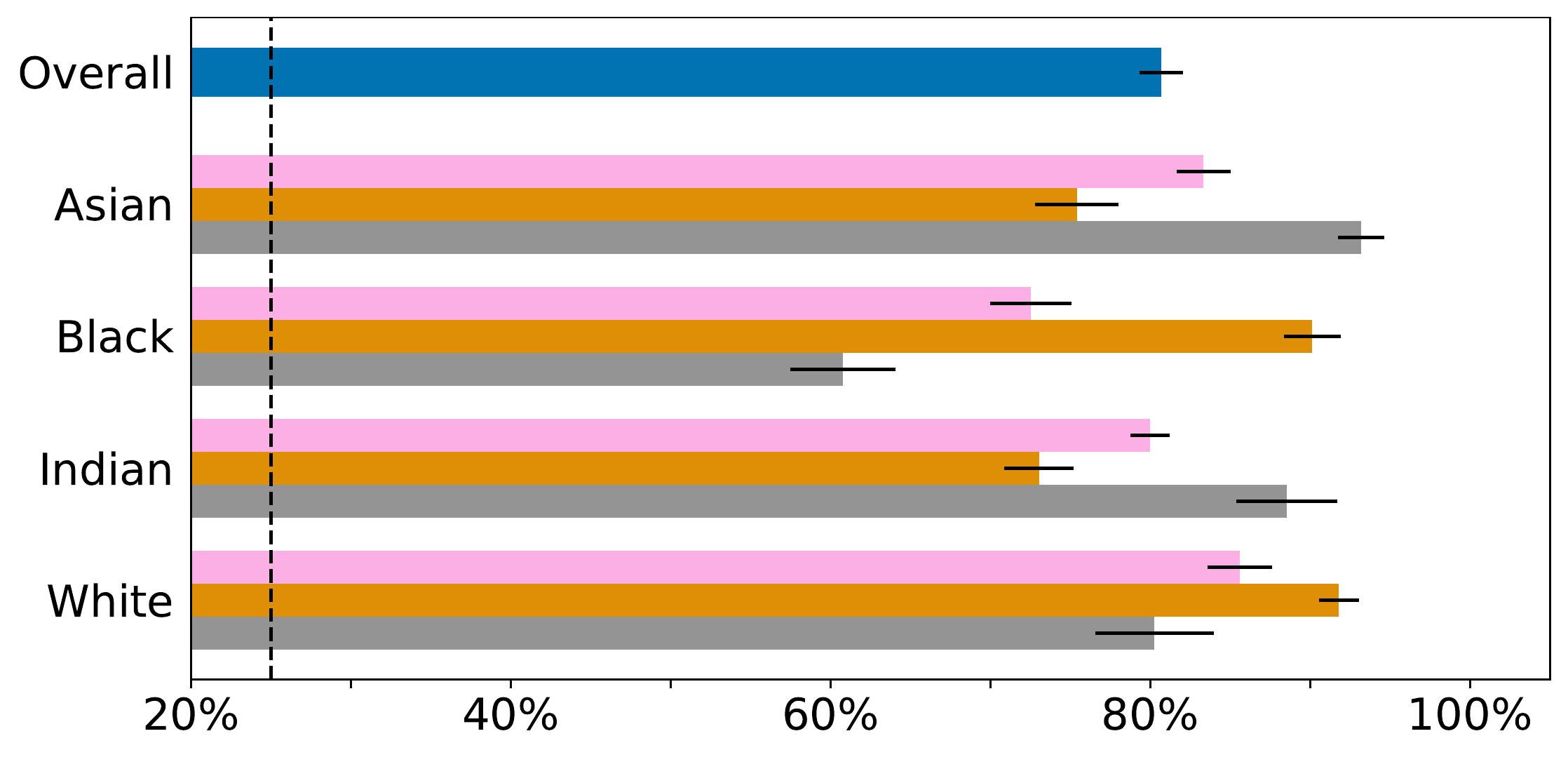}
         \caption{Racial Appearance (FFHQ)}
     \end{subfigure}
     \hfill
     \begin{subfigure}[b]{0.48\textwidth}
         \centering
         \includegraphics[width=\textwidth]{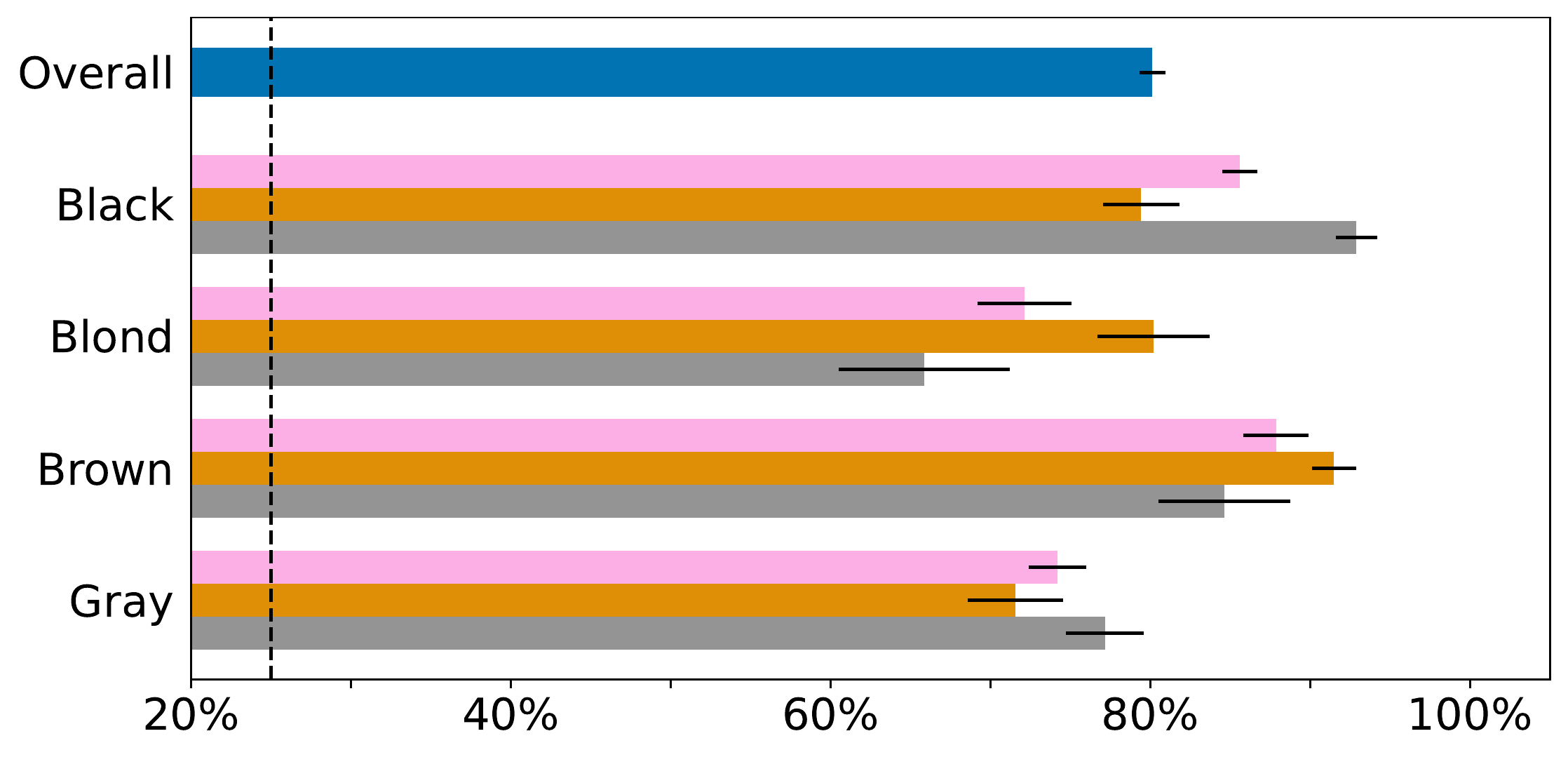}
         \caption{Racial Appearance (CelebAHQ)}
     \end{subfigure}

    \caption{Evaluation results for CAIA performed on ResNet-18 models to infer four different target attributes and \textbf{1000} identities. The black horizontal lines denote the standard deviation over nine runs. We further state random guessing (dashed line) for comparison.}
\end{figure*}
\clearpage

\subsection{ResNet-101 - CelebA}
\begin{figure*}[h!]
\centering
     \begin{subfigure}[c]{\textwidth}
         \centering
         \includegraphics[width=0.75\textwidth]{images/barplots/legend_small.pdf}
     \end{subfigure}
     
     \begin{subfigure}[b]{0.48\textwidth}
        \centering
         \includegraphics[width=\textwidth]{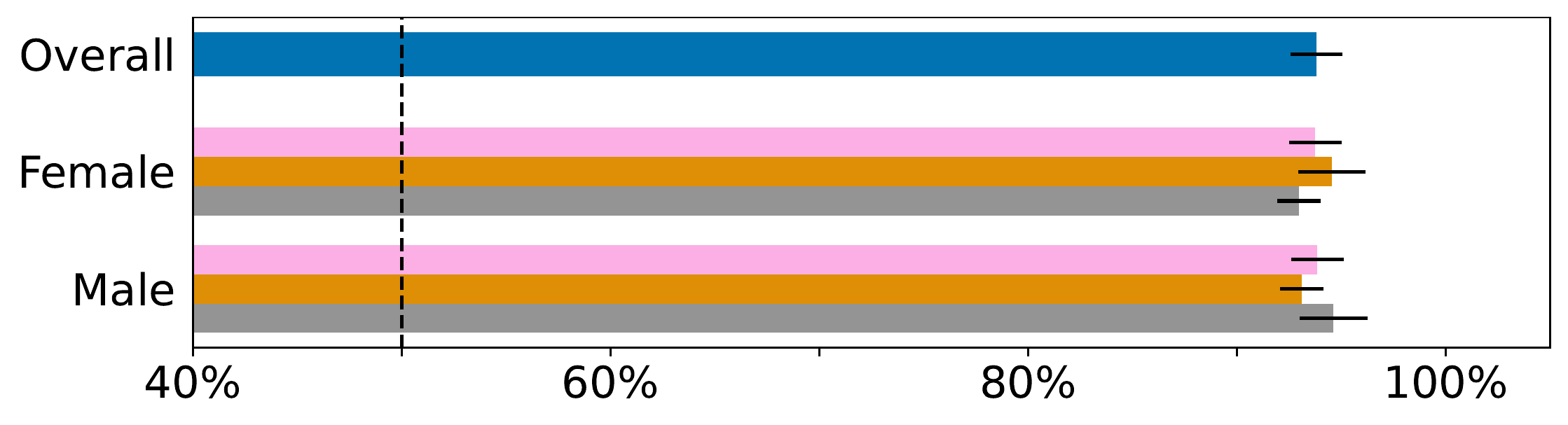}
         \caption{Gender (FFHQ)}
     \end{subfigure}
     \hfill
     \begin{subfigure}[b]{0.48\textwidth}
        \centering
         \includegraphics[width=\textwidth]{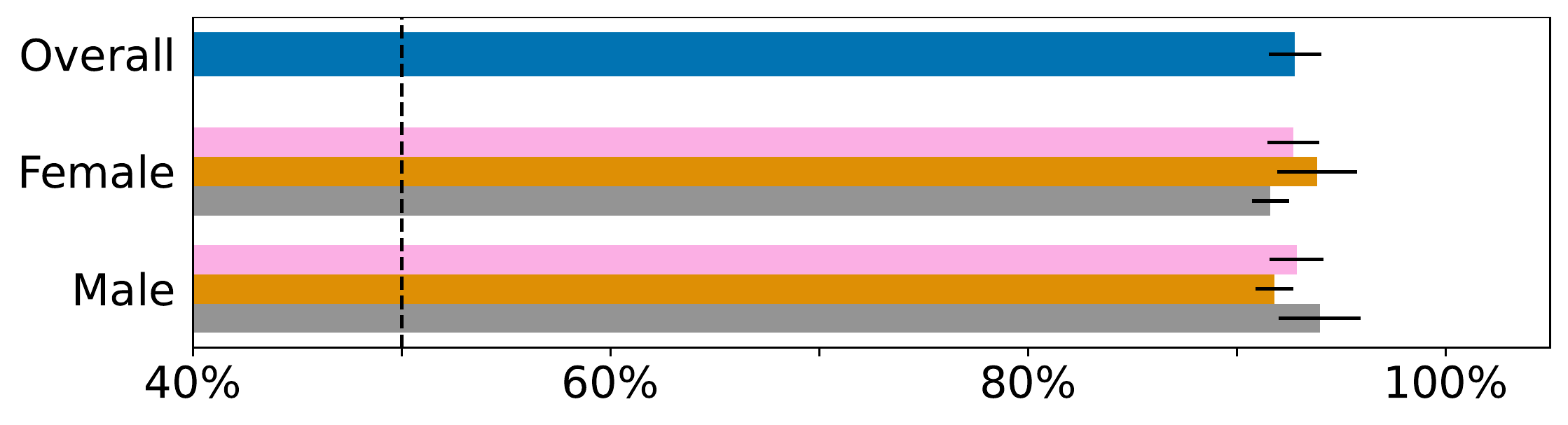}
         \caption{Gender (CelebAHQ)}
     \end{subfigure}

     \par\medskip
     \begin{subfigure}[b]{0.48\textwidth}
         \centering
         \includegraphics[width=\textwidth]{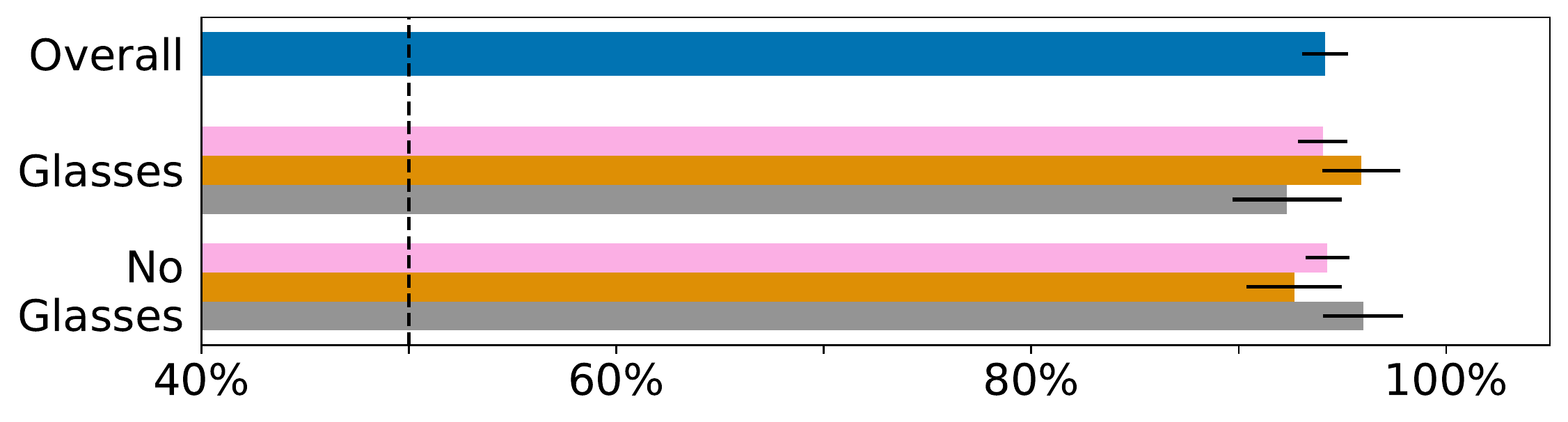}
         \caption{Eyeglasses (FFHQ)}
     \end{subfigure}
     \hfill
     \begin{subfigure}[b]{0.48\textwidth}
         \centering
         \includegraphics[width=\textwidth]{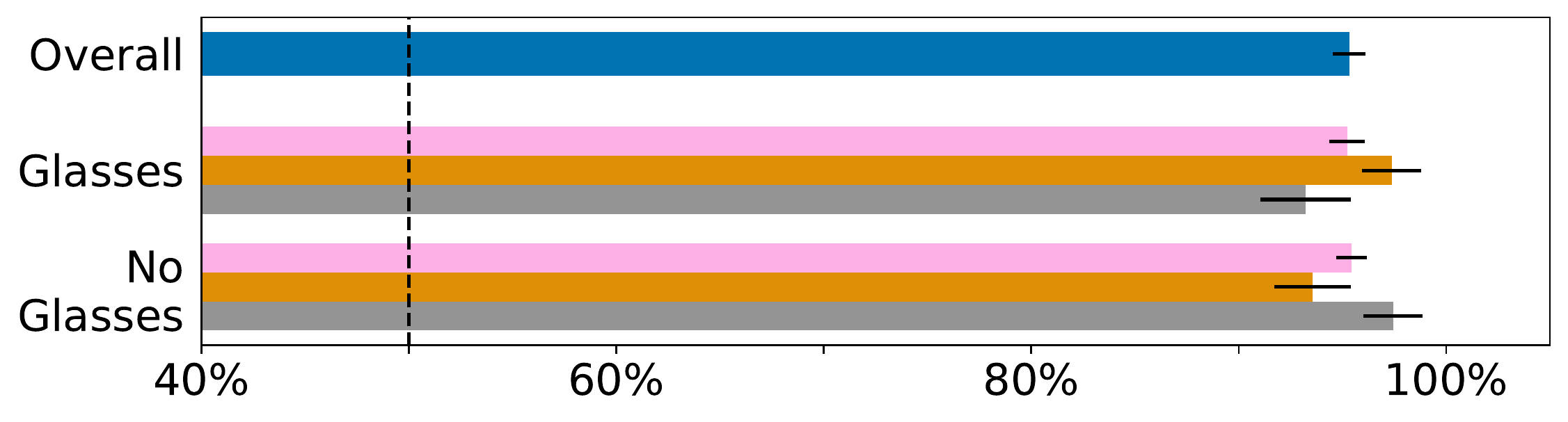}
         \caption{Eyeglasses (CelebAHQ)}
     \end{subfigure}

     \par\medskip
     \begin{subfigure}[b]{0.48\textwidth}
         \centering
         \includegraphics[width=\textwidth]{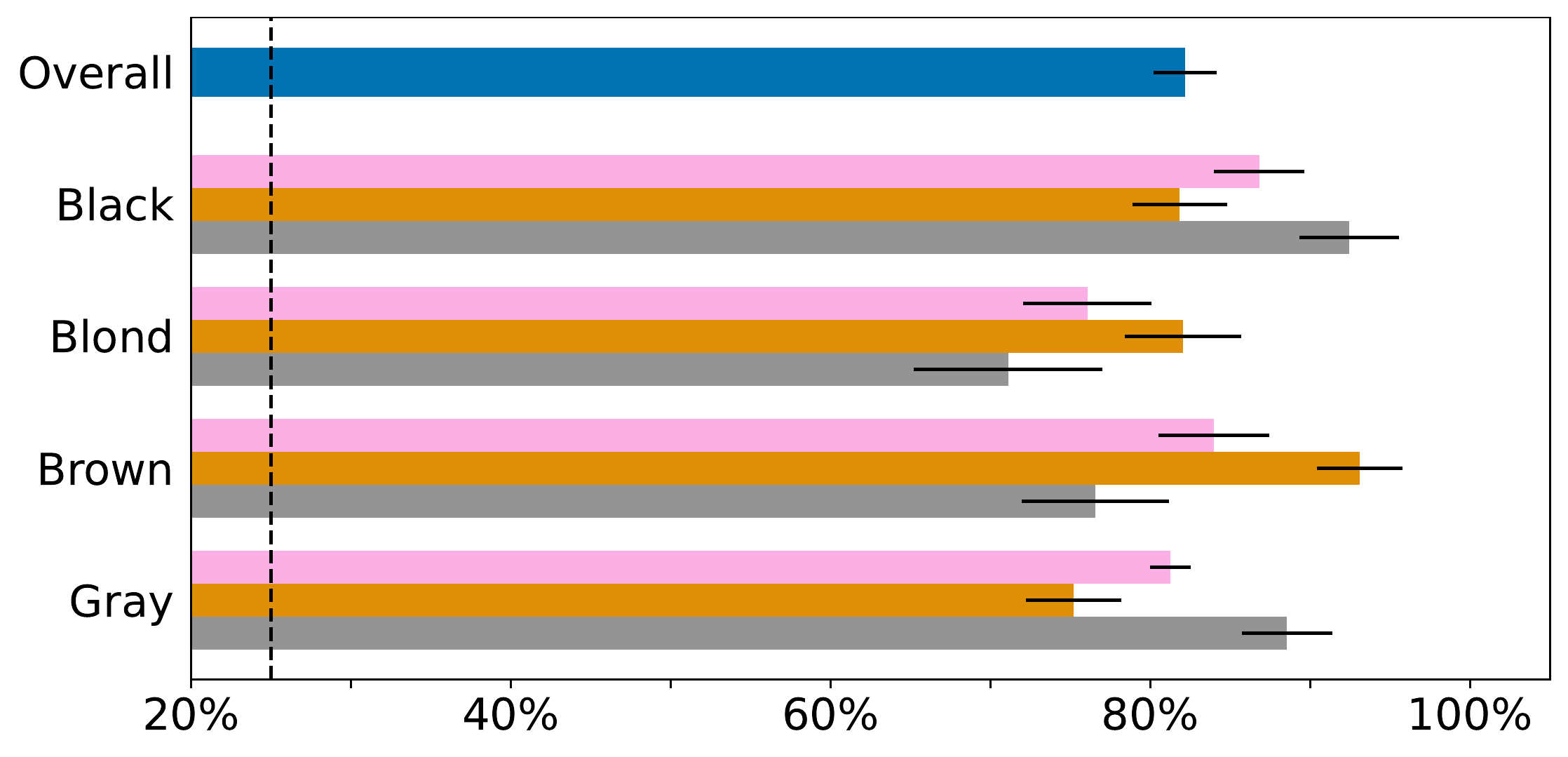}
         \caption{Hair Color (FFHQ)}
     \end{subfigure}
     \hfill
     \begin{subfigure}[b]{0.48\textwidth}
         \centering
         \includegraphics[width=\textwidth]{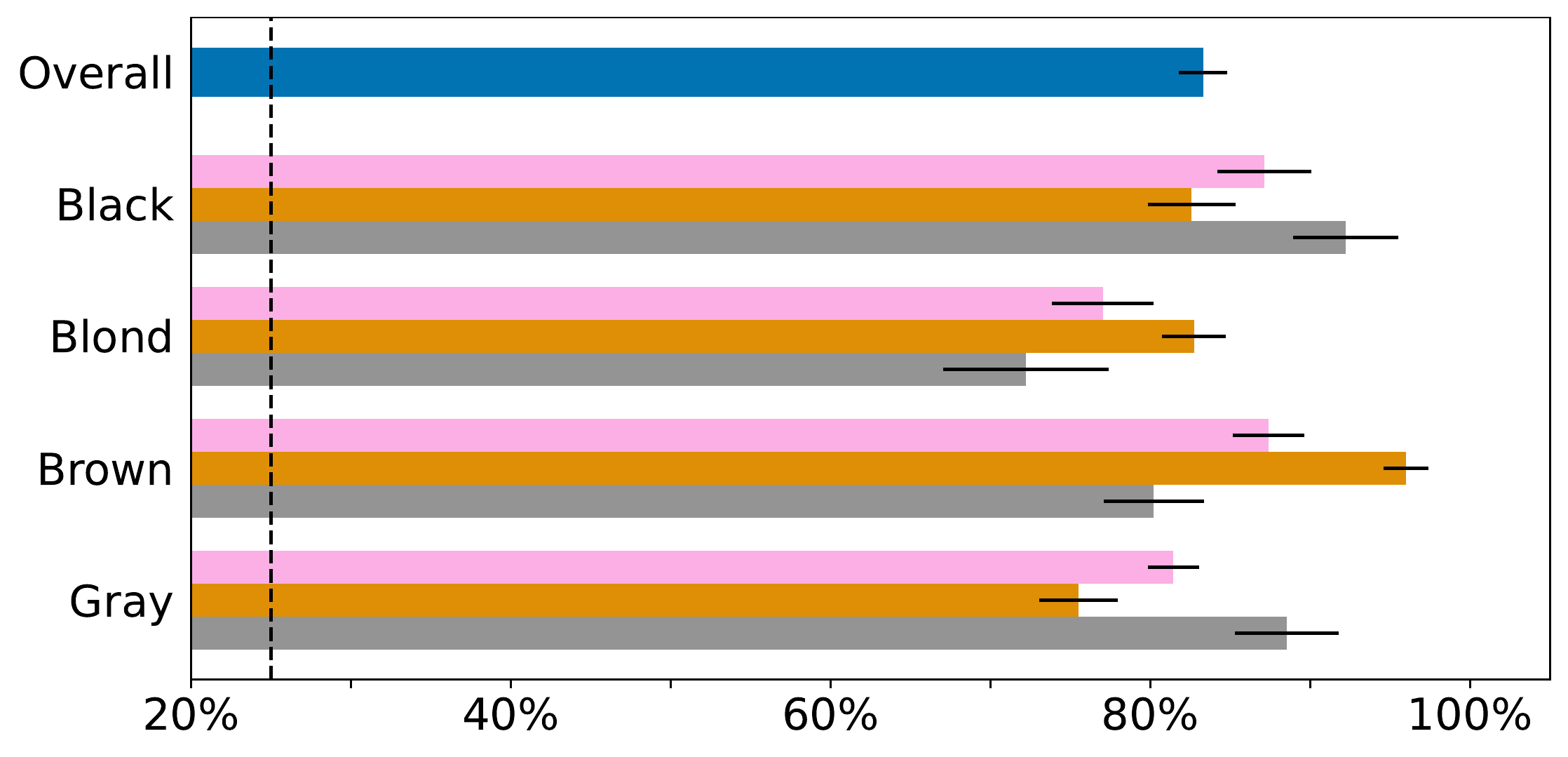}
         \caption{Hair Color (CelebAHQ)}
     \end{subfigure}

     \par\medskip
     \begin{subfigure}[b]{0.48\textwidth}
         \centering
         \includegraphics[width=\textwidth]{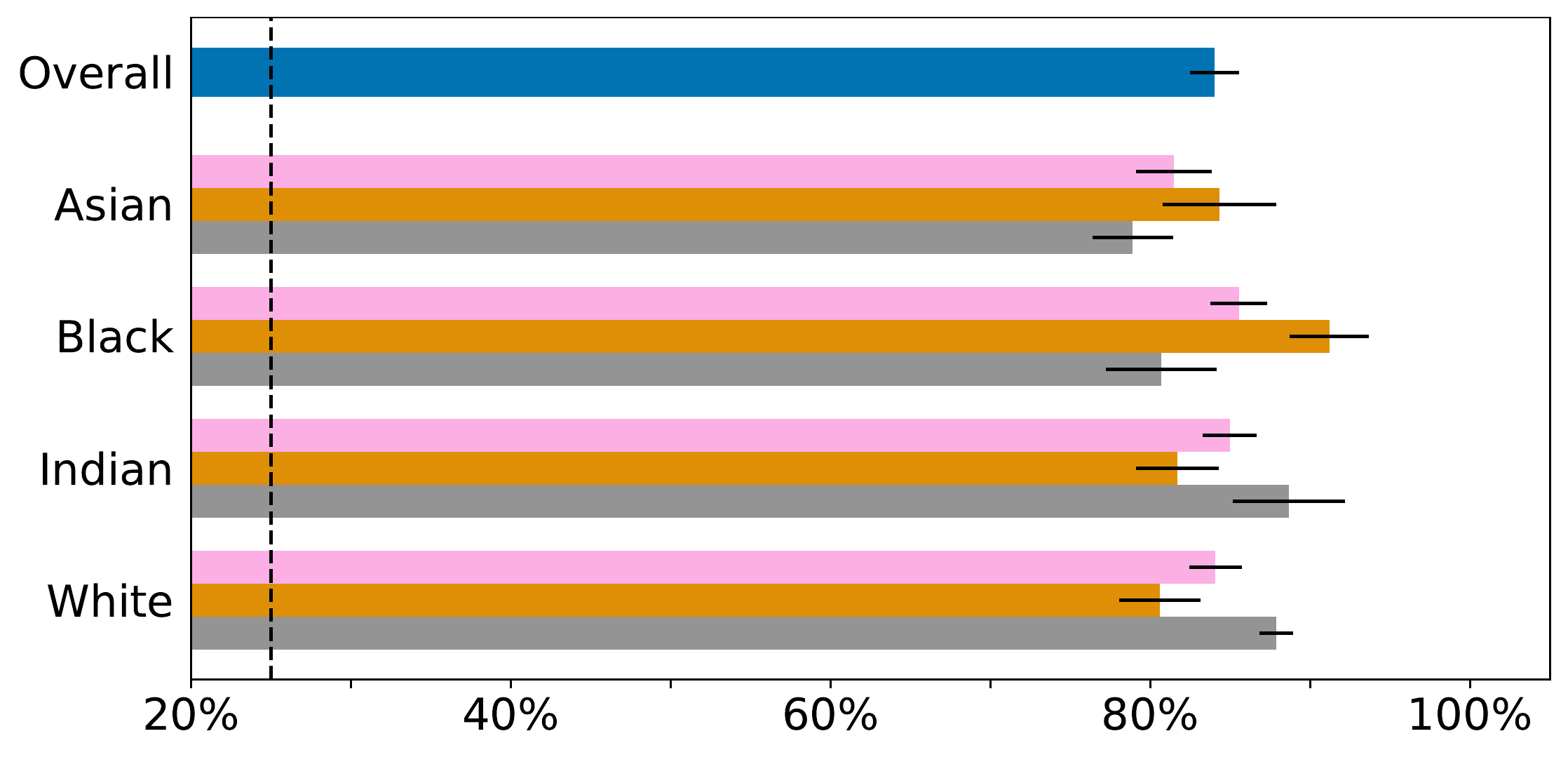}
         \caption{Racial Appearance (FFHQ)}
     \end{subfigure}
     \hfill
     \begin{subfigure}[b]{0.48\textwidth}
         \centering
         \includegraphics[width=\textwidth]{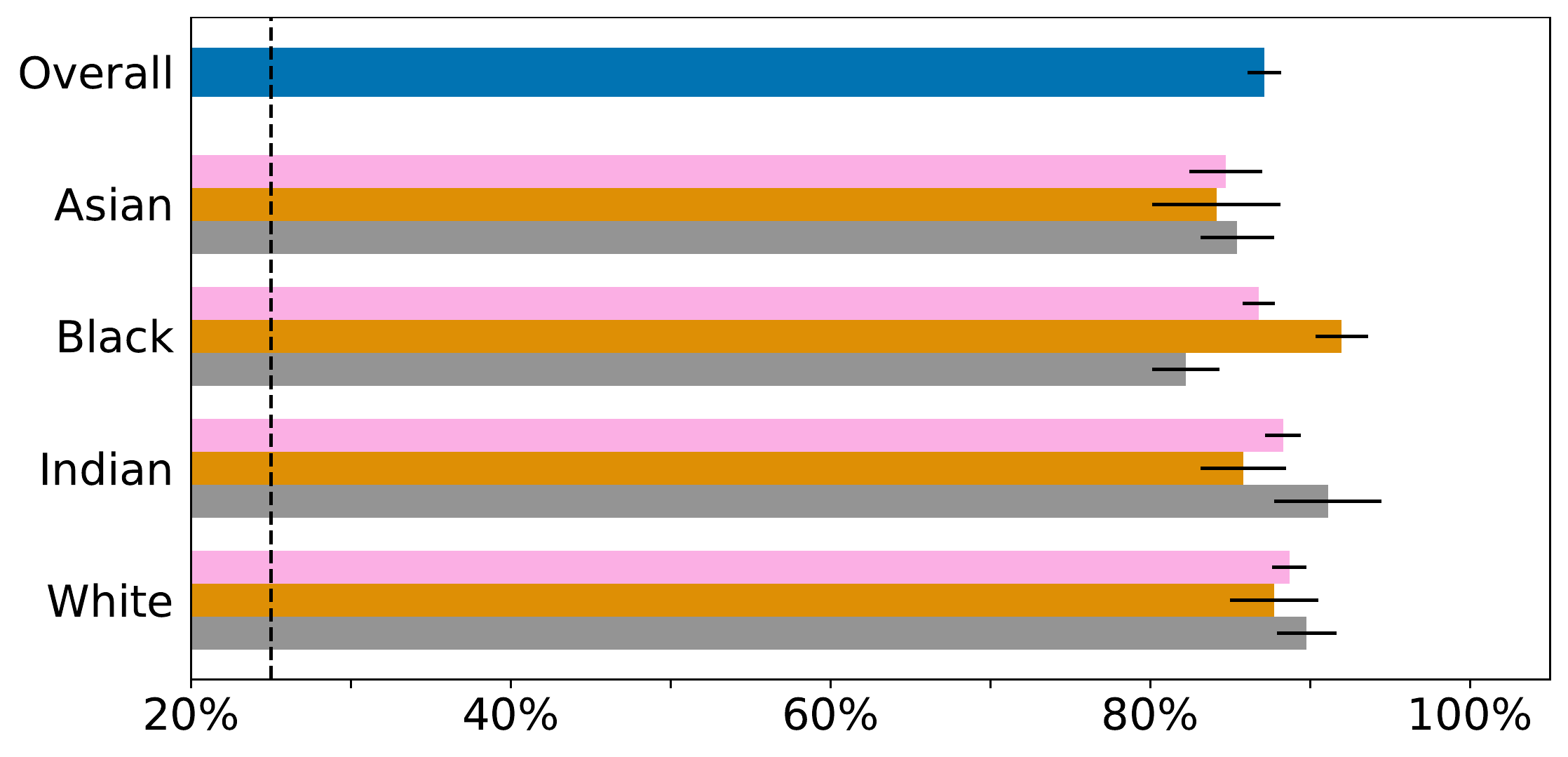}
         \caption{Racial Appearance (CelebAHQ)}
     \end{subfigure}
    \caption{Evaluation results for CAIA performed on ResNet-101 models to infer four different target attributes. The black horizontal lines denote the standard deviation over nine runs. We further state random guessing (dashed line) for comparison.}
\end{figure*}
\clearpage

\subsection{ResNet-101 - CelebA (1000 identities)}
\begin{figure*}[h!]
\centering
     \begin{subfigure}[c]{\textwidth}
         \centering
         \includegraphics[width=0.75\textwidth]{images/barplots/legend_small.pdf}
     \end{subfigure}
     
     \begin{subfigure}[b]{0.48\textwidth}
        \centering
         \includegraphics[width=\textwidth]{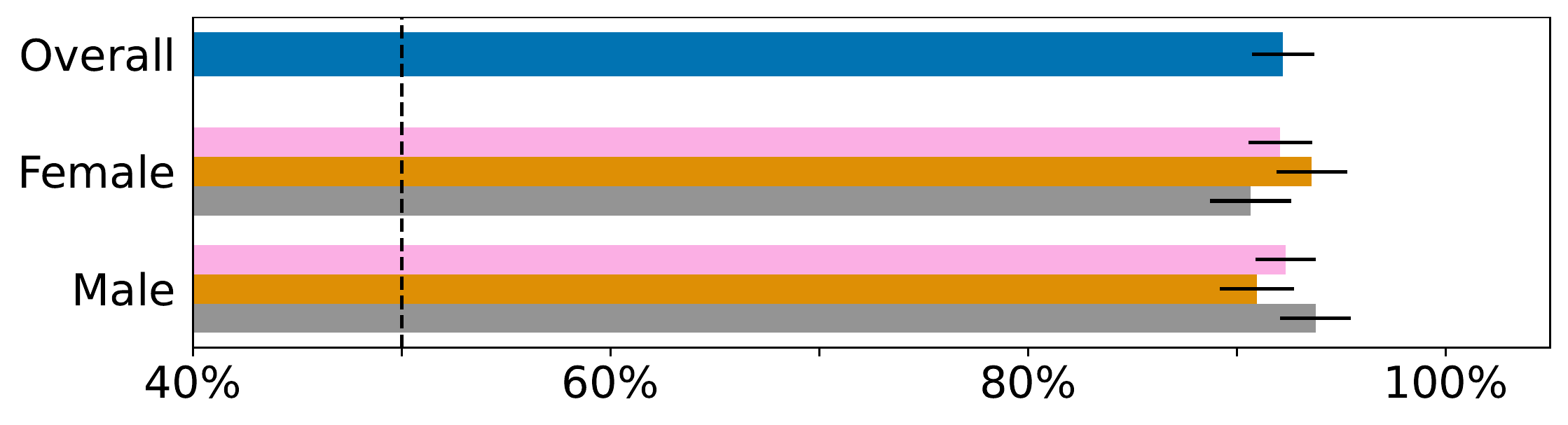}
         \caption{Gender (FFHQ)}
     \end{subfigure}
     \hfill
     \begin{subfigure}[b]{0.48\textwidth}
        \centering
         \includegraphics[width=\textwidth]{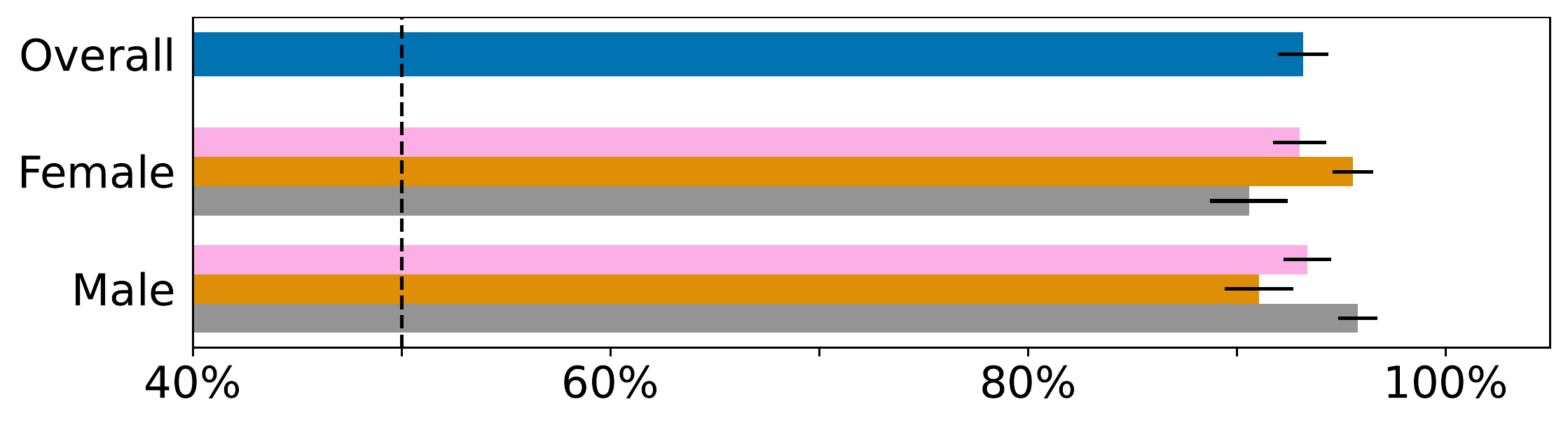}
         \caption{Gender (CelebAHQ)}
     \end{subfigure}

     \par\medskip
     \begin{subfigure}[b]{0.48\textwidth}
         \centering
         \includegraphics[width=\textwidth]{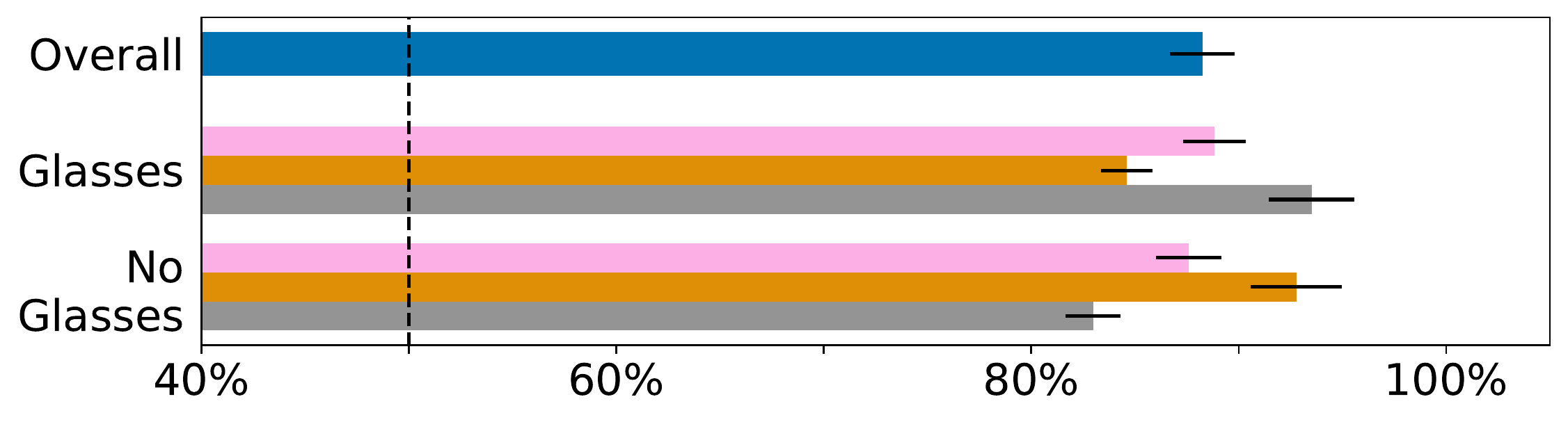}
         \caption{Eyeglasses (FFHQ)}
     \end{subfigure}
     \hfill
     \begin{subfigure}[b]{0.48\textwidth}
         \centering
         \includegraphics[width=\textwidth]{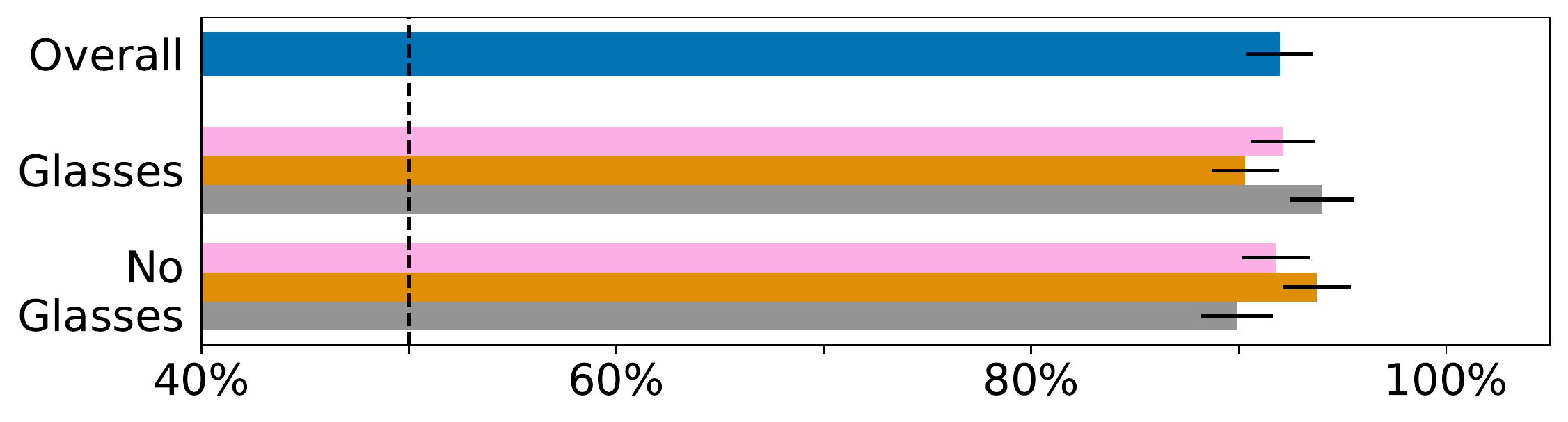}
         \caption{Eyeglasses (CelebAHQ)}
     \end{subfigure}

     \par\medskip
     \begin{subfigure}[b]{0.48\textwidth}
         \centering
         \includegraphics[width=\textwidth]{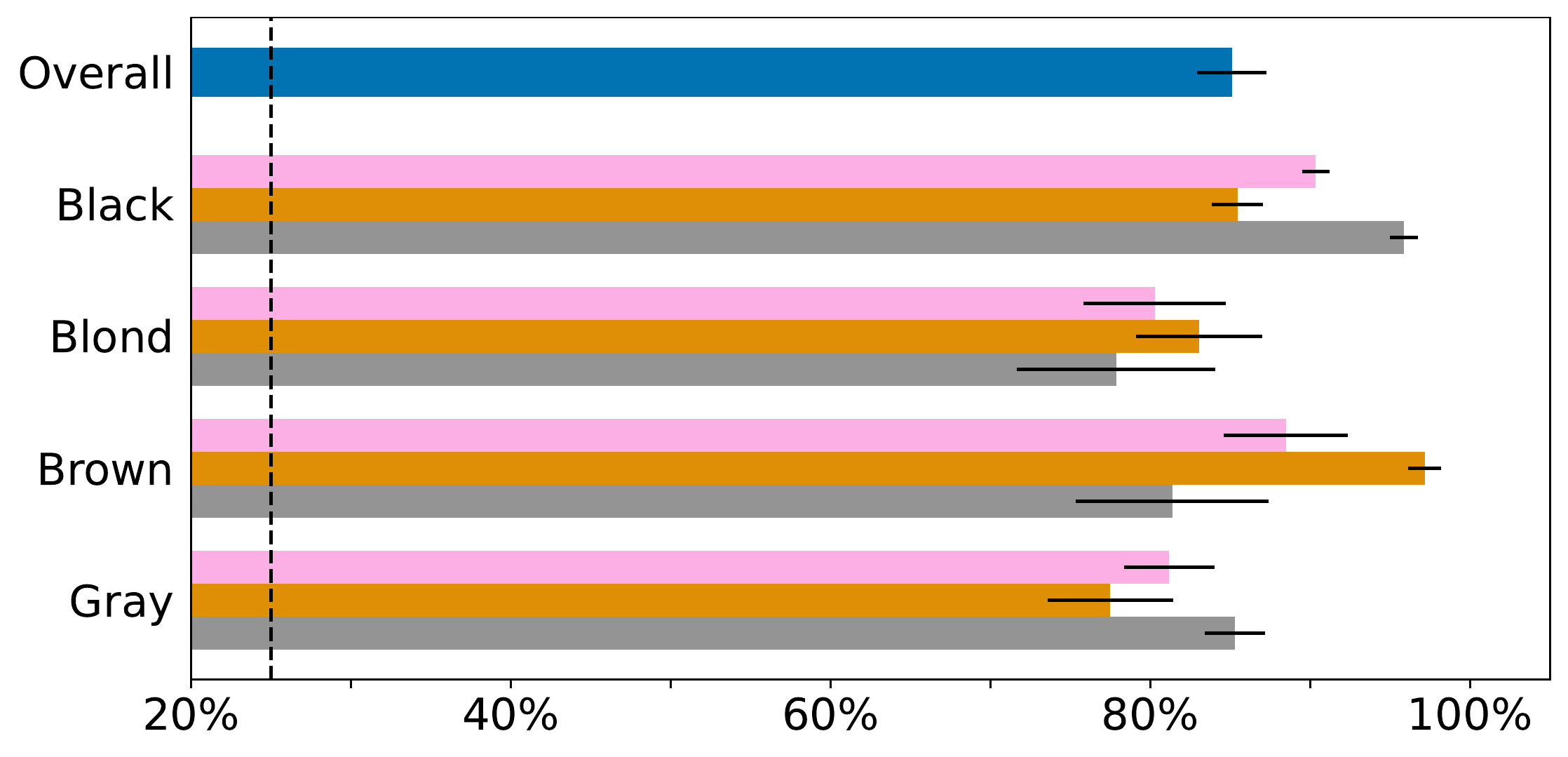}
         \caption{Hair Color (FFHQ)}
     \end{subfigure}
     \hfill
     \begin{subfigure}[b]{0.48\textwidth}
         \centering
         \includegraphics[width=\textwidth]{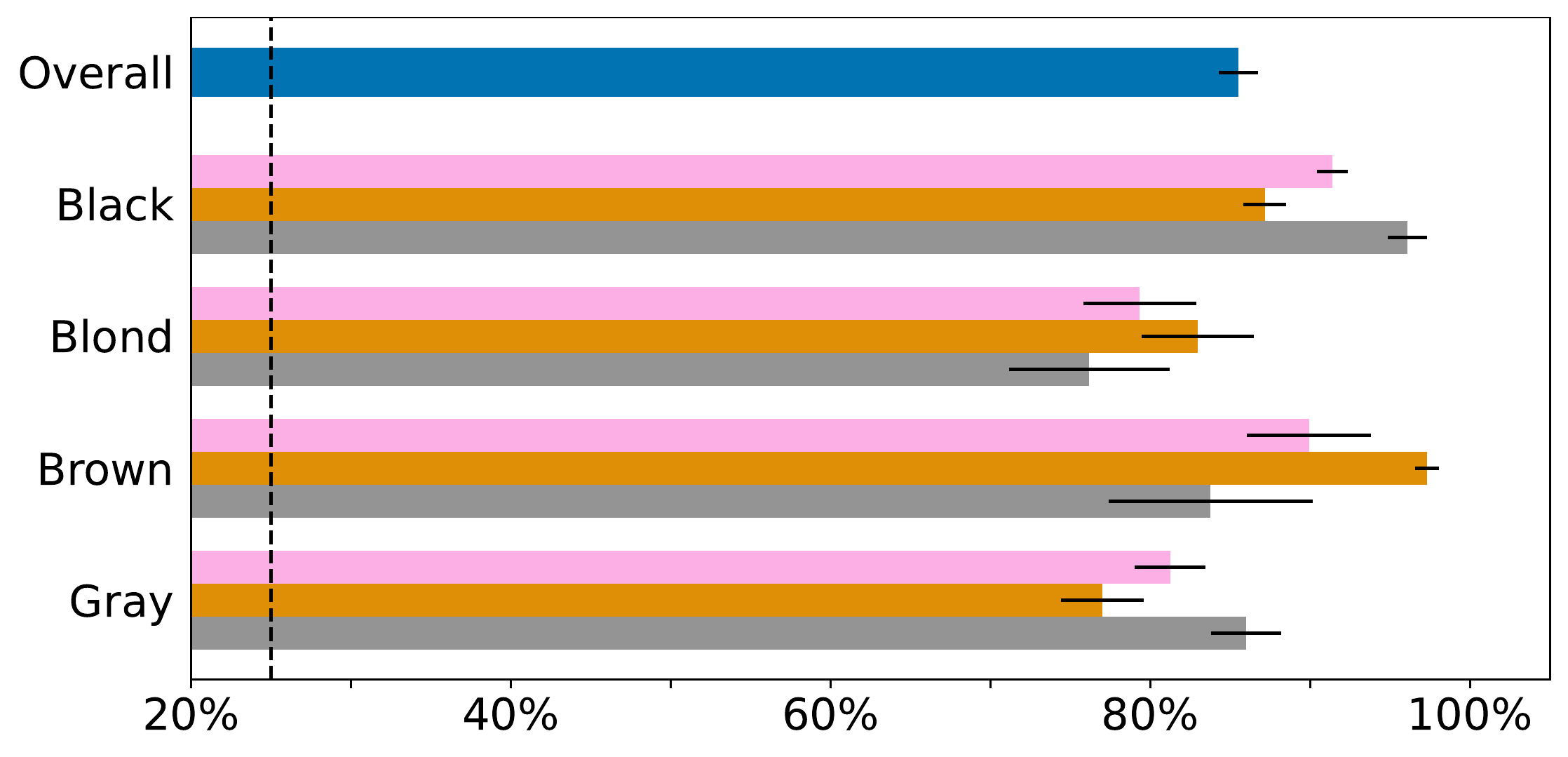}
         \caption{Hair Color (CelebAHQ)}
     \end{subfigure}

     \par\medskip
     \begin{subfigure}[b]{0.48\textwidth}
         \centering
         \includegraphics[width=\textwidth]{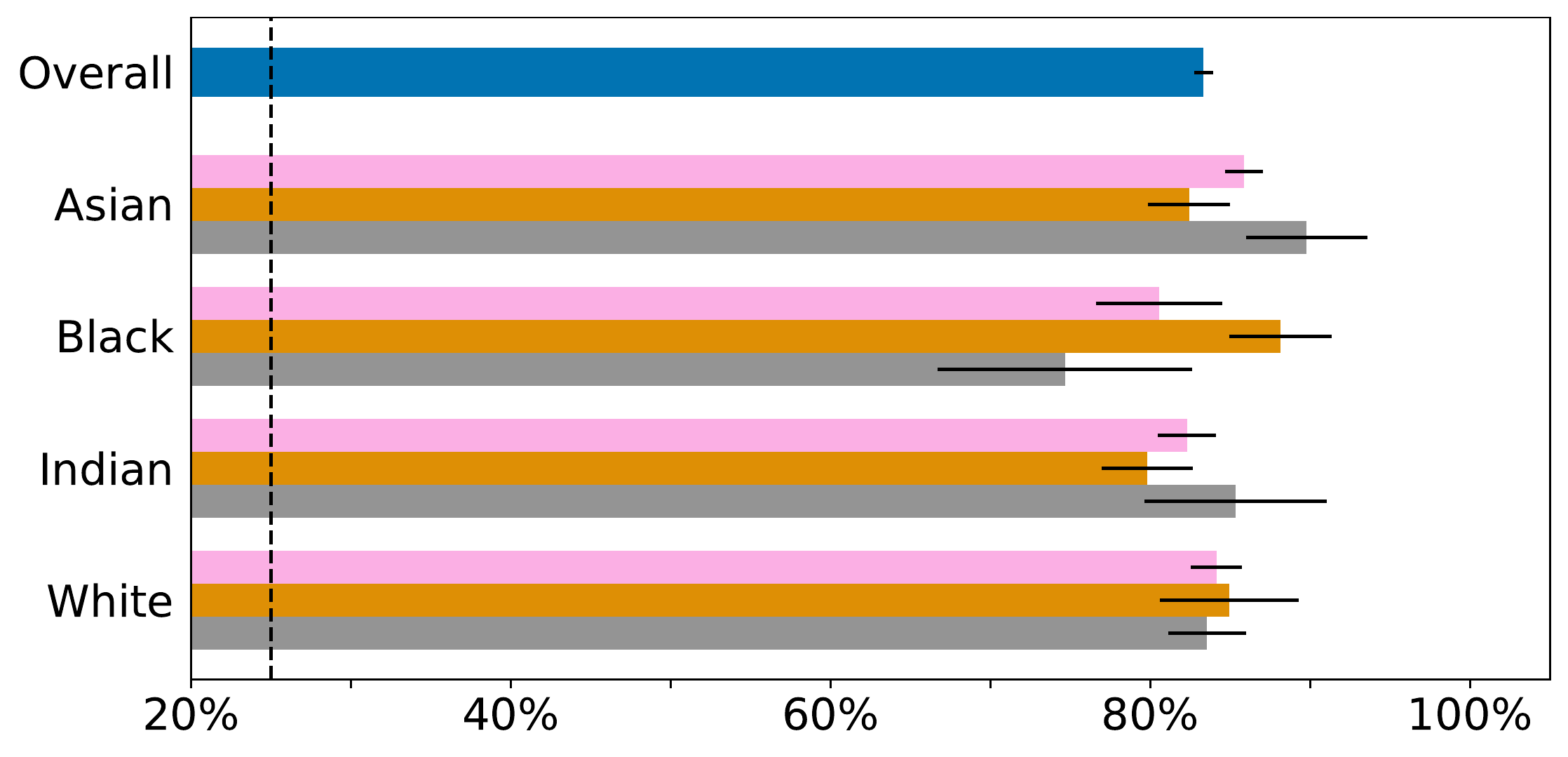}
         \caption{Racial Appearance (FFHQ)}
     \end{subfigure}
     \hfill
     \begin{subfigure}[b]{0.48\textwidth}
         \centering
         \includegraphics[width=\textwidth]{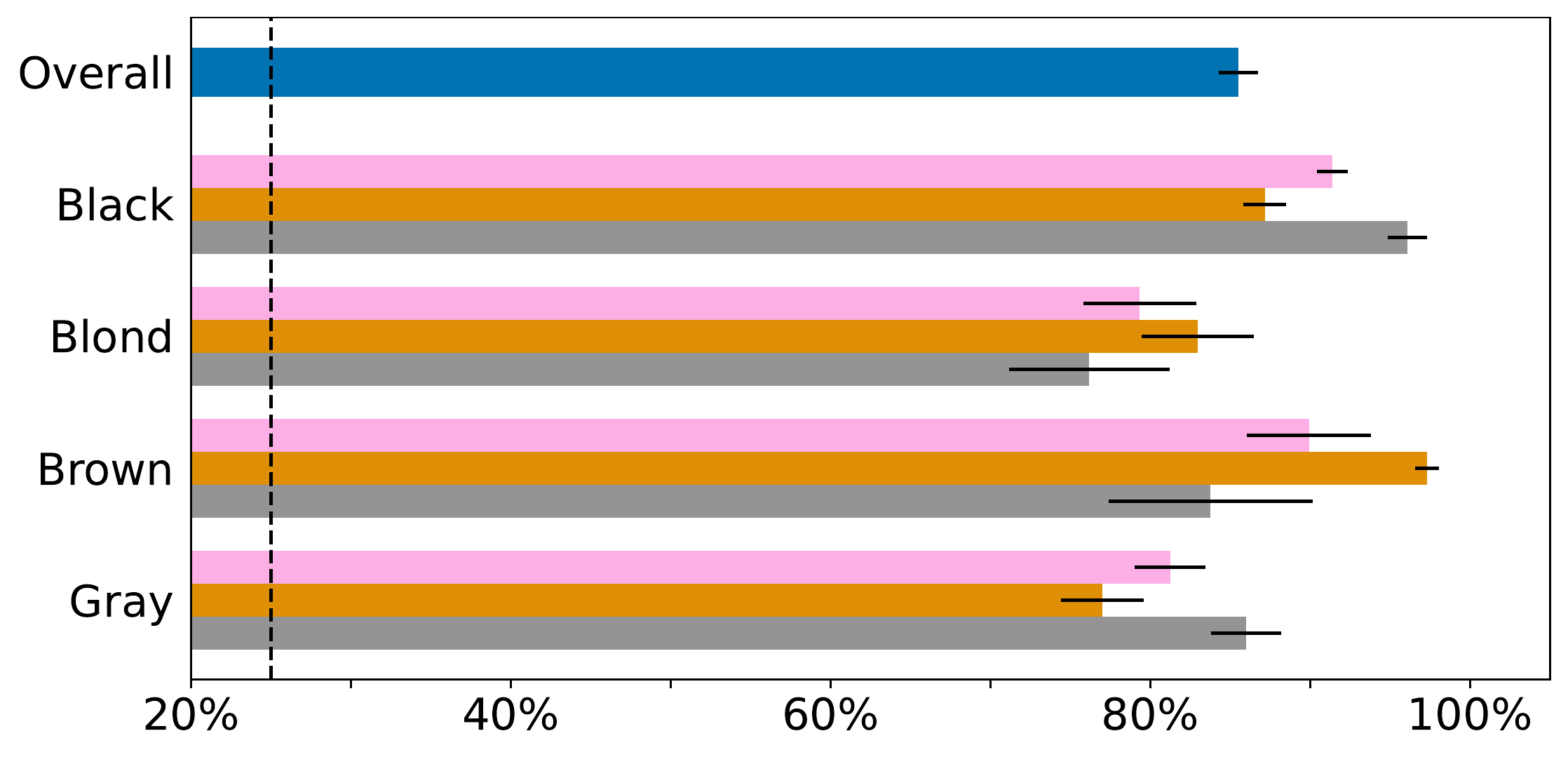}
         \caption{Racial Appearance (CelebAHQ)}
     \end{subfigure}
    \caption{Evaluation results for CAIA performed on ResNet-101 models to infer four different target attributes for \textbf{1000} identities. The black horizontal lines denote the standard deviation over nine runs. We further state random guessing (dashed line) for comparison.}
\end{figure*}
\clearpage

\subsection{ResNet-152 - CelebA}
\begin{figure*}[h!]
\centering
     \begin{subfigure}[c]{\textwidth}
         \centering
         \includegraphics[width=0.75\textwidth]{images/barplots/legend_small.pdf}
     \end{subfigure}
     
     \begin{subfigure}[b]{0.48\textwidth}
        \centering
         \includegraphics[width=\textwidth]{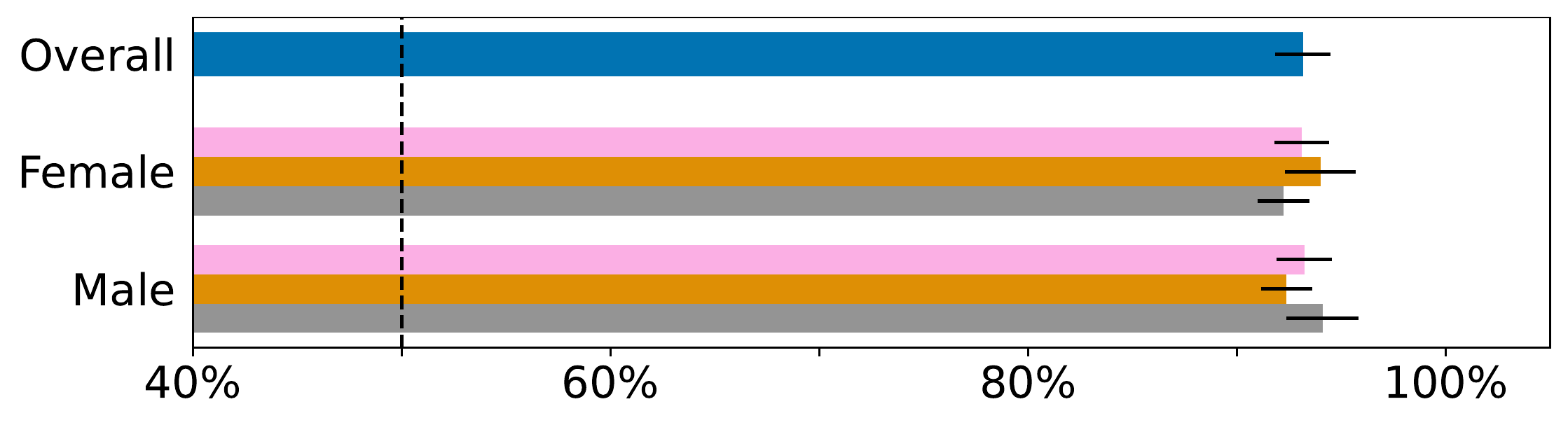}
         \caption{Gender (FFHQ)}
     \end{subfigure}
     \hfill
     \begin{subfigure}[b]{0.48\textwidth}
        \centering
         \includegraphics[width=\textwidth]{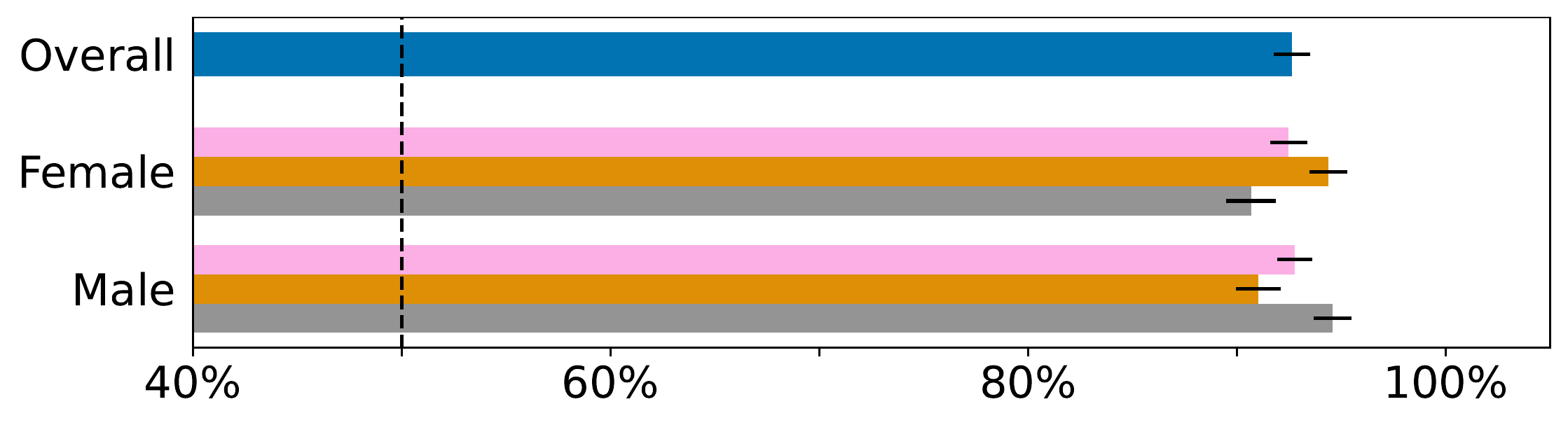}
         \caption{Gender (CelebAHQ)}
     \end{subfigure}

     \par\medskip
     \begin{subfigure}[b]{0.48\textwidth}
         \centering
         \includegraphics[width=\textwidth]{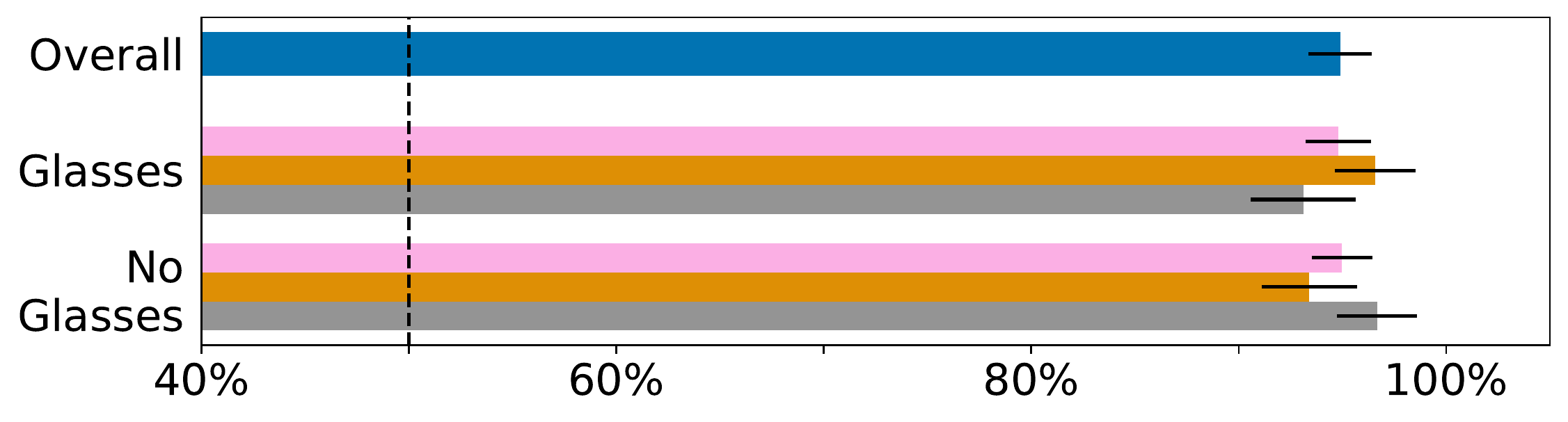}
         \caption{Eyeglasses (FFHQ)}
     \end{subfigure}
      \hfill
     \begin{subfigure}[b]{0.48\textwidth}
         \centering
         \includegraphics[width=\textwidth]{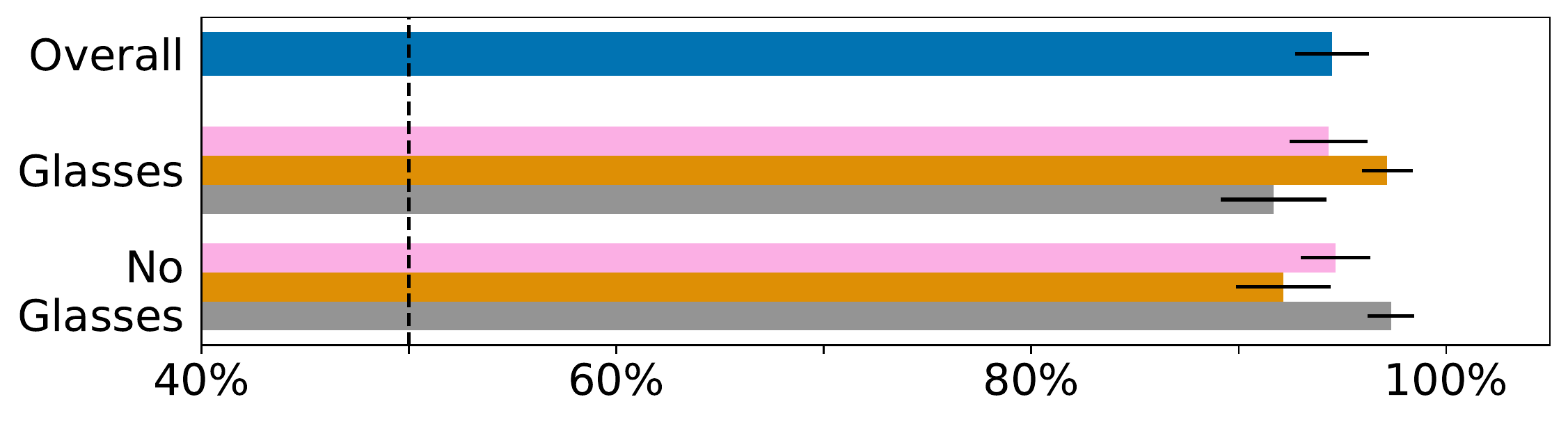}
         \caption{Eyeglasses (CelebAHQ)}
     \end{subfigure}
     
     \par\medskip
     \begin{subfigure}[b]{0.48\textwidth}
         \centering
         \includegraphics[width=\textwidth]{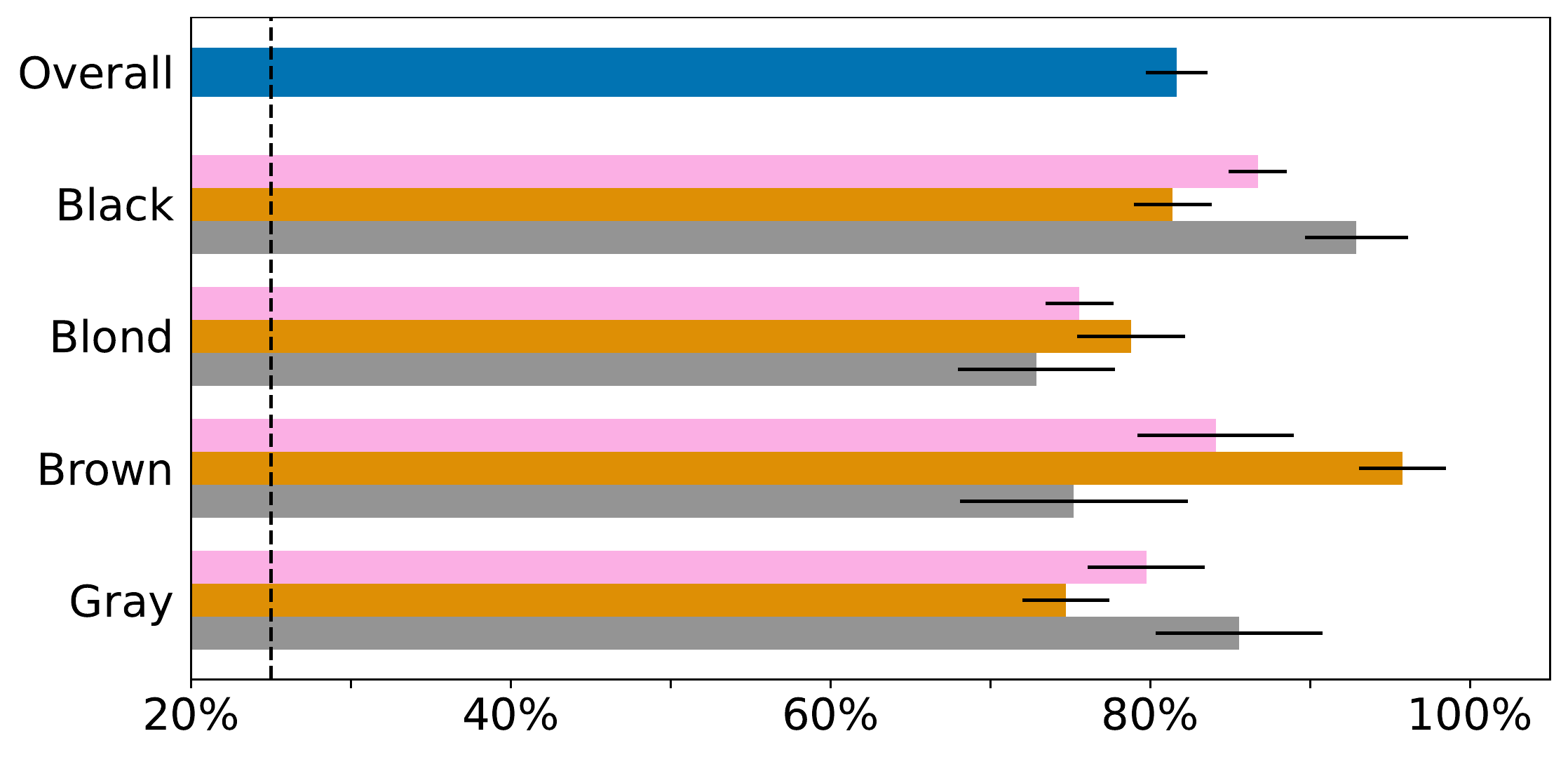}
         \caption{Hair Color (FFHQ)}
     \end{subfigure}
     \hfill
     \begin{subfigure}[b]{0.48\textwidth}
         \centering
         \includegraphics[width=\textwidth]{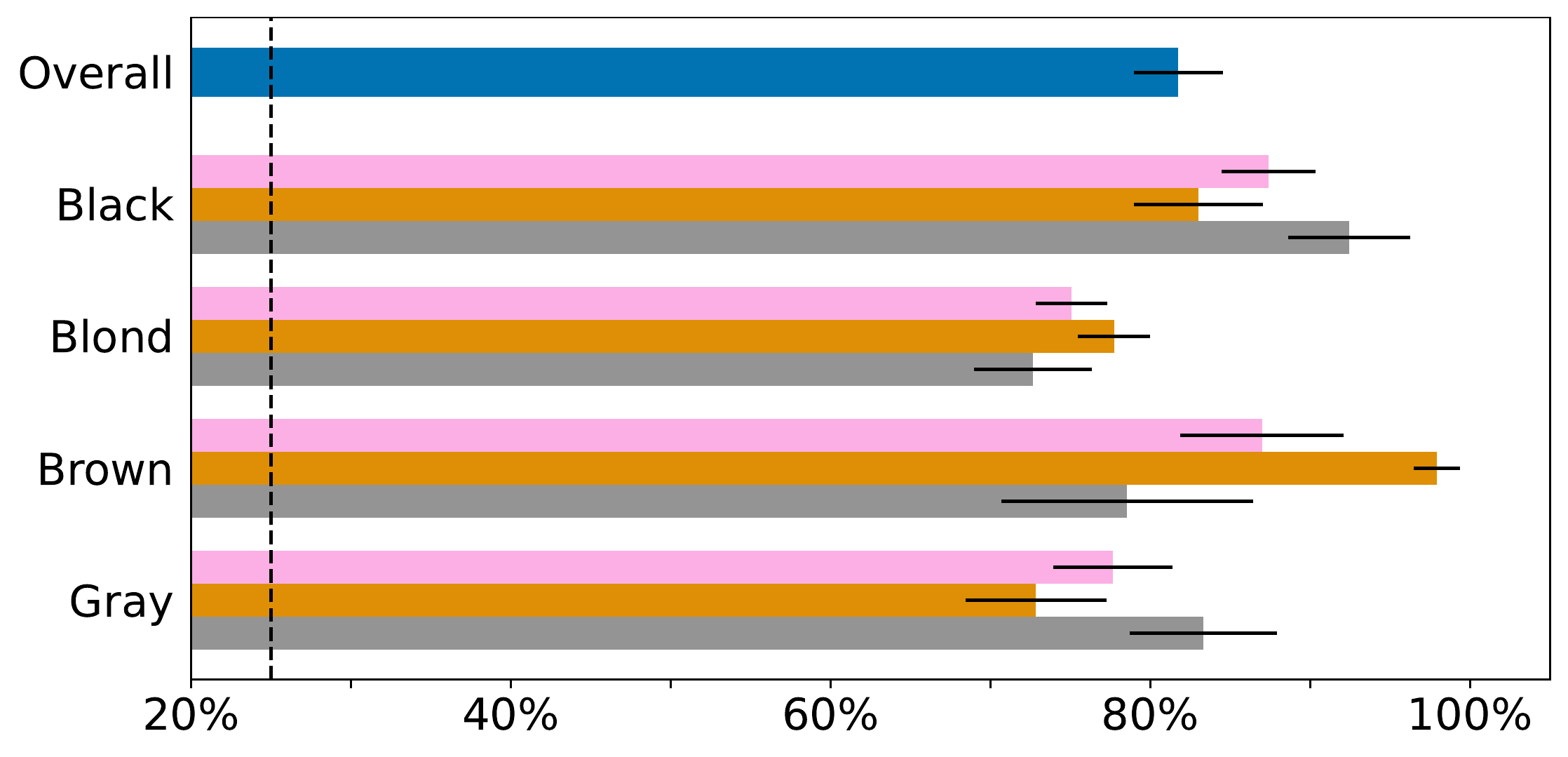}
         \caption{Hair Color (CelebAHQ)}
     \end{subfigure}

     \par\medskip

     \begin{subfigure}[b]{0.48\textwidth}
         \centering
         \includegraphics[width=\textwidth]{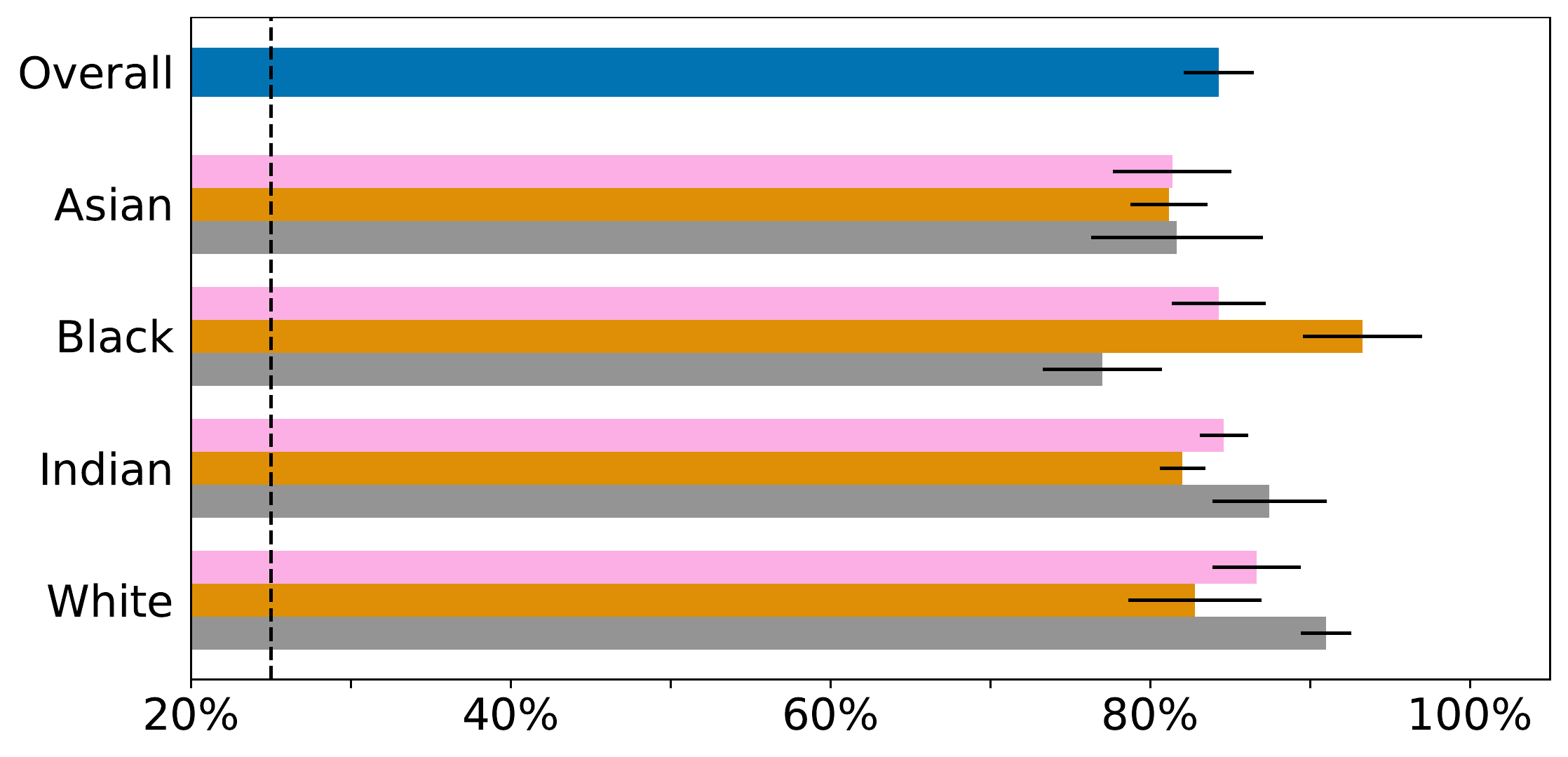}
         \caption{Racial Appearance (FFHQ)}
     \end{subfigure}
     \hfill
     \begin{subfigure}[b]{0.48\textwidth}
         \centering
         \includegraphics[width=\textwidth]{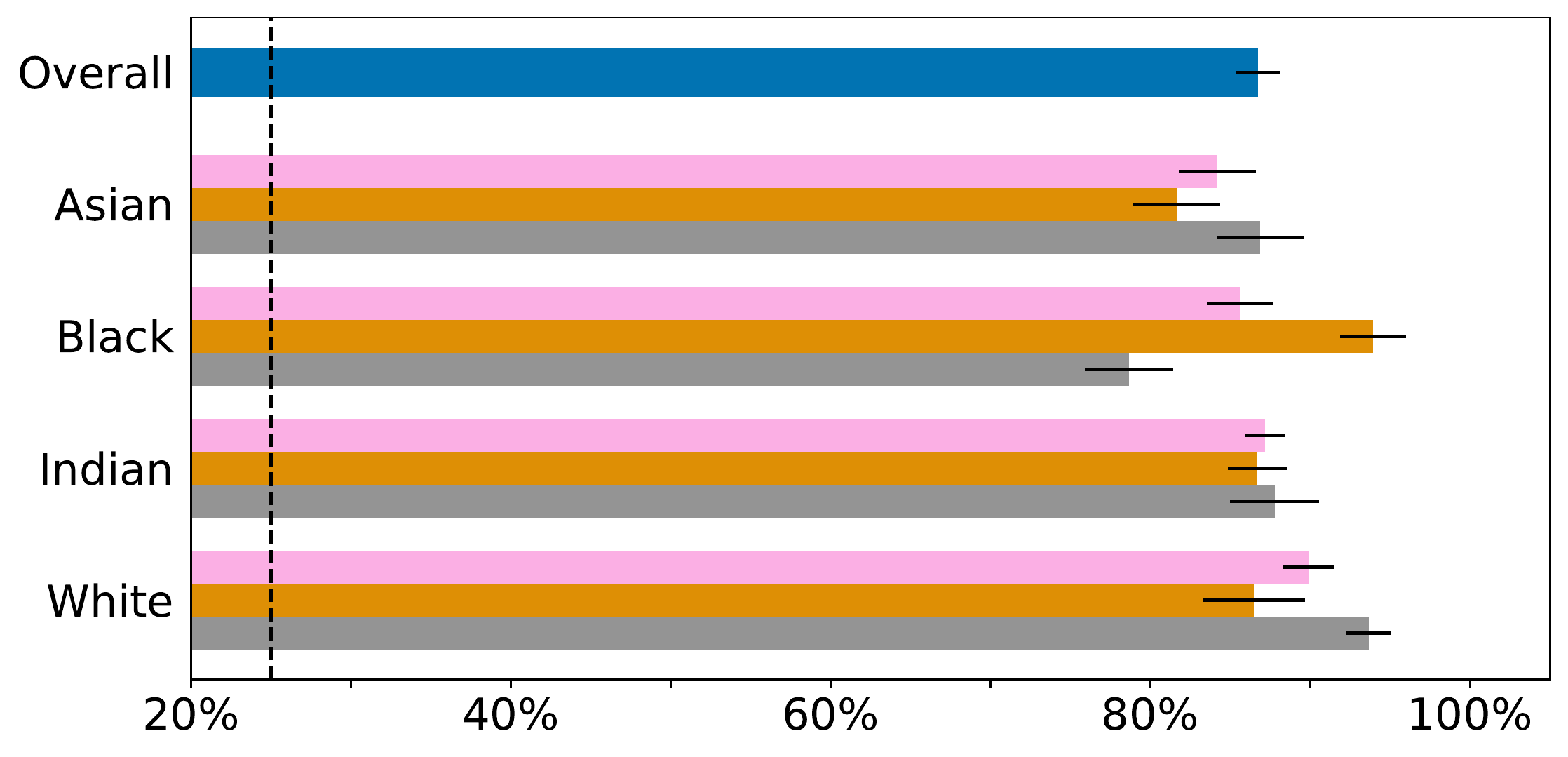}
         \caption{Racial Appearance (CelebAHQ)}
     \end{subfigure}

    \caption{Evaluation results for CAIA performed on ResNet-152 models to infer four different target attributes. The black horizontal lines denote the standard deviation over nine runs. We further state random guessing (dashed line) for comparison.}
\end{figure*}
\clearpage

\subsection{ResNet-152 - CelebA (1000 identities)}
\begin{figure*}[h!]
\centering
     \begin{subfigure}[c]{\textwidth}
         \centering
         \includegraphics[width=0.75\textwidth]{images/barplots/legend_small.pdf}
     \end{subfigure}
     
     \begin{subfigure}[b]{0.48\textwidth}
        \centering
         \includegraphics[width=\textwidth]{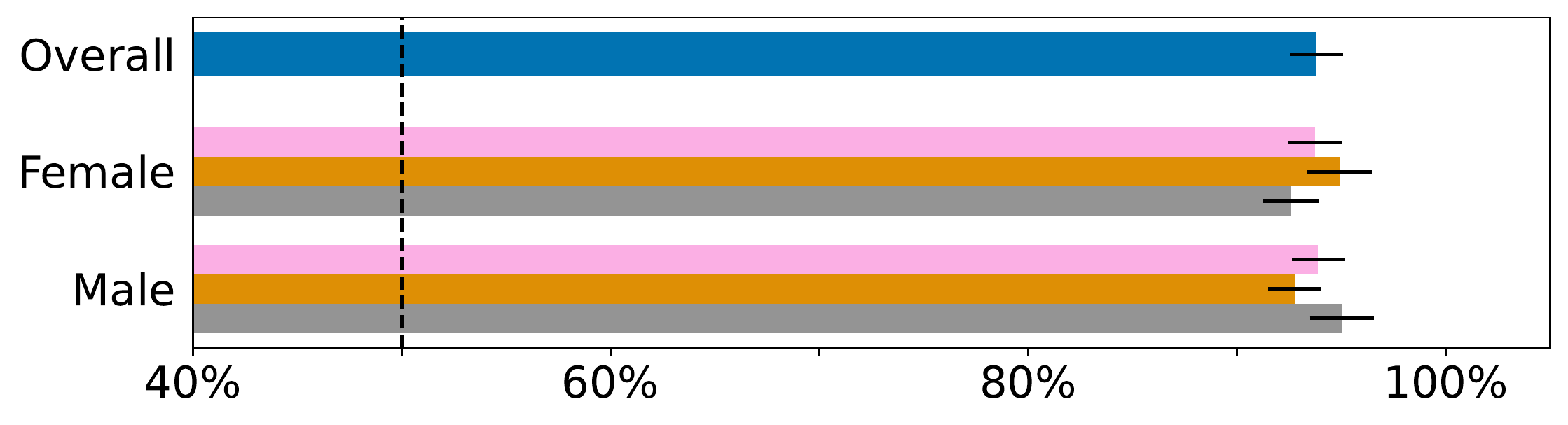}
         \caption{Gender (FFHQ)}
     \end{subfigure}
     \hfill
     \begin{subfigure}[b]{0.48\textwidth}
        \centering
         \includegraphics[width=\textwidth]{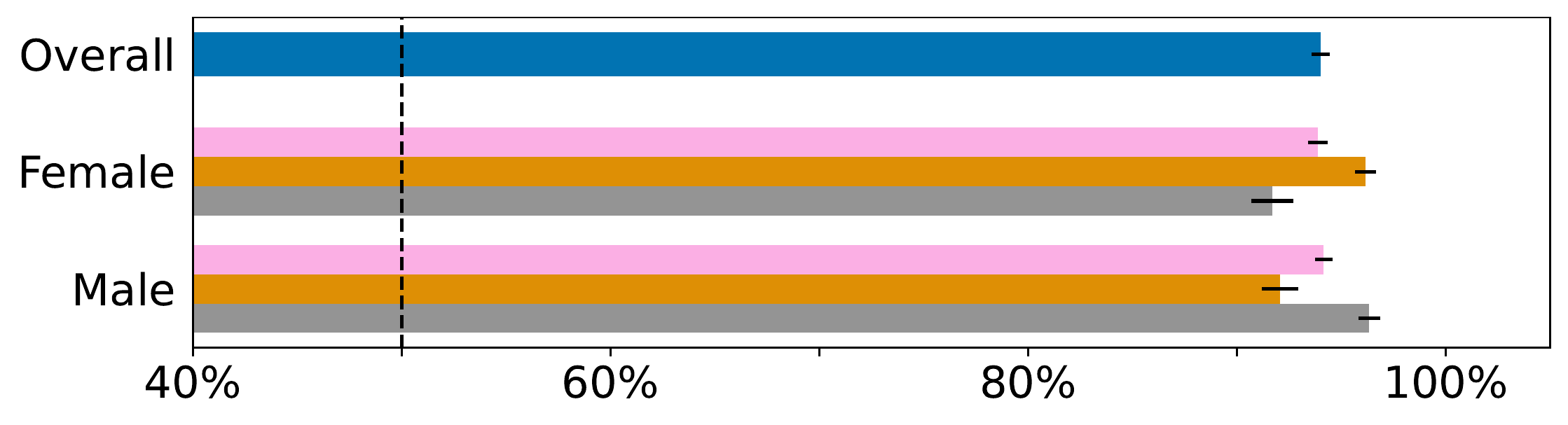}
         \caption{Gender (CelebAHQ)}
     \end{subfigure}

     \par\medskip
     \begin{subfigure}[b]{0.48\textwidth}
         \centering
         \includegraphics[width=\textwidth]{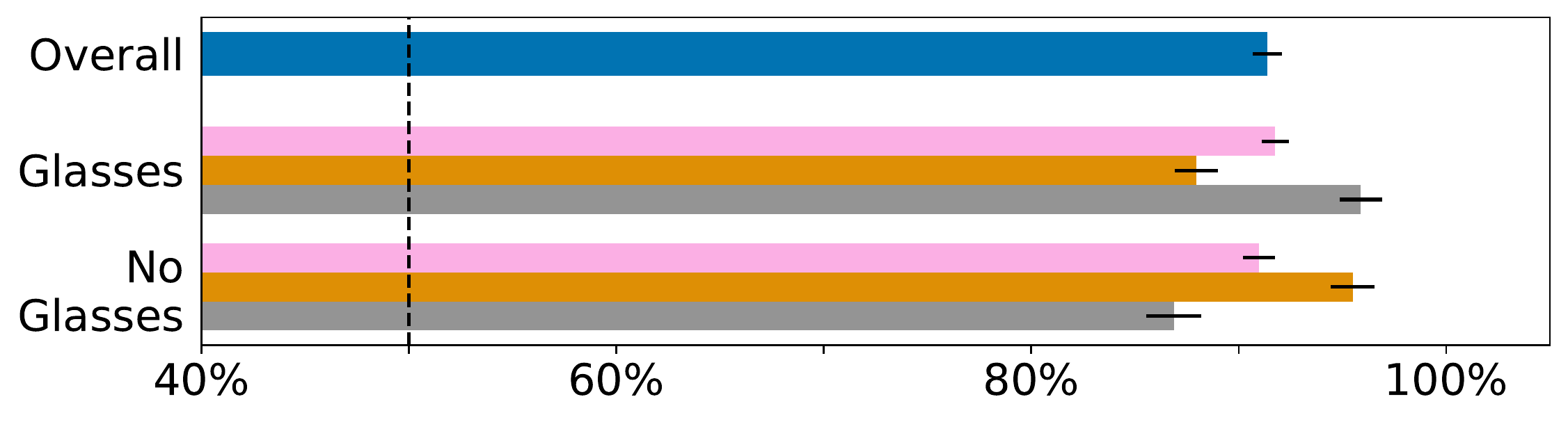}
         \caption{Eyeglasses (FFHQ)}
     \end{subfigure}
      \hfill
     \begin{subfigure}[b]{0.48\textwidth}
         \centering
         \includegraphics[width=\textwidth]{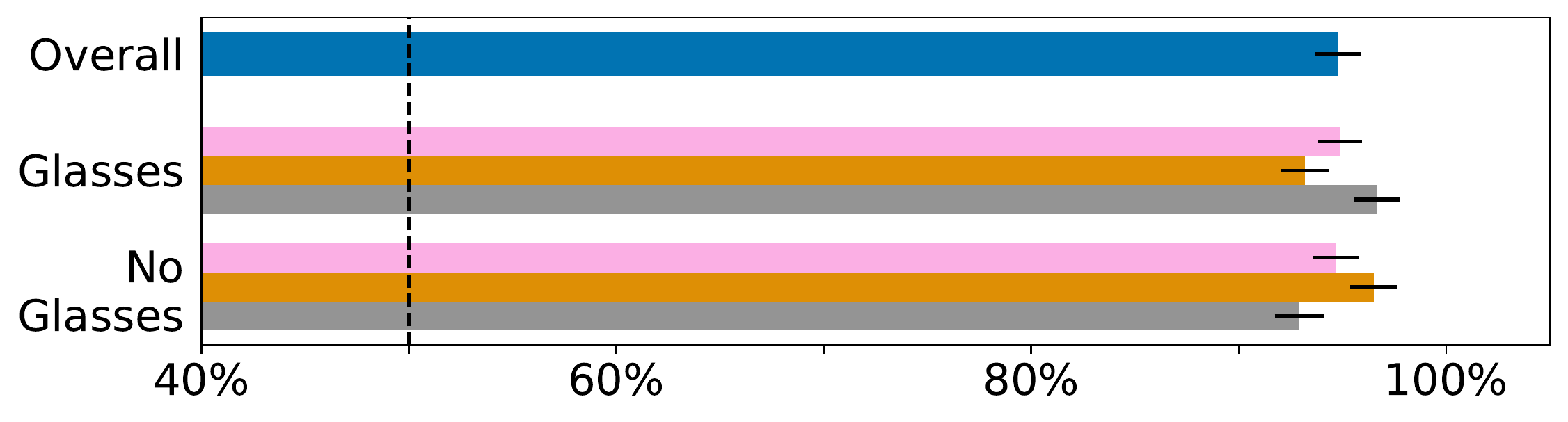}
         \caption{Eyeglasses (CelebAHQ)}
     \end{subfigure}
     
     \par\medskip
     \begin{subfigure}[b]{0.48\textwidth}
         \centering
         \includegraphics[width=\textwidth]{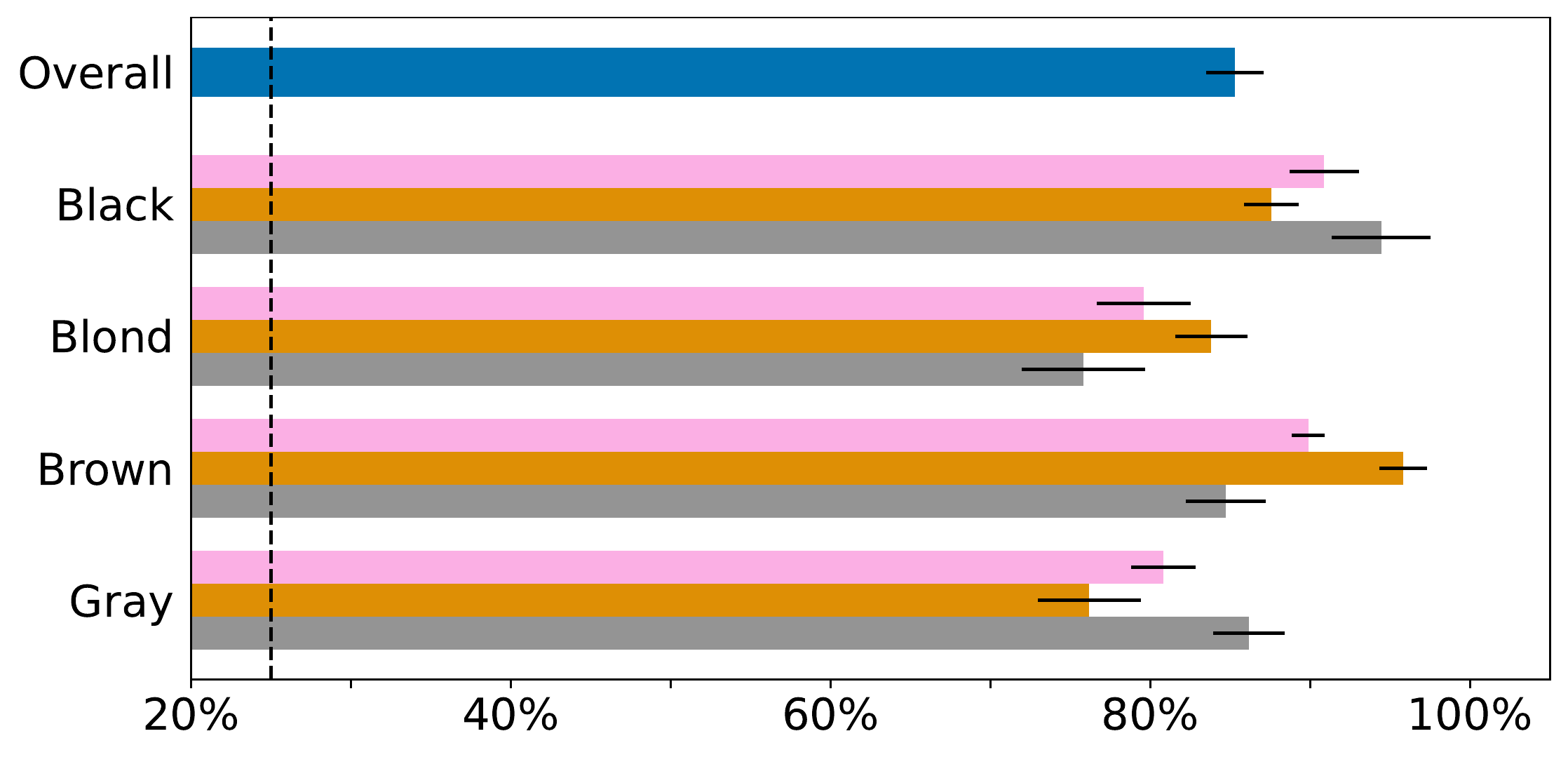}
         \caption{Hair Color (FFHQ)}
     \end{subfigure}
     \hfill
     \begin{subfigure}[b]{0.48\textwidth}
         \centering
         \includegraphics[width=\textwidth]{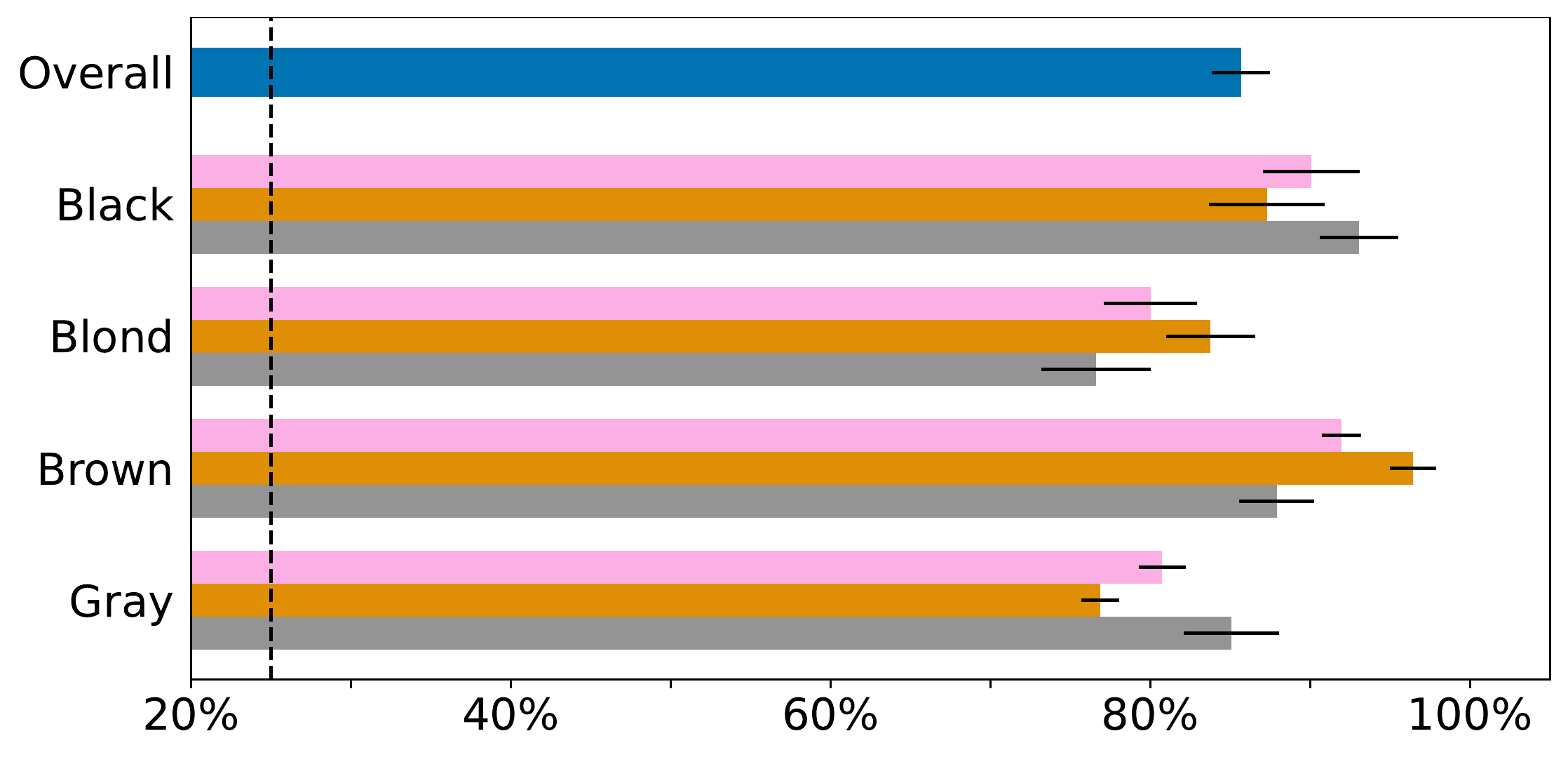}
         \caption{Hair Color (CelebAHQ)}
     \end{subfigure}

     \par\medskip

     \begin{subfigure}[b]{0.48\textwidth}
         \centering
         \includegraphics[width=\textwidth]{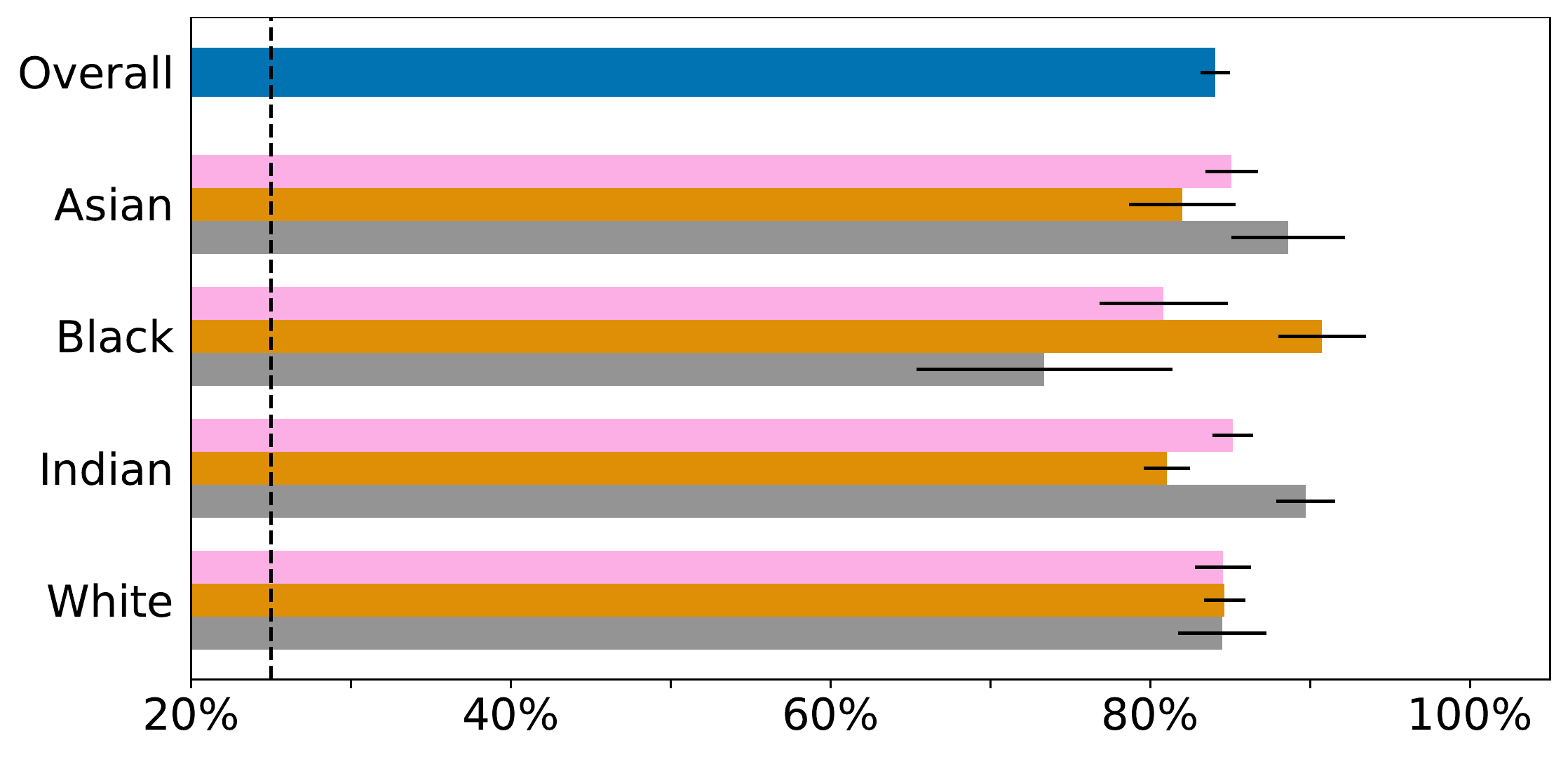}
         \caption{Racial Appearance (FFHQ)}
     \end{subfigure}
     \hfill
     \begin{subfigure}[b]{0.48\textwidth}
         \centering
         \includegraphics[width=\textwidth]{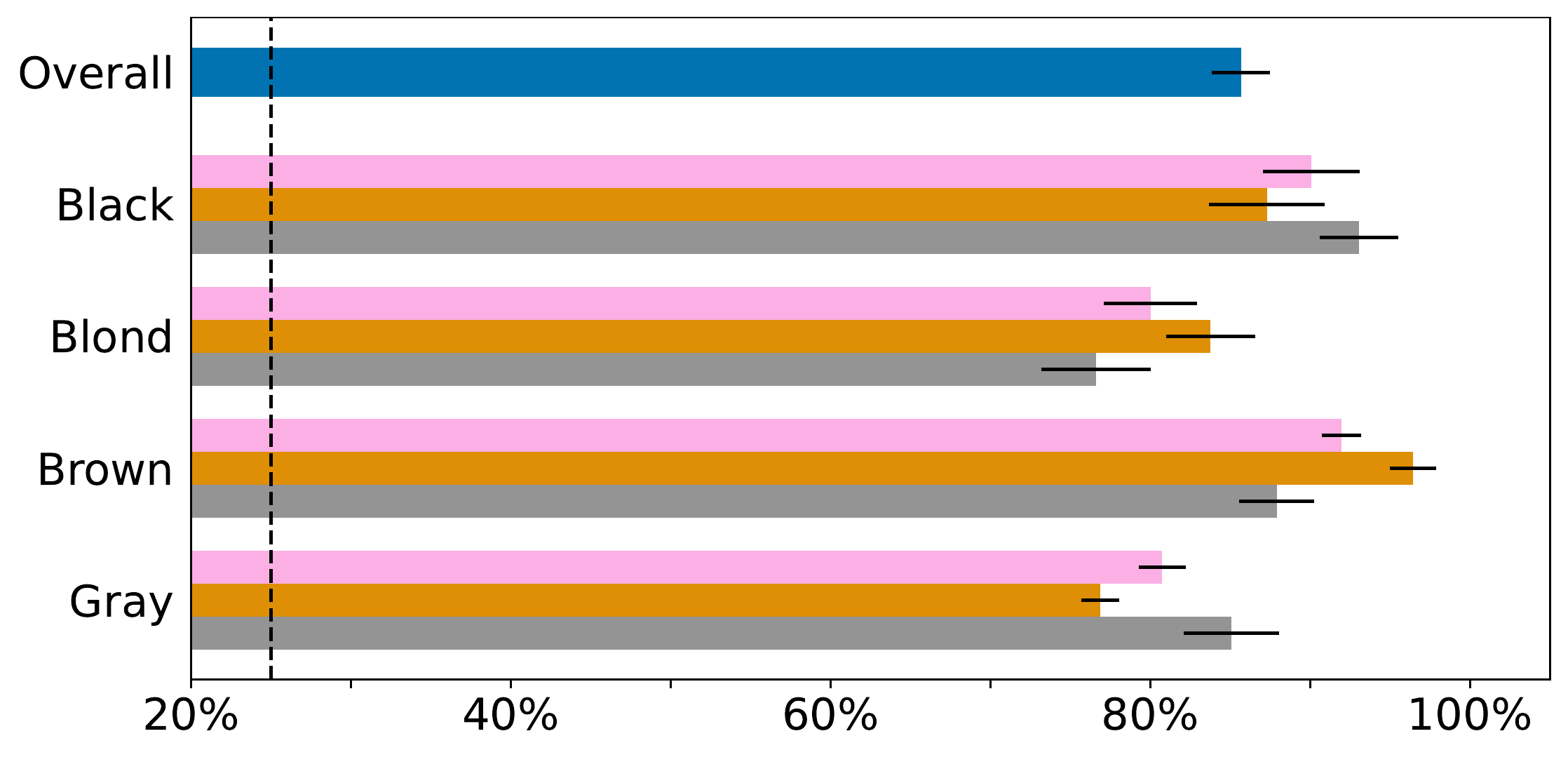}
         \caption{Racial Appearance (CelebAHQ)}
     \end{subfigure}

    \caption{Evaluation results for CAIA performed on ResNet-152 models to infer four different target attributes for \textbf{1000} identities. The black horizontal lines denote the standard deviation over nine runs. We further state random guessing (dashed line) for comparison.}
\end{figure*}
\clearpage

\subsection{DenseNet-169 - CelebA}
\begin{figure*}[h!]
\centering
     \begin{subfigure}[c]{\textwidth}
         \centering
         \includegraphics[width=0.75\textwidth]{images/barplots/legend_small.pdf}
     \end{subfigure}
     
     \begin{subfigure}[b]{0.48\textwidth}
        \centering
         \includegraphics[width=\textwidth]{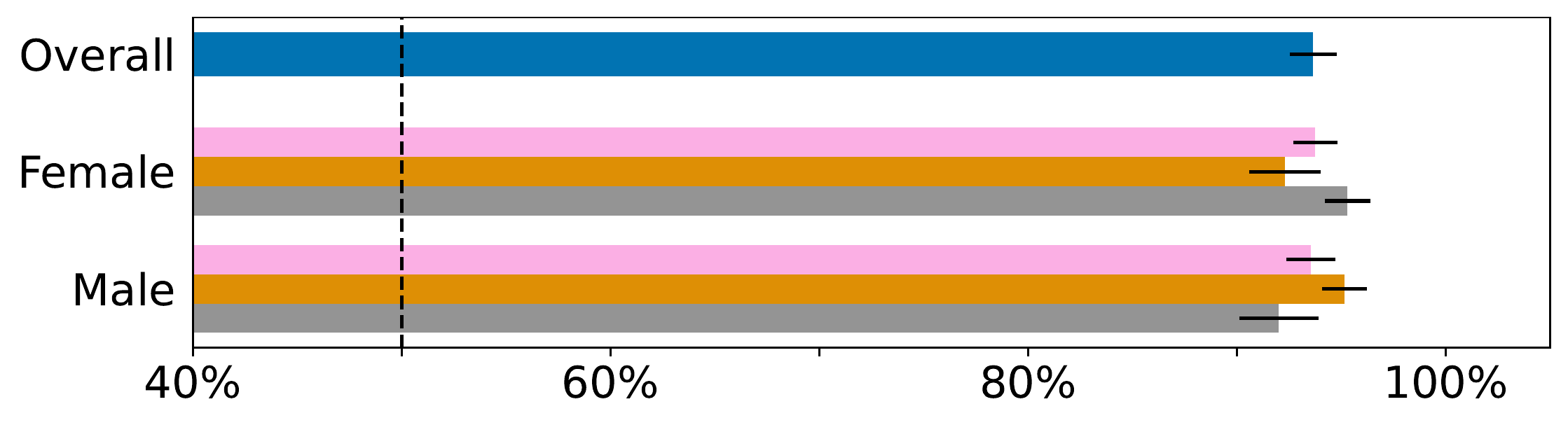}
         \caption{Gender (FFHQ)}
     \end{subfigure}
     \hfill
     \begin{subfigure}[b]{0.48\textwidth}
        \centering
         \includegraphics[width=\textwidth]{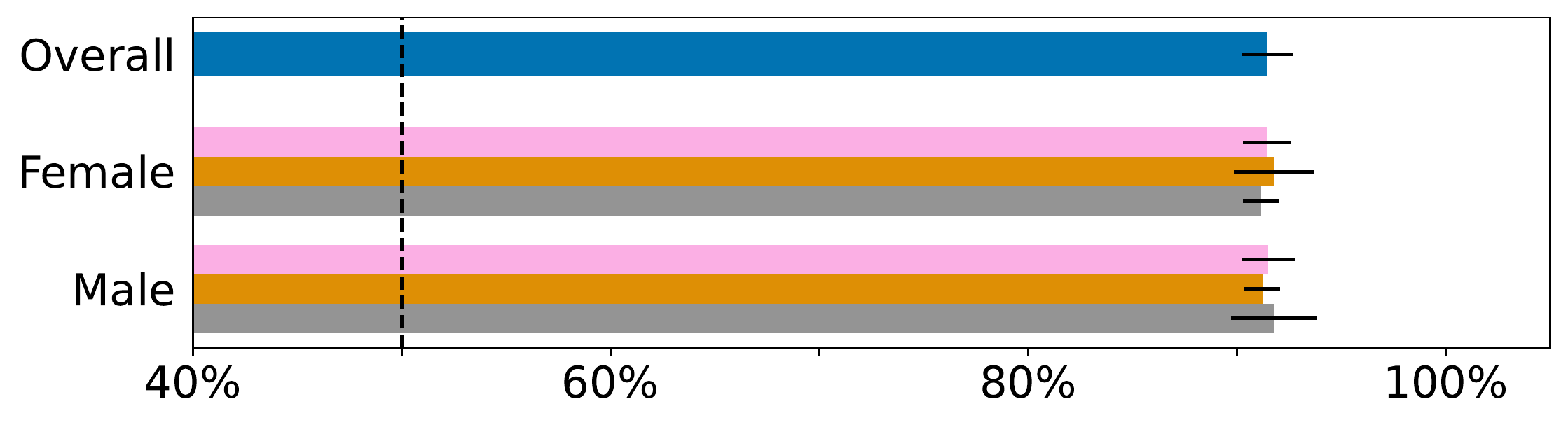}
         \caption{Gender (CelebAHQ)}
     \end{subfigure}
     
     \par\medskip
     \begin{subfigure}[b]{0.48\textwidth}
         \centering
         \includegraphics[width=\textwidth]{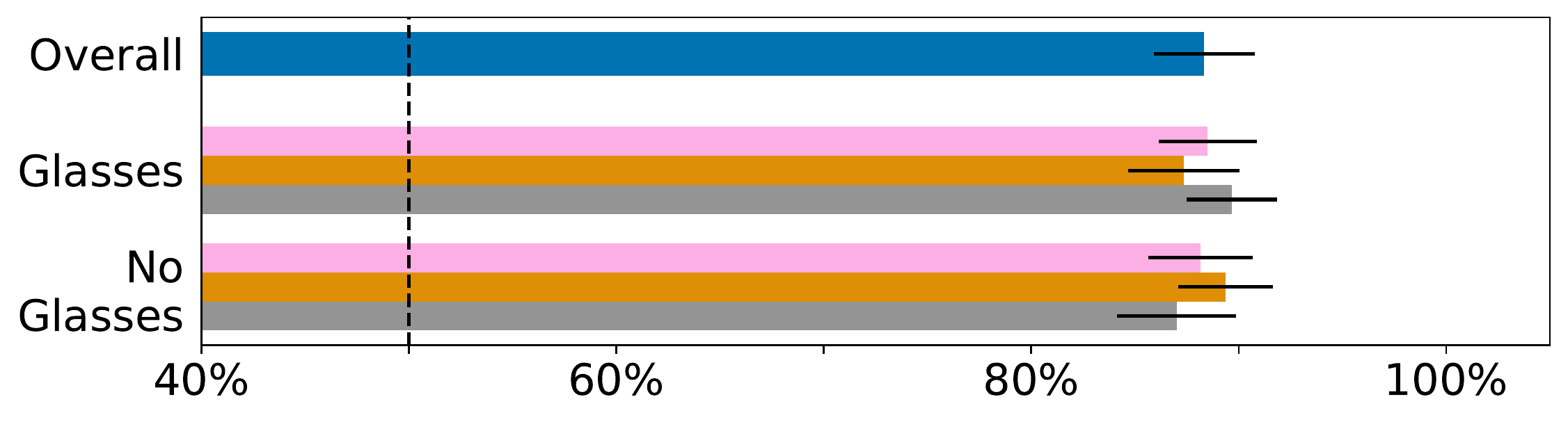}
         \caption{Eyeglasses (FFHQ)}
     \end{subfigure}
     \hfill
     \begin{subfigure}[b]{0.48\textwidth}
         \centering
         \includegraphics[width=\textwidth]{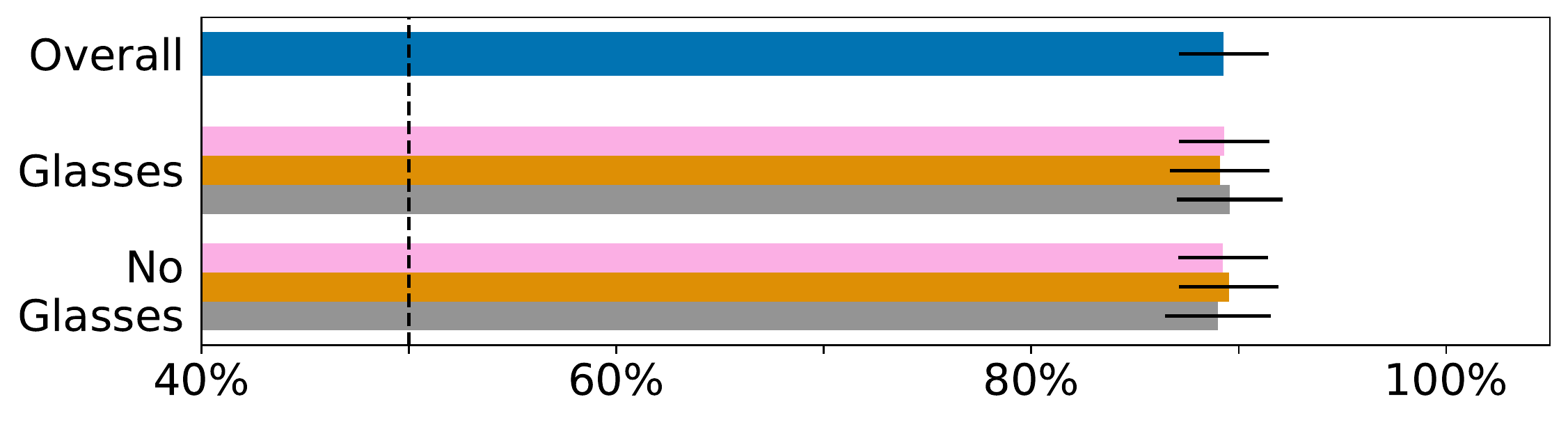}
         \caption{Eyeglasses (CelebAHQ)}
     \end{subfigure}

     \par\medskip
     \begin{subfigure}[b]{0.48\textwidth}
         \centering
         \includegraphics[width=\textwidth]{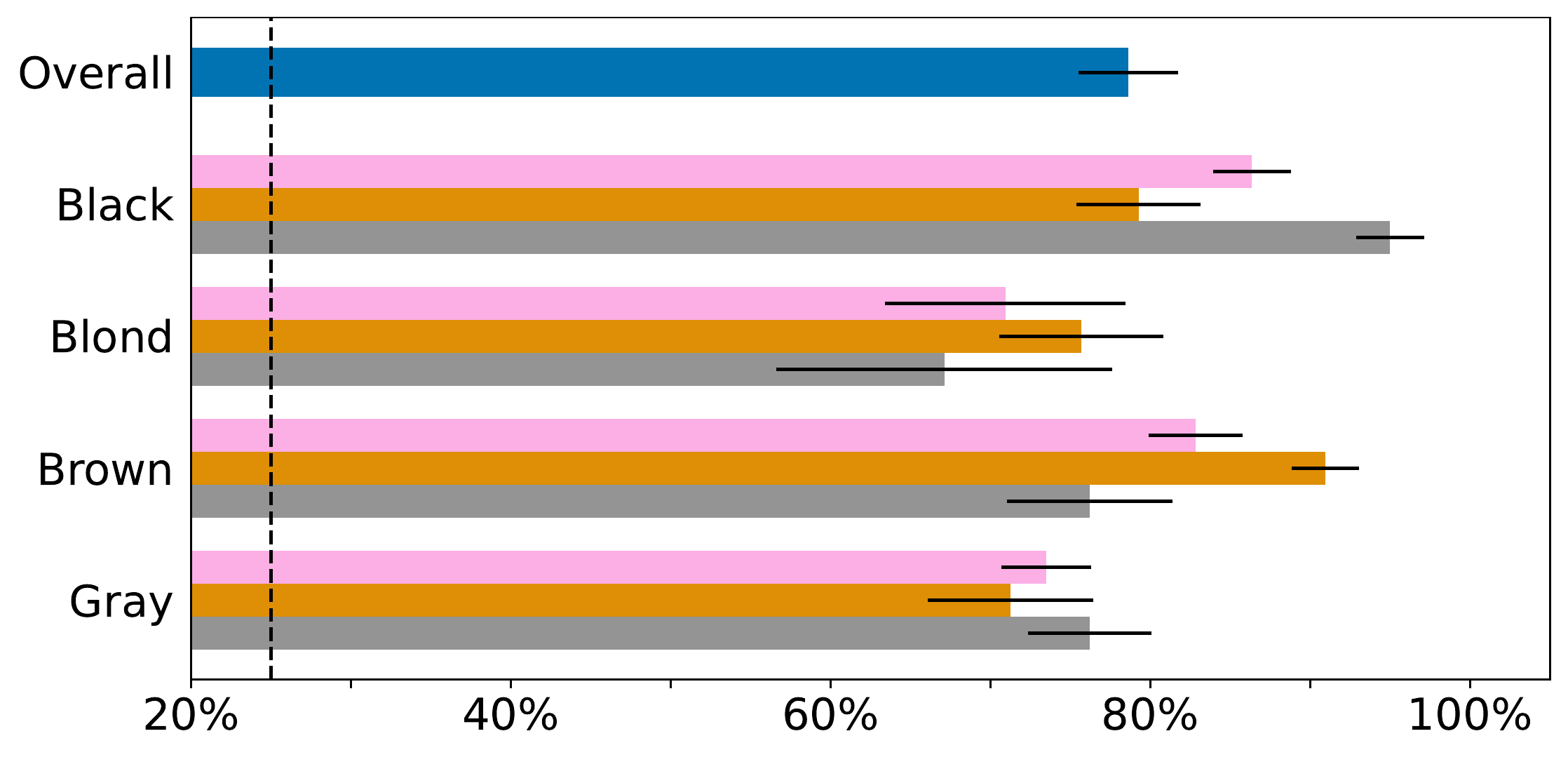}
         \caption{Hair Color (FFHQ)}
     \end{subfigure}
     \hfill
     \begin{subfigure}[b]{0.48\textwidth}
         \centering
         \includegraphics[width=\textwidth]{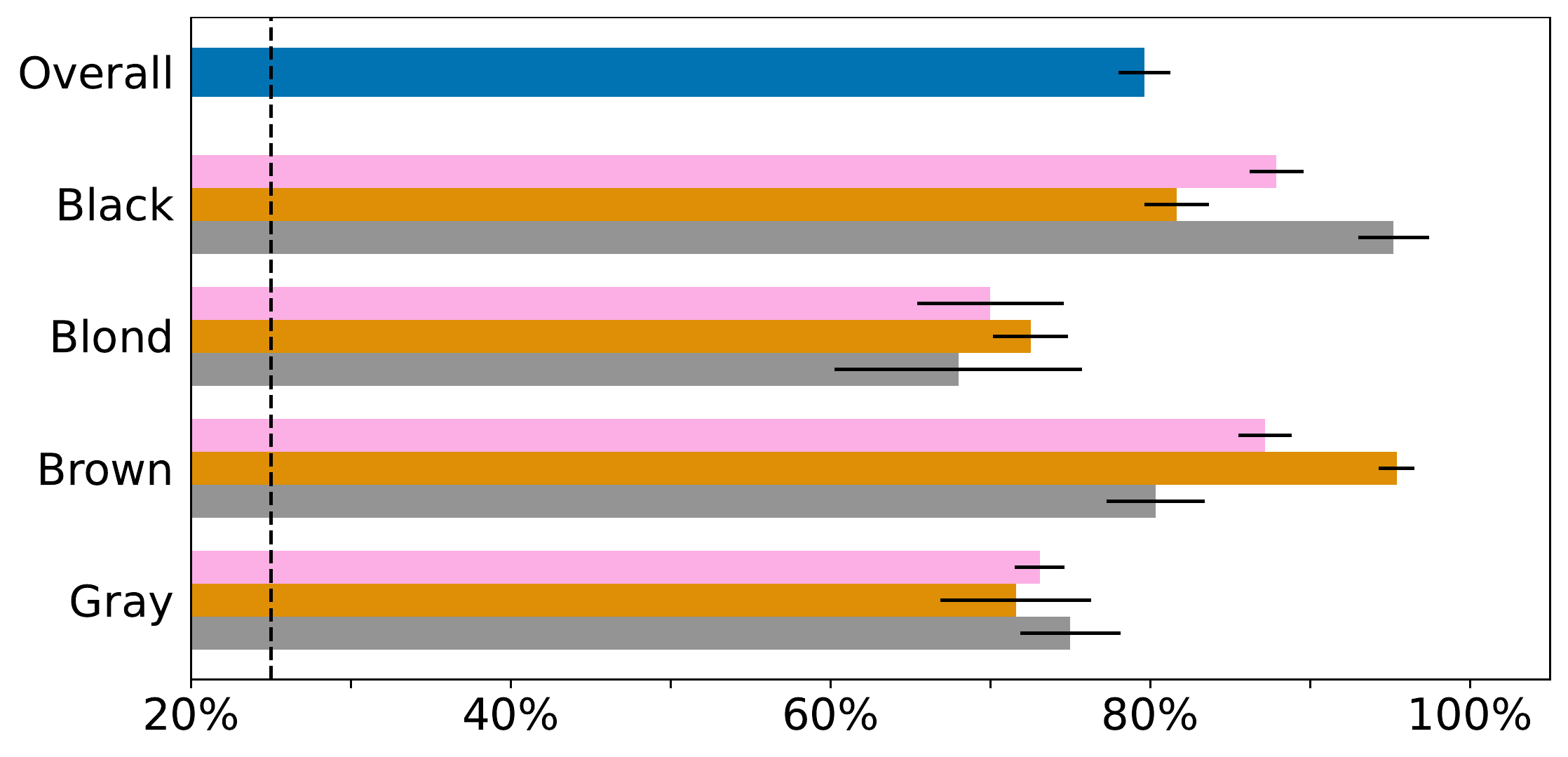}
         \caption{Hair Color (CelebAHQ)}
     \end{subfigure}
     \par\medskip

     \begin{subfigure}[b]{0.48\textwidth}
         \centering
         \includegraphics[width=\textwidth]{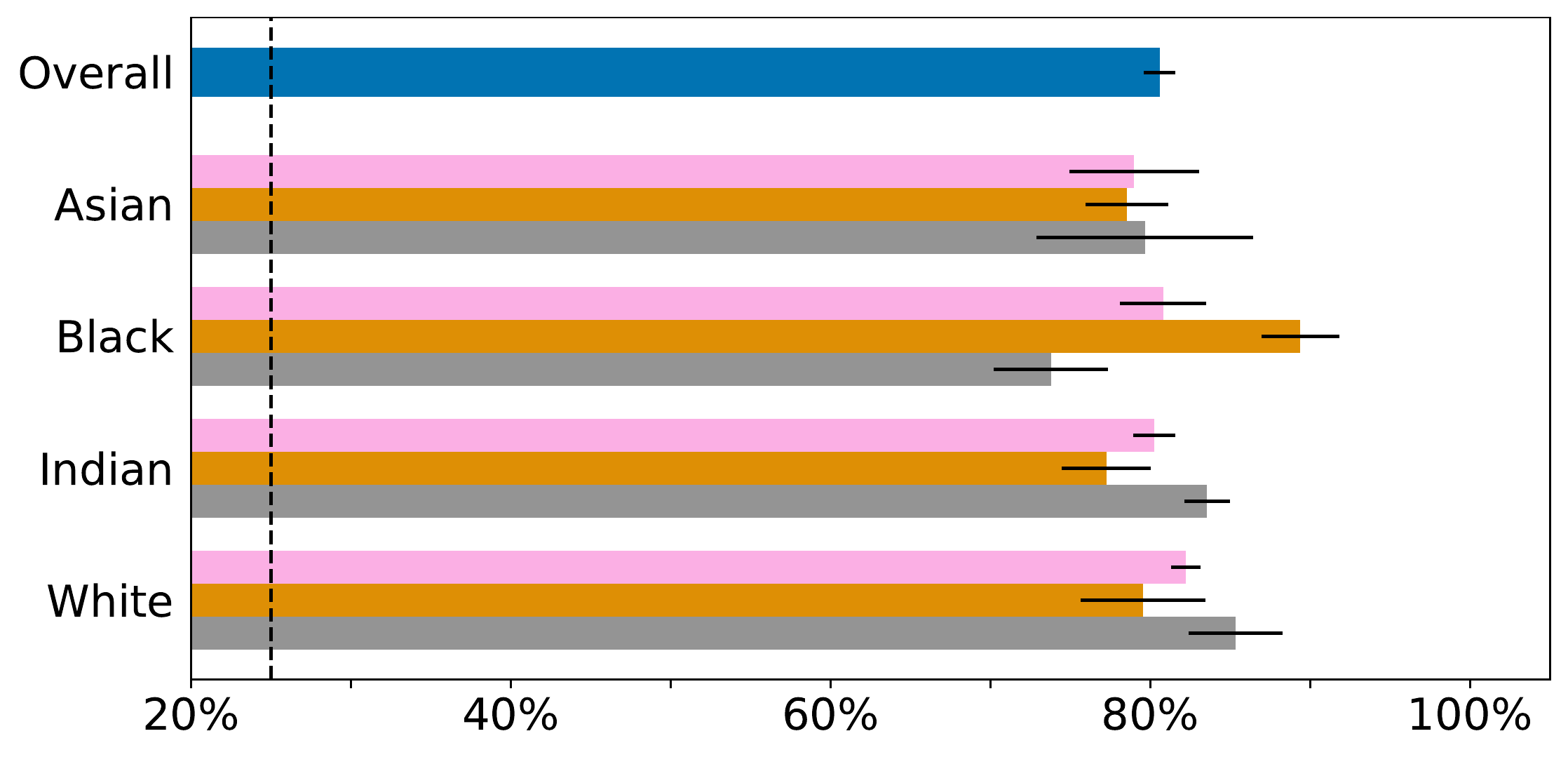}
         \caption{Racial Appearance (FFHQ)}
     \end{subfigure}
     \hfill
     \begin{subfigure}[b]{0.48\textwidth}
         \centering
         \includegraphics[width=\textwidth]{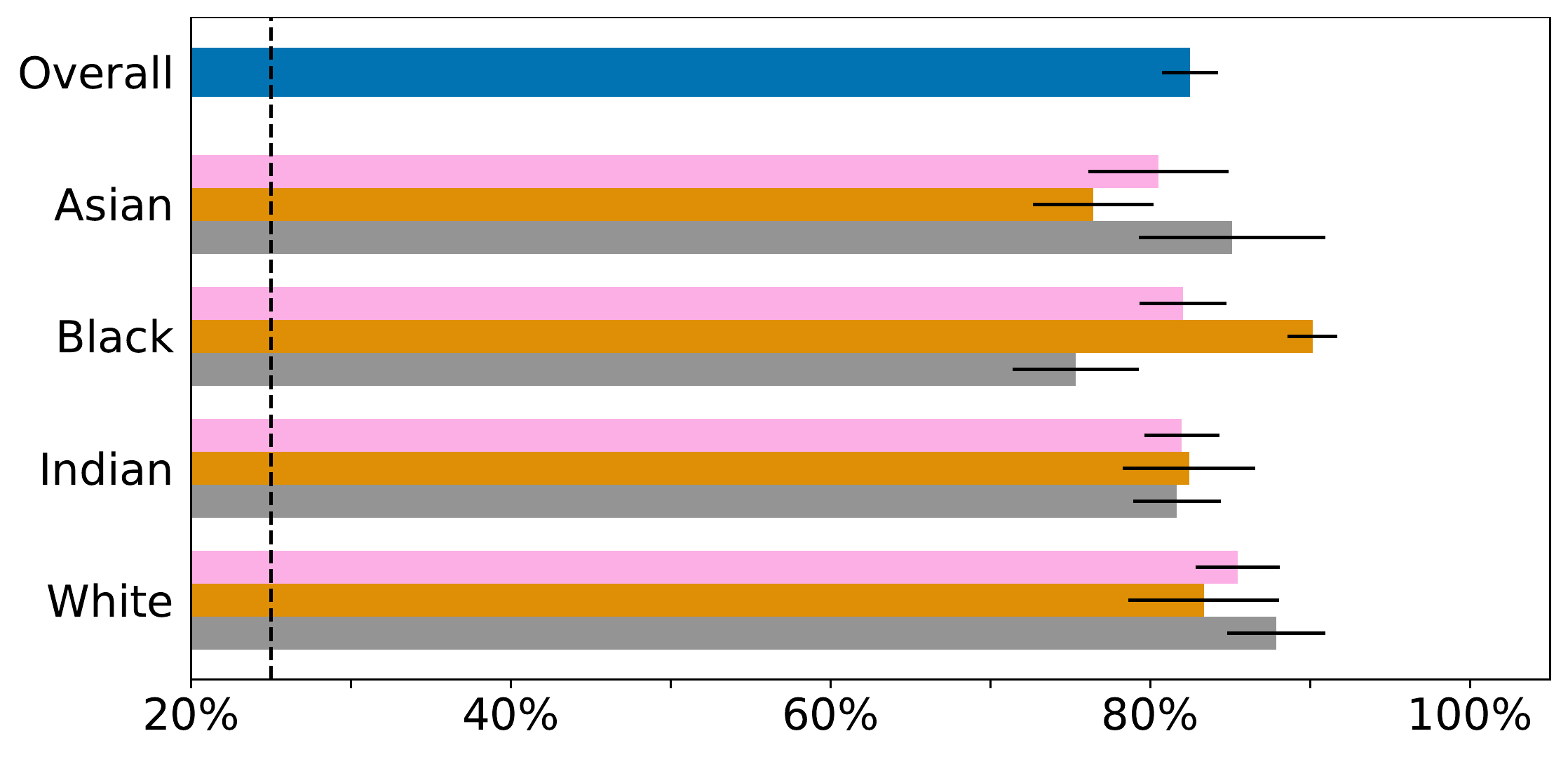}
         \caption{Racial Appearance (CelebAHQ)}
     \end{subfigure}

    \caption{Evaluation results for CAIA performed on DenseNet-169 models to infer four different target attributes. The black horizontal lines denote the standard deviation over nine runs. We further state random guessing (dashed line) for comparison.}
\end{figure*}
\clearpage

\subsection{DenseNet-169 - CelebA (1000 identities)}
\begin{figure*}[h!]
\centering
     \begin{subfigure}[c]{\textwidth}
         \centering
         \includegraphics[width=0.75\textwidth]{images/barplots/legend_small.pdf}
     \end{subfigure}
     
     \begin{subfigure}[b]{0.48\textwidth}
        \centering
         \includegraphics[width=\textwidth]{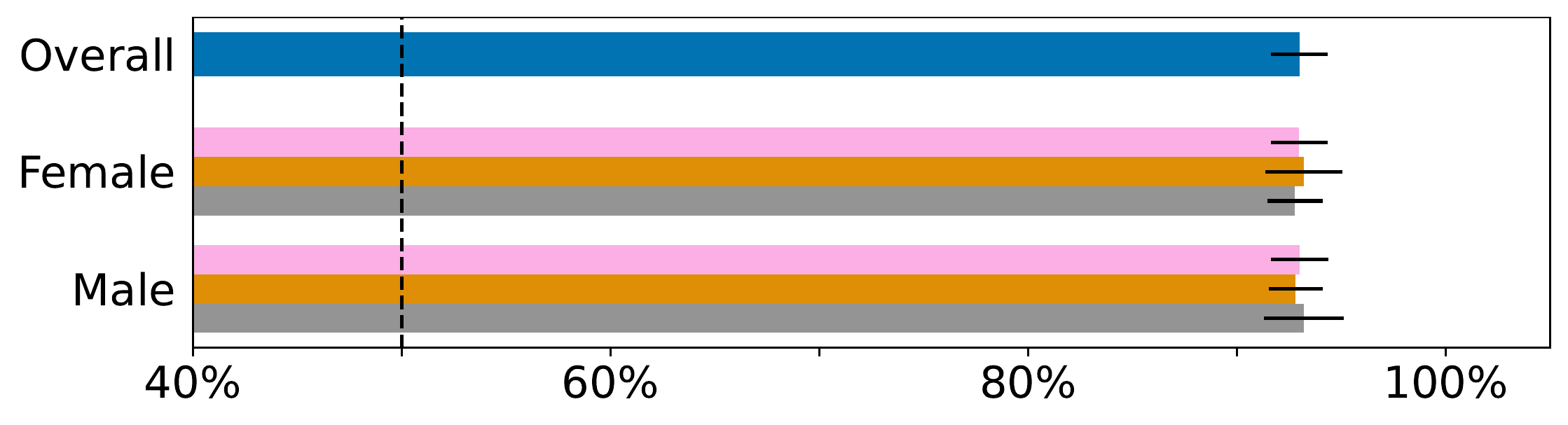}
         \caption{Gender (FFHQ)}
     \end{subfigure}
     \hfill
     \begin{subfigure}[b]{0.48\textwidth}
        \centering
         \includegraphics[width=\textwidth]{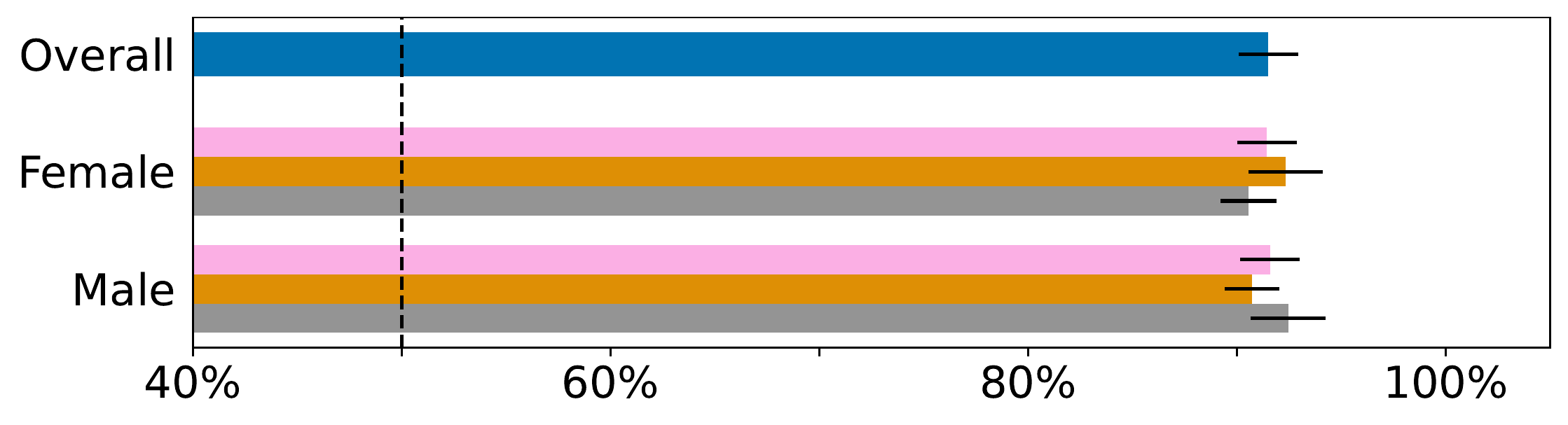}
         \caption{Gender (CelebAHQ)}
     \end{subfigure}
     
     \par\medskip
     \begin{subfigure}[b]{0.48\textwidth}
         \centering
         \includegraphics[width=\textwidth]{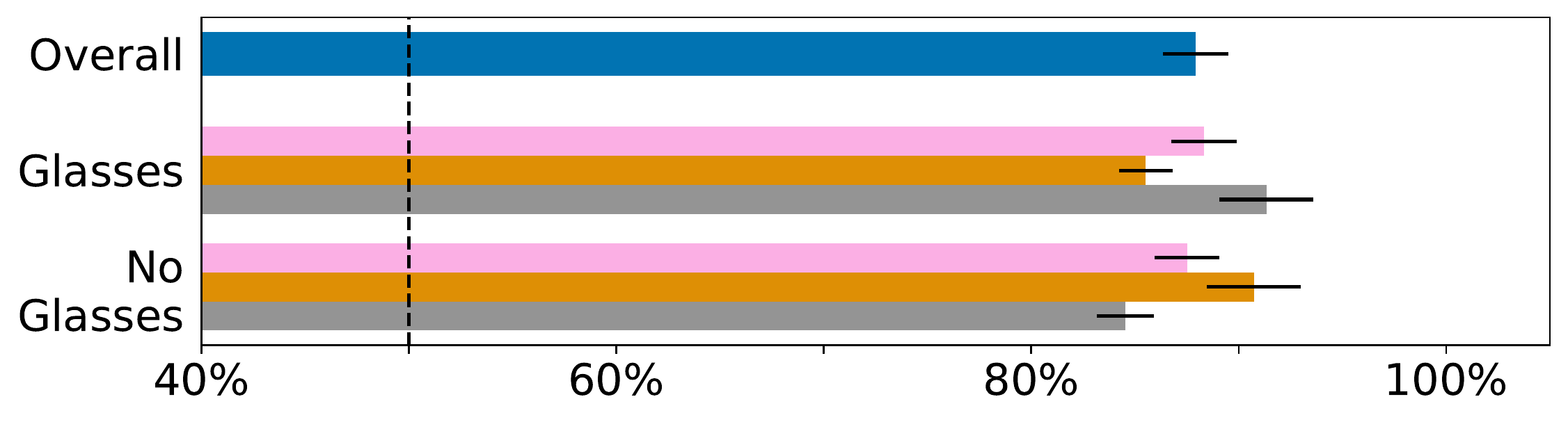}
         \caption{Eyeglasses (FFHQ)}
     \end{subfigure}
     \hfill
     \begin{subfigure}[b]{0.48\textwidth}
         \centering
         \includegraphics[width=\textwidth]{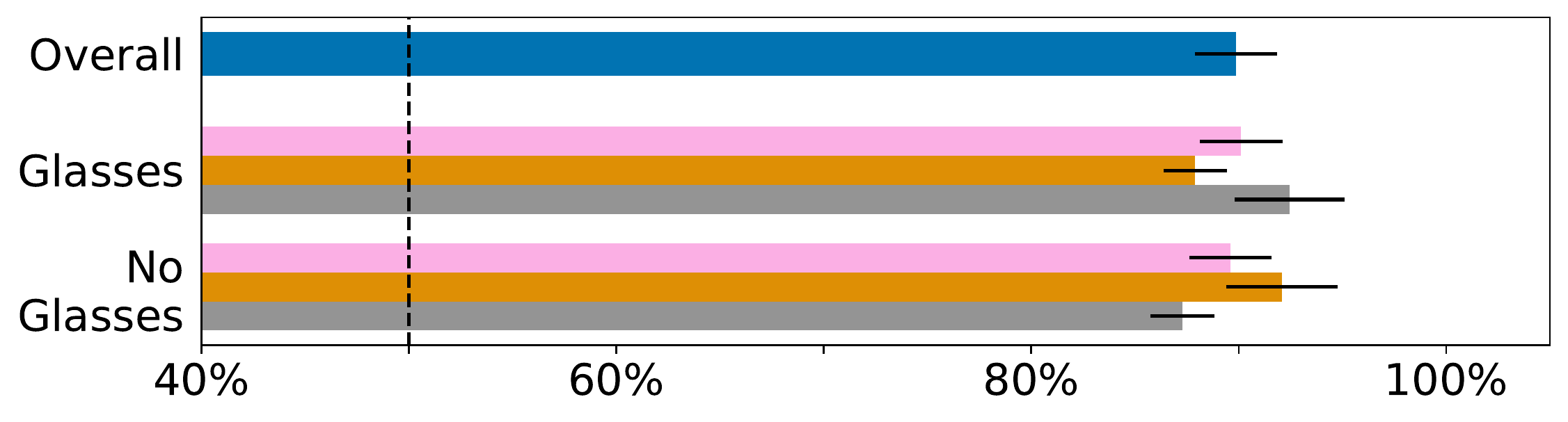}
         \caption{Eyeglasses (CelebAHQ)}
     \end{subfigure}

     \par\medskip
     \begin{subfigure}[b]{0.48\textwidth}
         \centering
         \includegraphics[width=\textwidth]{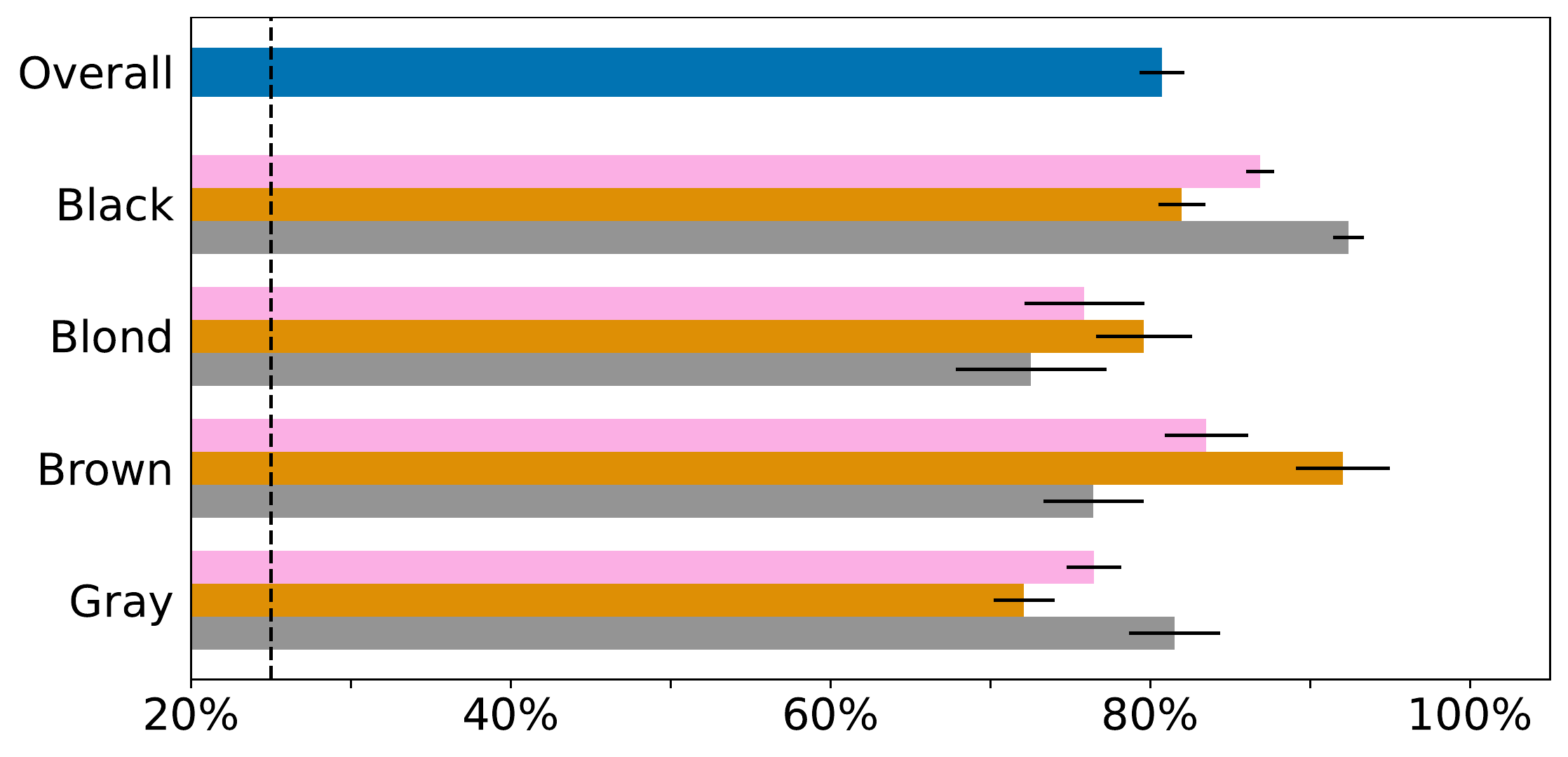}
         \caption{Hair Color (FFHQ)}
     \end{subfigure}
     \hfill
     \begin{subfigure}[b]{0.48\textwidth}
         \centering
         \includegraphics[width=\textwidth]{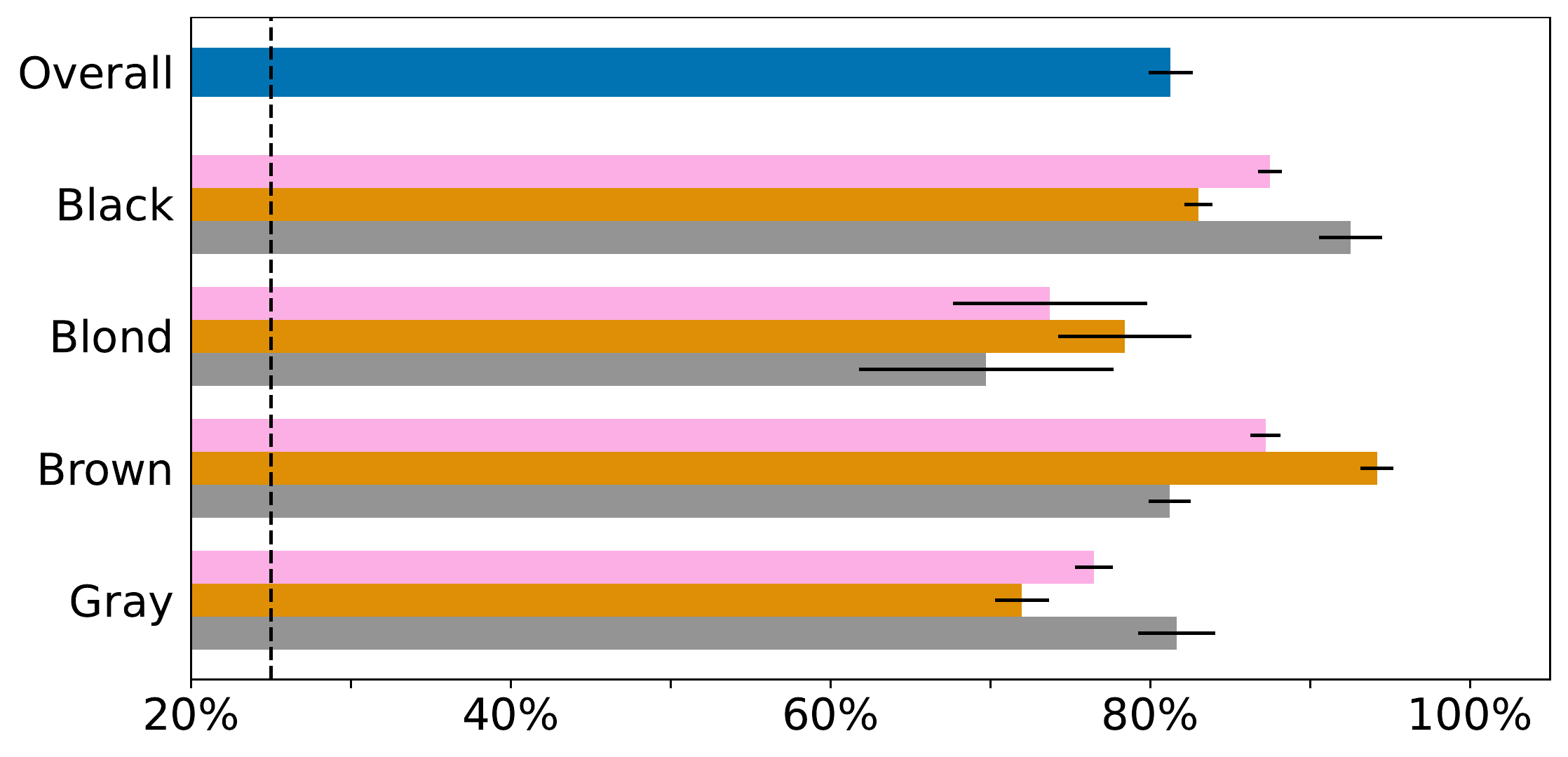}
         \caption{Hair Color (CelebAHQ)}
     \end{subfigure}
     \par\medskip

     \begin{subfigure}[b]{0.48\textwidth}
         \centering
         \includegraphics[width=\textwidth]{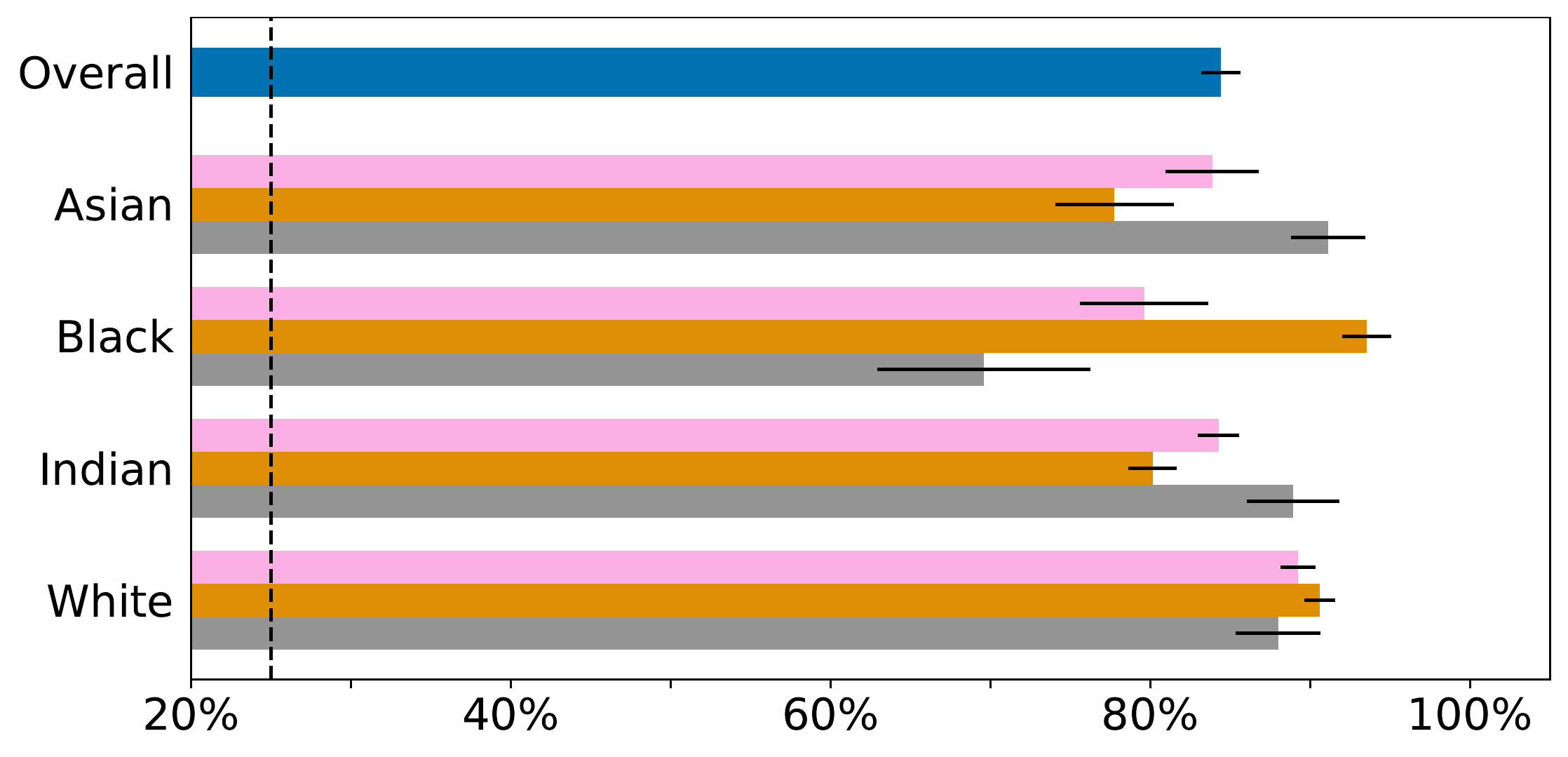}
         \caption{Racial Appearance (FFHQ)}
     \end{subfigure}
     \hfill
     \begin{subfigure}[b]{0.48\textwidth}
         \centering
         \includegraphics[width=\textwidth]{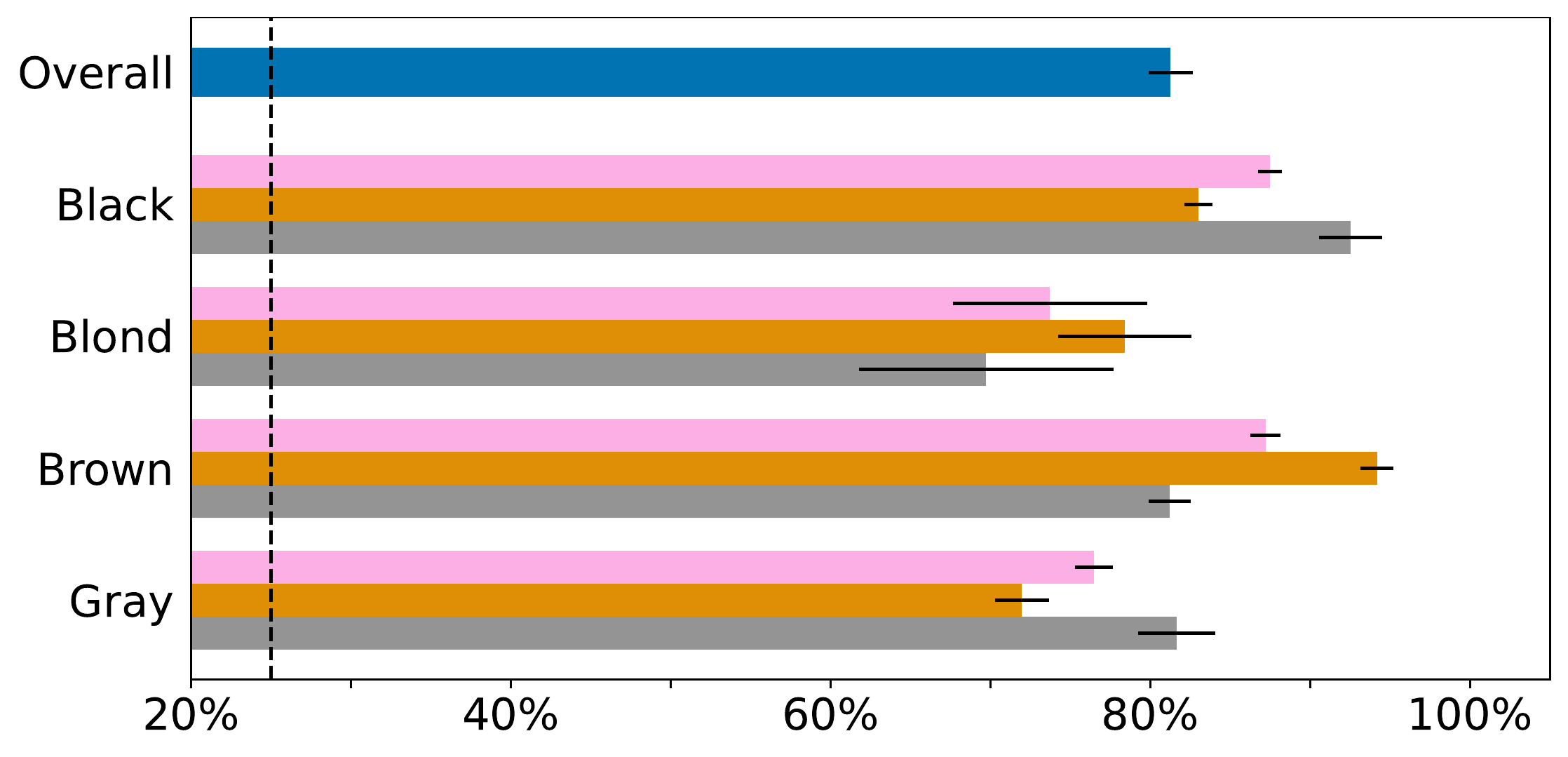}
         \caption{Racial Appearance (CelebAHQ)}
     \end{subfigure}

    \caption{Evaluation results for CAIA performed on DenseNet-169 models to infer four different target attributes for \textbf{1000} identities. The black horizontal lines denote the standard deviation over nine runs. We further state random guessing (dashed line) for comparison.}
\end{figure*}
\clearpage

\subsection{ResNeSt-101 - CelebA}
\begin{figure*}[h!]
\centering
     \begin{subfigure}[c]{\textwidth}
         \centering
         \includegraphics[width=0.75\textwidth]{images/barplots/legend_small.pdf}
     \end{subfigure}
     
     \begin{subfigure}[b]{0.48\textwidth}
        \centering
         \includegraphics[width=\textwidth]{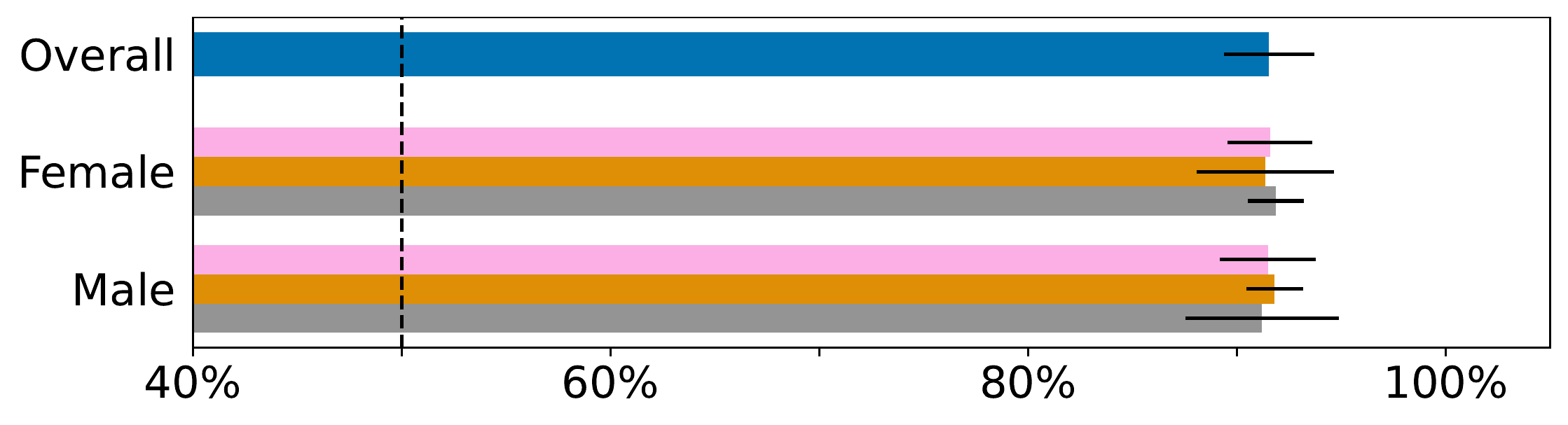}
         \caption{Gender (FFHQ)}
     \end{subfigure}
     \hfill
     \begin{subfigure}[b]{0.48\textwidth}
        \centering
         \includegraphics[width=\textwidth]{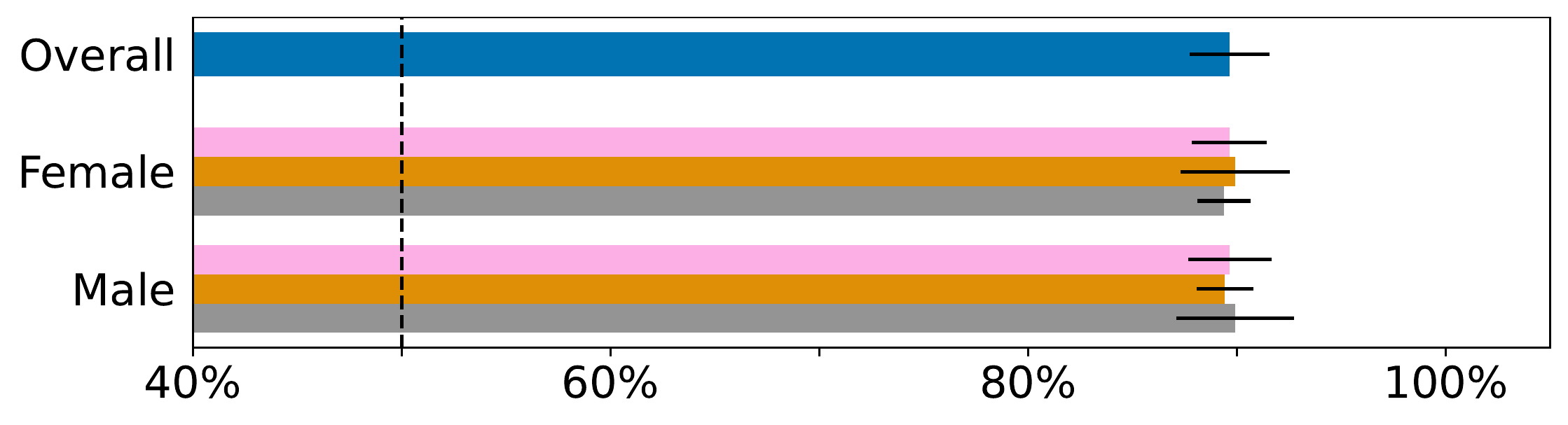}
         \caption{Gender (CelebAHQ)}
     \end{subfigure}

     \par\medskip
     \begin{subfigure}[b]{0.48\textwidth}
         \centering
         \includegraphics[width=\textwidth]{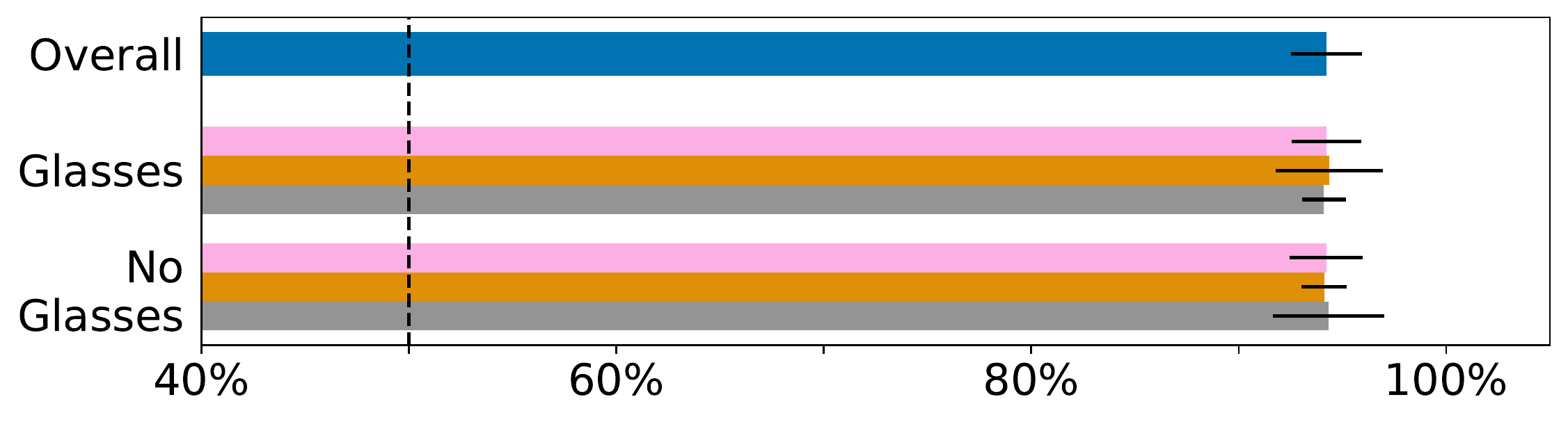}
         \caption{Eyeglasses (FFHQ)}
     \end{subfigure}
     \hfill
     \begin{subfigure}[b]{0.48\textwidth}
         \centering
         \includegraphics[width=\textwidth]{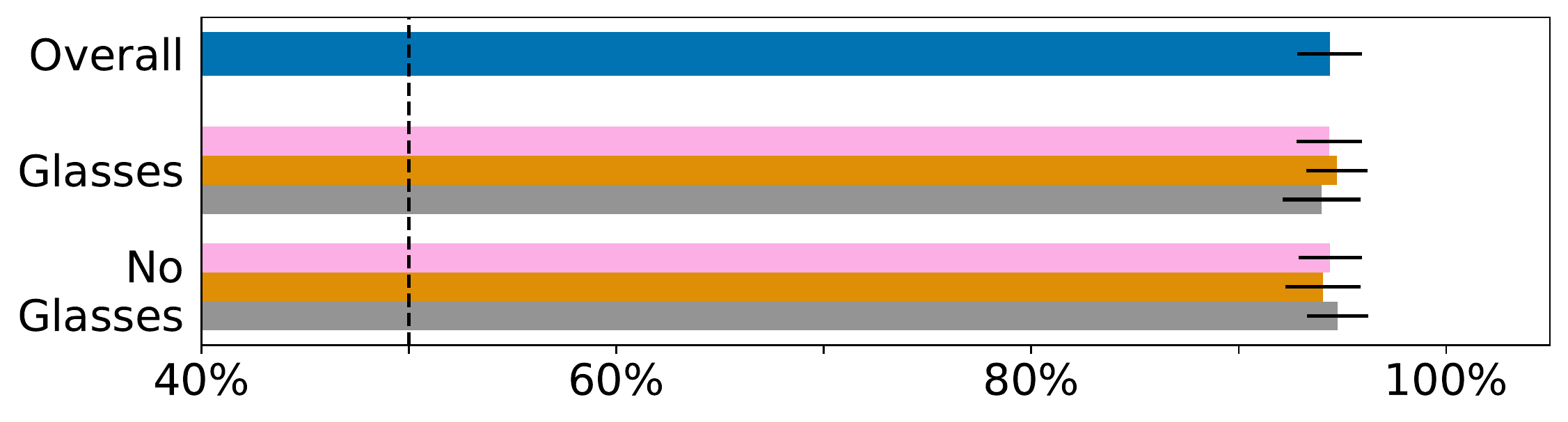}
         \caption{Eyeglasses (CelebAHQ)}
     \end{subfigure}
     
     \par\medskip
     \begin{subfigure}[b]{0.48\textwidth}
         \centering
         \includegraphics[width=\textwidth]{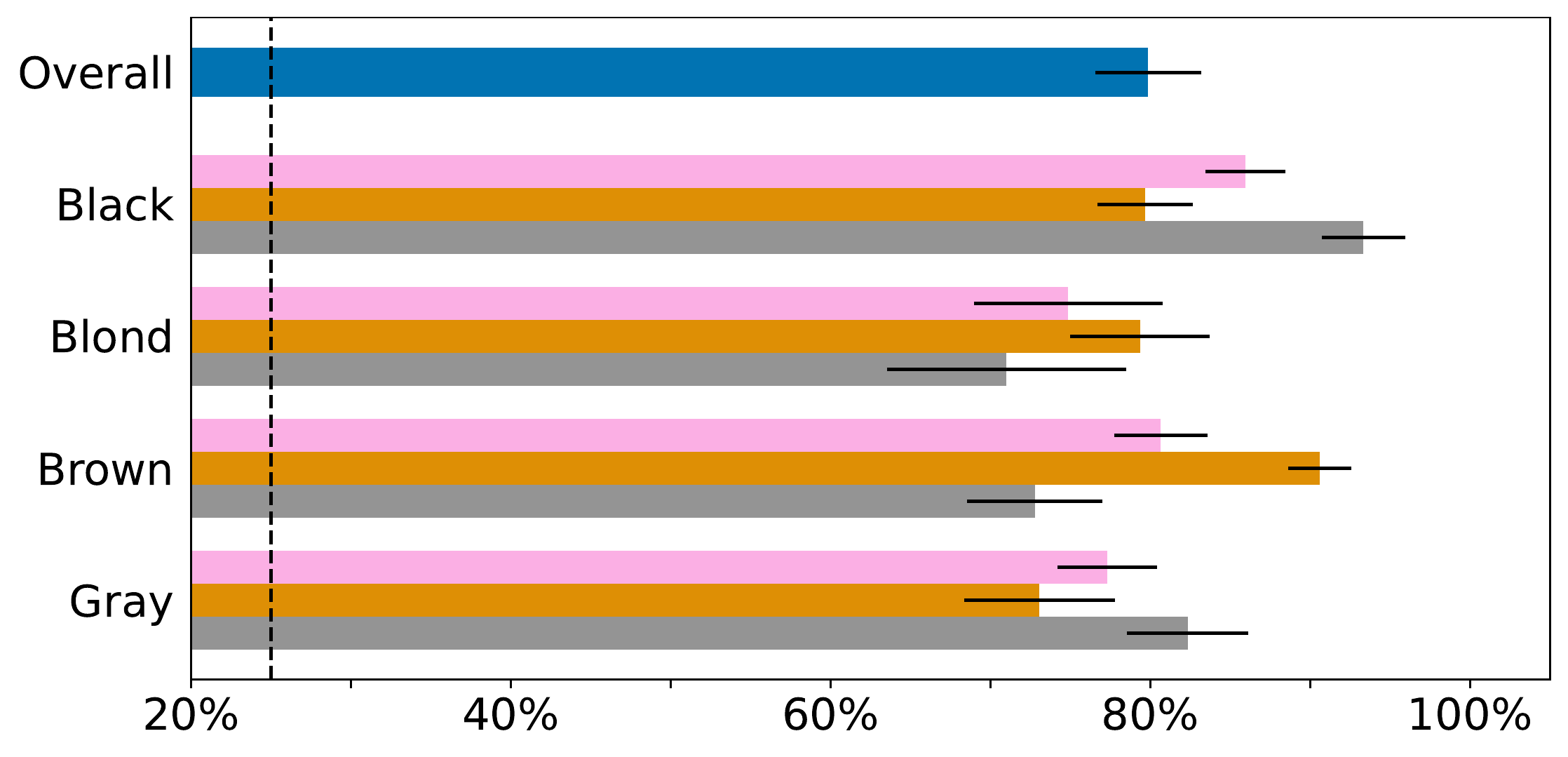}
         \caption{Hair Color (FFHQ)}
     \end{subfigure}
     \hfill
     \begin{subfigure}[b]{0.48\textwidth}
         \centering
         \includegraphics[width=\textwidth]{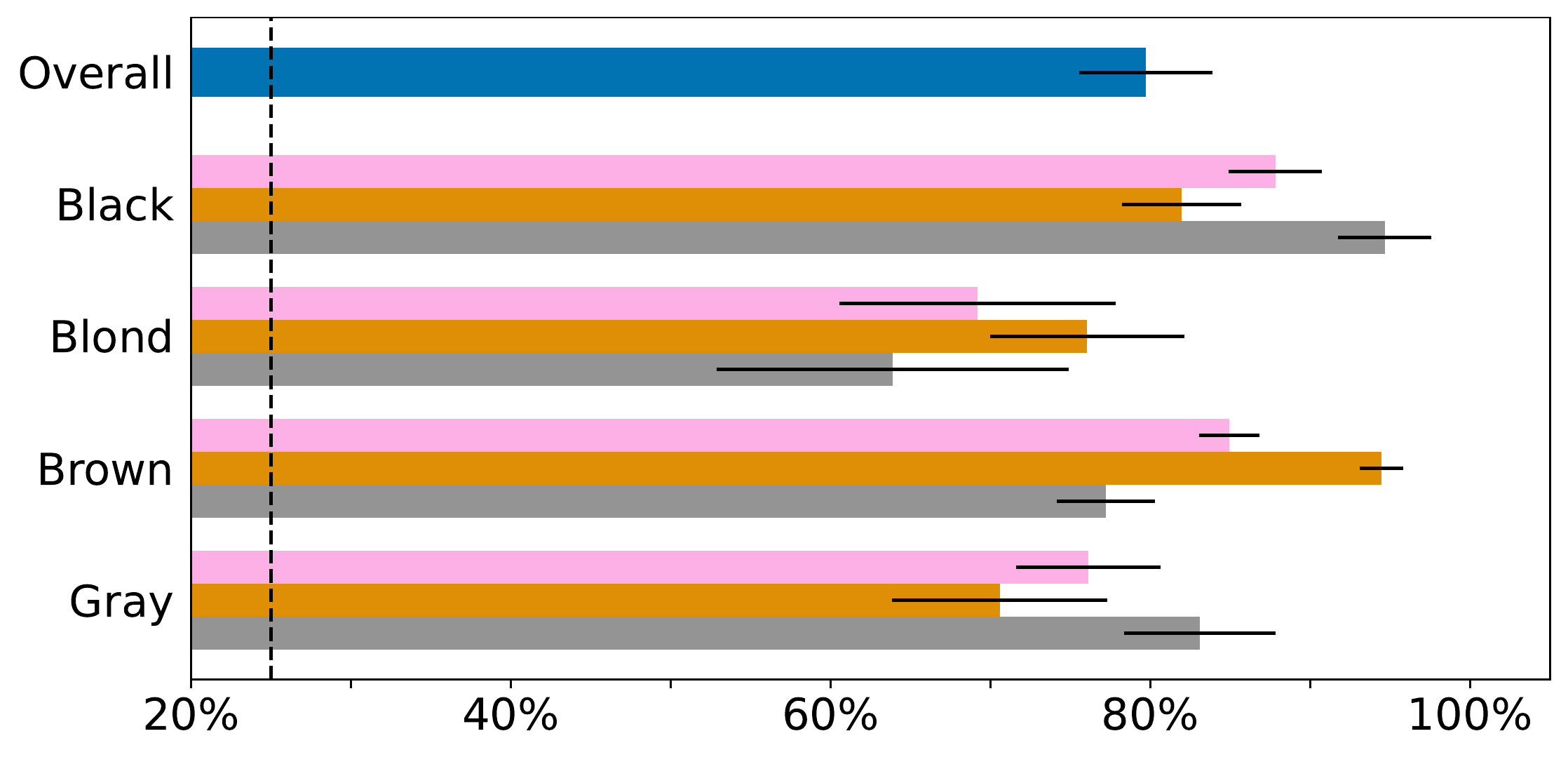}
         \caption{Hair Color (CelebAHQ)}
     \end{subfigure}

     \par\medskip
     \begin{subfigure}[b]{0.48\textwidth}
         \centering
         \includegraphics[width=\textwidth]{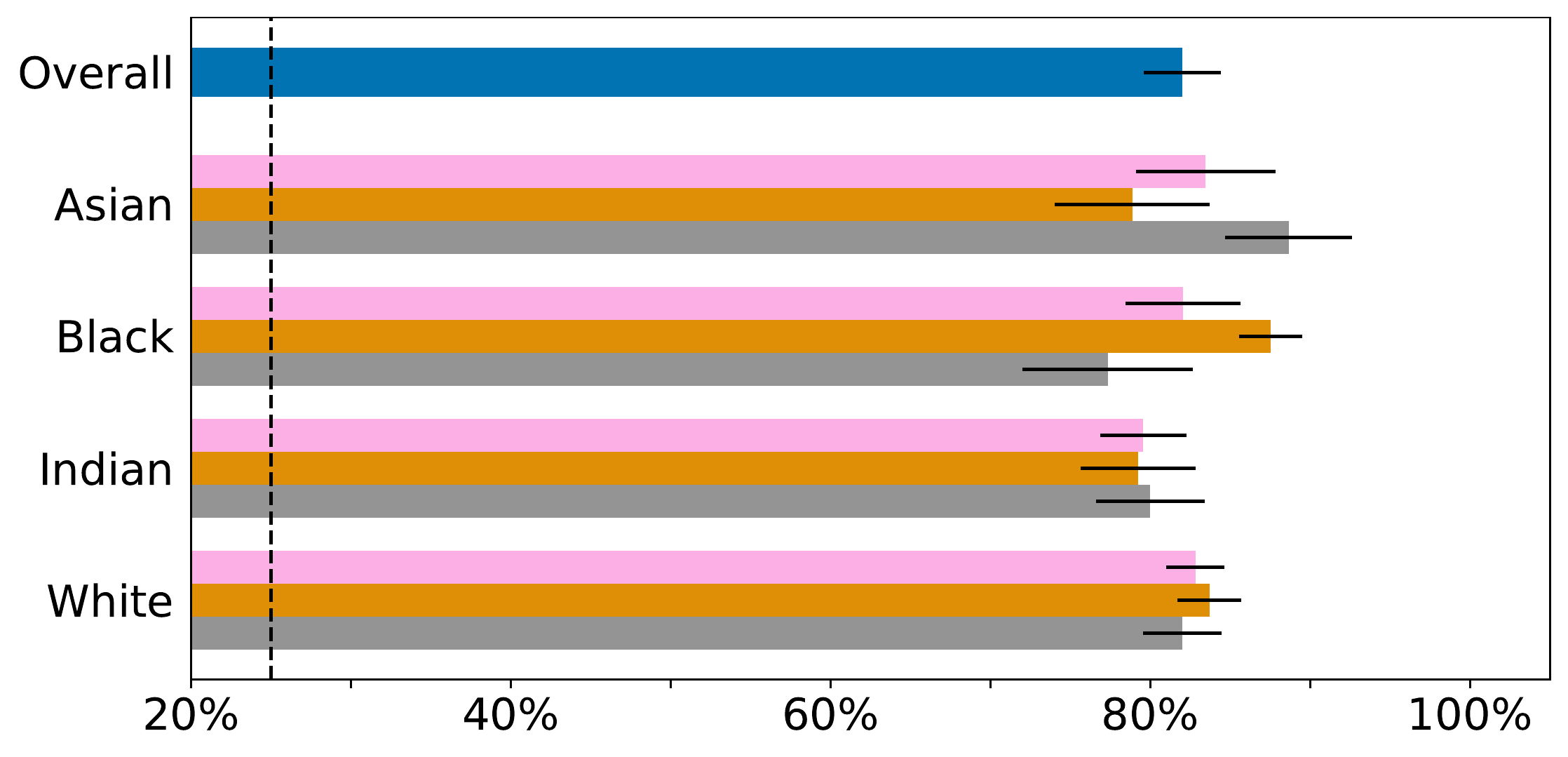}
         \caption{Racial Appearance (FFHQ)}
     \end{subfigure}
    \hfill
     \begin{subfigure}[b]{0.48\textwidth}
         \centering
         \includegraphics[width=\textwidth]{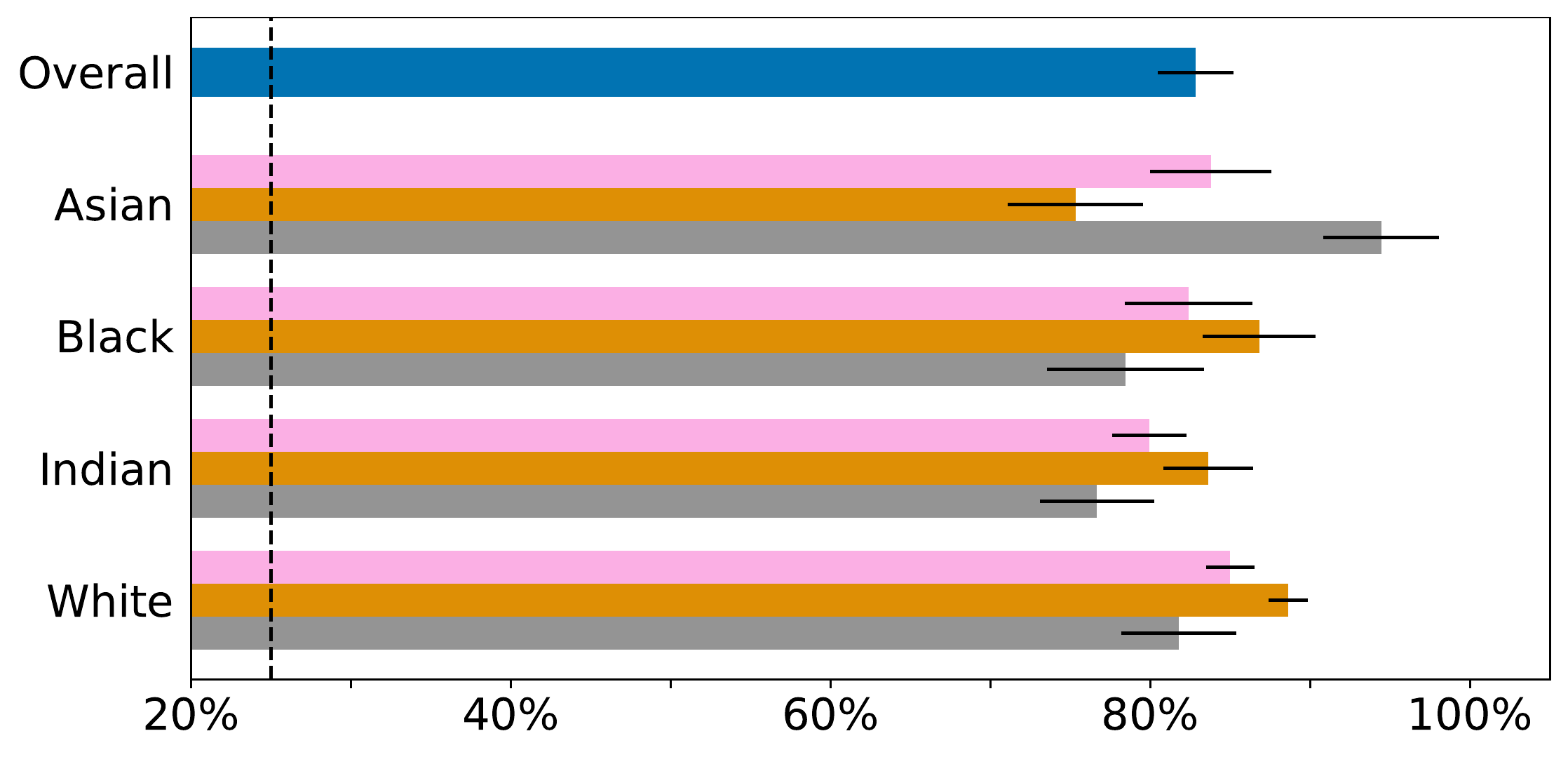}
         \caption{Racial Appearance (CelebAHQ)}
     \end{subfigure}

    \caption{Evaluation results for CAIA performed on ResNeSt-101 models to infer four different target attributes. The black horizontal lines denote the standard deviation over nine runs. We further state random guessing (dashed line) for comparison.}
\end{figure*}
\clearpage

\subsection{ResNeSt-101 - CelebA (1000 identities)}
\begin{figure*}[h!]
\centering
     \begin{subfigure}[c]{\textwidth}
         \centering
         \includegraphics[width=0.75\textwidth]{images/barplots/legend_small.pdf}
     \end{subfigure}
     
     \begin{subfigure}[b]{0.48\textwidth}
        \centering
         \includegraphics[width=\textwidth]{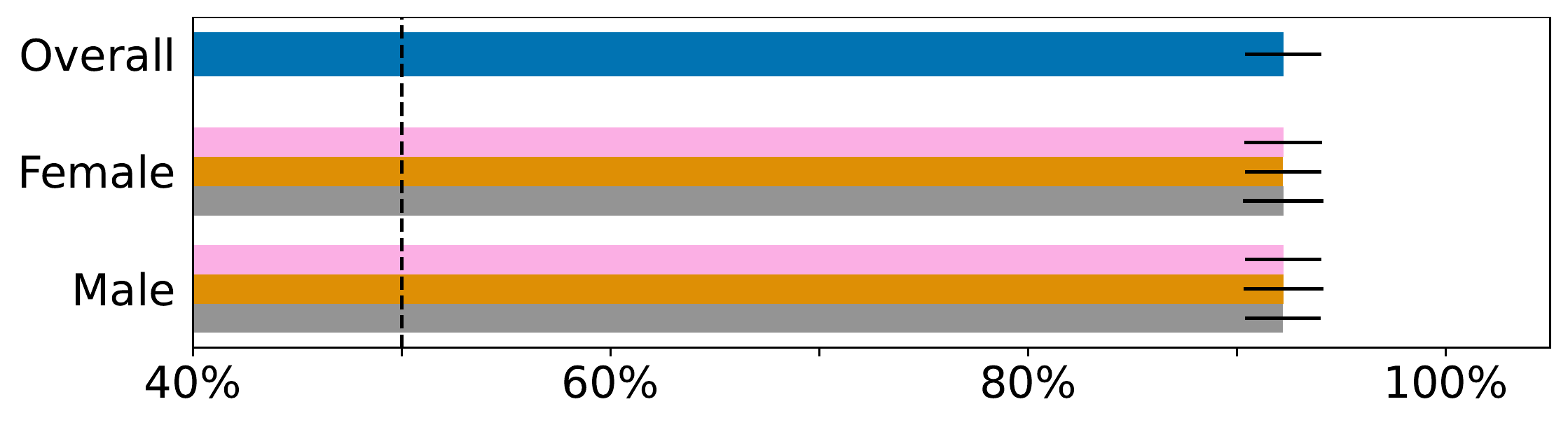}
         \caption{Gender (FFHQ)}
     \end{subfigure}
     \hfill
     \begin{subfigure}[b]{0.48\textwidth}
        \centering
         \includegraphics[width=\textwidth]{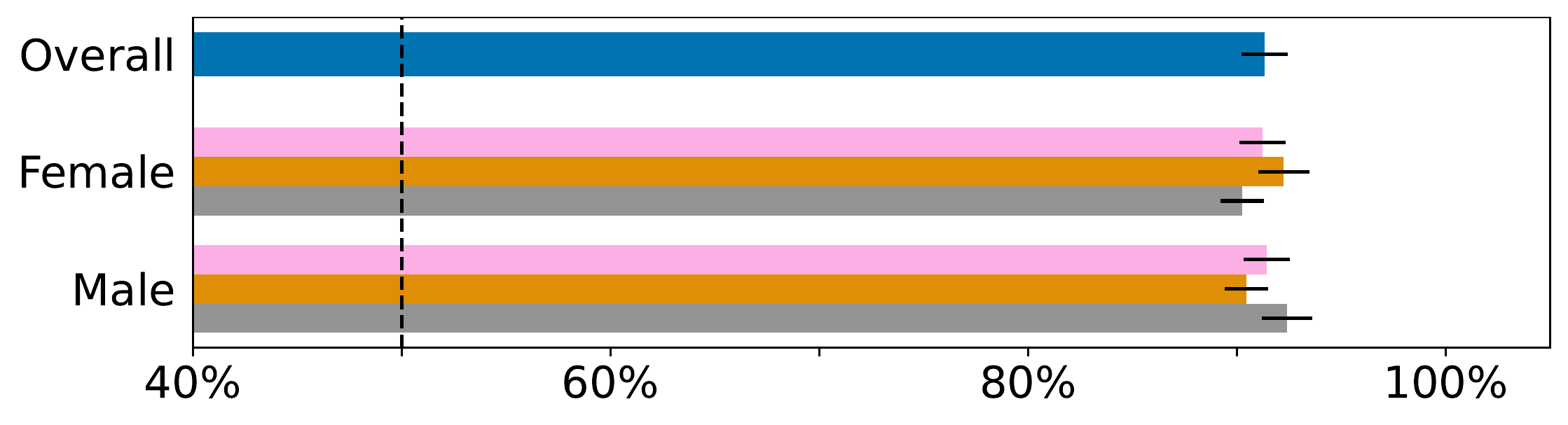}
         \caption{Gender (CelebAHQ)}
     \end{subfigure}

     \par\medskip
     \begin{subfigure}[b]{0.48\textwidth}
         \centering
         \includegraphics[width=\textwidth]{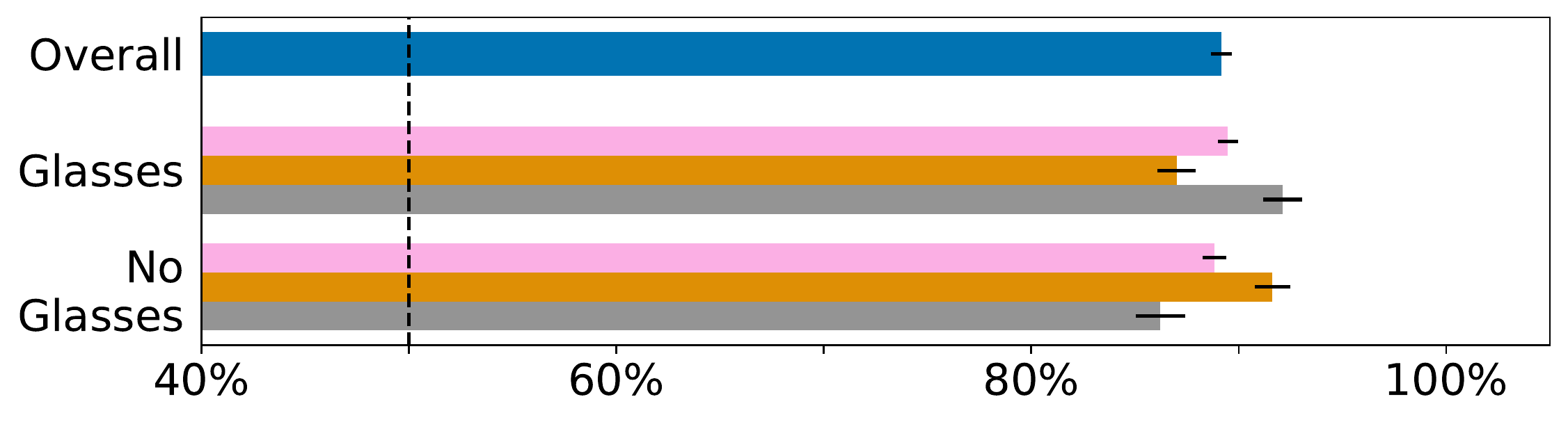}
         \caption{Eyeglasses (FFHQ)}
     \end{subfigure}
     \hfill
     \begin{subfigure}[b]{0.48\textwidth}
         \centering
         \includegraphics[width=\textwidth]{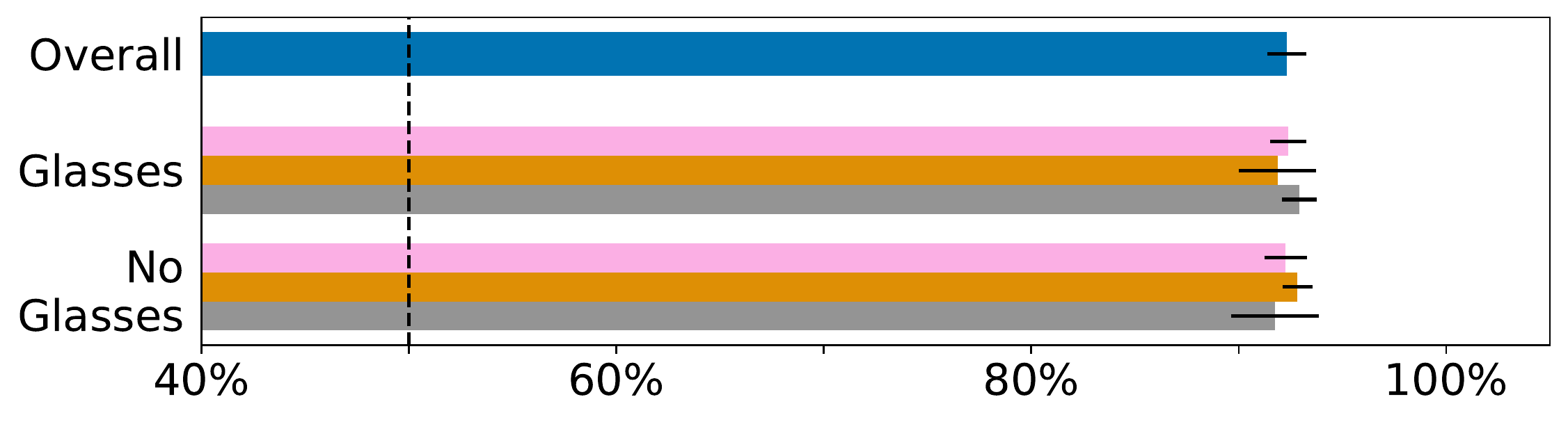}
         \caption{Eyeglasses (CelebAHQ)}
     \end{subfigure}
     
     \par\medskip
     \begin{subfigure}[b]{0.48\textwidth}
         \centering
         \includegraphics[width=\textwidth]{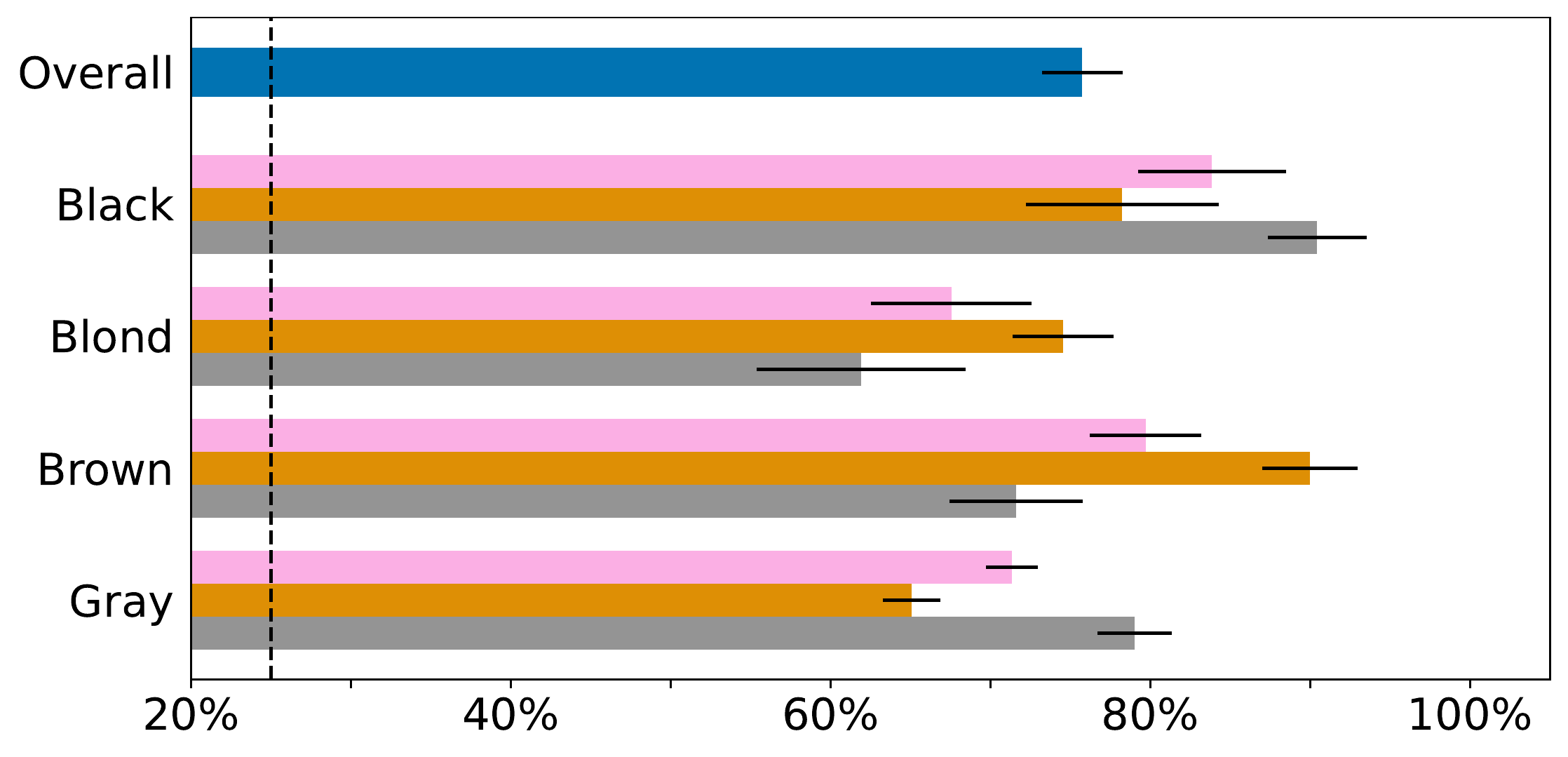}
         \caption{Hair Color (FFHQ)}
     \end{subfigure}
     \hfill
     \begin{subfigure}[b]{0.48\textwidth}
         \centering
         \includegraphics[width=\textwidth]{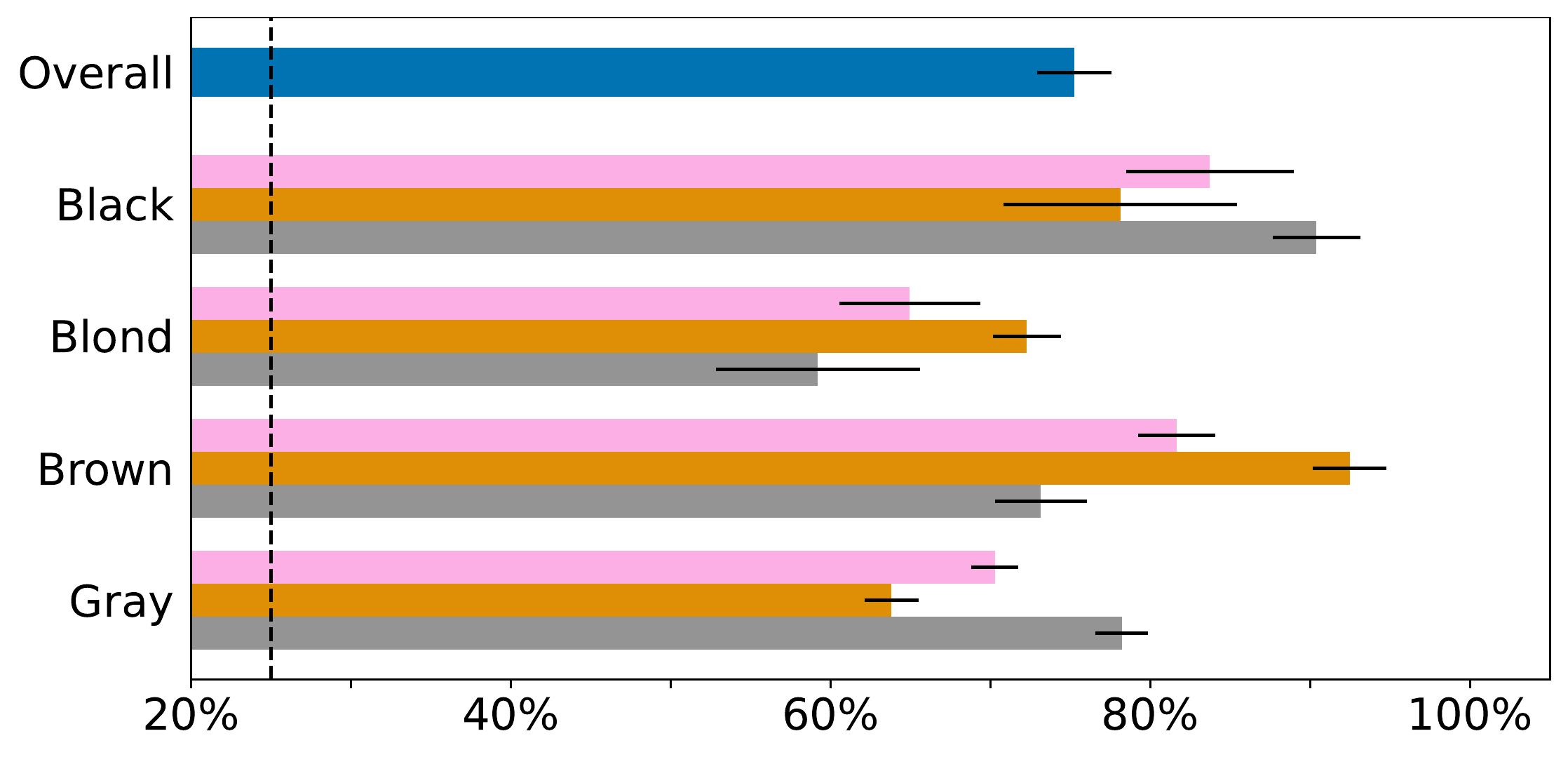}
         \caption{Hair Color (CelebAHQ)}
     \end{subfigure}

     \par\medskip
     \begin{subfigure}[b]{0.48\textwidth}
         \centering
         \includegraphics[width=\textwidth]{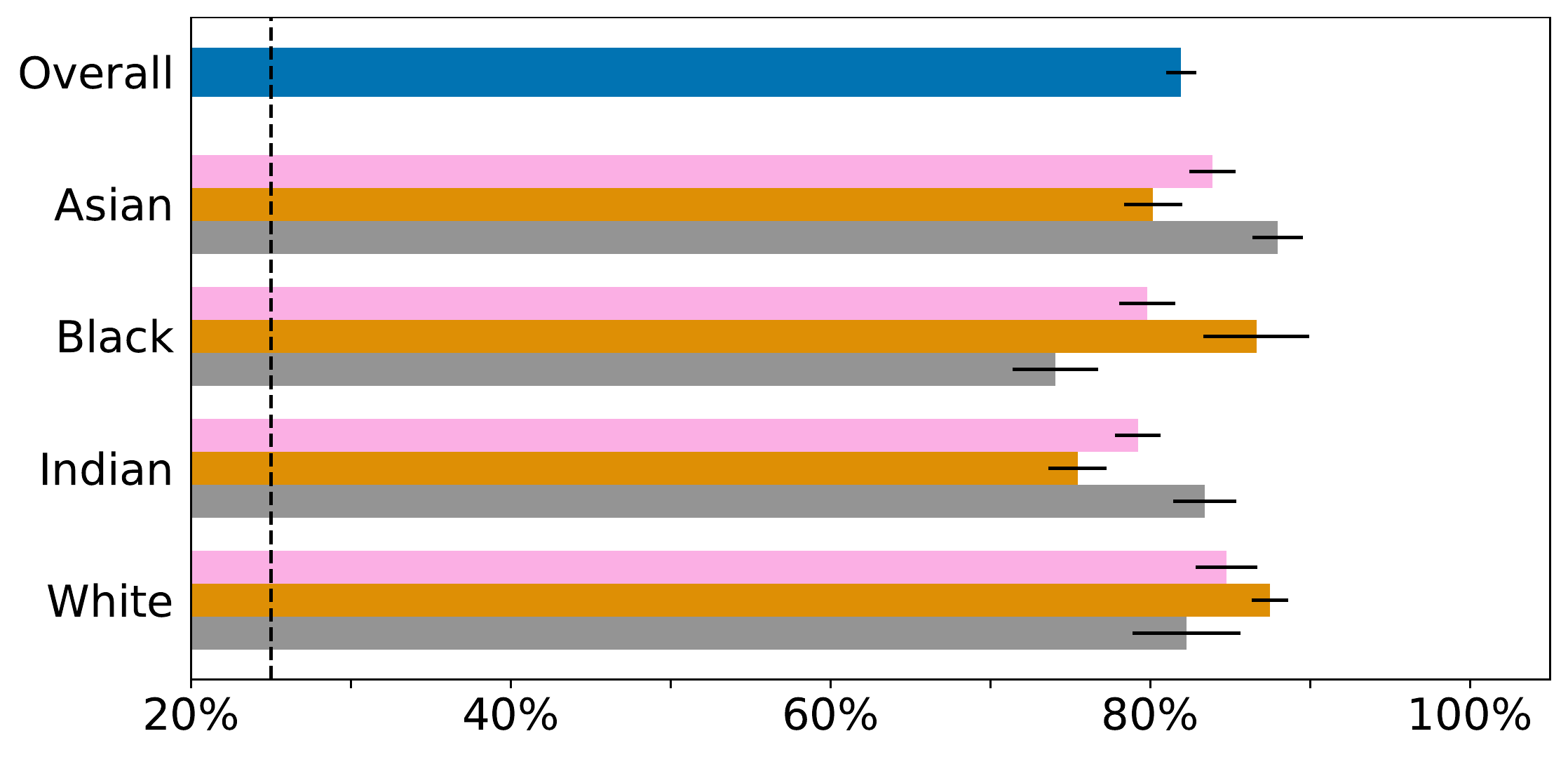}
         \caption{Racial Appearance (FFHQ)}
     \end{subfigure}
    \hfill
     \begin{subfigure}[b]{0.48\textwidth}
         \centering
         \includegraphics[width=\textwidth]{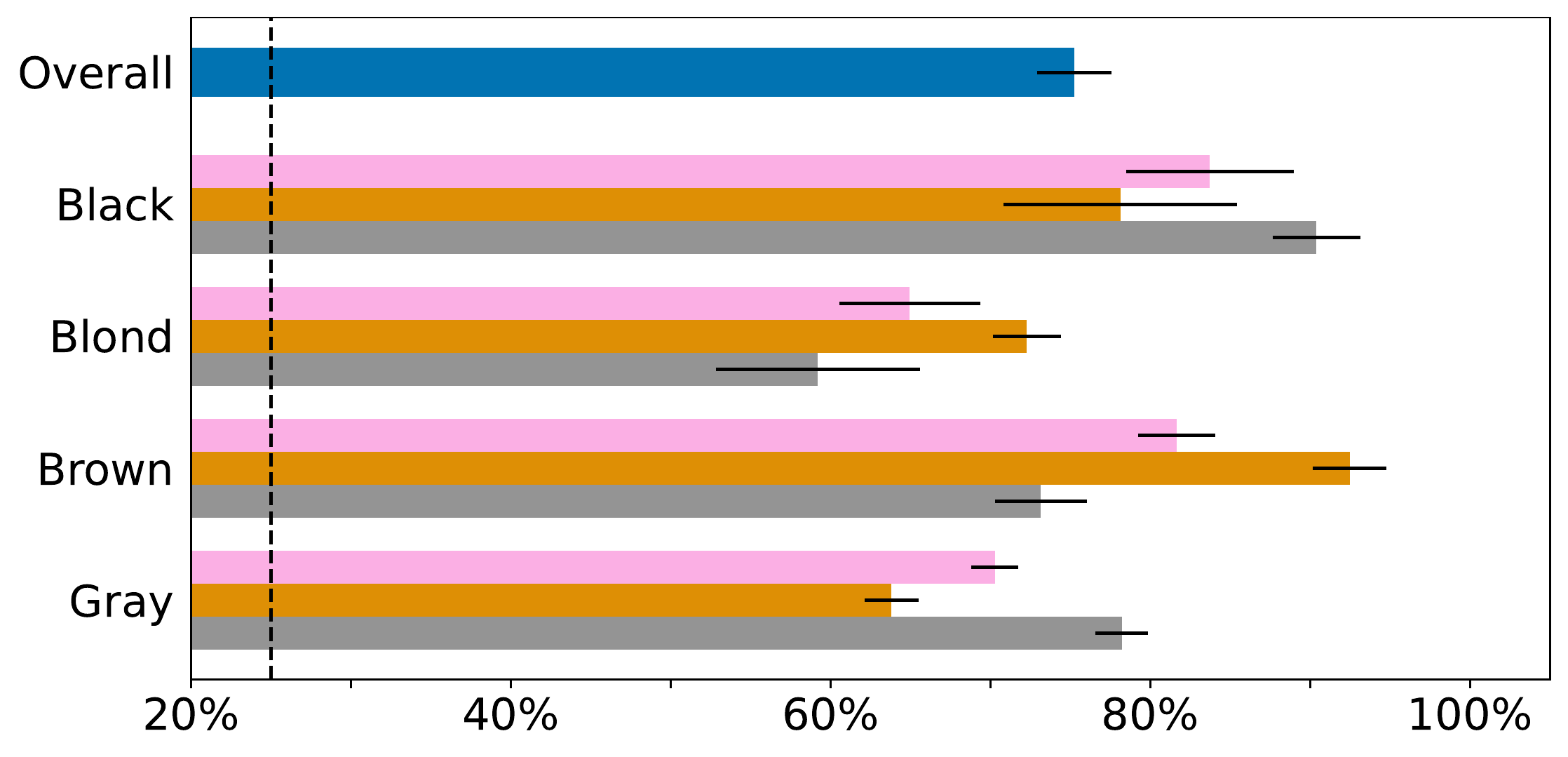}
         \caption{Racial Appearance (CelebAHQ)}
     \end{subfigure}

    \caption{Evaluation results for CAIA performed on ResNeSt-101 models to infer four different target attributes for \textbf{1000} identities. The black horizontal lines denote the standard deviation over nine runs. We further state random guessing (dashed line) for comparison.}
\end{figure*}
\clearpage

\clearpage
\section{Additional FaceScrub Experimental Results}\label{appx:add_results_Facescrub}

\subsection{ResNet-18 - FaceScrub (Cropped)}
\begin{figure*}[h!]
\centering
     \begin{subfigure}[c]{\textwidth}
         \centering
         \includegraphics[width=\textwidth]{images/robust_legend_long.pdf}
     \end{subfigure}
     
     \begin{subfigure}[b]{0.7\textwidth}
        \centering
         \includegraphics[width=\textwidth]{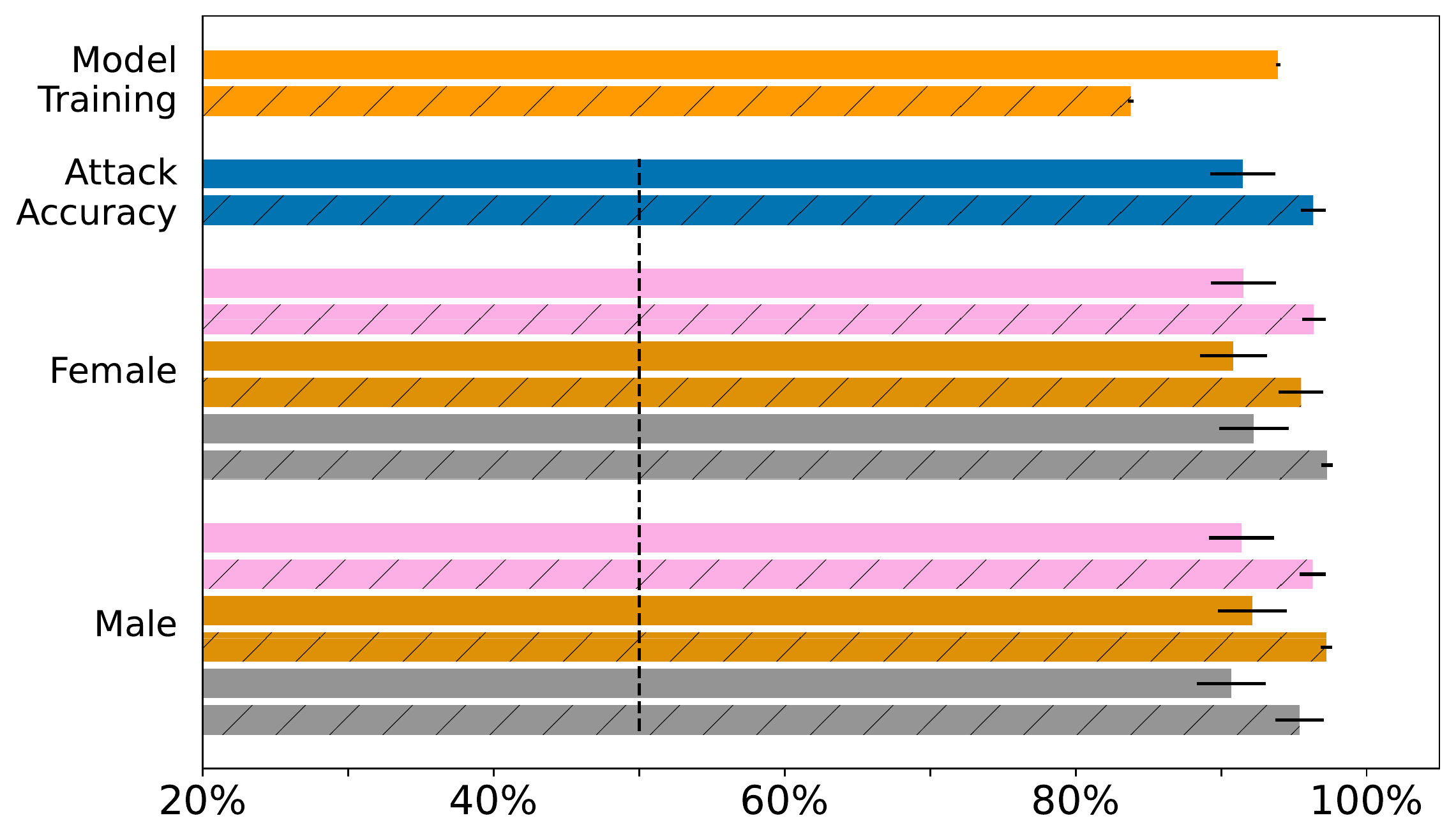}
         \caption{Gender (FFHQ)}
     \end{subfigure}
     
     \begin{subfigure}[b]{0.7\textwidth}
        \centering
         \includegraphics[width=\textwidth]{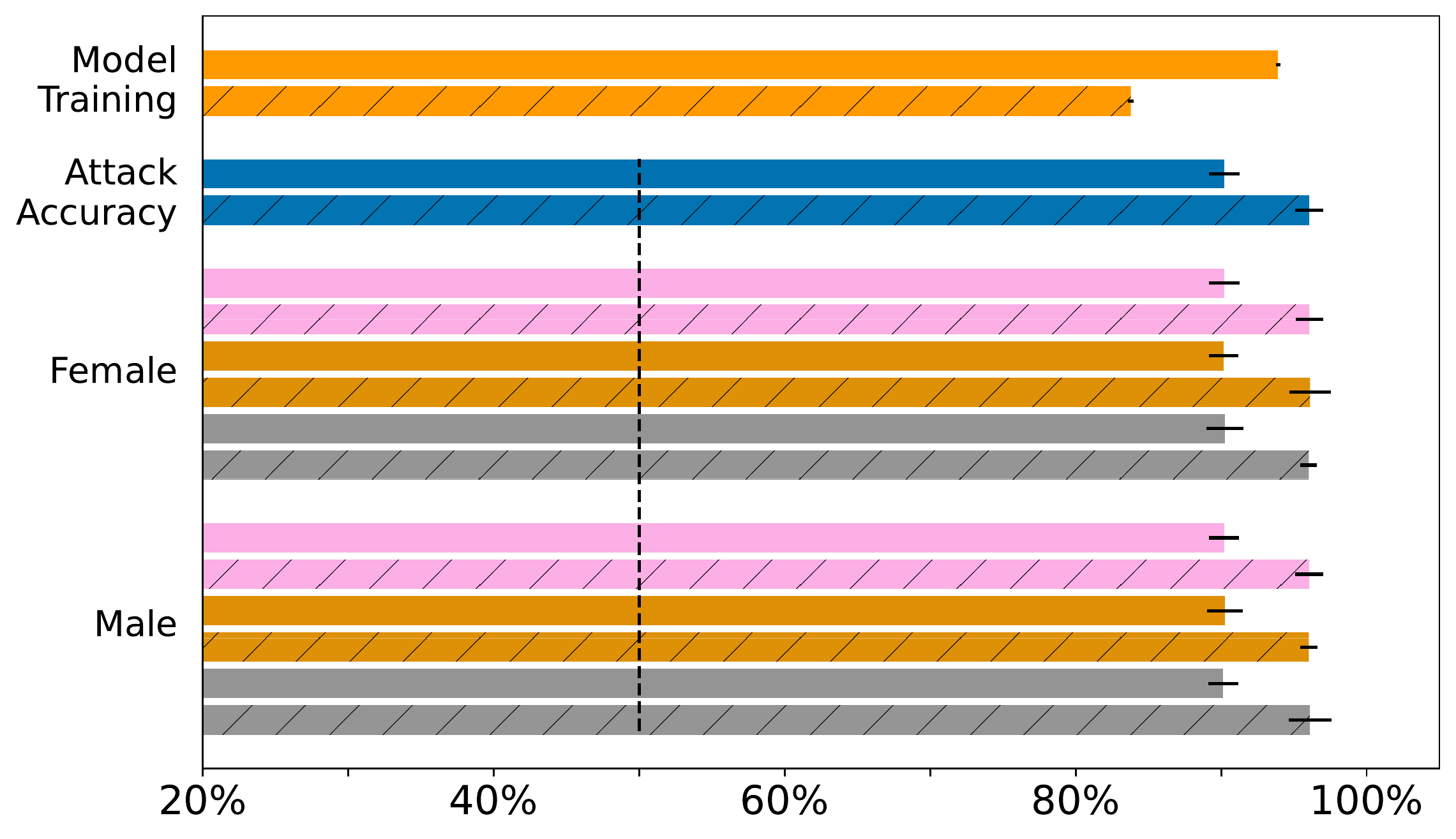}
         \caption{Gender (CelebAHQ)}
     \end{subfigure}

     \caption{Evaluation results for CAIA performed on ResNet-18 models to infer the gender appearance. The black horizontal lines denote the standard deviation over nine runs. We further state random guessing (dashed line) for comparison. The models were trained on the cropped FaceScrub dataset.}

\end{figure*}
\clearpage

\subsection{ResNet-101 - FaceScrub (Cropped)}
\begin{figure*}[h!]
\centering
     \begin{subfigure}[c]{\textwidth}
         \centering
         \includegraphics[width=\textwidth]{images/robust_legend_long.pdf}
     \end{subfigure}
     
     \begin{subfigure}[b]{0.7\textwidth}
        \centering
         \includegraphics[width=\textwidth]{images/facescrub_results/facescrub_robust_ffhq_resnet101.pdf}
         \caption{Gender (FFHQ)}
     \end{subfigure}
     \begin{subfigure}[b]{0.7\textwidth}
        \centering
         \includegraphics[width=\textwidth]{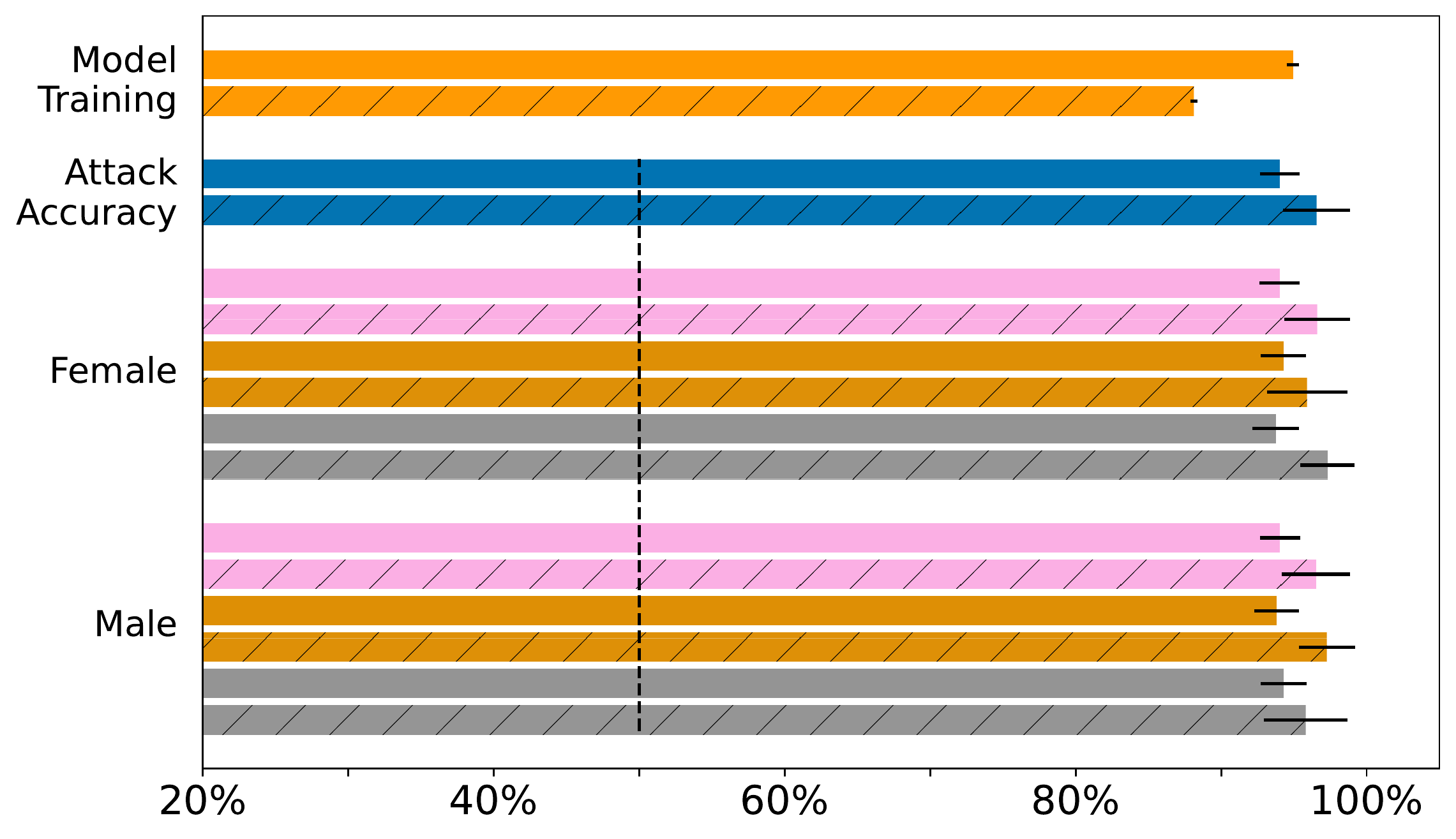}
         \caption{Gender (CelebAHQ)}
     \end{subfigure}
         
     \caption{Evaluation results for CAIA performed on ResNet-101 models to infer the gender appearance. The black horizontal lines denote the standard deviation over nine runs. We further state random guessing (dashed line) for comparison. The models were trained on the cropped FaceScrub dataset.}

\end{figure*}
\clearpage

\subsection{ResNet-152 - FaceScrub (Cropped)}
\begin{figure*}[h!]
\centering
     \begin{subfigure}[c]{\textwidth}
         \centering
         \includegraphics[width=\textwidth]{images/robust_legend_long.pdf}
     \end{subfigure}
     
     \begin{subfigure}[b]{0.7\textwidth}
        \centering
         \includegraphics[width=\textwidth]{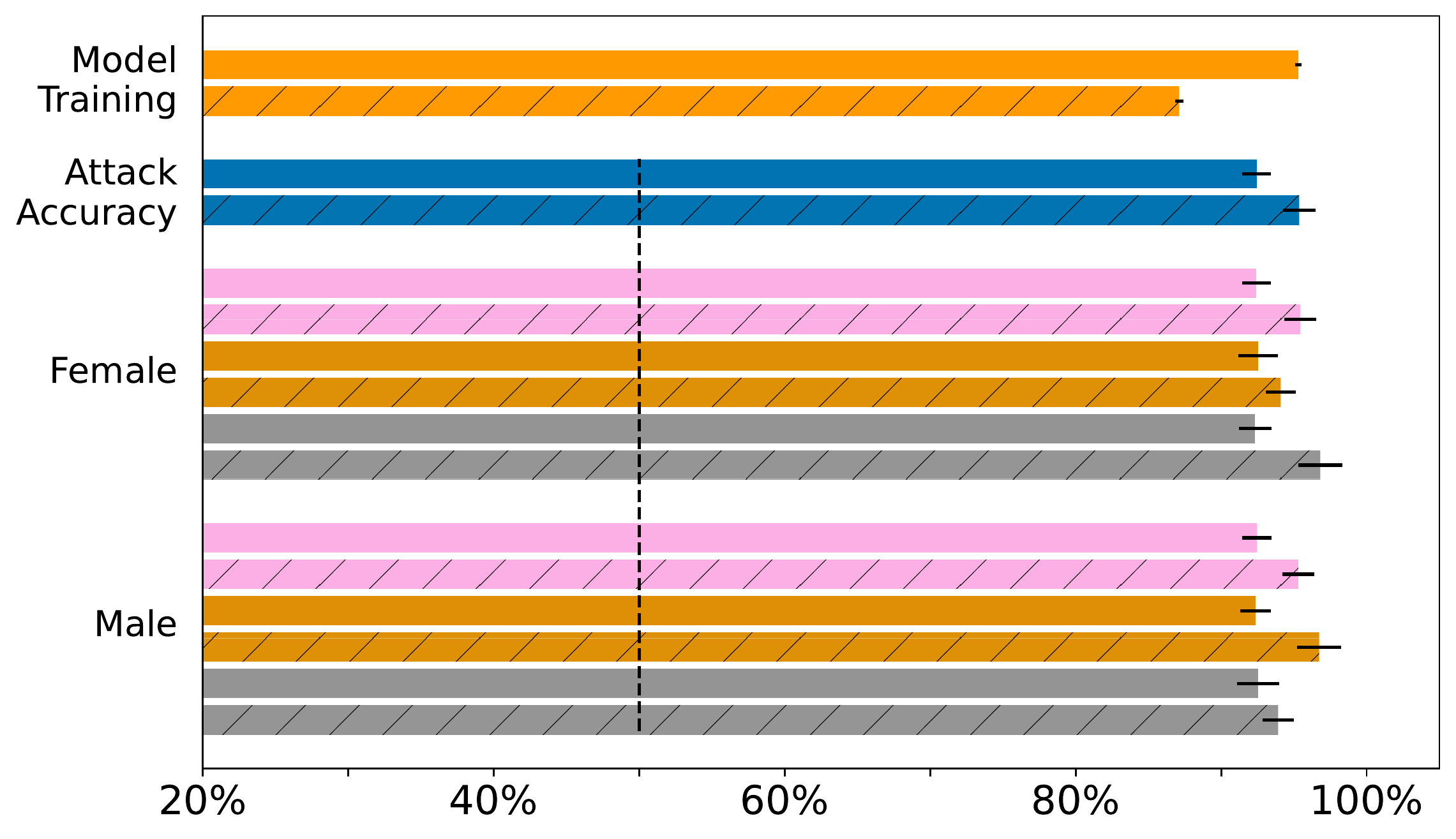}
         \caption{Gender (FFHQ)}
     \end{subfigure}
     
     \begin{subfigure}[b]{0.7\textwidth}
        \centering
         \includegraphics[width=\textwidth]{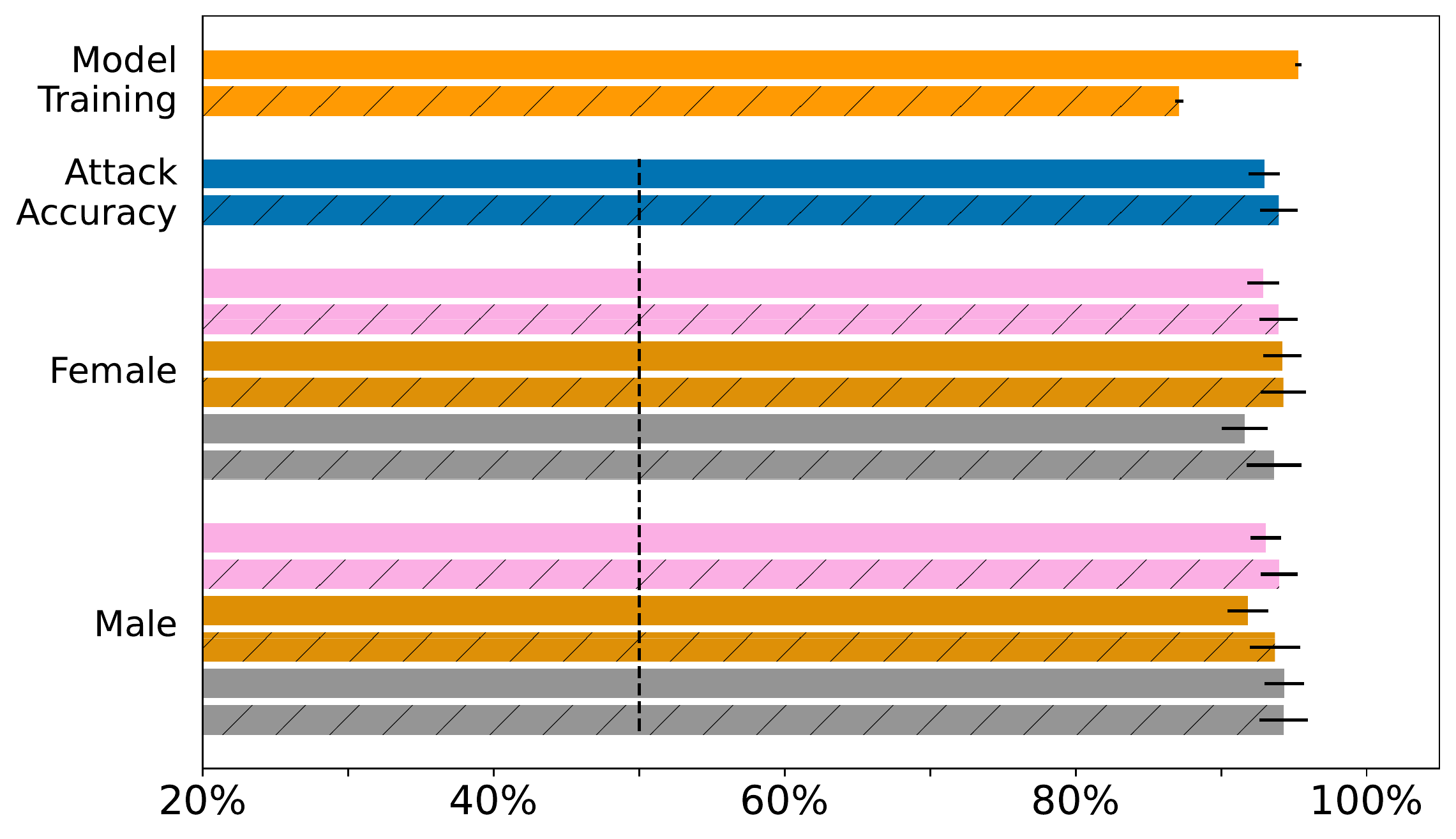}
         \caption{Gender (CelebAHQ)}
     \end{subfigure}

     \caption{Evaluation results for CAIA performed on ResNet-152 models to infer the gender appearance. The black horizontal lines denote the standard deviation over nine runs. We further state random guessing (dashed line) for comparison. The models were trained on the cropped FaceScrub dataset.}
\end{figure*}
\clearpage

\subsection{DenseNet-169 - FaceScrub (Cropped)}
\begin{figure*}[h!]
\centering
     \begin{subfigure}[c]{\textwidth}
         \centering
         \includegraphics[width=\textwidth]{images/robust_legend_long.pdf}
     \end{subfigure}
     
     \begin{subfigure}[b]{0.7\textwidth}
        \centering
         \includegraphics[width=\textwidth]{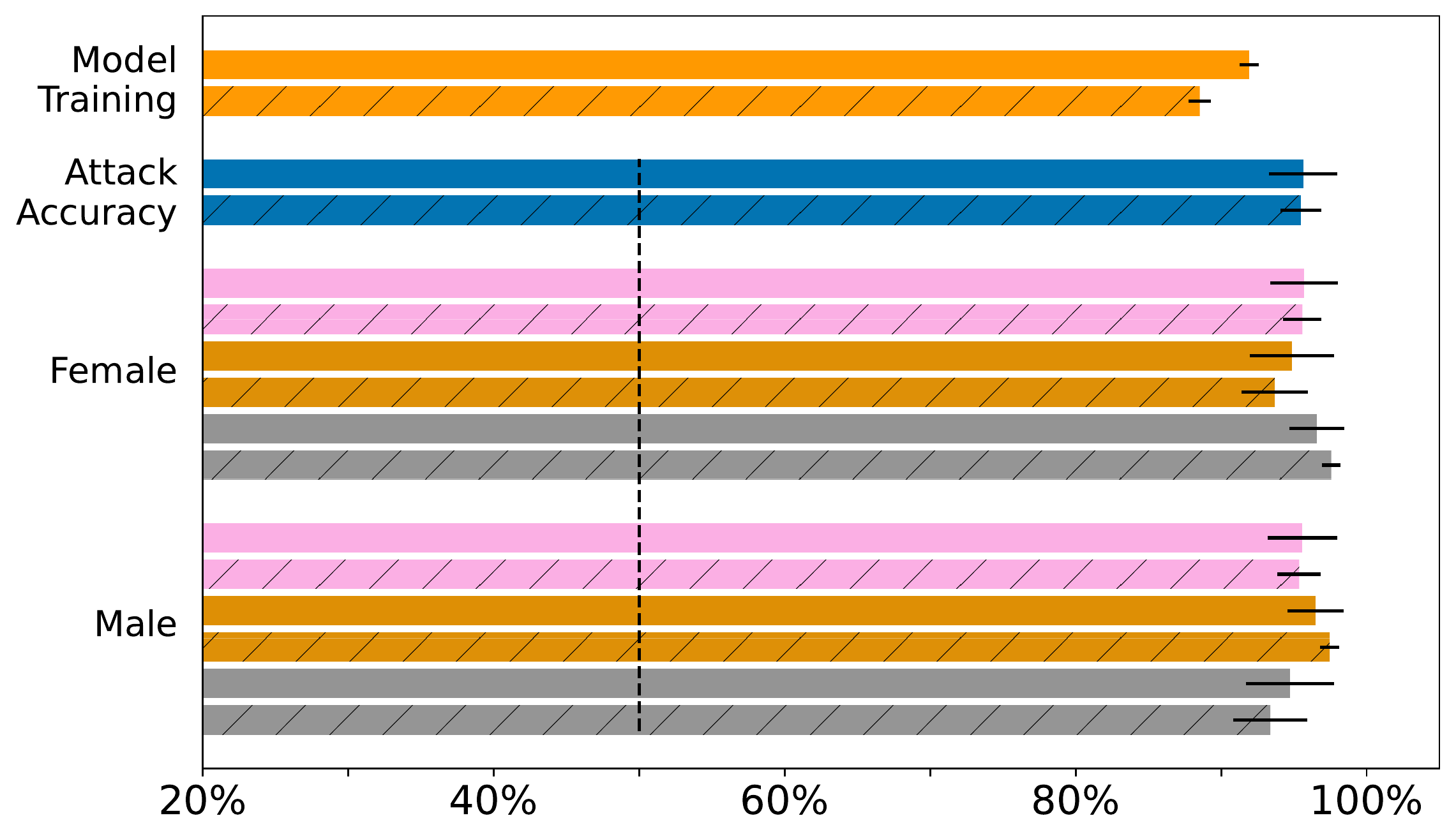}
         \caption{Gender (FFHQ)}
     \end{subfigure}
     
     \begin{subfigure}[b]{0.7\textwidth}
        \centering
         \includegraphics[width=\textwidth]{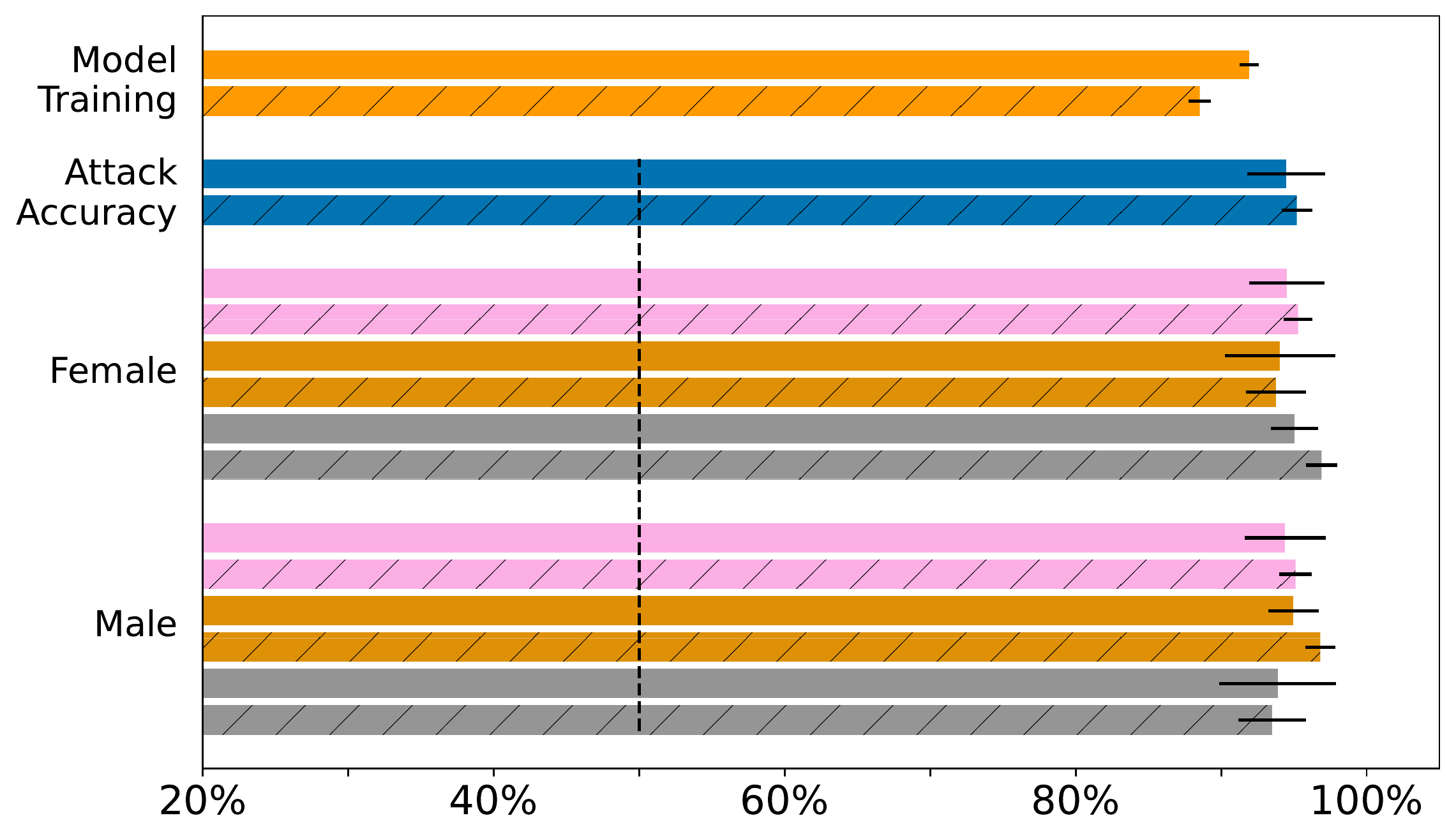}
         \caption{Gender (CelebAHQ)}
     \end{subfigure}
     
     \caption{Evaluation results for CAIA performed on DenseNet-169 models to infer the gender appearance. The black horizontal lines denote the standard deviation over nine runs. We further state random guessing (dashed line) for comparison. The models were trained on the cropped FaceScrub dataset.}
\end{figure*}
\clearpage

\subsection{ResNeSt-101 - FaceScrub (Cropped)}
\begin{figure*}[h!]
\centering
     \begin{subfigure}[c]{\textwidth}
         \centering
         \includegraphics[width=\textwidth]{images/robust_legend_long.pdf}
     \end{subfigure}
     
     \begin{subfigure}[b]{0.7\textwidth}
        \centering
         \includegraphics[width=\textwidth]{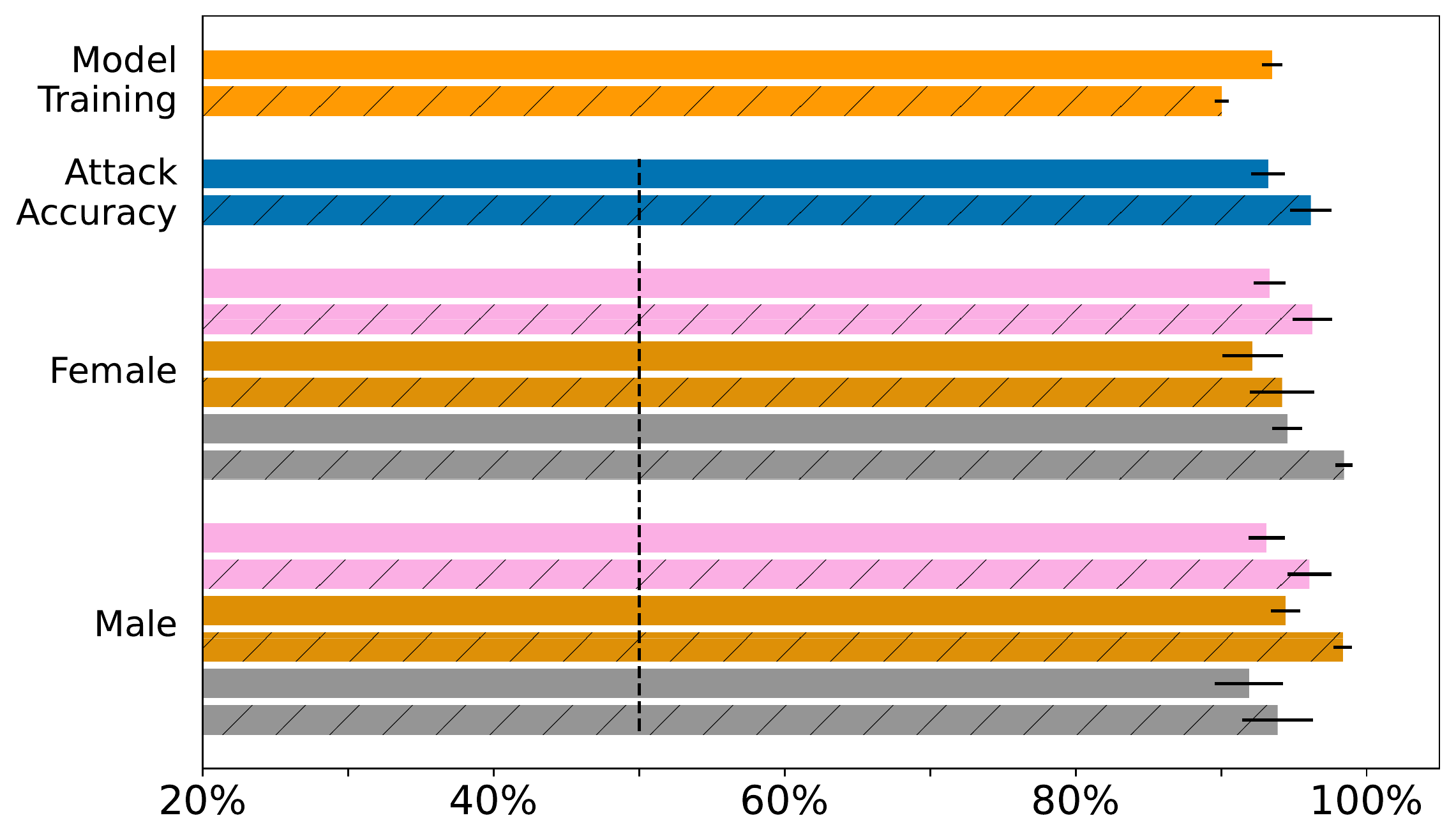}
         \caption{Gender (FFHQ)}
     \end{subfigure}
     
     \begin{subfigure}[b]{0.7\textwidth}
        \centering
         \includegraphics[width=\textwidth]{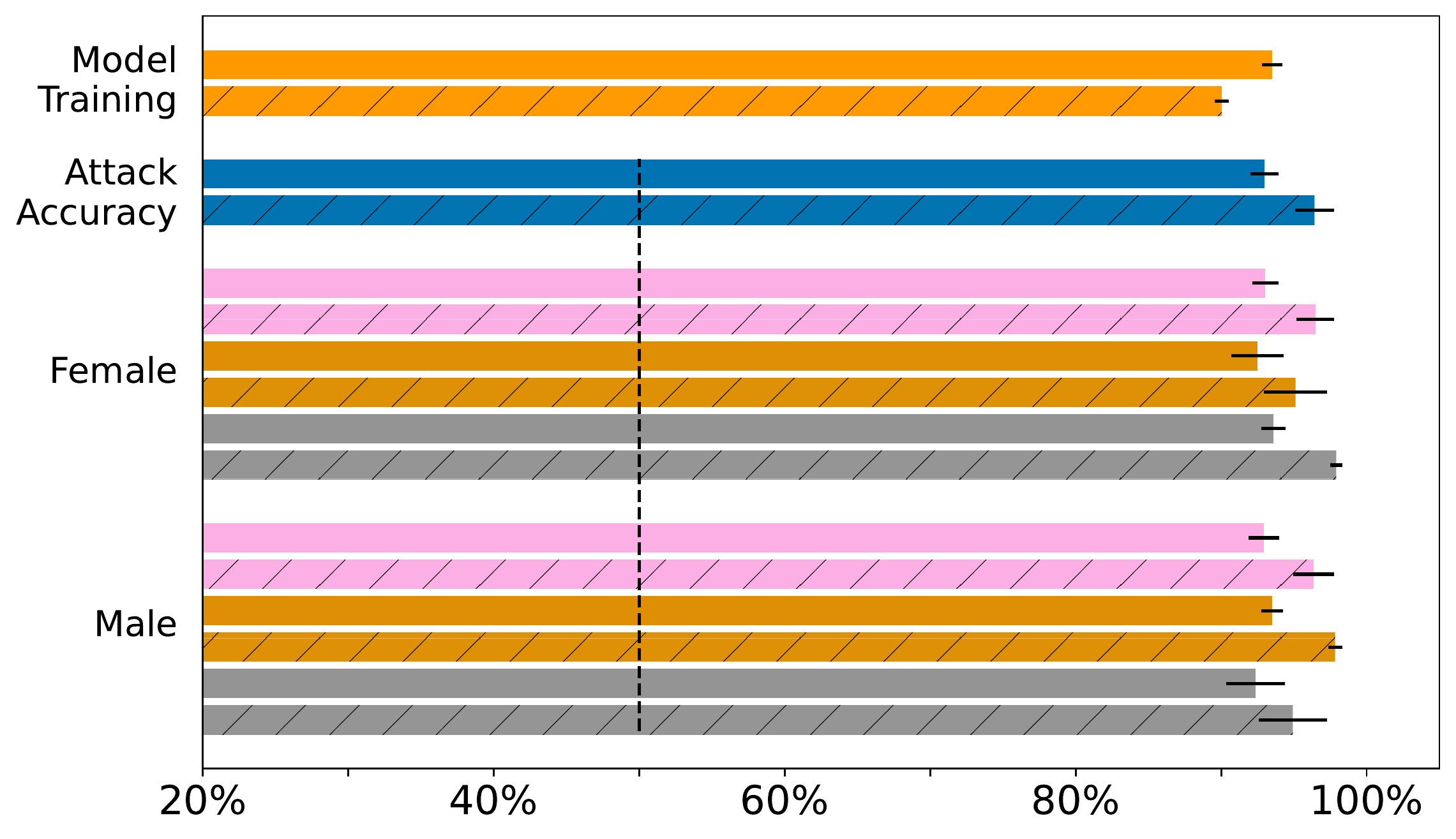}
         \caption{Gender (CelebAHQ)}
     \end{subfigure}
     
     \caption{Evaluation results for CAIA performed on ResNeSt-101 models to infer the gender appearance. The black horizontal lines denote the standard deviation over nine runs. We further state random guessing (dashed line) for comparison. The models were trained on the cropped FaceScrub dataset.}
\end{figure*}
\clearpage

\subsection{ResNet-18 - FaceScrub (Uncropped)}

\begin{figure*}[h!]
\centering
     \begin{subfigure}[c]{\textwidth}
         \centering
         \includegraphics[width=\textwidth]{images/robust_legend_long.pdf}
     \end{subfigure}
     
     \begin{subfigure}[b]{0.7\textwidth}
        \centering
         \includegraphics[width=\textwidth]{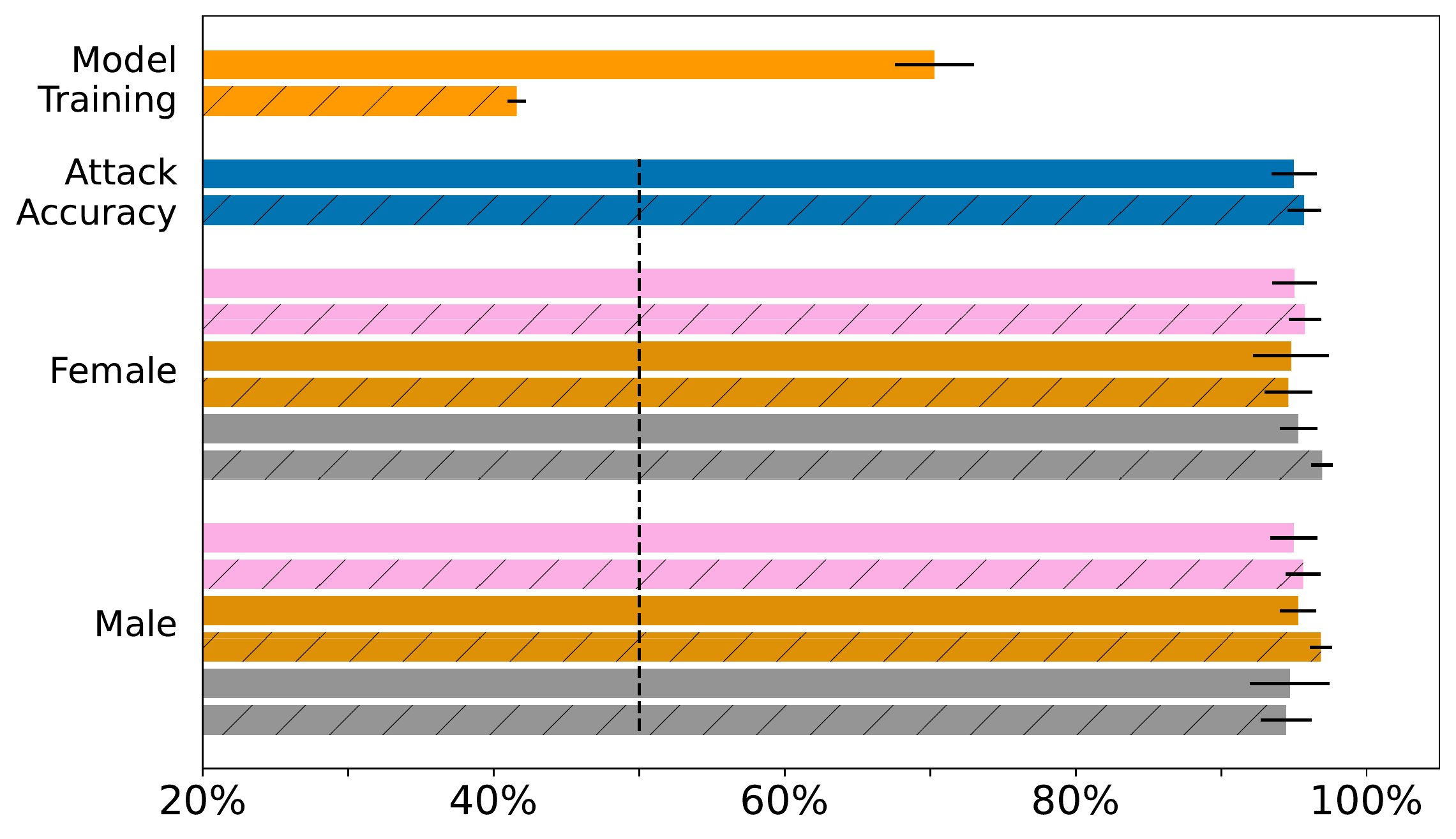}
         \caption{Gender (FFHQ)}
     \end{subfigure}

     \begin{subfigure}[b]{0.7\textwidth}
        \centering
         \includegraphics[width=\textwidth]{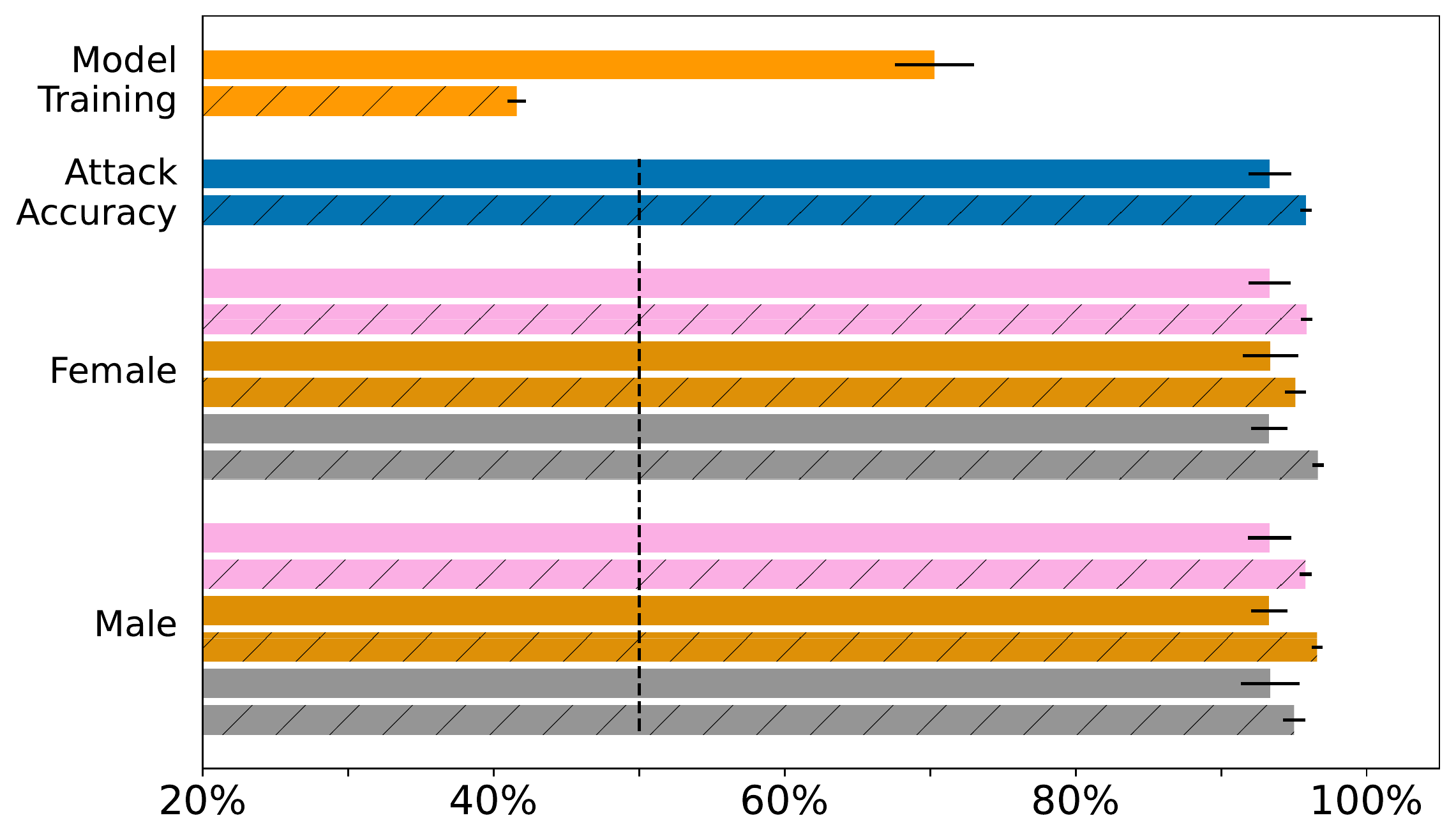}
         \caption{Gender (CelebAHQ)}
     \end{subfigure}

     \caption{Evaluation results for CAIA performed on ResNet-18 models to infer the gender appearance. The black horizontal lines denote the standard deviation over nine runs. We further state random guessing (dashed line) for comparison. The models were trained on the cropped FaceScrub dataset.}
\end{figure*}
\clearpage

\subsection{ResNet-101 - FaceScrub (Uncropped)}

\begin{figure*}[h!]
\centering
     \begin{subfigure}[c]{\textwidth}
         \centering
         \includegraphics[width=\textwidth]{images/robust_legend_long.pdf}
     \end{subfigure}
     
     \begin{subfigure}[b]{0.7\textwidth}
        \centering
         \includegraphics[width=\textwidth]{images/facescrub_results/facescrub_uncropped_robust_ffhq_resnet101.pdf}
         \caption{Gender (FFHQ)}
     \end{subfigure}
     
     \begin{subfigure}[b]{0.7\textwidth}
        \centering
         \includegraphics[width=\textwidth]{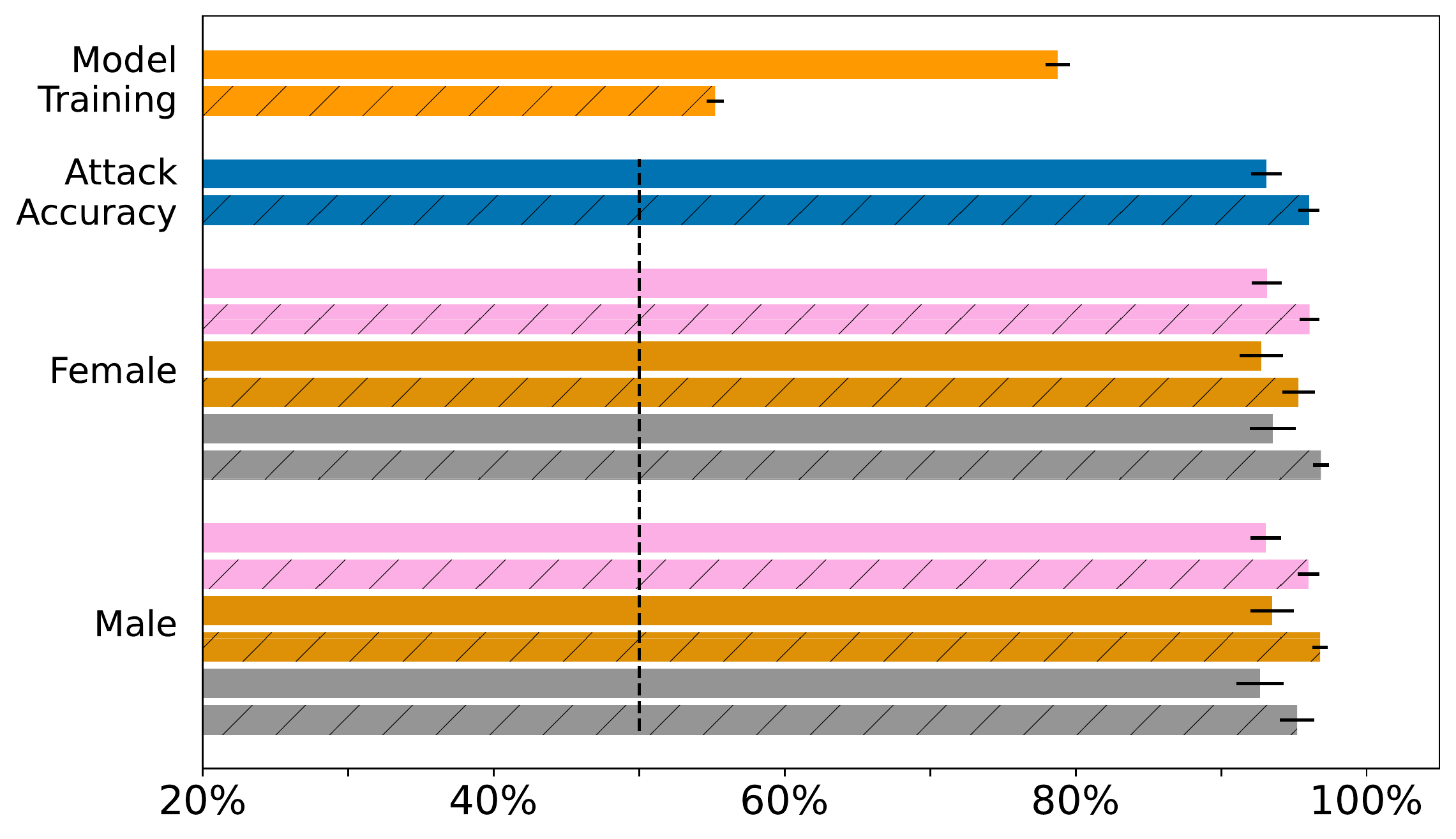}
         \caption{Gender (CelebAHQ)}
     \end{subfigure}

     \caption{Evaluation results for CAIA performed on ResNet-101 models to infer the gender appearance. The black horizontal lines denote the standard deviation over nine runs. We further state random guessing (dashed line) for comparison. The models were trained on the cropped FaceScrub dataset.}
\end{figure*}
\clearpage

\subsection{ResNet-152 - FaceScrub (Uncropped)}

\begin{figure*}[h!]
\centering
     \begin{subfigure}[c]{\textwidth}
         \centering
         \includegraphics[width=\textwidth]{images/robust_legend_long.pdf}
     \end{subfigure}
     
     \begin{subfigure}[b]{0.7\textwidth}
        \centering
         \includegraphics[width=\textwidth]{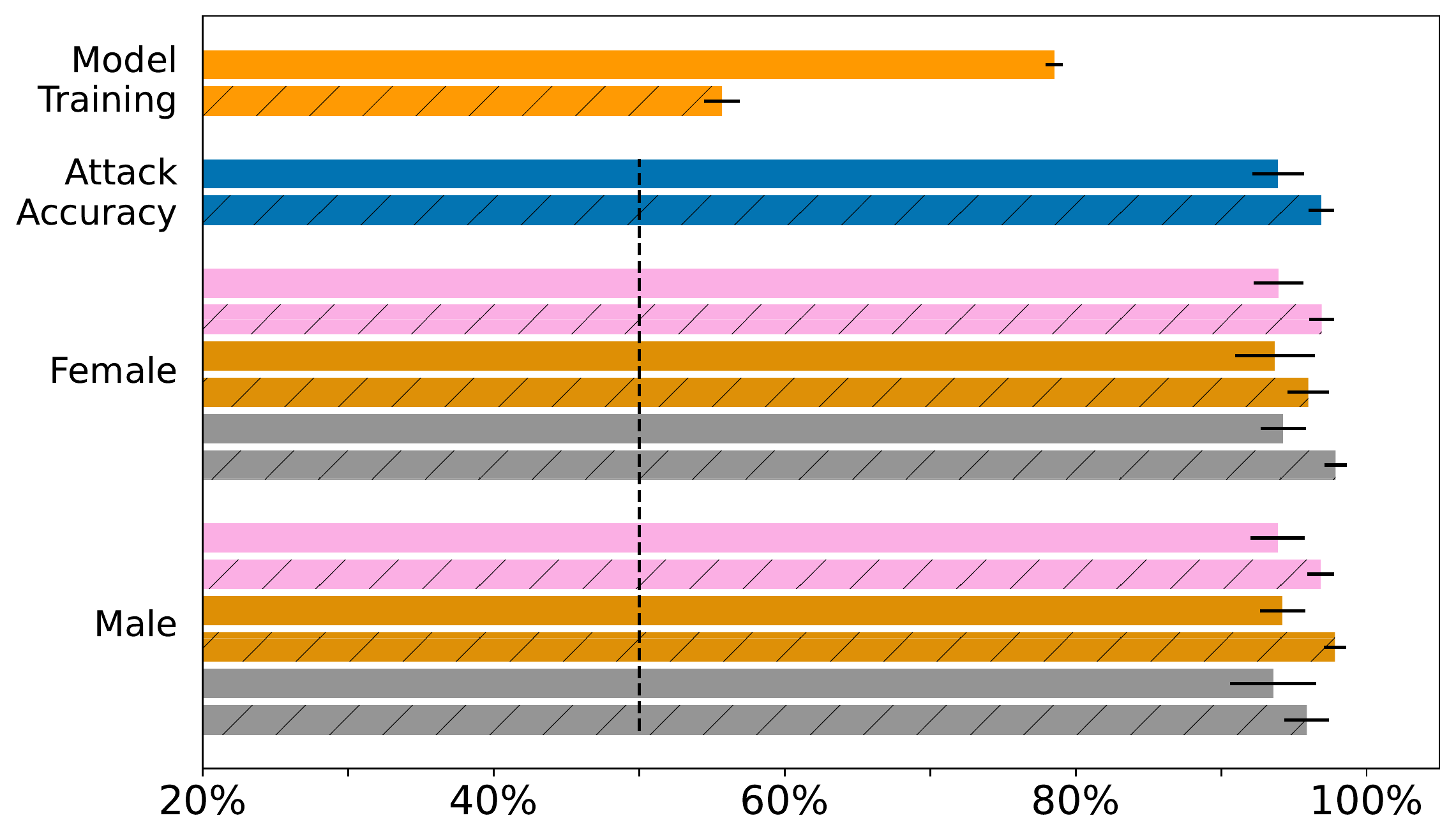}
         \caption{Gender (FFHQ)}
     \end{subfigure}
     \begin{subfigure}[b]{0.7\textwidth}
        \centering
         \includegraphics[width=\textwidth]{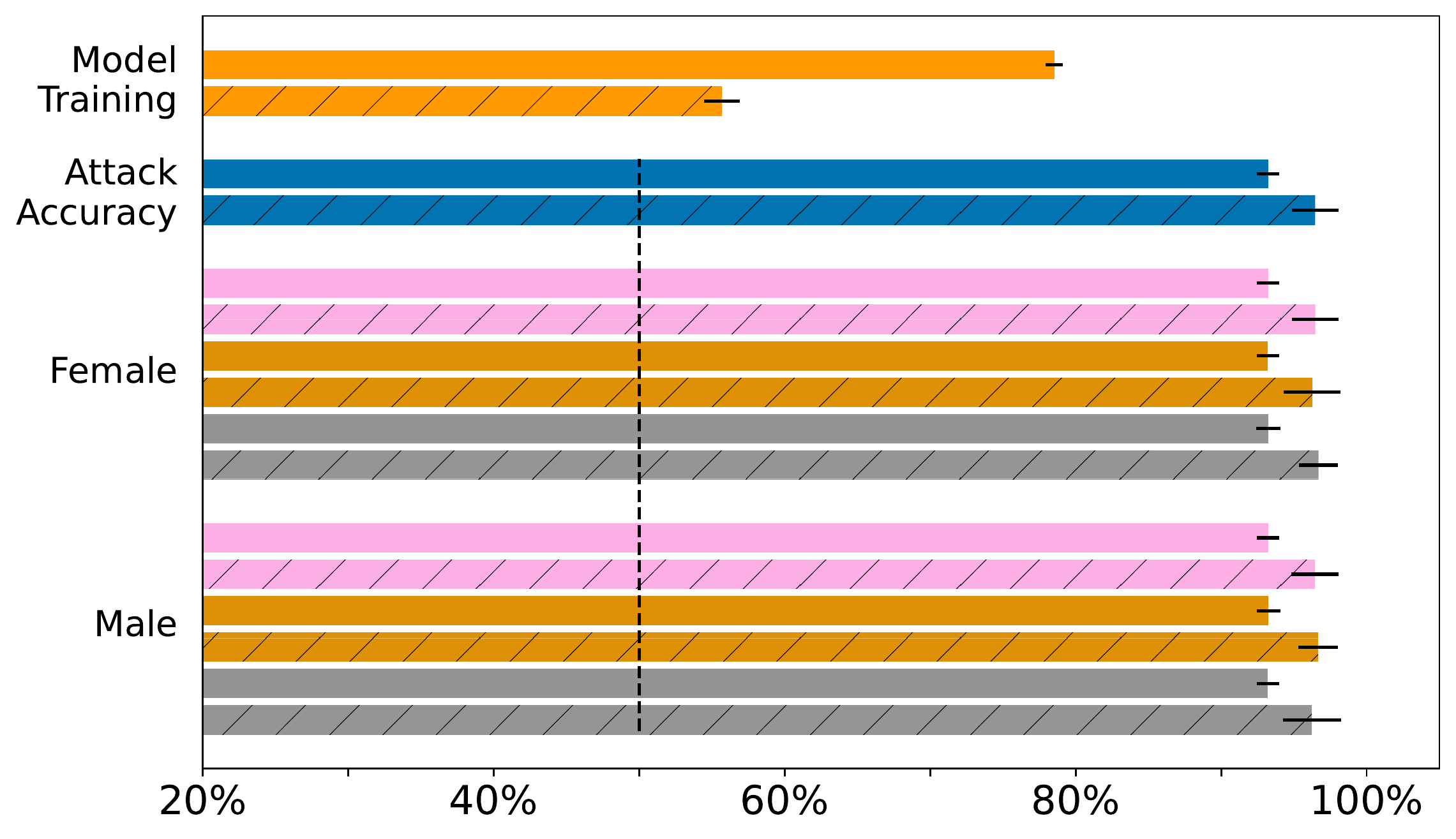}
         \caption{Gender (CelebAHQ)}
     \end{subfigure}

     \caption{Evaluation results for CAIA performed on ResNet-152 models to infer the gender appearance. The black horizontal lines denote the standard deviation over nine runs. We further state random guessing (dashed line) for comparison. The models were trained on the cropped FaceScrub dataset.}
\end{figure*}
\clearpage

\subsection{DenseNet-169 - FaceScrub (Uncropped)}

\begin{figure*}[h!]
\centering
     \begin{subfigure}[c]{\textwidth}
         \centering
         \includegraphics[width=\textwidth]{images/robust_legend_long.pdf}
     \end{subfigure}
     
     \begin{subfigure}[b]{0.7\textwidth}
        \centering
         \includegraphics[width=\textwidth]{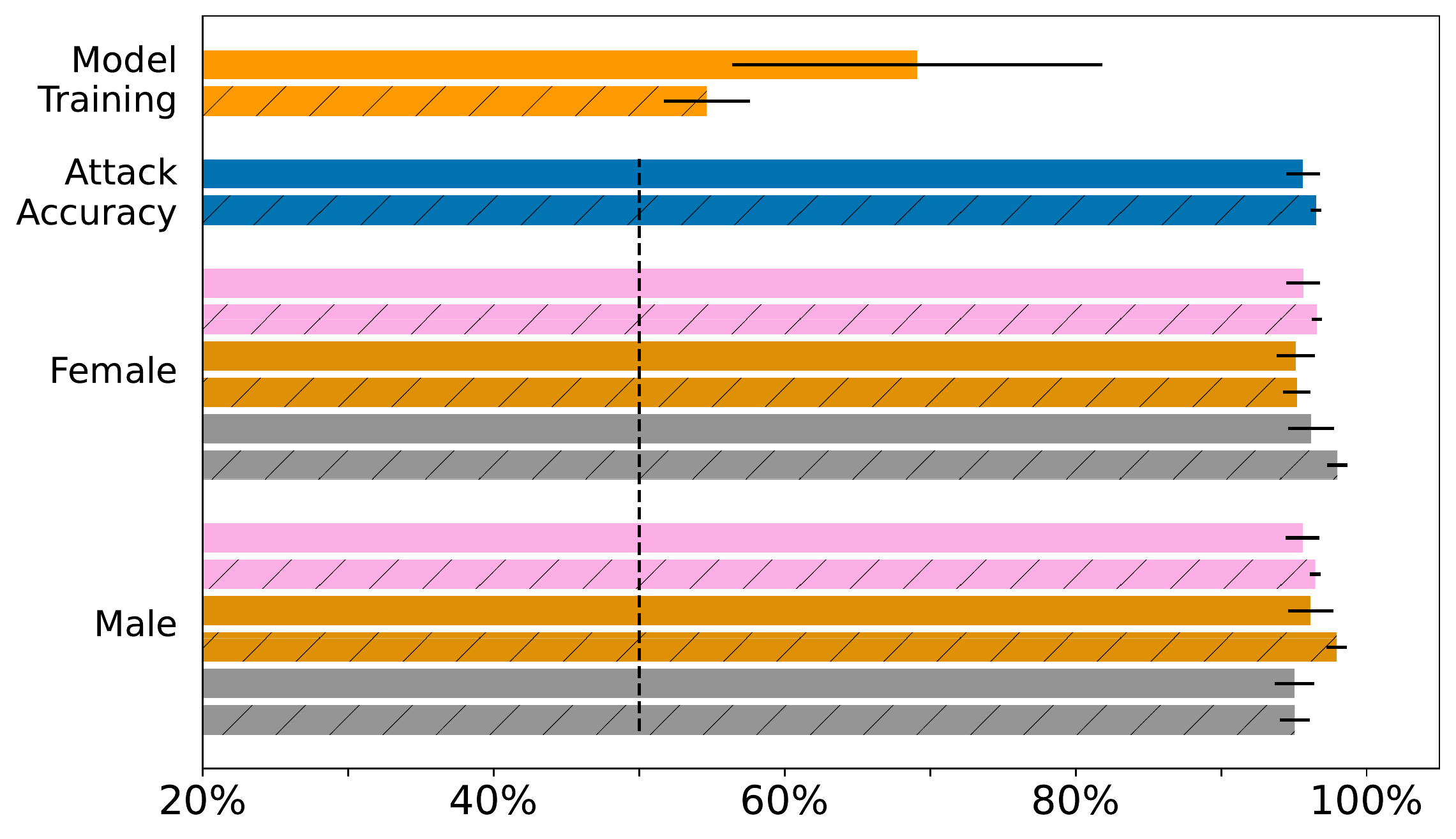}
         \caption{Gender (FFHQ)}
     \end{subfigure}
     \begin{subfigure}[b]{0.7\textwidth}
        \centering
         \includegraphics[width=\textwidth]{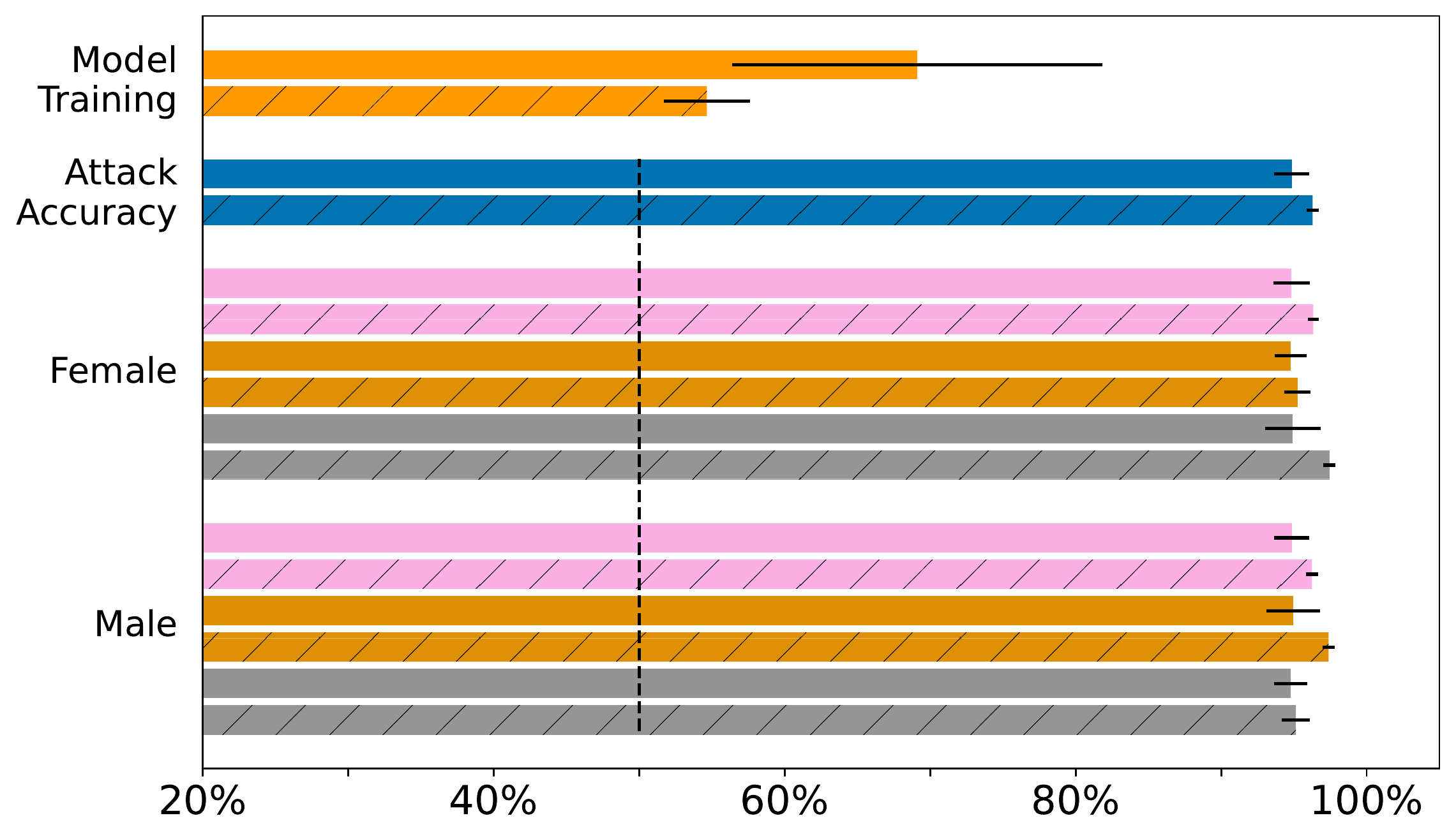}
         \caption{Gender (CelebAHQ)}
     \end{subfigure}

     \caption{Evaluation results for CAIA performed on DenseNet-169 models to infer the gender appearance. The black horizontal lines denote the standard deviation over nine runs. We further state random guessing (dashed line) for comparison. The models were trained on the cropped FaceScrub dataset.}
\end{figure*}
\clearpage

\subsection{ResNeSt-101 - FaceScrub (Uncropped)}

\begin{figure*}[h!]
\centering
     \begin{subfigure}[c]{\textwidth}
         \centering
         \includegraphics[width=\textwidth]{images/robust_legend_long.pdf}
     \end{subfigure}
     
     \begin{subfigure}[b]{0.7\textwidth}
        \centering
         \includegraphics[width=\textwidth]{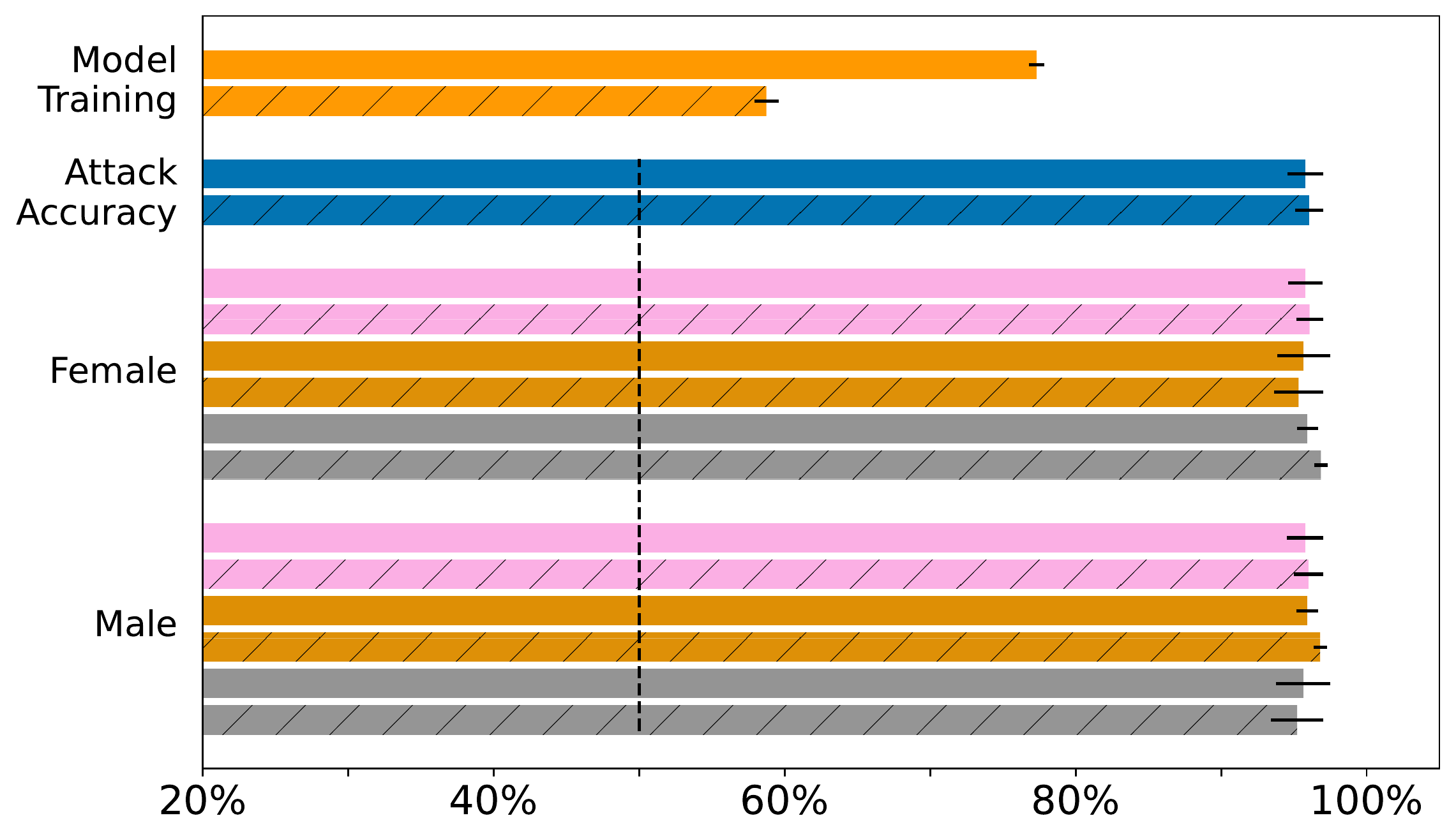}
         \caption{Gender (FFHQ)}
     \end{subfigure}
     \begin{subfigure}[b]{0.7\textwidth}
        \centering
         \includegraphics[width=\textwidth]{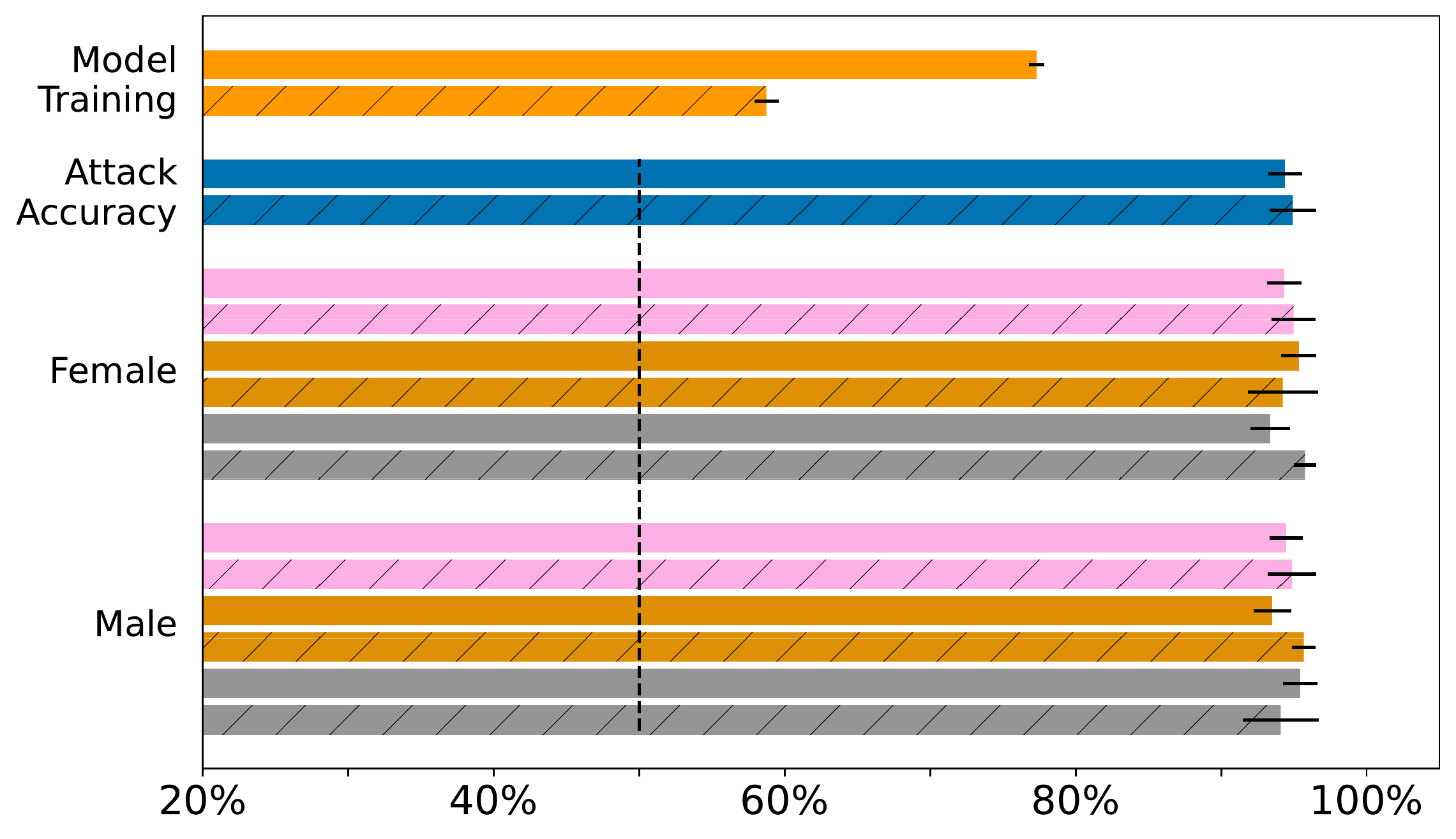}
         \caption{Gender (CelebAHQ)}
     \end{subfigure}

     \caption{Evaluation results for CAIA performed on ResNeSt-101 models to infer the gender appearance. The black horizontal lines denote the standard deviation over nine runs. We further state random guessing (dashed line) for comparison. The models were trained on the cropped FaceScrub dataset.}
\end{figure*}
\clearpage
\clearpage
\onecolumn

\section{Sample Visualization}\label{appx:sample_visualizations}

\subsection{Dataset Samples}\label{appx:dataset_samples}
\begin{figure*}[h!]
    \centering
    \includegraphics[width=0.85\linewidth]{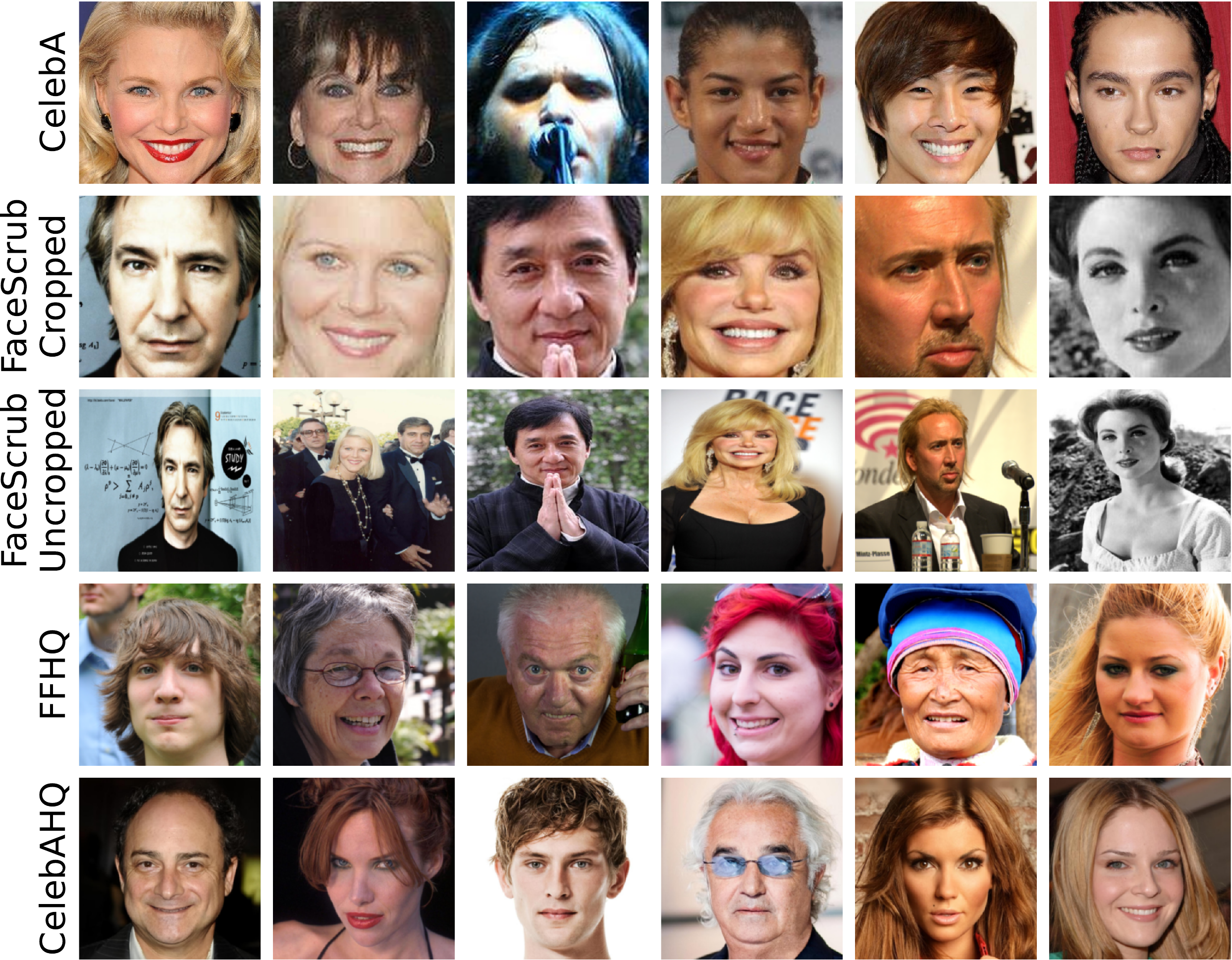}
    \caption{Random samples from the different datasets used throughout this work. Note that CelebA and FaceScrub images have much lower resolution than FFHQ and CelebAHQ. Also, the distribution of the images differs significantly.}
    \label{fig:dataset_samples}
\end{figure*}
\clearpage

\subsection{Attack Samples for Attribute Gender}
\begin{figure*}[h!]
    \centering
    \includegraphics[width=0.85\linewidth]{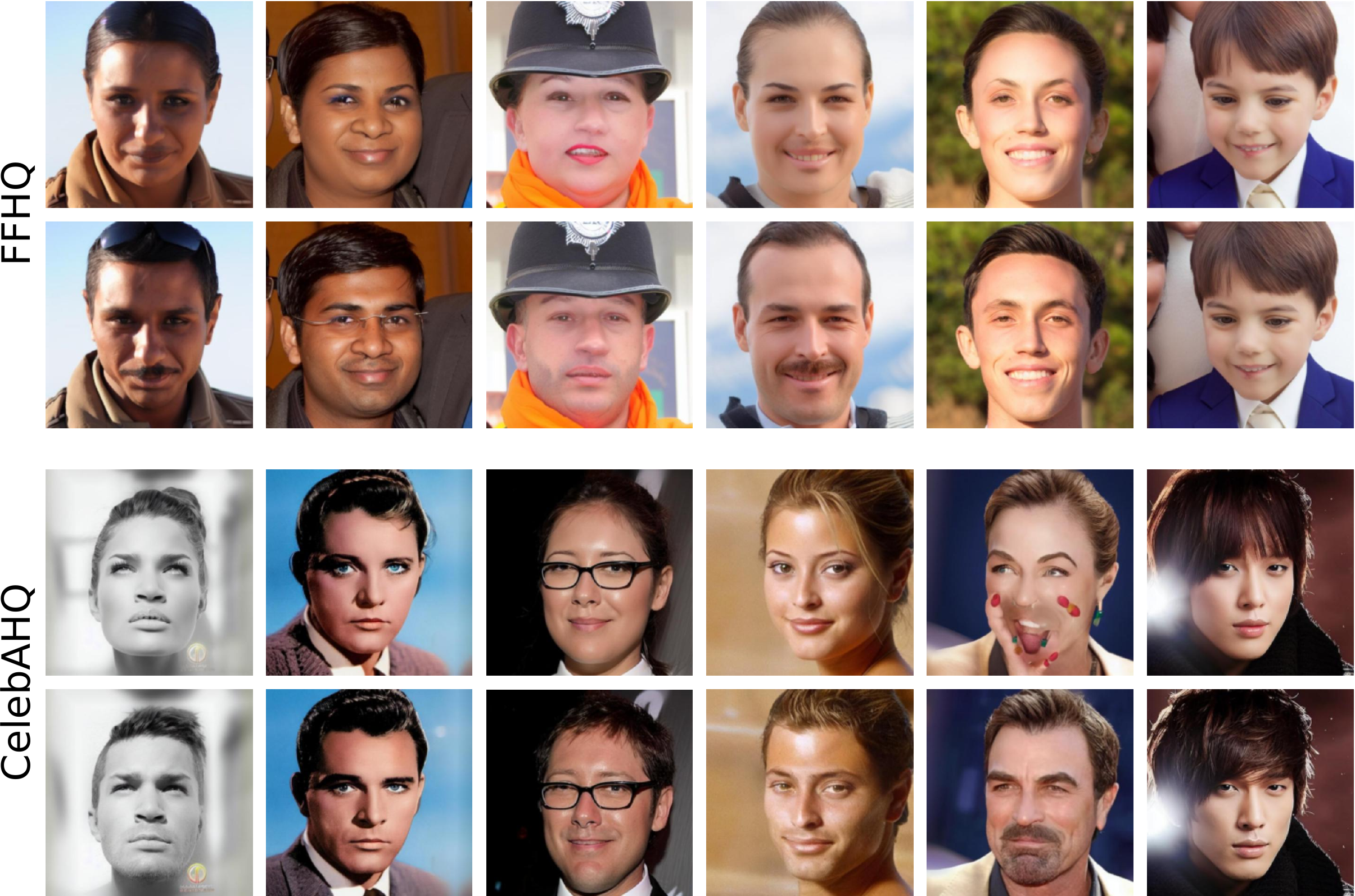}
    \caption{Attack samples to infer the attribute \textit{Gender}.}
    \label{fig:gender_samples}
\end{figure*}
\clearpage

\subsection{Attack Samples for Attribute Eyeglasses}
\begin{figure*}[h!]
    \centering
    \includegraphics[width=0.85\linewidth]{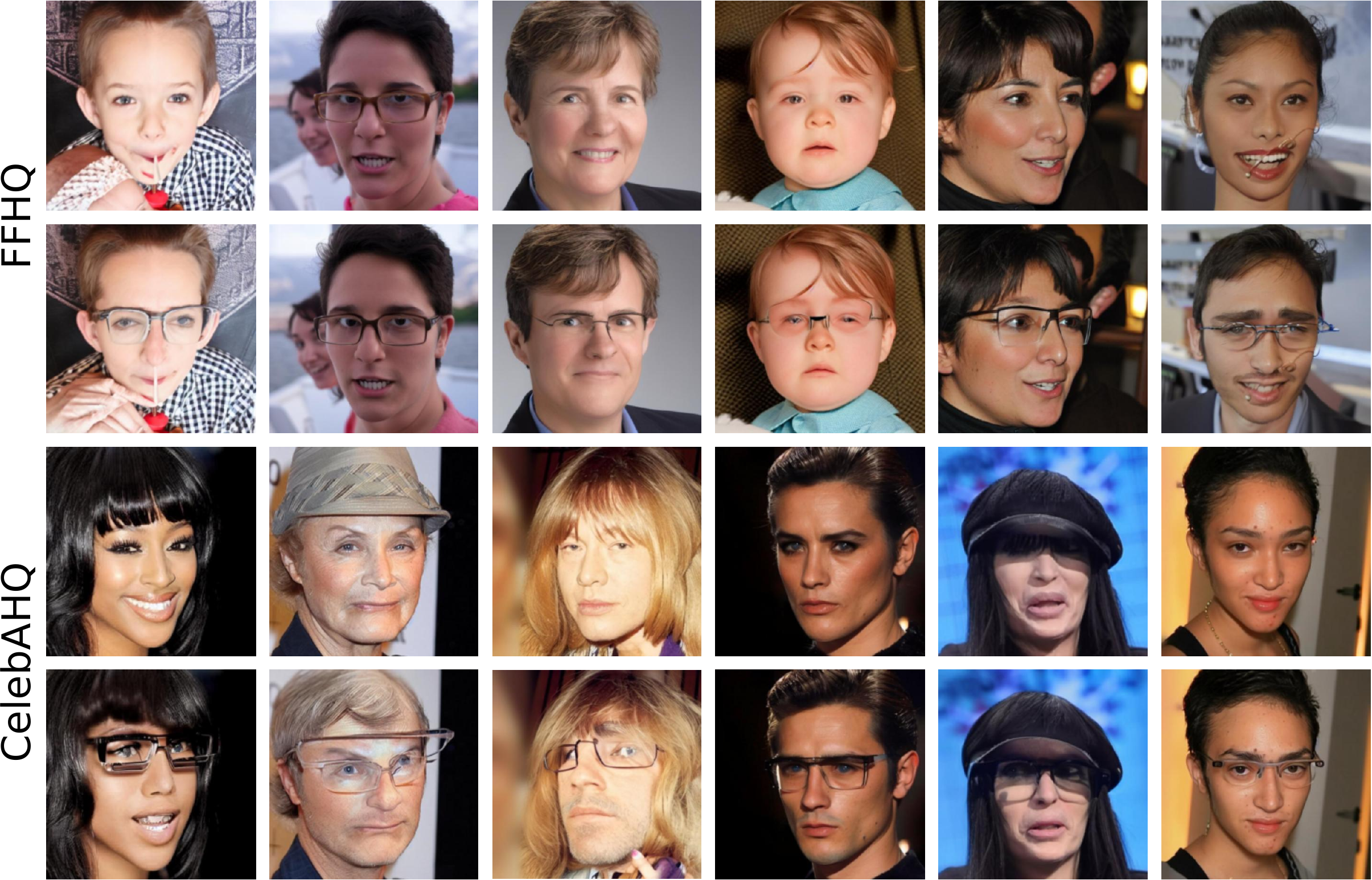}
    \caption{Attack samples to infer the attribute \textit{Eyeglasses}.}
    \label{fig:eyeglasses_samples}
\end{figure*}
\clearpage

\subsection{Attack Samples for Attribute Hair Color}
\begin{figure*}[h!]
    \centering
    \includegraphics[width=0.85\linewidth]{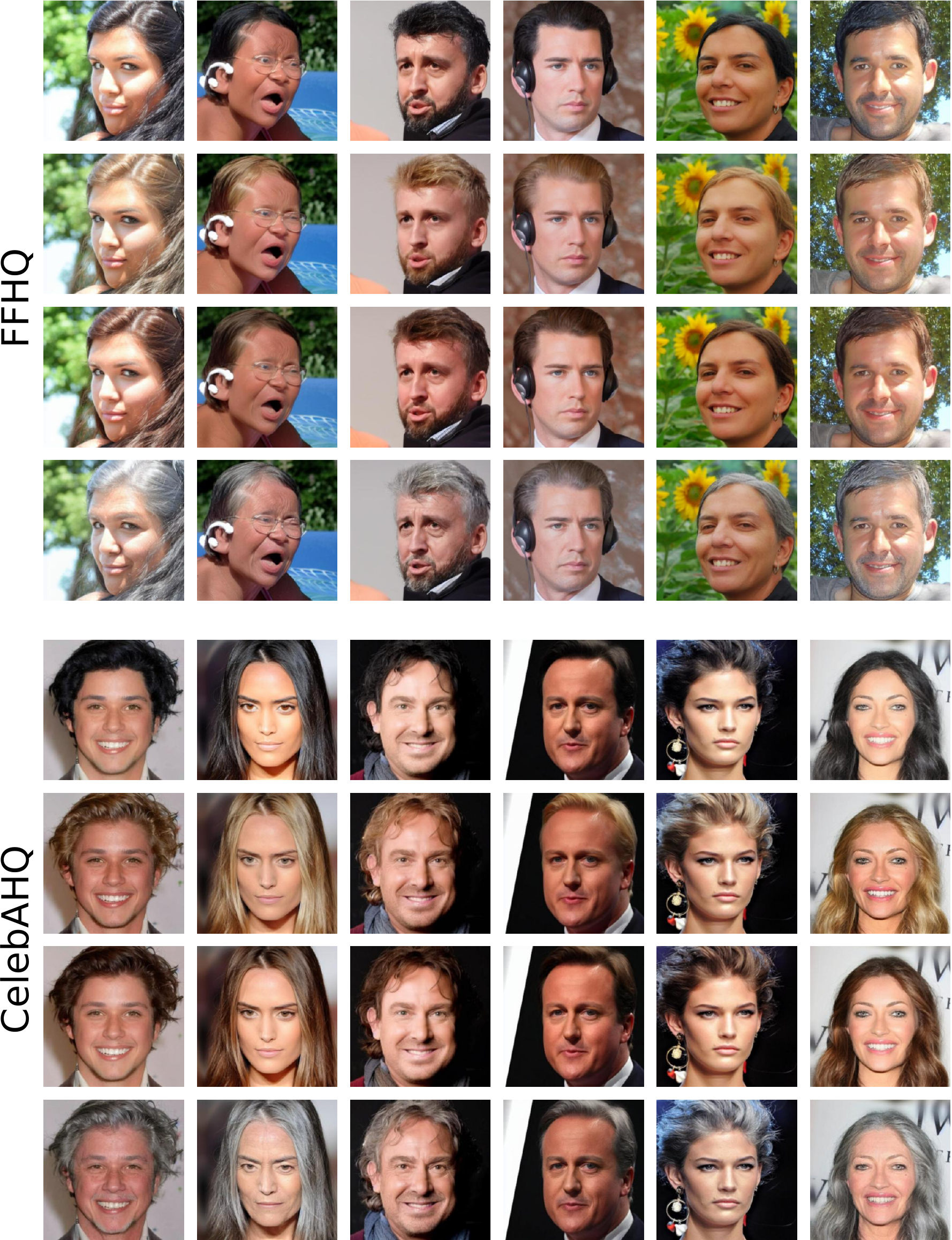}
    \caption{Attack samples to infer the attribute \textit{Hair Color}.}
    \label{fig:hair_color_samples}
\end{figure*}
\clearpage

\subsection{Attack Samples for Attribute Racial Appearance}
\begin{figure*}[h!]
    \centering
    \includegraphics[width=0.85\linewidth]{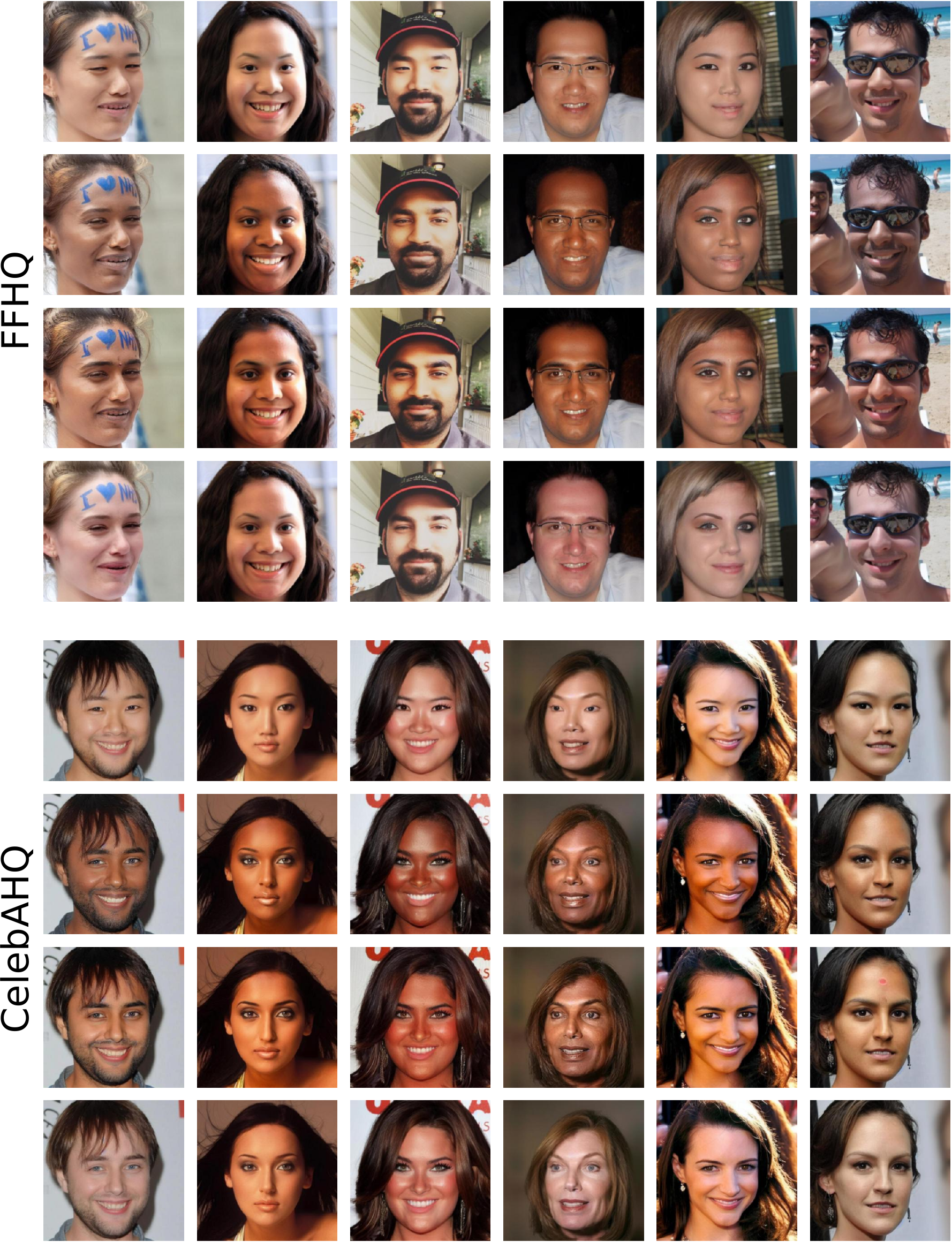}
    \caption{Attack samples to infer the attribute \textit{Racial Appearance}.}
    \label{fig:race_samples}
\end{figure*}

\end{document}